\documentclass{article}

\usepackage[nonatbib,main,final]{neurips_2026}
\usepackage[numbers,sort&compress]{natbib}

\usepackage[utf8]{inputenc} 
\usepackage[T1]{fontenc}    
\usepackage{hyperref}       
\usepackage{url}            
\usepackage{booktabs}       
\usepackage{amsfonts}       
\usepackage{nicefrac}       
\usepackage{microtype}      
\usepackage{xcolor}         

\usepackage[export]{adjustbox}
\usepackage{float}
\usepackage{tikz}
\usepackage{placeins}
\usepackage{pifont}
\newcommand{\cmark}{\ding{51}}
\newcommand{\xmark}{\ding{55}}
\usetikzlibrary{calc,matrix}
\usepackage{multirow}
\usepackage{hyperref}       
\usepackage[normalem]{ulem}
\usepackage{cancel}
\usepackage{bbm}
\usepackage{amssymb}            
\usepackage{mathtools}          
\usepackage{afterpage}
\usepackage{wrapfig}
\usepackage{nicefrac}

\usepackage{stackengine}

\usepackage{amsmath} 
\usepackage{array}

\usepackage{caption}
\usepackage{graphicx}           
\usepackage{adjustbox}
\usepackage{subcaption}         
\usepackage{url}                

\usepackage{amsmath,amsfonts,bm}

\def\ShortFigref#1{Fig.~\ref{#1}}

\def\eqref#1{equation~\ref{#1}}
\def\Eqref#1{Eq.~\ref{#1}}

\def\1{\bm{1}}

\def\vtheta{{\bm{\theta}}}

\DeclareMathAlphabet{\mathsfit}{\encodingdefault}{\sfdefault}{m}{sl}
\SetMathAlphabet{\mathsfit}{bold}{\encodingdefault}{\sfdefault}{bx}{n}

\DeclareMathOperator*{\argmax}{arg\,max}

\usepackage[ruled,vlined,linesnumbered]{algorithm2e}

\newcommand{\qsel}{Q_{\text{sel}}}
\newcommand{\qeval}{Q_{\text{eval}}}

\title{Deep Double Q-learning}

\author{%
	Prabhat Nagarajan, Martha White\textsuperscript{1}, Marlos C. Machado\textsuperscript{1} \\
	Department of Computing Science\\
	University of Alberta\\
    Alberta Machine Intelligence Institute \\
    \textsuperscript{1} Canada CIFAR AI Chair \\
	Edmonton, AB, Canada \\
	\texttt{nagarajan@ualberta.ca, whitem@ualberta.ca, machado@ualberta.ca} 
}

\begin{document}

\maketitle

\begin{abstract}
    Double Q-learning is a classical control algorithm that mitigates the maximization bias of Q-learning.
To do so, it explicitly trains two independent action-value functions and uses them to decouple action-selection from action-evaluation when computing bootstrap targets.
Double DQN adapts this decoupling to deep reinforcement learning (RL) but directly trains only a single action-value function and does not fully decouple its estimators.
Consequently, the two estimators remain correlated, and overestimation persists.
In this paper, we introduce Deep Double Q-learning (DDQL), a deep RL algorithm that explicitly trains two Q-functions through the Double Q-learning objective. 
DDQL stabilizes training through a combination of techniques, including lower replay ratios, longer target network update intervals, and shared layers.
Across 57 Atari 2600 games, DDQL improves aggregate performance over Double DQN, outperforming it on 47 games while further reducing overestimation.
We study key design choices of DDQL, including its network architecture, replay ratio, and minibatch sampling strategies.
\end{abstract}

\section{Introduction}
\label{Introduction}

In the classical Q-learning algorithm~\citep{q_learning}, the agent updates the value estimate of a state-action pair in part towards the value of the maximum estimated action-value in the subsequent state.
Formally, it updates its current estimate, $Q(s,a)$, towards the \textit{bootstrap target} $G = r + \gamma \max_{a'}Q(s',a')$,  where $r$ is the received reward, $\gamma \in [0, 1)$ is the discount factor, and $Q(s',a')$ is the estimated value of state-action pair $(s', a')$.
The computation of $G$ can be rewritten as
\begin{equation} \label{bootstrap_sel_est}
    G = r + \gamma \qeval\Big(s',\argmax_{a'}\qsel(s',a')\Big),
\end{equation}
where $\qeval = \qsel = Q$.
That is, the same $Q$ is used to both select the greedy action whose value should be bootstrapped and then evaluate that greedy action to produce the bootstrap target.

This coupling of action-selection and action-evaluation in the bootstrap target, combined with the maximization operation, gives rise to the so-called \textit{optimizer's curse}~\citep{smith2006optimizer}, or \textit{maximization bias}~\citep{sutton2018reinforcement}.
If we estimate the highest action-value by using a single estimator to select and evaluate the best action, then this estimate is non-negatively biased~\citep{smith2006optimizer,double_q_learning}.
This bias causes overestimation, where the estimated value (i.e., discounted return) of a state-action pair exceeds its true value.
These biased updates create a feedback loop where overestimated values are propagated backward to other action-value estimates.

Double Q-learning~\citep{double_q_learning} is a classical algorithm that explicitly addresses the maximization bias in Q-learning.
Double Q-learning maintains two independent estimators of the action-value function, $Q_1$ and $Q_2$, which are used to decouple action-selection and action-evaluation in the bootstrap target.
Specifically, these Q-functions are trained through \textit{reciprocal bootstrapping}, that is, the action-value functions bootstrap from one another.
If a Q-function, say $Q_1$, is being updated, that Q-function is used to select the value-maximizing action, $\hat{a}^{*} = \argmax_{a'}Q_1(s',a')$. The other Q-function, $Q_2$, evaluates the selected action, $Q_2(s',\hat{a}^{*})$, to assign credit backward through bootstrapping.
By doing so, Double Q-learning mitigates the maximization bias and consequent overestimation of Q-learning.

The Deep Q-network (DQN) algorithm~\citep{dqn}, the analog of Q-learning in deep RL, also suffers from overestimation, and Double DQN~\citep{ddqn} was introduced as a solution.
Like DQN, Double DQN trains a single Q-network to estimate action-values.
To address the maximization bias, it leverages the idea of using two different estimators in the bootstrap target by using the target network, a time-delayed copy of the Q-network, as its second estimator.
However, the target network remains correlated to the Q-network by virtue of being its time-delayed copy, and thus overestimation persists.
Although Double DQN partially mitigates the maximization bias by integrating the ideas of double learning into DQN, adapting Double Q-learning to deep RL should reduce overestimation further.
While a variety of approaches have been proposed to reduce overestimation~\citep{averaged_dqn,crossdqn,redq,waltz2024addressing,scdqn,peer2021ensemble}, including the use of two Q-functions~\citep{td3,decorrelatedddqn}, none have adapted Double Q-learning to deep RL.

In this paper, we introduce Deep Double Q-learning (DDQL), the first adaptation of Double Q-learning to value-based deep RL that learns two action-value functions through reciprocal bootstrapping.
DDQL outperforms Double DQN in aggregate across 57 Atari 2600 games~\citep{ale,revisitingale} and reduces overestimation.
In addition, we study the impact of different design choices in adapting Double Q-learning to deep RL, including the network architecture, maintaining separate replay buffers for each Q-function, and the replay ratio.

\section{Preliminaries}
\label{background}

We formulate the task as a Markov decision process (MDP).
A finite MDP is a tuple, $(\mathcal{S}, \mathcal{A}, \mathcal{R}, P, \gamma)$, where $\mathcal{S}$, $\mathcal{A}$, and $\mathcal{R}$ are finite sets of states, actions, and scalar rewards, respectively.
$P$ specifies the probability $P(s'| s,a)$ of transitioning from state $s$ to $s'$ after taking action $a$.
At each transition, the agent receives a scalar reward $r \in \mathcal{R}$. The value $\gamma \in [0,1)$ is the discount factor.

A policy $\pi$ specifies the probability, $\pi(a|s)$, of taking action $a$ in state $s$.
The \emph{expected} \emph{discounted return} of following $\pi$ after taking action $a$ in state $s$ is $q_{\pi}(s,a) := \mathbb{E}\left[\sum_{t=0}^\infty \gamma^{t} R_{t+1} | S_0 = s, A_0 = a, A_{t \ge 1} \sim \pi \right],$
where $R_t$, $S_t$ and $A_t$ are random variables representing the reward, state, and action at time $t$, respectively.
We call $q_\pi$ the \textit{action-value function} or Q-function.
An optimal policy is denoted $\pi^{*}$. Its Q-function is denoted by $q^{*}$.

\textbf{Q-learning} \citep{watkins_thesis, q_learning} maintains an estimated Q-function $Q$ to learn $q^{*}$ through experience.
Given an arbitrary transition $(s, a, r, s')$, it performs the update $Q(s,a) \leftarrow Q(s,a) + \alpha \left[G - Q(s,a)\right]$, for step size $\alpha > 0$ and bootstrap target $G$.
Q-learning is known to suffer from \textit{overestimation} \citep{issuesfuncapprox,double_q_learning}.
Formally, for some policy $\pi$, given an estimate $Q$ of $q_\pi$, we say that $Q$ overestimates the action-value of state-action pair $(s,a)$ when $Q(s,a) > q_{\pi}(s,a)$.
Q-learning is a special case of off-policy Expected Sarsa~\citep{expected_sarsa,sutton2018reinforcement}, which learns a Q-function for \textit{target policy} $\pi$ with the modified target $G' = r + \gamma \sum_{a'} \pi(a'|s') Q(s',a')$.
When the target policy is the \textit{greedy policy with respect to the current estimated action-value function}, $Q$, we recover Q-learning.
Thus, if $\mathfrak{g}_{Q}$ denotes the greedy policy with respect to $Q$, then overestimation in Q-learning is present when 
$Q(s,a) > q_{\mathfrak{g}_{Q}}(s,a)$.

Q-learning's systematic overestimation stems from the maximization bias in its target computation when bootstrapping~\citep{double_q_learning}.
Suppose that for some $s'$, all actions $a'$ have the same true action-value $q_{\mathfrak{g}_{Q}}(s',a')$.
As estimates are likely inexact, each $Q(s',a')$ can be written $Q(s',a') = q_{\mathfrak{g}_{Q}}(s',a') + \epsilon_{a'}$, for some error $\epsilon_{a'}$.
If for some $a'$, $\epsilon_{a'}$ is positive, then $Q(s',a') >  q_{\mathfrak{g}_{Q}}(s',a')$ is an overestimate.
Combined with maximization, the estimate of the maximum action-value is an overestimate.
That is, $\text{max}_{a'}\text{ }Q(s',a') > \text{max}_{a'}\text{ } q_{\mathfrak{g}_{Q}}(s',a')$.
Consequently, when $\text{max}_{a'}\text{ }Q(s',a')$ is bootstrapped, the update is biased towards an overestimated value. 
This biased update on $Q(s,a)$ then has downstream overestimation effects when predecessors to $s$ are updated.

\textbf{Double Q-learning} \citep{double_q_learning} is designed to address the maximization bias in Q-learning.
It learns two Q-functions, denoted $Q_1$ and $Q_2$. 
At each timestep, one of these Q-functions is selected uniformly at random to be updated, and the other is used to compute the target of the update.
Formally, if $Q_1$ is chosen to be updated, then the update is:
\begin{align} \label{double_q_update}
    Q_1(s,a) \leftarrow Q_1(s,a) + \alpha \left[r + \gamma \text{ }Q_2\Big(s', \underset{a'}{\text{argmax }} Q_{1}(s', a')\Big) - Q_1(s,a)\right].
\end{align}
$Q_1$ is updated to be closer to the target value, where $\qsel = Q_1$ selects an action greedily for evaluation, and $\qeval = Q_2$ evaluates the selected action.
$Q_2$ is updated similarly, where $Q_2$ is used for action selection and $Q_1$ for evaluation in \Eqref{double_q_update}.
By having separate value functions selecting and evaluating the action in the bootstrap target, Double Q-learning is able to ameliorate overestimation.
Double Q-learning also converges under the same standard technical conditions as Q-learning~\citep{double_q_learning}.

\textbf{The Deep Q-network (DQN)} algorithm~\citep{dqn} casts Q-learning as a sequence of deep supervised regression problems.
The agent stores its recent interactions in an experience replay buffer \citep{lin_replay_aaai, linreplay}, which serves as a dataset.
Minibatches are sampled from this buffer to train a Q-function, represented as a neural network with parameters $\vtheta$, through a regression loss on the semi-gradient temporal difference (TD) error \citep{td_learning}.
In this paper, we use the squared error loss $(y - \hat{y})^2$.

To train the Q-network parameters $\vtheta$ on transition $(s,a,r,s')$, the prediction $\hat{y}$ is the Q-network's output $Q(s,a; \vtheta)$ and the target follows $G$ from \Eqref{bootstrap_sel_est}, $y = r + \gamma Q(s',\text{argmax}_{a'}Q(s',a';\vtheta_{\text{sel}}); \vtheta_{\text{eval}})$.
DQN computes these targets using $\vtheta_{\text{sel}} = \vtheta_{\text{eval}} = \vtheta^{-}$, where
$\vtheta^{-}$ are the parameters of the \textit{target network}.
The target network's parameters are \textit{not} directly learned.
They are periodically copied from the Q-network, $\vtheta^{-} \gets \vtheta$, and held fixed for a \textit{target network update interval} (TNU interval).
By remaining fixed for an interval, target networks provide stable, stationary regression targets.

Like Q-learning, DQN also suffers from the maximization bias~\citep{ddqn}, since $\vtheta_{\text{sel}} = \vtheta_{\text{eval}}$.
Double DQN~\citep{ddqn} is mostly identical to DQN except that it uses $\vtheta_{\text{sel}} = \vtheta$.
In other words, instead of having the target network evaluate its own greedy action, Double DQN has the target network evaluate the Q-network's greedy action, acting analogously to $Q_2$ in \Eqref{double_q_update}.
By decoupling action selection and action evaluation in the bootstrap target, Double DQN partially mitigates the maximization bias.

\section{Deep Double Q-learning}
\label{ddql_algs}

This section introduces Deep Double Q-learning (DDQL), our adaptation of Double Q-learning to value-based deep RL.
DDQL differs from Double DQN in that it adapts Double Q-learning to deep RL, whereas Double DQN integrates double updating into DQN.
Desirably, DDQL introduces no additional hyperparameters.
Appendix~\ref{appendix:efficient_implementation} describes how DDQL can be implemented.
Appendix~\ref{appendix:related} describes how DDQL relates to other methods in the literature, including clipped double Q-learning, which has been popularized by actor-critic methods like TD3~\citep{td3} and Soft Actor-Critic~\citep{soft_actor_critic}.

To adapt Double Q-learning to deep RL, we identify three defining features of Double Q-learning, which stem from the double estimator~\citep{double_q_learning} upon which Double Q-learning is built.
These features are (1) \textit{target bootstrap decoupling}, (2) \textit{double estimation with reciprocal bootstrapping}, and (3) \textit{dataset partitioning}.
Each defining feature can be viewed as an incremental step upon the previous towards decoupling action-selection and action-evaluation in the bootstrap target.
The first feature, target bootstrap decoupling, simply refers to the use of different Q-functions $\qsel$ and $\qeval$ in \Eqref{bootstrap_sel_est}, i.e., $\qsel \neq \qeval$.
This feature is the only feature of the three integrated by Double DQN, with $\qsel = Q(\cdot, \cdot; \vtheta)$ and $\qeval = Q(\cdot, \cdot; \vtheta^-)$.
DDQL additionally integrates the latter two features.

\subsection{Double estimation and reciprocal bootstrapping}

\textit{Double estimation with reciprocal bootstrapping} refers to the explicit training of two Q-functions where one is used to select actions in the bootstrap target and the other is used to evaluate the selected action.
DDQL does this by maintaining two sets of Q-network parameters, $\vtheta_1$ and $\vtheta_2$, which are both trained.
As in most DQN-style setups~\citep{dqn, ddqn, per, duelingnets, distributionaldqn, rainbow}, we also maintain target networks for stability, with parameters $\vtheta_1^{-}$ and $\vtheta_2^{-}$.
Suppose that $\vtheta_1$ is updated for some transition $(s, a, r, s')$. Then the DDQL target is:
\begin{equation} \label{ddql_update}
    y_1(s') = r + \gamma Q(s',  \underset{a'}{\arg\max}\, Q(s',a'; \vtheta_1^{-}); \vtheta_2^{-}).
\end{equation}
If parameters $\vtheta_1$ are being updated, their corresponding target network parameters $\vtheta_1^-$ are used to select a greedy action, and parameters $\vtheta_2^-$ are used to evaluate this action to produce the target value.
We use target networks both for action-selection and action-evaluation in the bootstrap target to ensure fixed, stationary targets for the duration of the TNU interval, as is done in DQN.
The presence of two Q-networks and two target networks creates several alternative ways to use two different networks in the bootstrap target.
These alternatives are discussed in Appendix~\ref{appendix:adaptation}.

For DDQL, we use the squared error loss, although in principle any loss that is a function of the TD error can be used.
Given a minibatch $\mathcal{B} = \{(s_i,a_i,r_{i},s_i')\}_{i=1}^{N}$ of $N$ transitions, we have the loss
\begin{equation} \label{ddql_single_q_loss}
    \mathcal{L}_{1}(\mathcal{B}) = \frac{1}{N}\sum_{i=1}^N \big(y_1(s_i') - Q(s_i,a_i; \vtheta_1)\big)^2.
\end{equation}
When updating $\vtheta_2$, $y_2(s')$ and $\mathcal{L}_2$ are defined analogously to Equations~\ref{ddql_update} and~\ref{ddql_single_q_loss}, respectively.

In tabular Double Q-learning, only one of the two Q-functions is updated at a time using the online transition.
When sampling from a replay buffer, which is relatively offline and stationary, we can instead sample two separate minibatches from the buffer and update each Q-function simultaneously in a single gradient update by minimizing the loss $\mathcal{L}_{\text{DDQL}} = \mathcal{L}_1 + \mathcal{L}_2$.

\subsection{Dataset partitioning}

In Double Q-learning, each experience transition is used to update only one of the two Q-functions, i.e., the sets of transitions used to train each Q-function are non-overlapping.
We refer to this feature of Double Q-learning as \textit{dataset partitioning}.
Dataset partitioning further decouples the two Q-functions by separating their training data.
To approximate dataset partitioning, we sample distinct minibatches from a single buffer to train each Q-function.
Section~\ref{dataset_partitioning_experiments} explores a strict form of dataset partitioning that  uses separate replay buffers to train each Q-function.

\subsection{Network architecture}
\begin{figure}[t]
    \centering
    \includegraphics[width=0.9\linewidth]{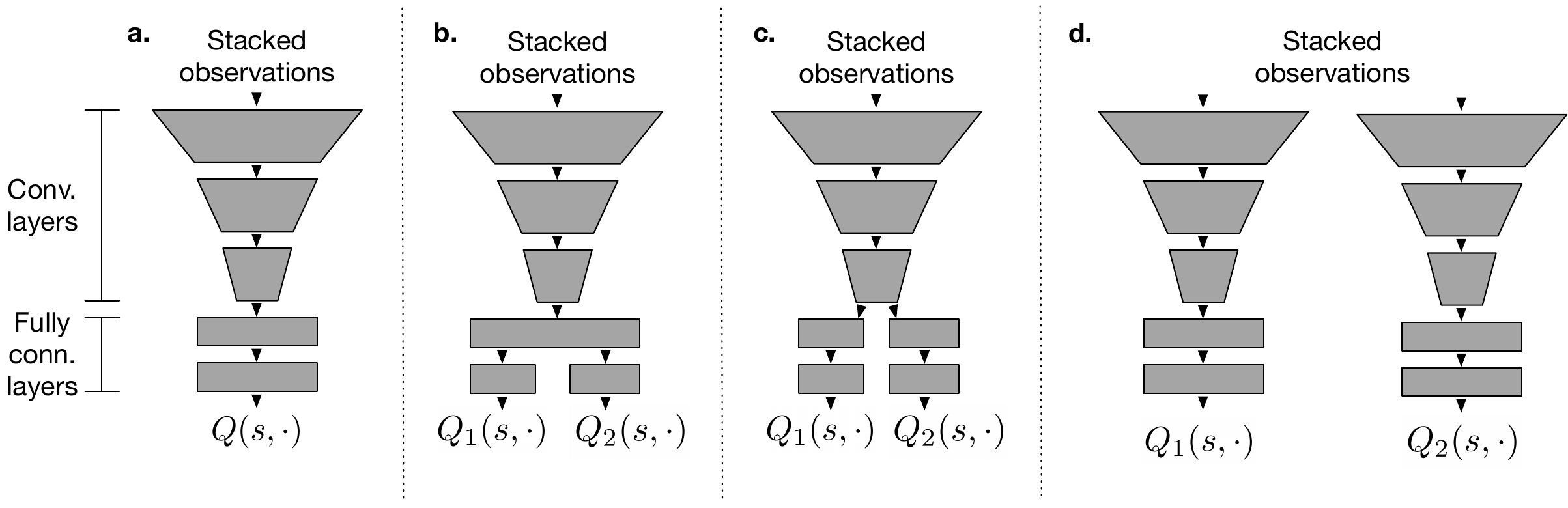}
        \caption{\textbf{DDQL architectures}.
    \textbf{a)}\ DQN/Double DQN, \textbf{b)}\ DDQL, \textbf{c)}\ DM-DDQL, and \textbf{d)}\ DN-DDQL.}
    \label{fig:architectures}
\end{figure}

\begin{wraptable}[11]{R}{0.66\textwidth}
\centering
\caption{Different adaptations of Double Q-learning to deep RL in terms of its defining features.}
\label{double_q_faithfulness}
\setlength{\tabcolsep}{4pt} 
\small 
\begin{tabular}{@{}l l|c|c|c@{}}
\toprule
\multicolumn{2}{c|}{\textbf{\begin{tabular}[c]{@{}c@{}}Double Q-learning \\ Defining Feature\end{tabular}}} & \textbf{\begin{tabular}[c]{@{}c@{}}Double \\ DQN\end{tabular}} & \textbf{\begin{tabular}[c]{@{}c@{}}DDQL \& \\ DM-DDQL\end{tabular}} & \textbf{\begin{tabular}[c]{@{}c@{}}DN-DDQL\end{tabular}} \\ 
\midrule
\textbf{(1)}   & Target bootstrap decoupling    & \cmark & \cmark & \cmark \\
\textbf{(2) }  & Double estimation              & -      & \cmark & \cmark \\
\textbf{(2a)}  & Exclusive parameters  & -      & \xmark & \cmark \\
\textbf{(3)}   & Dataset partitioning           & -      & \cmark & \cmark \\
\textbf{(3a)}  & Separate buffers               & -      & \xmark & \xmark \\
\bottomrule
\end{tabular}
\end{wraptable}
When bridging the gap from tabular to deep RL, we transition from two lookup-table Q-functions to neural networks.
DDQL uses a single network with two output heads, each representing a Q-function.
Doing so allows our two Q-functions to benefit from a shared representation and potentially benefit from auxiliary task effects~\citep{aux_tasks} while still maintaining separate parameters.

We explore two additional architectures, which we call double-MLP DDQL (DM-DDQL) and double-network DDQL (DN-DDQL).
DM-DDQL uses a single shared trunk of convolution layers, branching into two two-layer multi-layer perceptrons (MLPs), each representing a Q-function.
In DN-DDQL, the two Q-functions are parameterized by separate networks.
In all DDQL variants, $\vtheta_1$ and $\vtheta_2$ refer to the parameters of the respective Q-functions, even when parameters are shared, as in DDQL and DM-DDQL.
\ShortFigref{fig:architectures} depicts the three DDQL architectures.
Algorithm~\ref{alg:ddql} provides the pseudocode for DDQL.
Table~\ref{double_q_faithfulness} highlights how different algorithms and architectures incorporate the defining ideas of Double Q-learning. 

\begin{algorithm}[t]
\caption{Deep Double Q-learning}
\label{alg:ddql}
\SetKwInput{Input}{Input}
\Input{target network update interval $C$, gradient update frequency $k$}
Initialize replay buffer $\mathcal{D}$, Q-network parameters $\vtheta_1$ and $\vtheta_2$, $\emph{reset} \gets \text{True}$\;
\For{$t = 0$ to $T$}{
    \If{$\text{reset} == \text{True}$}{
            $s_t \gets $ sample start state\;
        }
    \If{$t \pmod{kC} == 0$}{
        $\vtheta_1^- \leftarrow \vtheta_1$, $\vtheta_2^- \leftarrow \vtheta_2$\;
    }
    Select and execute action $a_t$ and observe $s_{t+1}, r_{t+1}$\;
    Store $(s_t, a_t, r_{t+1}, s_{t+1})$ in $\mathcal{D}$\;
    $\emph{reset} \gets (s_{t+1}$ is terminal$)$\;
    \If{$t \pmod k == 0$}{
        Sample distinct minibatches $\mathcal{B}_1, \mathcal{B}_2$ from $\mathcal{D}$\;
        Update $\vtheta_1, \vtheta_2$ through minibatch gradient descent on $\mathcal{L}_{\text{DDQL}}$\;
    }
}
\end{algorithm}

\subsection{Key algorithm details} \label{algorithm_details}

We \textbf{initialize} the independent parts of the Q-functions identically. 
That is, in DDQL, two output heads share the same initialization; in DM-DDQL, the two MLPs are initialized identically; and in DN-DDQL, the networks are initialized identically.
The reason for this is twofold.
First, in tabular Double Q-learning, the Q-functions are typically  both initialized to zeros.
Second, there is no apparent benefit to having different initial action-value predictions, as the two Q-functions will, by design, deviate as they train on different experiences.

The \textbf{replay ratio} is the number of gradient updates per environment transition~\citep{replay_fundamentals}.
Our DDQL agents use a replay ratio of $\nicefrac{1}{8}$, as opposed to DQN and Double DQN's $\nicefrac{1}{4}$.
Specifically, both Q-functions are updated \textit{simultaneously} every eight timesteps.
Though this replay ratio induces half as many parameter updates as Double DQN, it ensures that the total number of updates to individual Q-functions matches Q-learning, as two Q-functions are updated simultaneously.
This ratio is also consistent with Double Q-learning, which on average updates each Q-function half as often as a Q-learner.
The use of simultaneous updates is relevant for DDQL to mitigate gradient oscillation.
Section~\ref{replay_ratio_experiments} studies the role of the replay ratio in DDQL.

The \textbf{target network update interval} (TNU interval) is another important hyperparameter.
While typically described in terms of environment timesteps, this interval should be a function of the number of parameter updates, as noted by \citet{parametricmodels}.
DDQL uses a TNU interval of 7,500 updates, inherited from the tuned version of Double DQN~\citep{ddqn}, which uses a 3x longer interval than DQN.
As DDQL performs gradient updates half as often as Double DQN, DDQL refreshes its target network half as often in terms of environment timesteps.
Despite doubling the batch size, halving the update frequency results in minimal slowdowns.
DDQL and DN-DDQL are respectively 11\% and 25\% slower than Double DQN.

\subsection{Double DQN versus Deep Double Q-learning}

The core similarity between Double DQN and DDQL is that they both use target bootstrap decoupling.
That is, they both use one Q-function to select the highest-valued action in the bootstrap target and another Q-function to evaluate the selected action.
This use of two Q-functions in DQN's bootstrap target gives the name Double Deep Q-networks.
Double DQN borrows one particular idea of Double Q-learning and applies it to a Q-learning style algorithm, DQN.

By contrast, DDQL takes the Double Q-learning algorithm itself and adapts it to deep RL.
That is, DDQL is Double Q-learning using replay buffers and target networks.
Double DQN only performs gradient updates on a single Q-function with a single loss function, and uses the target network as the second Q-function in the bootstrap target.
Importantly, DDQL trains two Q-functions, each with its own loss, and uses each Q-function in the target computation of the other Q-function (reciprocal bootstrapping).
It also trains the two Q-functions on different transitions (dataset partitioning).
Training two Q-functions introduces several design choices, including the choice of network architecture and the minibatch sampling strategy.
We integrate the three defining features of Double Q-learning into our algorithm, and accordingly name it Deep Double Q-learning.

\section{Evaluating Deep Double Q-learning}
\label{evaluating_ddql}

This section summarizes our empirical methodology, which is comprehensively outlined in Appendix~\ref{appendix:experiments}.
We then show that DDQL reduces overestimation and outperforms Double DQN.

\subsection{Experimental setup} \label{experimental_setup}

\textbf{Evaluation.} We conduct experiments in the Arcade Learning Environment \citep{ale,revisitingale}, the platform on which Double DQN was originally evaluated.\footnote{Each run takes 2 to 3.5 GPU days on average. We estimate our experiments require between 8 and 13 GPU years.}
We evaluate Double DQN, DDQL, DM-DDQL, and DN-DDQL across the Atari-57 set of environments~\citep{ddqn}.
To study specific aspects of DDQL, we perform additional ablations on a set of games we call \textit{Ablation-11}, which combine the Atari-5 games~\citep{atari5}---a representative subset of Atari-57---and six additional games that have been used to study overestimation in DQN~\citep{ddqn,averaged_dqn}.
We use the environment settings proposed by \citet{revisitingale}.
That is, we use sticky actions, the full action set, and game-over termination.

\textbf{Double DQN.} All algorithms\footnote{\href{https://github.com/prabhatnagarajan/ddql}{github.com/prabhatnagarajan/ddql}.} are built upon the PFRL library~\citep{pfrl}.
Our Double DQN uses the same architecture and hyperparameters as the original paper's tuned Double DQN~\citep{ddqn}.
Reflecting recent advances~\citep{revisiting_rainbow,statistical_precipice}, we use the Adam optimizer~\citep{adam} with the squared error loss instead of the original RMSprop with the Huber loss.
Our settings for Adam follow those of~\citet{rainbow}, following standard practice in the DQN literature~\citep{optimisticperspective,statistical_precipice,revisiting_rainbow, stopregressing}.

\textbf{DDQL.} Following \citet{ddqn}'s approach of maintaining the same hyperparameters as the baseline, all DDQL variants share Double DQN's hyperparameters, unless stated otherwise in Section~\ref{ddql_algs}.
DDQL's Q-functions share Double DQN's hidden layers and branch into two architecturally identical output heads that match Double DQN's output layer.
DM-DDQL's Q-functions share Double DQN's convolutional layers with two architecturally identical two-layer MLPs that match Double DQN's final two layers.
Each of DN-DDQL's two networks matches Double DQN's architecture.
The behavior policy for the DDQL variants during training is an $\epsilon$-greedy policy with respect to the average of the two Q-functions: $\frac{1}{2}Q(s,a; \vtheta_1) + \frac{1}{2}Q(s,a; \vtheta_2)$, following Double DQN's $\epsilon$-decay schedule.
Each algorithm is run for five seeds for each evaluated game unless stated otherwise.

\textbf{Measuring Overestimation.}
Double DQN learns a Q-function that estimates the expected discounted return of the greedy policy with respect to that very Q-function.
Thus, overestimation occurs when the Q-value estimate exceeds the expected return of its greedy policy.
Similar to~\citet{ddqn}, we measure overestimation during periodic evaluation phases.
For all state-action pairs of all completed episodes in this evaluation phase, we compute the average difference between the action-value predictions and their corresponding achieved returns to compute overestimation.
For DDQL, the action-value prediction is the average of its two Q-values.
Appendix~\ref{appendix:overestimation} outlines our overestimation computation, including how we handle episodes that are truncated due to time limits.

\subsection{Results: overestimation and performance}

\begin{figure}[t!]
    \centering
        \includegraphics[width=\linewidth]{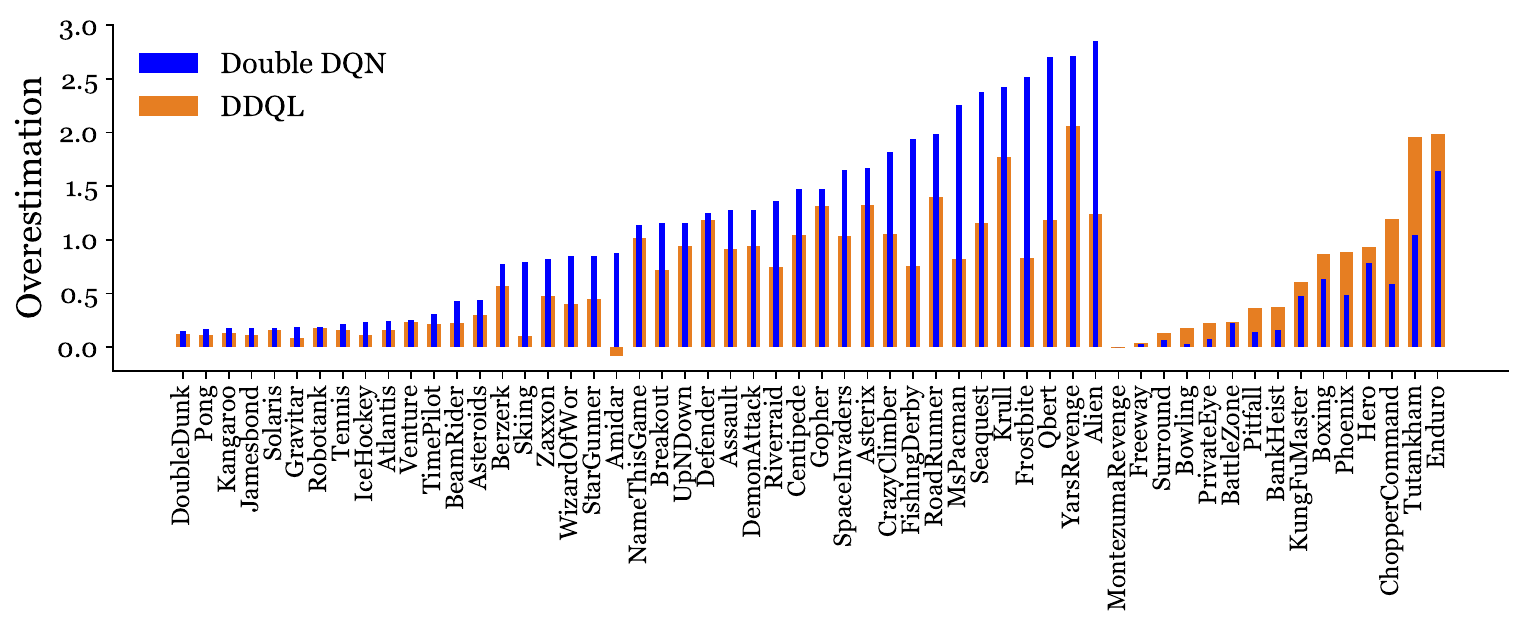}
    \caption{\textbf{Per-game DDQL overestimation.} Final overestimations averaged across five seeds of Double DQN and DDQL (Video Pinball is omitted for visibility). 
    }
    \label{fig:main_overestimations}
\end{figure}

\ShortFigref{fig:main_overestimations} depicts the average final overestimation of Double DQN and DDQL across games (per-game overestimation curves are in \ShortFigref{Atari57:Overestimation:page_2} in Appendix \ref{appendix:more_results}).
While the degree of overestimation is game-dependent, as the scale of returns can vary between games, Double DQN overestimates in all games other than Montezuma's Revenge.
This overestimation suggests that its target bootstrap decoupling with the Q-network and target network is insufficient for eliminating overestimation.
This result is to be expected: because the target network is a time-delayed copy of the Q-network, the two networks remain heavily correlated.
DDQL also overestimates, but it reduces overestimation compared to Double DQN in 42 out of 57 environments.

\begin{figure}[t]
    \centering
    \includegraphics[width=0.32\textwidth, valign=m]{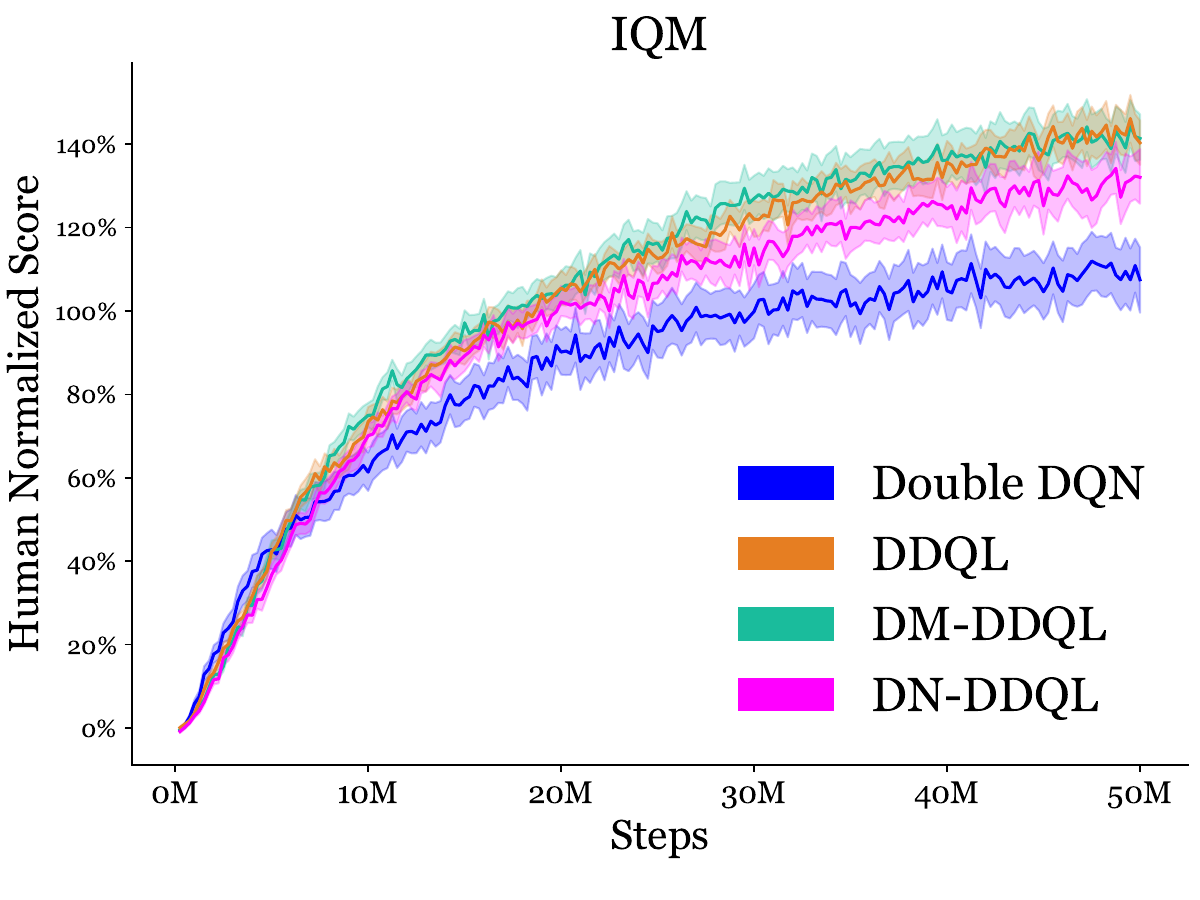}\hfill
    \includegraphics[width=0.67\textwidth, valign=m]{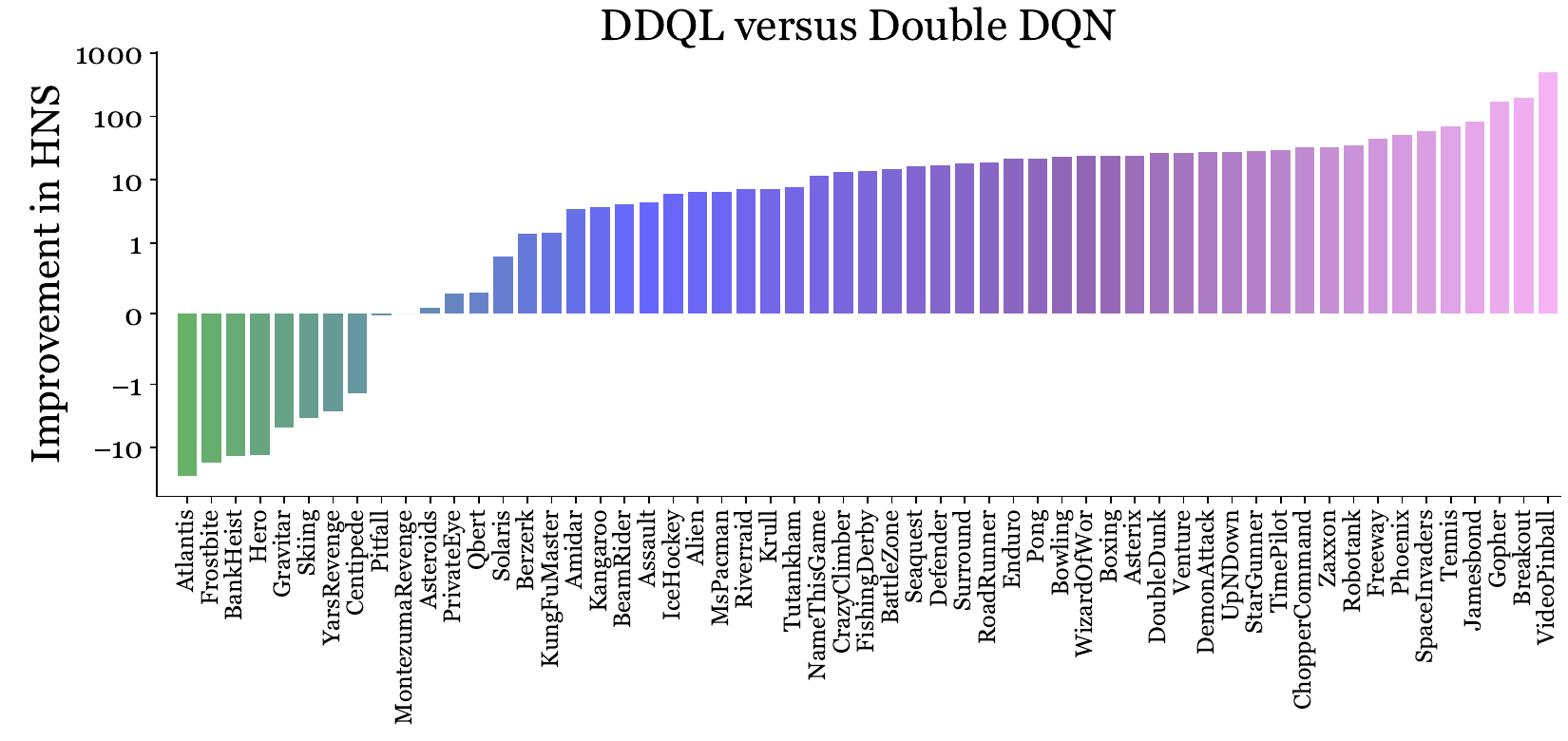}
    \caption{\textbf{Aggregate learning curves and per-game performance improvements.} (\textit{left}) Interquartile mean of the HNS throughout training. The shaded region represents a 95\% stratified bootstrap confidence interval~\citep{statistical_precipice}. (\textit{right}) Per-game improvement in HNS of DDQL over Double DQN, calculated as the average area under the curve across five seeds. The y-axis is on a log-scale.}
    \label{fig:combined_hns_results}
\end{figure}

\ShortFigref{fig:combined_hns_results} depicts DDQL's performance improvements over Double DQN.
\ShortFigref{fig:combined_hns_results} (\textit{left}) plots the standard metric of interquartile mean (IQM)~\citep{statistical_precipice} of the human-normalized score (HNS)~\citep{ddqn} across the 57 Atari games.
In Appendix~\ref{appendix:more_results}, we show the per-game learning curves in \ShortFigref{Atari57:Score:page_2} and the per-game scores in Table~\ref{table_results}.
As a percentage of Double DQN's IQM of the HNS, DDQL performs 31\% better than Double DQN.
\ShortFigref{fig:combined_hns_results} (\textit{right}) shows the per-game improvements, in terms of average area under the curve (AUC) of the HNS of DDQL over Double DQN.
DDQL outperforms Double DQN in 47 out of 57 games.
The log-scale y-axis shows that performance decreases are often small, while the gains can be quite high.

\section{Understanding Deep Double Q-learning}
\label{understanding_ddql}
In this section, we study the impact of different design choices when adapting Double Q-learning to deep RL.
In particular, we study the impact of maintaining separate parameters, of using two separate buffers for dataset partitioning, and of varying the replay ratio.

\subsection{Independent parameters}
\label{double_net_experiments}

\begin{figure}[t!]
    \centering
        \includegraphics[width=\linewidth]{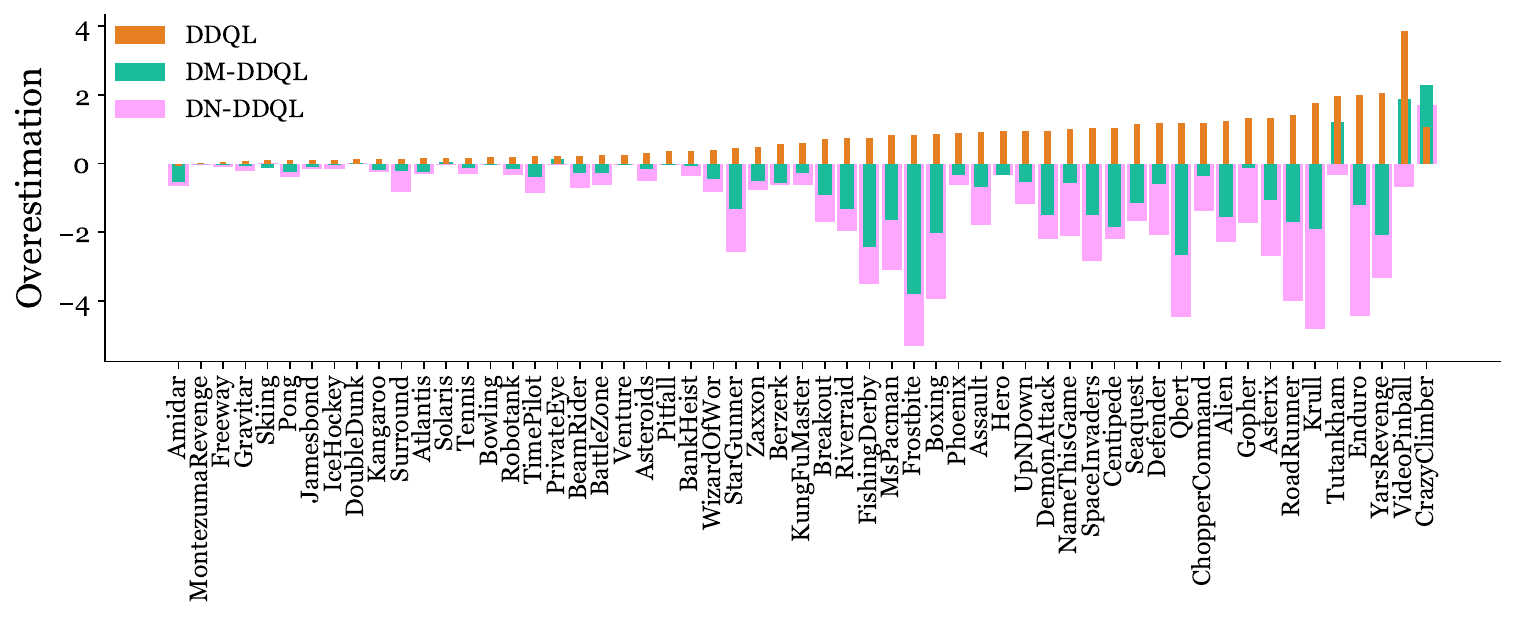}
    \caption{\textbf{Per-game overestimation for the DDQL variants.} Final overestimations averaged across five seeds per algorithm on each of the 57 games. DDQL generally overestimates the most, followed by DM-DDQL, then DN-DDQL. The latter two often underestimate.
    }
    \label{fig:dn_ddql_overestimation}
\end{figure}

By evaluating DM-DDQL and DN-DDQL, we can examine the impact of parameter sharing on overestimation.
DN-DDQL decouples the two Q-functions further by representing them using separate neural networks.
The result of this decoupling is large reductions in overestimation, to the point of underestimation, a phenomenon known to occur in Double Q-learning~\citep{double_q_learning}.
This underestimation is shown in \ShortFigref{fig:dn_ddql_overestimation}, which shows the per-game overestimations of all the DDQL variants.
The large gap in overestimation between DDQL and DN-DDQL suggests that the sharing of the hidden layers in the network of DDQL contributes substantially to overestimation.
This finding is reinforced by DM-DDQL, which represents an intermediate architecture between DDQL and DN-DDQL.
Like DDQL, DM-DDQL shares parameters, but fewer.
Correspondingly, DM-DDQL exhibits an intermediate amount of overestimation between DDQL and DN-DDQL.
These results suggest that structural independence in the Q-functions de-correlates the action-value functions.

\ShortFigref{fig:combined_hns_results} (\emph{left}) depicts DM-DDQL and DN-DDQL's performance improvements over Double DQN in aggregate across the 57 games (per-game improvements in \ShortFigref{fig:per_game_dn_ddql_perf} in Appendix~\ref{appendix:more_results}).
As a percentage of Double DQN's IQM of the HNS, DM-DDQL and DN-DDQL perform 32\% and 23\% better than Double DQN, respectively.
Though DN-DDQL closely embodies the original Double Q-learning algorithm, using separate networks yields lower performance than partial parameter sharing.
DM-DDQL, by contrast, is just as effective as DDQL.

Sharing hidden layers, as in DDQL and DM-DDQL, strikes a middle ground between reducing overestimation and retaining good performance.
That said, all three architectures perform well, demonstrating that the DDQL algorithm is compatible with different architectures and effective under different degrees of parameter-sharing.
If overestimation is a major concern, DM-DDQL may be the best option.
In general we recommend DDQL due to its simplicity and generality.

\subsection{Partitioned datasets} 
\label{dataset_partitioning_experiments}

\begin{figure}[t]
    \centering
    \includegraphics[width=\linewidth]{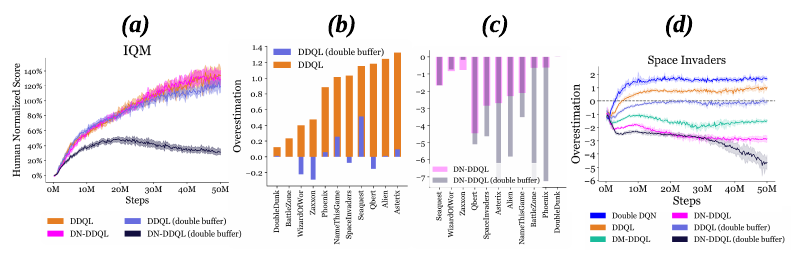}
    \caption{\textbf{Dataset partitioning experiments.} (\emph{a}) The IQM of the HNS on the Ablation-11 games of DDQL and its double-buffer counterparts. Shaded region: 95\% stratified bootstrap confidence interval. (\emph{b} \& \emph{c}) Final overestimation, averaged over five seeds of DDQL and DN-DDQL, respectively, and their double-buffer counterparts. (\emph{d}) Overestimation throughout training for different algorithms averaged across five seeds on Space Invaders. Translucent lines depict individual seeds.}
    \label{fig:dataset_partitioning_experiments}
\end{figure}

The third defining feature of Double Q-learning is dataset partitioning, where each experience transition is used to train only a single Q-function.
DDQL maintains a single buffer and samples distinct minibatches to update each Q-function.
Sampling distinct minibatches, however, does not ensure strict dataset partitioning.
Experiences are sampled multiple times from the buffer and so transitions are inevitably used to train both Q-functions.

To study the impact of strict dataset partitioning, we modify DDQL to maintain two separate buffers with half the capacity of a standard replay buffer, leaving the algorithm otherwise unchanged.
Transitions are added uniformly at random to one of the two buffers.
Each buffer generates minibatches exclusively for one of the two Q-functions to ensure that transitions are used to train only one Q-function.
By separating the data used to train each Q-function, we further decouple the two Q-functions.
Thus, we hypothesize reduced overestimation in this setting.

We compare DDQL to its double-buffer counterparts on the Ablation-11 environments in \ShortFigref{fig:dataset_partitioning_experiments} (\emph{a}), reporting the IQM of the HNS.
DDQL is slightly better than its double-buffer counterpart.
Using separate buffers in DN-DDQL substantially harms performance.
Training on different buffers may require better generalization from the Q-networks, a requirement that may be softer when parameters are shared.
\ShortFigref{DatasetPartitioningDH:Ablation11:Score} and \ShortFigref{DatasetPartitioningDN:Ablation11:Score} in Appendix~\ref{appendix:more_results} depict the per-game learning curves for DDQL and DN-DDQL respectively.

\ShortFigref{fig:dataset_partitioning_experiments} (\emph{b}) and (\emph{c}) show the average final overestimation of DDQL and DN-DDQL and their double-buffer counterparts (per-game curves for DDQL shown in \ShortFigref{DatasetPartitioningDH:Ablation11:Overestimation} in Appendix~\ref{appendix:more_results}).
As hypothesized, the use of separate buffers reduces overestimation substantially for both DDQL variants.
DDQL's overestimation goes closer to 0, providing further evidence that separating the data used to train each estimator, as is done in Double Q-learning, reduces overestimation.
Applied to DN-DDQL, which already separates parameters, it reduces overestimation even further, albeit at the cost of poor performance.

The impact of integrating Double Q-learning's defining features on overestimation is demonstrated in \ShortFigref{fig:dataset_partitioning_experiments} (\emph{d}), which shows individual overestimation for different algorithms in the game Space Invaders.
Double DQN, which simply decouples estimators in the bootstrap target using its Q-network and target network, has the most overestimation.
DDQL, DM-DDQL, and DN-DDQL, the methods which integrate double estimation with reciprocal bootstrapping and a weaker form of dataset partitioning, reduce overestimation.
The gaps between DDQL, DM-DDQL, and DN-DDQL demonstrate that the lower the degree of parameter-sharing, the greater the reduction in overestimation.
Lastly, the double-buffer variants show that strict dataset partitioning, where we use two buffers to explicitly separate the data used to train each Q-function, reduces overestimation further yet.

\subsection{Replay ratio} \label{replay_ratio_experiments}

\begin{wrapfigure}{R}{0.65\textwidth}
    \vspace{-20pt}
    \centering
    \includegraphics[width=0.48\linewidth, valign=m]{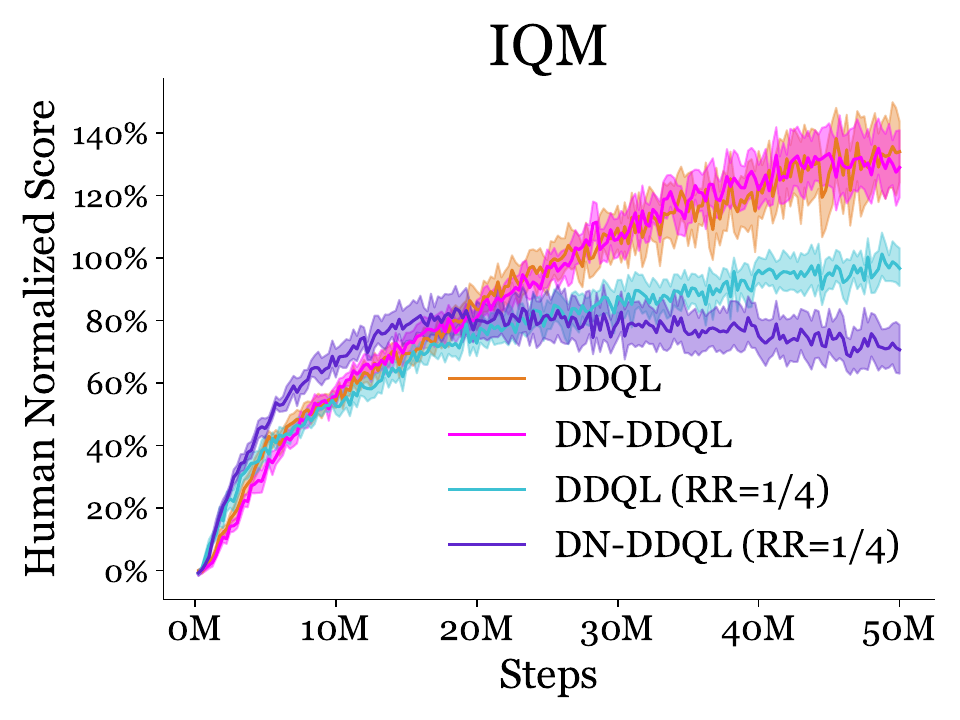}
    \hfill
    \includegraphics[width=0.48\linewidth, valign=m]{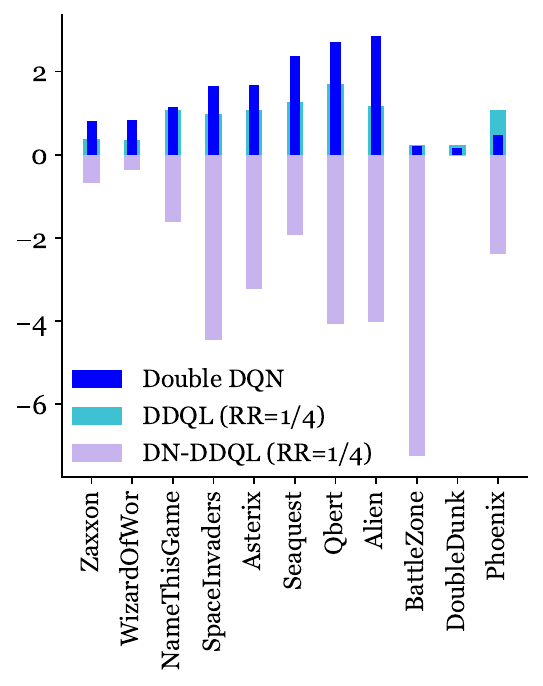}
    \caption{\textbf{Replay Ratio Experiments.} (\emph{left}) IQM of the HNS across five seeds on the Ablation-11 games. Shaded region: 95\% stratified bootstrap confidence interval. Increasing the replay ratio harms DDQL's performance. (\emph{right}) Compares the final overestimation of DDQL with a replay ratio of $\nicefrac{1}{4}$ to Double DQN, averaged over five seeds, on the Ablation-11 games.}
    \label{fig:rr_experiments}
    \vspace{-10pt}
\end{wrapfigure}
DDQL uses a replay ratio of $\nicefrac{1}{8}$ instead of the $\nicefrac{1}{4}$ used by DQN and Double DQN on Atari 2600 games.
To study the importance of this difference, we evaluate DDQL and DN-DDQL with double the replay ratio at $\nicefrac{1}{4}$ on the Ablation-11 environments.
This change ensures that each Q-function receives as many updates as the single Q-function in Double DQN.
All other algorithm details, hyperparameters, and experiment configurations are unchanged.

\ShortFigref{fig:rr_experiments} (\emph{left}) shows the IQM of the human-normalized score across the Ablation-11 games.
Doubling the replay ratio reduces performance in both DDQL variants.
The lower replay ratio performs better.

To understand why this occurs, consider that each parameter update can change the behavior policy, usually substantially~\citep{replay_fundamentals,schaul2022phenomenon}.
A lower replay ratio implies less frequent updates, ensuring that the buffer gets more transitions per policy.
Thus, transitions are re-sampled fewer times, reducing potential overfitting.
Moreover, the buffer has data from fewer and more recent policies in terms of parameter updates, slowing the change of its data distribution.
In short, lower replay ratios create a more stationary learning problem.
DDQL benefits from this stationarity, and is able to outperform Double DQN with half as many parameter updates.

\ShortFigref{fig:rr_experiments} (\emph{right}) contrasts Double DQN's overestimation to that of DDQL with double the replay ratio.
As DDQL with twice as many updates still reduces overestimation over Double DQN, the argument that DDQL's reduced overestimation is solely a consequence of having fewer updates---for example, by slowing the compounding of overestimation errors---is eliminated.
Additionally, we ran three seeds of Double DQN on all 57 environments with a replay ratio of $\nicefrac{1}{8}$, shown in \ShortFigref{fig:half_rr_perf} in Appendix~\ref{appendix:additional_results}.
In short, we find that the lower replay ratio does not benefit Double DQN's performance, indicating that DDQL's improved performance is not solely attributable to having a lower replay ratio.

\subsection{On initializations and targets}
Appendix~\ref{appendix:additional_results} contains additional results and discussion on other design choices that contribute to performance.
Appendix~\ref{init_ablation} investigates the impact of initializing the two Q-functions identically.
We find that in DN-DDQL with shorter TNU intervals, identical initializations can be critical for good performance.
Appendix~\ref{dqn_vs_ddqn} investigates the benefits of using a target network for action-selection in the bootstrap target rather than a Q-network, which is continually changing.
We find that using stationary targets, that is, using two target networks as opposed to a target network and a Q-network, improves performance in some environments.

\section{Discussion and conclusion}
\label{discussion}

In this paper, we introduced Deep Double Q-learning (DDQL), an adaptation of Double Q-learning featuring double estimation with reciprocal bootstrapping and dataset partitioning.
DDQL reduces overestimation and improves performance over Double DQN, without additional hyperparameters.
With DDQL, we show that Double Q-learning can be adapted into a high-performing algorithm in the value-based deep RL setting in a manner that closely resembles the original algorithm.
We studied multiple aspects of DDQL, including three architectures, two kinds of dataset partitioning, and the importance of the replay ratio.

This paper has two main limitations.
The first is that DDQL is restricted to discrete-action, value-based RL.
Though in principle DDQL can be extended to continuous control actor-critic methods such as Soft Actor-Critic~\citep{soft_actor_critic} or TD3~\citep{td3}, the integration is not seamless.
In Appendix~\ref{appendix:related}, we discuss how the Clipped Double Q-learning used in actor-critic methods differs substantially from Double Q-learning and DDQL.
The second limitation is that it is yet unclear how DDQL integrates with other deep RL advances such as those in Rainbow~\citep{rainbow}.

Though a general study on the relationship between overestimation and performance is outside the scope of this paper, our results shed some light on this topic.
Broadly, our results are consistent with other works that have shown that the impact of overestimation can be beneficial or harmful~\citep{maxmin,waltz2024addressing}, depending on the situation.
We observed that the overestimation reduction accompanying Double Q-learning's defining features does not necessarily translate to performance improvements.
Deep learning has several interacting parts.
Thus, mechanisms to decouple the Q-functions must be balanced against their impact on the training process.
Overestimation is impacted by the architecture, training data, and the replay ratio, yet these all impact learning dynamics differently.

In fact, this paper can largely be viewed as outlining how to stabilize training in spite of these decoupling strategies.
Techniques to improve optimization and representation learning, such as identical initialization, simultaneous updates, and shared hidden layers, benefit learning.
Moreover, stabilizing mechanisms to induce stationarity, including longer target network update intervals, lower replay ratios, and fixed targets, all contribute to stable learning and help realize better performance.

\begin{ack}
We thank the anonymous reviewers for their feedback, which improved the paper.
We thank Andrew Patterson for useful discussions, Brett Daley for reviewing parts of the code, and
Abhishek Naik for making helpful suggestions.
We thank Anna Hakhverdyan and Alex Lewandowski for providing feedback on drafts of the paper.
The authors especially thank Khurram Javed, Emma Jordan, Aditya Ganeshan, and Jens Tuyls for extensive discussions and feedback on the paper.

This research was supported in part by the Natural Sciences and Engineering Research
Council of Canada (NSERC) and the Canada CIFAR AI Chair Program.
Prabhat Nagarajan was supported by the Alberta Innovates Graduate Student Scholarship during the bulk of this research. Computational resources were provided in part by the Digital Research Alliance of Canada.
\end{ack}

\bibliography{main.bib}
\bibliographystyle{abbrvnat}


\clearpage
\appendix

\section{Related work} \label{appendix:related}

As we study adaptations of Double Q-learning in this paper, measuring DDQL's impact on overestimation has naturally been a central point.
Though DDQL indeed reduces overestimation, our objective was more to adapt Double Q-learning than it was to find the best algorithm that reduces overestimation.
Nonetheless, we should acknowledge the broader literature in value-based deep RL that aims to address or manage overestimation.
Averaged-DQN~\citep{averaged_dqn} reduces overestimation by performing Q-learning updates with the Q-learning target averaged over multiple historical target networks. 
Self-correcting DQN~\citep{scdqn} aims to balance overestimation and underestimation, but does not implement double estimation.
Like Double DQN, it too leverages the target network instead of a second independent Q-function.

There are value-based ensemble methods that aim to manage overestimation.
Maxmin DQN \citep{maxmin} aims to allow and control for different amounts of overestimation or underestimation.
Ensemble Bootstrapped Q-Learning (EBQL)~\citep{peer2021ensemble} and cross Q-learning~\citep{crossdqn} both generalize Double Q-learning to ensembles.
Both papers only investigate multi-head architectures, and only investigate ensembles of at least five members.
\citet{crossdqn}'s study is also restricted to the small environments of CartPole and Lunar Lander~\citep{towers2024gymnasium}.
While EBQL is evaluated on Atari 2600 games, its evaluation is restricted to 11 random environments as opposed to the 57 games, and overestimation is not measured in these environments. 
\citet{waltz2024addressing} also work in the ensemble setting, building on Bootstrapped DQN~\citep{bootstrapped_dqn}, with the aim of balancing overestimation and underestimation.

Though our focus in this paper is on adapting Double Q-learning for value-based deep RL, it is important to note an adaptation commonly used in deep off-policy actor-critic settings known as Clipped Double Q-learning (CDQ)~\citep{td3}.
CDQ adapts some ideas of Double Q-learning and has been popularized by algorithms like TD3 \citep{td3} and Soft Actor-Critic~\citep{soft_actor_critic}.
CDQ is distinct from Double Q-learning and DDQL in several ways.
First, CDQ computes a \textit{single} target Q-value which is used to update both Q-functions, rather than having each Q-function update towards different targets.
Second, this single target is computed by bootstrapping the minimum estimated Q-value across two Q-functions.
Third, each Q-function evaluates the \textit{same} target action from a single policy, rather than having two different policies select the actions for each respective Q-function.
Fourth, these Q-functions are trained on the same minibatches, i.e., they are trained on the same data unlike Double Q-learning or DDQL.
In the tabular setting, this is akin to updating both Q-functions with the same experience and same target value, which fundamentally departs from Double Q-learning.
There are also other approaches for reducing overestimation that have been applied to the off-policy actor-critic setting.
Decorrelated Double Q-learning~\citep{decorrelatedddqn} introduces a regularization term to de-correlate the two Q-functions.
Randomized Ensembled Double Q-Learning~\citep{redq} maintains an ensemble of Q-functions and minimizes across them to compute targets.
These works all aim to mitigate overestimation, but they neither study variants that resemble the original Double Q-learning algorithm, nor do they study the pure value-based setting.

\section{DDQL implementation} \label{appendix:efficient_implementation}

When computing gradients for DDQL on two separate minibatches, one option is to use two separate forward and backward passes to compute the loss and gradients for each Q-function.
Another option is to compute these simultaneously by masking minibatches.
Rather than sampling two separate minibatches of size 32, we can first sample a larger minibatch of size 64.
On this larger batch, we can compute the predictions and targets for each Q-function, giving us 128 total prediction-target pairs.
Only 32 prediction-target pairs are relevant for $Q_1$ and a different 32 pairs are relevant for $Q_2$.
We can simply mask the irrelevant pairs before computing the loss.
Depending on how element-wise losses are aggregated, some scaling may be needed.
For example, if using the mean squared error loss, the batch size is doubled with our masking, so the computed loss should be adjusted.

\section{Experimental details} \label{appendix:experiments}

In this appendix, we provide details regarding environments, training, evaluation, and overestimation measurement.

\subsection{Environments, training, and evaluation}
\paragraph{Environments}
In this paper, we use the environment settings proposed by \citet{revisitingale}.
All agents use the full action set of 18 actions for all games.
We use ``sticky'' actions, where the simulator repeats the action executed at the previous frame with probability 0.25, regardless of the agent's selected action.
We use the actual termination of the game as the termination signal to the learning agent.
These details are outlined in Table~\ref{env_details}.

The Ablation-11 set of environments contains the Atari-5 subset, a subset of Atari-57 games that are said to be predictive of Atari-57 performance in terms of median human-normalized score:  \textsc{BattleZone}, \textsc{DoubleDunk}, \textsc{NameThisGame}, \textsc{Phoenix}, and \textsc{Qbert}~\citep{atari5}.
It also contains six different Atari 2600 games that are used in the literature to study overestimation in DQN \citep{ddqn,averaged_dqn}: \textsc{Alien}, \textsc{Asterix}, \textsc{Seaquest}, \textsc{SpaceInvaders}, \textsc{WizardofWor}, and \textsc{Zaxxon}.

\paragraph{Training}
The agent is trained for 50M timesteps.
When training, the rewards are clipped to be between $-1$ and $1$.
The agents deploy an $\epsilon$-greedy policy during training, beginning at $\epsilon=1.0$ and linearly annealing to $0.01$ over 1M timesteps.
The agent's replay buffer has a capacity of 1M, and network updates are only performed after the agent has completed 50k timesteps.
The network architecture used is identical to the architecture used by \citet{ddqn}, i.e., the DQN architecture~\citep{dqn} with a single shared bias for all actions in the final layer as opposed to per-action biases.
Agents use the Adam optimizer with the MSE loss using a step size of ${6.25\mathrm{e}{-5}}$, $\epsilon={1.5\mathrm{e}{-4}}$, $\beta_1 = 0.9$, and $\beta_2 = 0.999$
~\citep{rainbow,optimisticperspective,revisiting_rainbow,stopregressing}.
Target networks are copied from the main Q-networks every 7,500 gradient updates, which corresponds to 30k timesteps for Double DQN and 60k timesteps for DDQL.
The preprocessing scheme follows that of~\citet{dqn}.
Table \ref{hypers} outlines our hyperparameters and training details.

\paragraph{Evaluation}
Agents are evaluated for 125k timesteps after every 250k timesteps of training.
Evaluation episodes are truncated at 30 minutes, or 27k timesteps.
Agents deploy an $\epsilon$-greedy policy during evaluation with $\epsilon=0.001$.
For DDQL agents, the $\epsilon$-greedy policy is with respect to the average action-values of the two Q-functions.
Table \ref{evals} outlines these details.

\paragraph{Human-normalized scores}
The human-normalized score~\citep{dqn} of an agent can be computed as
\begin{equation} \label{hns_eq}
    \text{score}_{\text{hns}} = \frac{\text{score}_{\text{agent}} - \text{score}_{\text{random}}}{\text{score}_{\text{human}} - \text{score}_{\text{random}}}
\end{equation}
Our human scores and random scores used to compute \Eqref{hns_eq} are taken from DQN Zoo~\citep{dqnzoo2020github}. 
Suppose we have $N$ environments and $M$ seeds per environment.
Let $X_{j,k}$ refer to the human-normalized score of the agent on the $j$th environment and $k$th seed.
To compute the interquartile mean of the human-normalized score, we follow~\citet{statistical_precipice}:
\begin{equation}
    \text{IQM HNS} = \text{IQM}\left(\{ X_{j,k} : j = 1,\dots,N \quad \text{and}\quad k = 1,\dots,M \} \right).
\end{equation}

\begin{table*}[] 
\caption{Environment details, following~\citet{revisitingale}.}
\centering
\begin{small}
\resizebox{\textwidth}{!}{%
\begin{tabular}{@{}l|c|p{0.65\linewidth}@{}}
\toprule
\textbf{Detail} & \textbf{Setting} & \textbf{Description} \\ 
\midrule
sticky action probability & 0.25 & Probability by which the simulator ignores the agent's selected action and repeats the action executed in the previous frame. \vspace{1mm} \\
termination criterion & end-of-game & How episode termination is signaled during training (either end-of-game or loss-of-life). \vspace{1mm} \\
action space & full & The full action space has 18 actions. The minimal action space uses the game-specific minimum required actions.\\
\bottomrule
\end{tabular}%
}
\end{small}
\label{env_details}
\end{table*}
\begin{table*}[] 
\centering
\caption{Hyperparameters and training details.}
\begin{small}
\resizebox{\columnwidth}{!}{%
\begin{tabular}{@{}p{0.3\linewidth}|p{0.1\linewidth}|p{0.5\linewidth}@{}}\toprule
  \textbf{Hyperparameter} & \textbf{Value} & \textbf{Description} \\ \midrule
minibatch size & 32 & Number of transitions used per update per Q-function. \vspace{1mm} \\
replay memory size & 1,000,000 & Number of transitions stored in the replay buffer. \vspace{0.5mm} \\
agent history length & 4 & Number of previous frames stacked in state representation. \vspace{1mm} \\
target network~{update frequency} & 7,500 & Frequency (in terms of parameter updates) of target network updates \vspace{1mm} \\
discount factor & 0.99 &  Value of $\gamma$ used in the target computation. \vspace{1mm} \\
action repeat & 4 &  The number of simulator frames for which an action is repeated in a single timestep. \vspace{1mm} \\
update frequency & 8 &  Frequency (in timesteps) of parameter updates. \vspace{1mm}\\
replay start size & 50,000 &  Minimum number of transitions in the replay buffer required before parameter updates begin. \vspace{1mm} \\
initial exploration & 1.0 &  Initial value of $\epsilon$ used for $\epsilon$-greedy exploration. \vspace{1mm} \\
final exploration & 0.01 &  Final value of $\epsilon$ used for $\epsilon$-greedy exploration. \vspace{1mm} \\
final exploration timestep & 1,000,000 &  The number of timesteps over which $\epsilon$ is linearly annealed to its final $\epsilon$. \vspace{1mm} \\
maximum episode length & 27,000 &  Timesteps after which an episode is truncated and the environment is reset. \vspace{1mm} \\
step size & 6.25e-5 &  Adam optimizer step size. \vspace{1mm} \\
Adam $\epsilon$ & 1.5e-4 &  The $\epsilon$ used by Adam. \vspace{1mm} \\
Adam $\beta_1$ & 0.9 &  $\beta_1$ hyperparameter value in Adam. \vspace{1mm} \\
Adam $\beta_2$ & 0.999 &  $\beta_2$ hyperparameter value in Adam. \\
\bottomrule
\end{tabular}
}
\end{small}
\label{hypers}
\end{table*}

\begin{table*}[] 
\centering
\caption{Evaluation details.}
\begin{small}
\resizebox{\columnwidth}{!}{%
\begin{tabular}{@{}p{0.3\linewidth}|p{0.1\linewidth}|p{0.5\linewidth}@{}}\toprule
  \textbf{Detail} & \textbf{Value} & \textbf{Description} \\ \midrule
evaluation $\epsilon$ & 0.001 & The $\epsilon$ used for the $\epsilon$-greedy policy used during evaluation. \\
maximum episode length & 27,000 & Number of timesteps in evaluation episodes. Corresponds to 30 minutes of gameplay.\footnotemark\\
evaluation phase length & 125,000 & Length of periodic evaluation phases in terms of number of timesteps.\\
evaluation frequency & 250,000 & Frequency (in terms of training timesteps) of evaluation phase. \\
\bottomrule
\end{tabular}
}
\end{small}
\label{evals}
\end{table*}

\footnotetext{{(30 mins = 30 minutes * 60 seconds/minute * 60 frames/second / 4 frames per timestep = 27,000 timesteps)}}

\subsection{Measuring overestimation} \label{appendix:overestimation}
As discussed in Section~\ref{experimental_setup}, we measure overestimation by comparing achieved returns of a greedy policy to the Q-value predictions of our networks.
Recall the greedy policy with respect to $Q$, $\mathfrak{g}_{Q}$.
Double DQN is trained so that 
\[
Q(s,a; \bm{\theta}) \approx \mathbb{E}\left[\sum_{t=0}^{\infty} \gamma^{t} R_{t+1} | S_0=s, A_0=a, A_{t \ge 1} \sim \mathfrak{g}_{Q(\cdot|\bm{\theta})} \right].
\]
During the periodic evaluations, we compute overestimation using the completed episodes.
The agents use $\epsilon$-greedy exploration with $\epsilon = 0.001$, which is near-greedy, in order to obtain a near-unbiased sample of a discounted return of the greedy policy.
This evaluation phase produces $k$ completed episodes (discarding incomplete episodes) $\tau_{1},...,\tau_{k}$ of lengths $T_1, ..., T_k$ respectively. 
For example, $\tau_1 = \{s_0, a_0, r_1, s_1, a_1,\dots, a_{T_1 - 1}, r_{T_1}, s_{T_1}\}$.
We then compute the average predicted state-action values across these state-action pairs across all the completed trajectories:
\begin{equation*}
    \hat{Q} = \frac{1}{\sum_{i=1}^k T_i}\sum_{i=1}^k \sum_{(s,a) \in \tau_i} Q(s, a; \bm{\theta}).
\end{equation*}

For an episode $\tau_i$, the discounted return for a state-action pair $(s_t,a_t)$ in the trajectory is ${\text{Return}}(\tau_i, s_t, a_t) = \sum_{r_j \in \tau_i; j \ge t+1}^{} \gamma^{j-t-1} r_j$.
Then the average return across all state-action pairs in the set of completed episodes is: 
\[
\hat{G} = \frac{1}{\sum_{i=1}^k T_i} \sum_{i=1}^k \sum_{(s,a) \in \tau_i} \text{Return}(\tau_i, s, a).
\]
$\hat{G}$ is a near-unbiased estimate of the expected discounted return under a greedy policy with respect to the Q-values.
We then compute the overestimation: $\hat{Q} - \hat{G}$.
For DDQL, we use the near-greedy policy with respect to the average of the two Q-functions, and $\hat{Q}$ is formed from the average Q-value:
\begin{equation*}
    \hat{Q} = \frac{1}{2\sum_{i=1}^k T_i}\sum_{i=1}^k \sum_{(s,a) \in \tau_i} \big(Q(s, a; \bm{\theta}_1) + Q(s, a; \bm{\theta}_2)\big).
\end{equation*}

Agents trained on Atari environments typically employ reward clipping, where rewards are clipped to the range $[-1,1]$.
Since agents are trained to predict return estimates under this reward, we also ensure that this clipping is applied when computing returns.
This clipping, combined with discounting, often causes discounted returns to be much smaller than the raw, unclipped scores.

Another point to note is that evaluation episodes are truncated after 30 minutes of play (or 27k timesteps) as is typical in ALE evaluations.
In these instances, to compute the return, we bootstrap the final Q-value of the non-terminal state at which the episode is truncated.
This can indeed have an impact on results, but this occurs infrequently.
Moreover, in the instances in which this does occur, this bootstrapped value is discounted to less than 0.005 for over 98\% of state-action pairs in that truncated episode.

We counted the incidences of truncations across all algorithms, seeds, and environments.
This totals well over 1k training runs, each with 200 evaluation phases of multiple episodes.
We found that in 37 out of the 57 environments, truncation was never once exercised during evaluation across all algorithms and seeds.
Upon inspecting the 20 environments for which truncation was exercised at least once, we found that in four of these environments truncation was exercised only in 1-2 episodes across all evaluation phases, seeds, and algorithms.
Moreover, in some environments, like Montezuma's Revenge, where our agents are unable to achieve rewards but avoid ending the game for 30 minutes of gameplay, truncations occur but have no impact on the overestimation.

The use of a near-greedy policy instead of a greedy policy introduces some bias.
However, policies with moderate stochasticity, even when sticky actions are used, generally improve performance.
To understand why, let us consider an example.
After a life loss in the game of Breakout, the agent must take a specific action (``Fire'' action) to launch the ball and continue to receive rewards.
If the agent selects any other action deterministically, even sticky action repeats cannot cause the agent to fire the ball.

\section{Target bootstrap decoupling and double estimation} \label{appendix:adaptation}

In this Appendix, we provide a broader discussion, largely appealing to prior literature, on the various ways in which target bootstrap decoupling and double estimation with reciprocal bootstrapping can be integrated into learning algorithms.

\subsection{Target bootstrap decoupling}
\begin{wraptable}[6]{R}{0.4\textwidth}
\caption{Single network target bootstrap decoupling.}
\centering
\begin{small}
\begin{tabular}{@{}l|c|c@{}}
\textbf{Algorithm} & $\bm{Q_{\textbf{eval}}}$ & $\bm{Q}_{\textbf{sel}}$ \\ \midrule
Double DQN & $\vtheta^{-}$  & $\vtheta^{\text{ }}$ \\
Inverse Double DQN & $\vtheta^{\text{ }}$ &  $\vtheta^{-}$ \\
\end{tabular}
\end{small}
\label{single_q_target_bootstrap_decoupling}
\end{wraptable}

In algorithms where only a single Q-network and target network are available, such as DQN and Double DQN, we can at most achieve target bootstrap decoupling.
To implement target bootstrap decoupling, one network must be used as $Q_{\text{sel}}$ and the other as $Q_{\text{eval}}$.
This admits two possible choices, which are Double DQN and inverse Double DQN~\citep{deadlytriad}.\footnote{We are introducing this terminology here. This algorithm is called \textit{inverse Double Q-learning} by~\citet{deadlytriad}.}

Double DQN sets $\vtheta_{\text{eval}} = \vtheta^-$ and $\vtheta_{\text{sel}} = \vtheta$.
Inverse Double DQN inverts this by setting $\vtheta_{\text{eval}} = \vtheta$ and $\vtheta_{\text{sel}} = \vtheta^-$.
Though inverse Double DQN implements target bootstrap decoupling, it has been investigated by~\citet{deadlytriad} and has been shown to be unstable relative to both DQN and Double DQN.
This makes Double DQN the clear choice as an algorithm that both performs well and reduces overestimation.
These two variants are highlighted in Table \ref{single_q_target_bootstrap_decoupling}.

\subsection{Double estimation} \label{appendix_double_estimation}
DDQL algorithms implement double estimation with reciprocal bootstrapping.
In particular, we now train two separate Q-functions with parameters $\vtheta_1$ and $\vtheta_2$, which may have corresponding target networks $\vtheta_1^{-}$ and $\vtheta_2^-$. 
In this setting too, we have multiple options for computing bootstrap targets.
If $\vtheta_1$ or $\vtheta_1^-$ is selecting the action in the target, then either $\vtheta_2$ or $\vtheta_2^-$ is used to evaluate the selected action, and vice versa.
Suppose $\vtheta_1$ is being updated.
We are again presented with four main algorithm variants, which are summarized in Table~\ref{ddql_variants}.

\begin{enumerate}
        \item $\vtheta_{\text{eval}} = \vtheta_2^-$ and $\vtheta_{\text{sel}} = \vtheta_1^-$. We denote this $\text{DDQL}_{\text{DQN}}$ because it uses target networks for both action-selection and action-evaluation as is done in DQN.
        This strategy ensures a stationary target for the interval between target network updates.
        \item $\vtheta_{\text{eval}} = \vtheta_2^-$ and $\vtheta_{\text{sel}} = \vtheta_1$. We denote this $\text{DDQL}_{\text{Double DQN}}$ because it selects an action for computing the target using a Q-network and evaluates it with a target network, as is done in Double DQN.
        \item $\vtheta_{\text{eval}} = \vtheta_2$ and $\vtheta_{\text{sel}} = \vtheta_1^-$. We denote this $\text{DDQL}_{\text{Inverse}}$. In some sense, this is the analog of inverse Double DQN in the setting where we train two networks.
        Action-selection is performed with a target network, and action-evaluation is performed with a Q-network.
        \item $\vtheta_{\text{eval}} = \vtheta_2$ and $\vtheta_{\text{sel}} = \vtheta_1$. We denote this $\text{DDQL}_{\text{No target}}$.
        In this variant, only the Q-networks are used to both select and evaluate actions. Target networks are not used.
\end{enumerate}

\begin{wraptable}{r}{0.4\textwidth}
\caption{Options for target bootstrap decoupling with reciprocal bootstrapping in DDQL.}
\centering
\begin{small}
\begin{tabular}{@{}l|c|c@{}}
\textbf{Algorithm} & $\vtheta_{\textbf{eval}}$ & $\vtheta_{\textbf{sel}}$ \\ \midrule
$\text{DDQL}_{\text{DQN}}$ & $\vtheta_2^{-}$ & $\vtheta_1^{-}$ \\
$\text{DDQL}_{\text{Double DQN}}$ & $\vtheta_2^{-}$ & $\vtheta_1^{\text{ }}$ \\
$\text{DDQL}_{\text{Inverse}}$ & $\vtheta_2^{\text{ }}$ & $\vtheta_1^{-}$ \\
$\text{DDQL}_{\text{No target}}$ & $\vtheta_2^{\text{ }}$ & $\vtheta_1^{\text{ }}$ \\
\end{tabular}
\end{small}
\label{ddql_variants}
\end{wraptable}

While all of these algorithms implement double estimation, our study is primarily on the first algorithm, $\text{DDQL}_{\text{DQN}}$.
$\text{DDQL}_{\text{DQN}}$ is compared to $\text{DDQL}_{\text{Double DQN}}$ in Appendix~\ref{dqn_vs_ddqn}.
When computing the bootstrap target, $\text{DDQL}_{\text{Inverse}}$ has stationary action-selection but nonstationary action-evaluation.
Moreover, since $\vtheta_2$ will also be updated with $\vtheta_1$ as its action-evaluator, the targets will be highly nonstationary with both Q-functions changing.
Furthermore, given that \citet{deadlytriad} found Inverse Double DQN to be unstable, we did not study $\text{DDQL}_{\text{Inverse}}$.

Additionally, we do not study $\text{DDQL}_{\text{No target}}$, which does not leverage any of the stationarity afforded by target networks.
It may seem that a separate Q-network can replace the role of a target network by providing a secondary estimate.
However, this secondary estimate is nonstationary, as this other Q-network is itself being updated frequently, leading to instability.
Our preliminary results found this to perform poorly, which is consistent with findings that show that entirely forgoing the use of target networks in DQN is generally worse~\citep{dqn}.

\section{Expanded discussion and results} \label{appendix:additional_results}

In this appendix, we discuss the use of simultaneous updates, identical initialization, and alternative strategies to compute bootstrap targets.
For the latter two, we also provide results.

\subsection{Simultaneous updates}
In the original Double Q-learning algorithm, at every timestep, the experience transition is used to update one of the two Q-functions.
The updates are not done simultaneously, and in expectation, a Q-function is updated once every two timesteps.
When we train Q-networks through experience replay, we do not typically use the online experiences to train.
Rather, we treat the replay buffer as a pseudo-offline dataset from which we sample minibatches for parameter updates.
This permits us to sample distinct minibatches and perform simultaneous updates. 
Moreover, simultaneous updates intuitively should benefit DDQL by avoiding the conflicting gradients that can arise in successive updates that optimize different losses.
Furthermore, optimizing both losses simultaneously can enrich the learned representation by offering a mutual auxiliary task effect.

\subsection{The importance of identical initialization} \label{init_ablation}

In tabular TD learning, we typically initialize the action-value lookup tables to zeros.
When training Q-networks, however, we randomly initialize the weights of our Q-networks.
When training two Q-networks (or heads) through DDQL, we are then presented with a choice of whether to initialize both Q-functions with the same random weights or with different random weights.
We opt for the former.

Our choice of identically initializing the two Q-functions is sensible for two reasons.
First, in tabular Double Q-learning, Q-functions are typically initialized identically, and the distinction between the two Q-functions initially emerges from using different experience transitions.
Second, in Double Q-learning, we eventually want the two Q-functions to converge. 
Since both Q-functions are trained through reciprocal bootstrapping, training two interdependent Q-functions that start from different initializations may make optimization more challenging.
Anecdotally, we did find that at lower target network update intervals, DDQL had fewer divergent or non-learning runs when identically initialized.

\begin{figure}[h]
    \centering
    \includegraphics[width=0.31\linewidth]{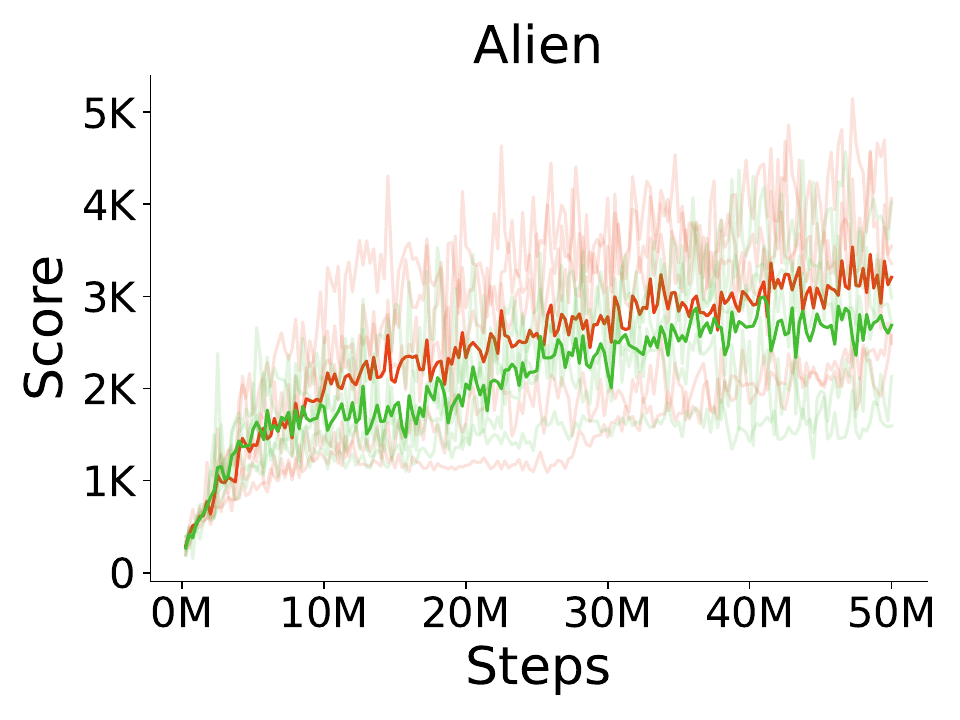}
    \includegraphics[width=0.31\linewidth]{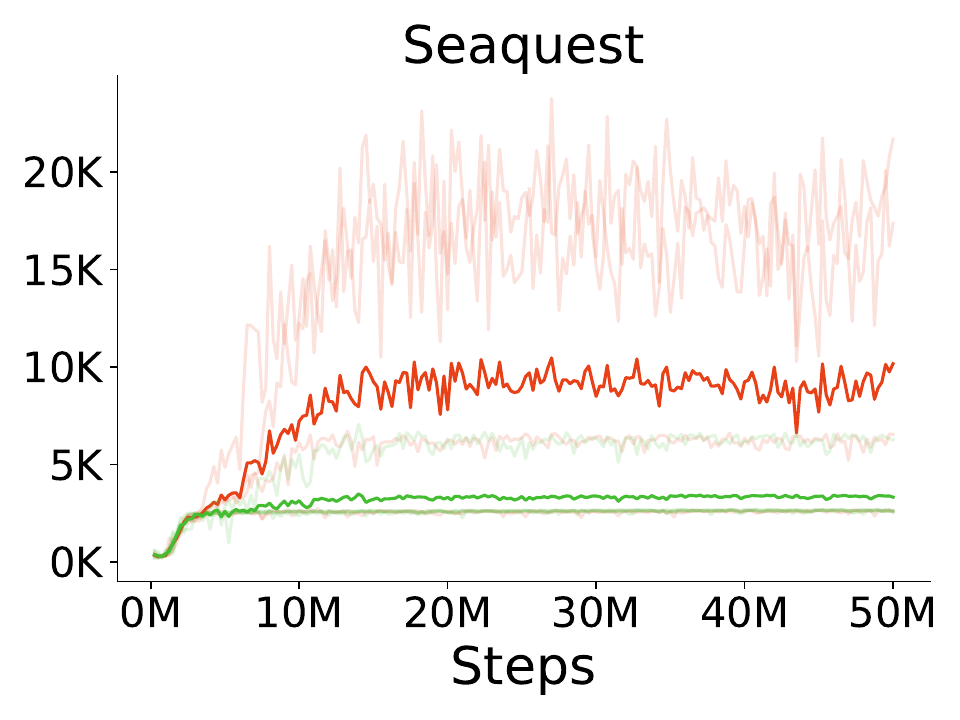}
    \includegraphics[width=0.31\linewidth]{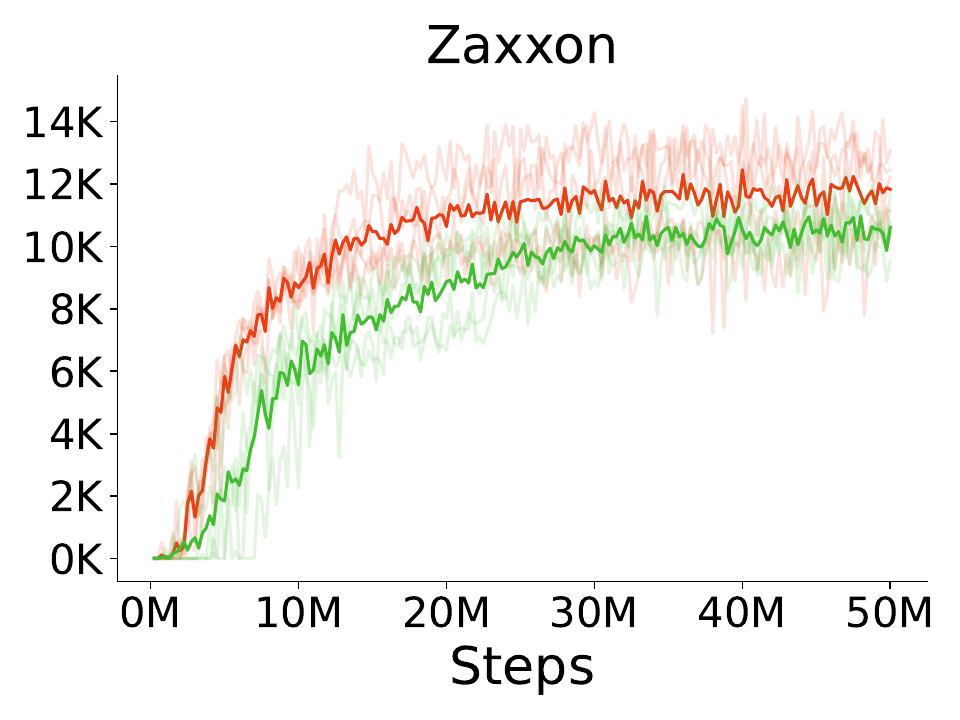}
    \includegraphics[width=0.6\linewidth]{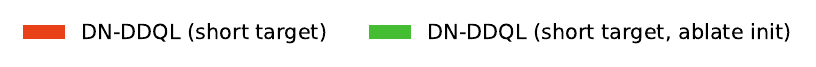}
    \caption{A comparison of DN-DDQL with a short target network update interval to DN-DDQL with a short target network update interval and non-identical initialization. Identical initialization can help performance.}
    \label{fig:init}
\end{figure}

Identical initialization is not absolutely essential, but it can have a strong effect on DN-DDQL in some environments, at smaller target network update intervals.
Figure~\ref{fig:init} depicts three environments in which a non-identical initialization performs worse than an identical initialization.
We run DN-DDQL with DQN's original shorter target network update interval of 2,500 gradient updates as opposed to 7,500 gradient updates.
We denote this variant ``short target''.
We compare it to DN-DDQL agents that share this shorter target network update interval and are initialized non-identically.
We denote these latter agents ``short target, ablate init''.
All other hyperparameters and training configurations are otherwise unchanged.
We can see a visible difference in performance between the variants, where the only difference between the curves emerges from whether or not the networks are initialized identically.
For many environments, the initialization is immaterial, especially at larger target network update intervals.
Nonetheless, these results are interesting even if only from the perspective of better understanding learning dynamics.

\subsection{Fixed DQN-style targets versus Double DQN-style targets} \label{dqn_vs_ddqn}

When training two Q-networks, each with its own target network, through reciprocal bootstrapping, we have several choices, as depicted in Table~\ref{ddql_variants} in Appendix~\ref{appendix_double_estimation}.
Appendix~\ref{appendix_double_estimation} discussed why $\text{DDQL}_{\text{No target}}$ and $\text{DDQL}_{\text{Inverse}}$ are discarded as choices.
Our primary candidate algorithms are $\text{DDQL}_{\text{DQN}}$ and $\text{DDQL}_{\text{Double DQN}}$.
We use the former in this paper.

Our choice can be motivated by the implicit goals underlying the development of DQN.
DQN essentially formulates the learning problem as a sequence of relatively stationary supervised regression tasks.
The replay buffer serves as a stationary dataset from which minibatches are sampled and used to minimize some regression loss over a fixed set of targets induced by the target network.
Every target network refresh begins a new regression task, as the targets change.
For the duration that the target network is held fixed, the targets are stationary and the only source of nonstationarity is the changing replay buffer, which is only a mild form of nonstationarity.

Double DQN forgoes stationary targets by using the actively changing Q-network to select actions for stationary evaluation.
The more the Q-function's greedy actions change, which they often do~\citep{schaul2022phenomenon}, the more nonstationary the targets become.
Double DQN finds a nice tradeoff where overestimation is reduced at the cost of some additional amount of nonstationarity in the selected actions in bootstrap targets, with stable action-evaluations.

When using two Q-networks, each with its own target network, we can avoid this tradeoff and can decouple both action-selection and action-evaluation in the target while retaining stationarity in both.
Moreover, our results throughout this paper are consistent with the hypothesis that slowing down nonstationarity is essential for reciprocal training of Q-functions.
As such, we use $\text{DDQL}_{\text{DQN}}$, where, when updating $\bm{\theta}_1$, we use  $\vtheta_{\text{sel}} = \bm{\theta}_1^-$ and $\vtheta_{\text{eval}} = \bm{\theta}_2^-$.
By doing so, we fix the target values for both Q-functions for the duration of the target network update interval.

\begin{figure}[h]
        \centering
    	\includegraphics[width=0.21\linewidth]{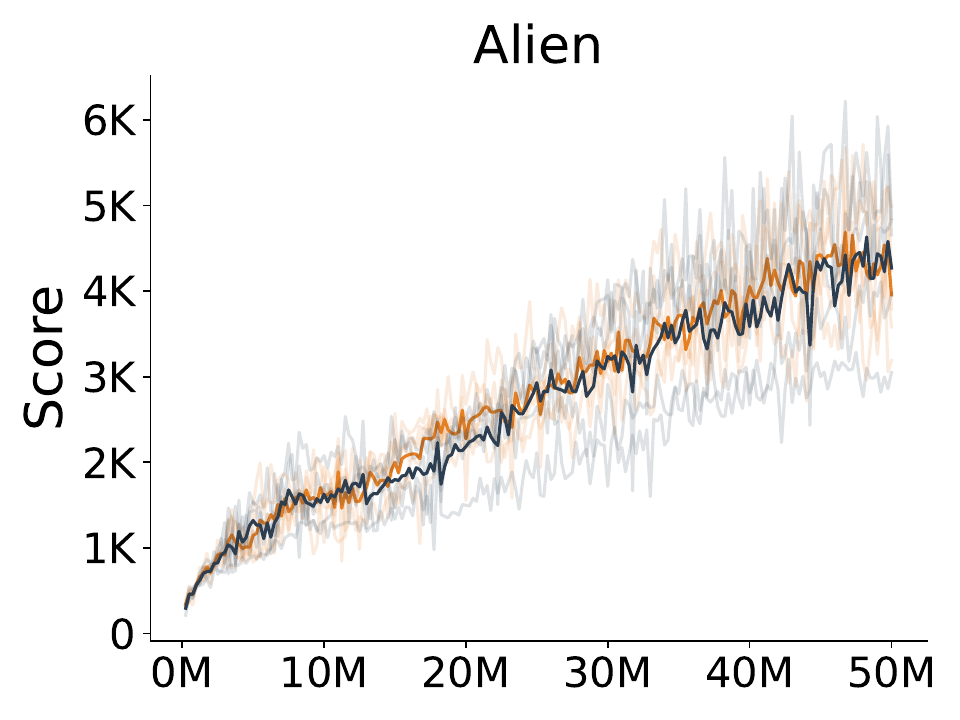} 
	\hspace{0.005\linewidth}
	\includegraphics[width=0.21\linewidth]{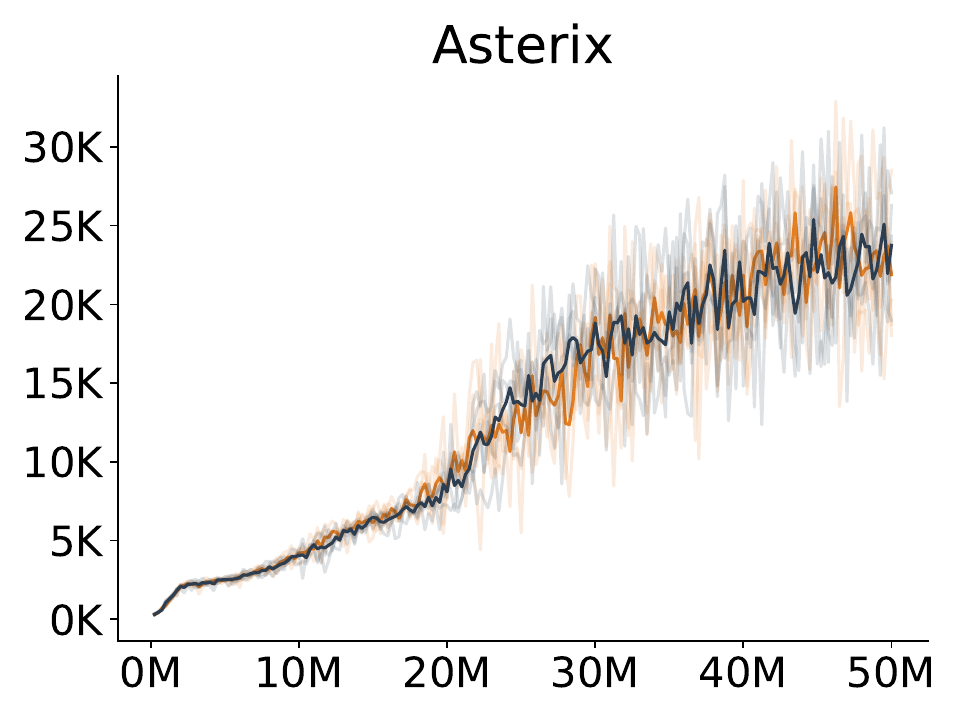} 
	\hspace{0.005\linewidth}
	\includegraphics[width=0.21\linewidth]{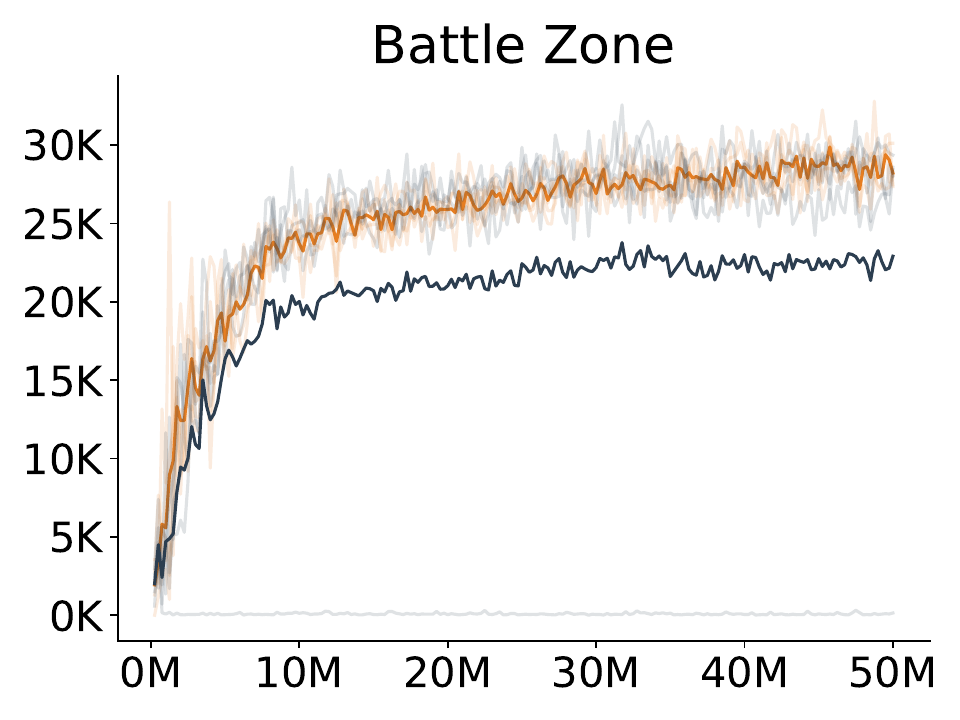} 
	\hspace{0.005\linewidth}
	\includegraphics[width=0.21\linewidth]{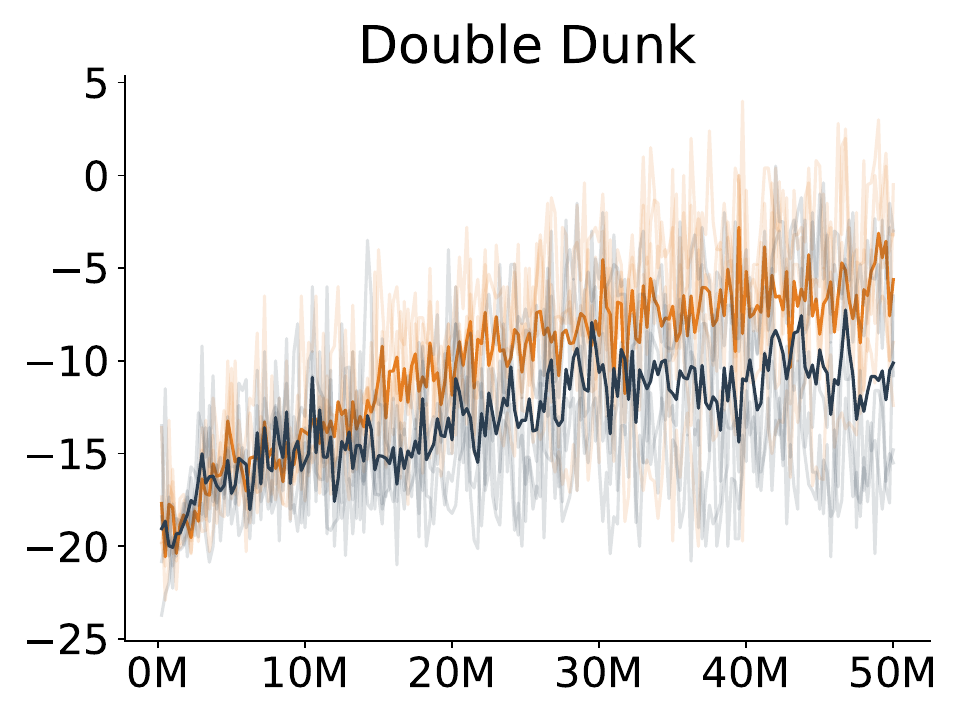} 
	\hspace{0.005\linewidth}
	\includegraphics[width=0.21\linewidth]{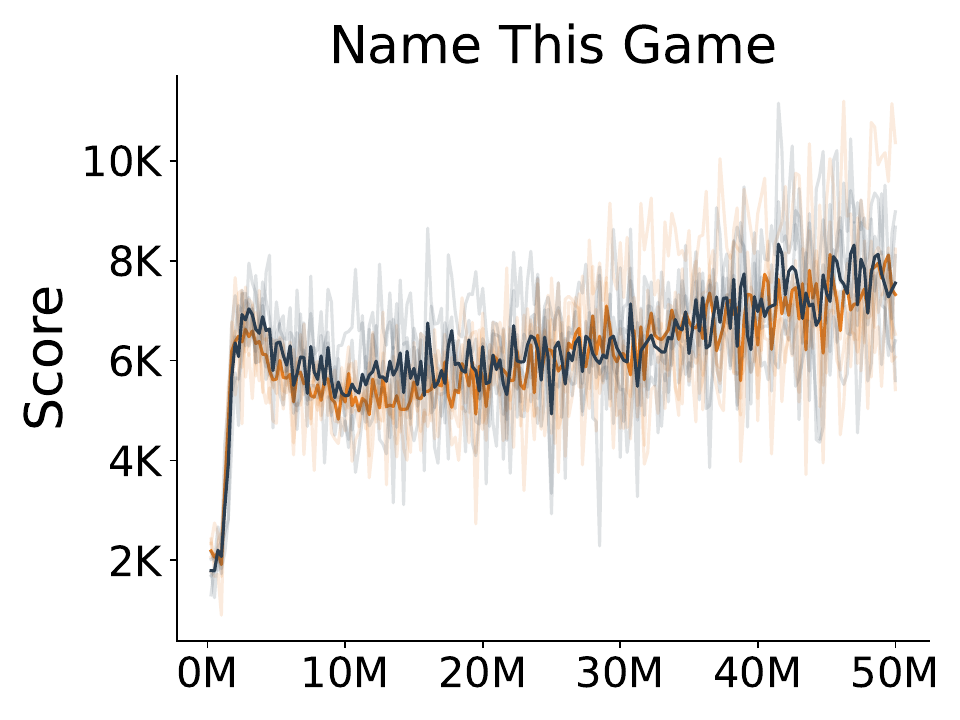} 
	\includegraphics[width=0.21\linewidth]{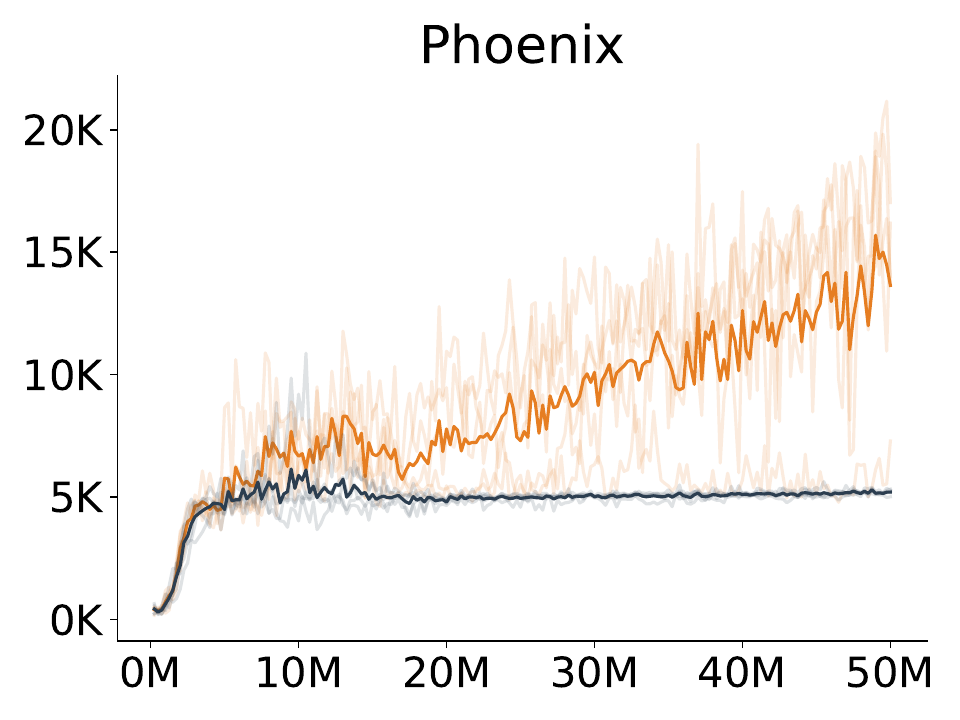} 
	\hspace{0.005\linewidth}
	\includegraphics[width=0.21\linewidth]{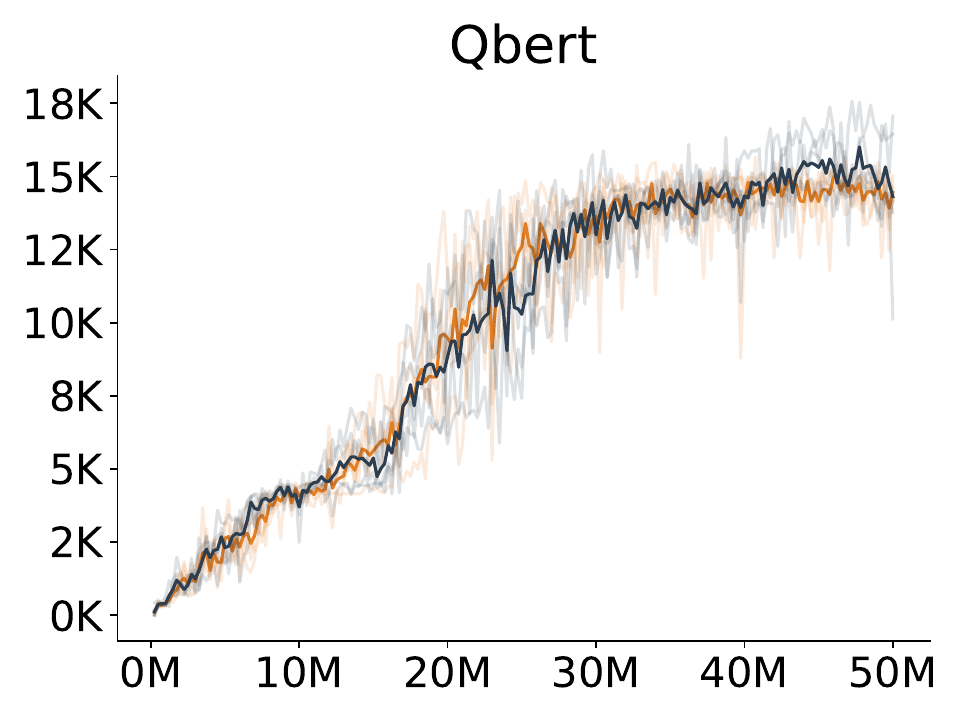} 
	\hspace{0.005\linewidth}
	\includegraphics[width=0.21\linewidth]{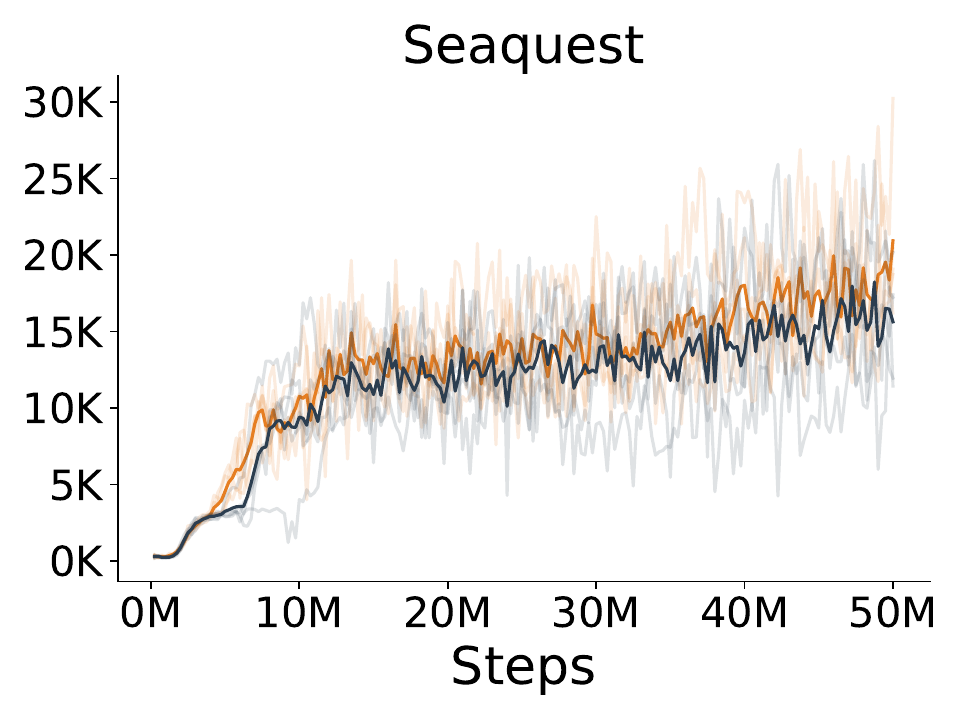} 
	\hspace{0.005\linewidth}
	\includegraphics[width=0.21\linewidth]{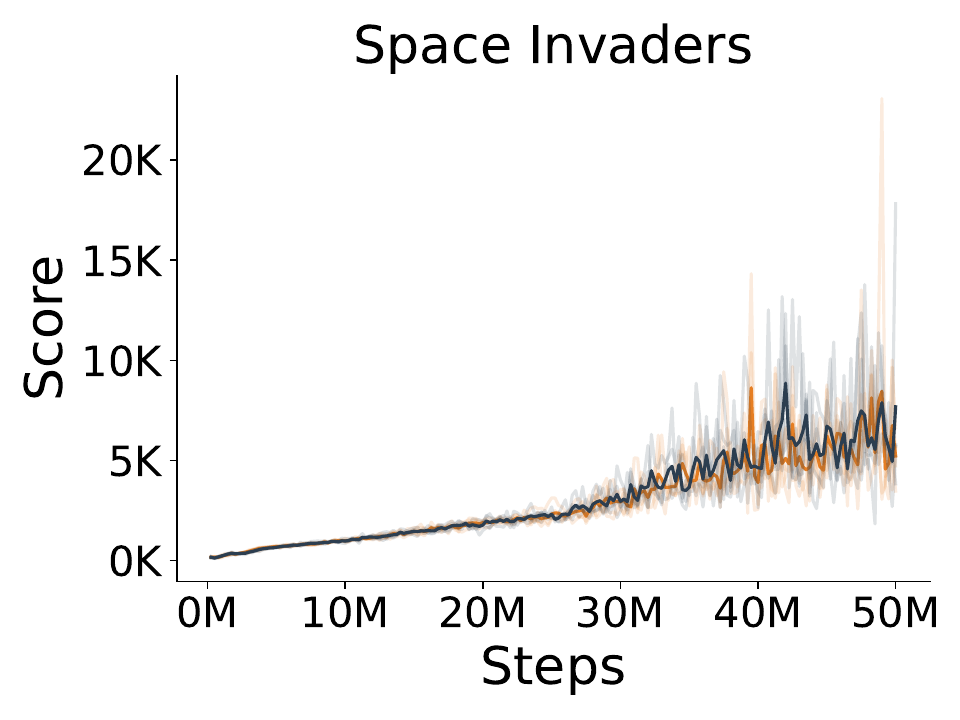} 
	\hspace{0.005\linewidth}
	\includegraphics[width=0.21\linewidth]{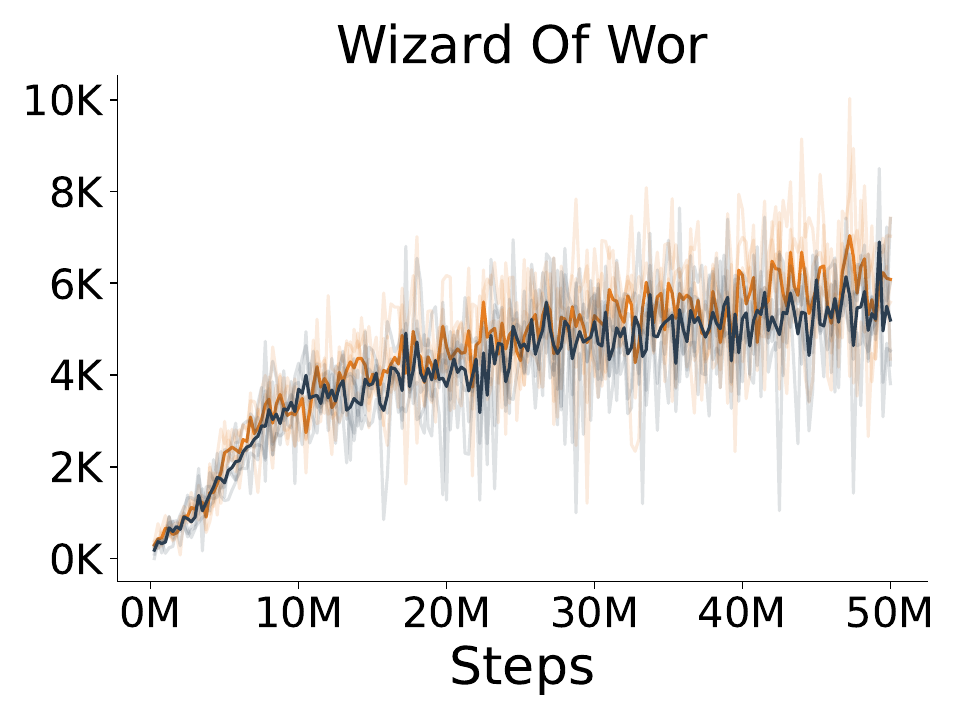} 
	\includegraphics[width=0.21\linewidth]{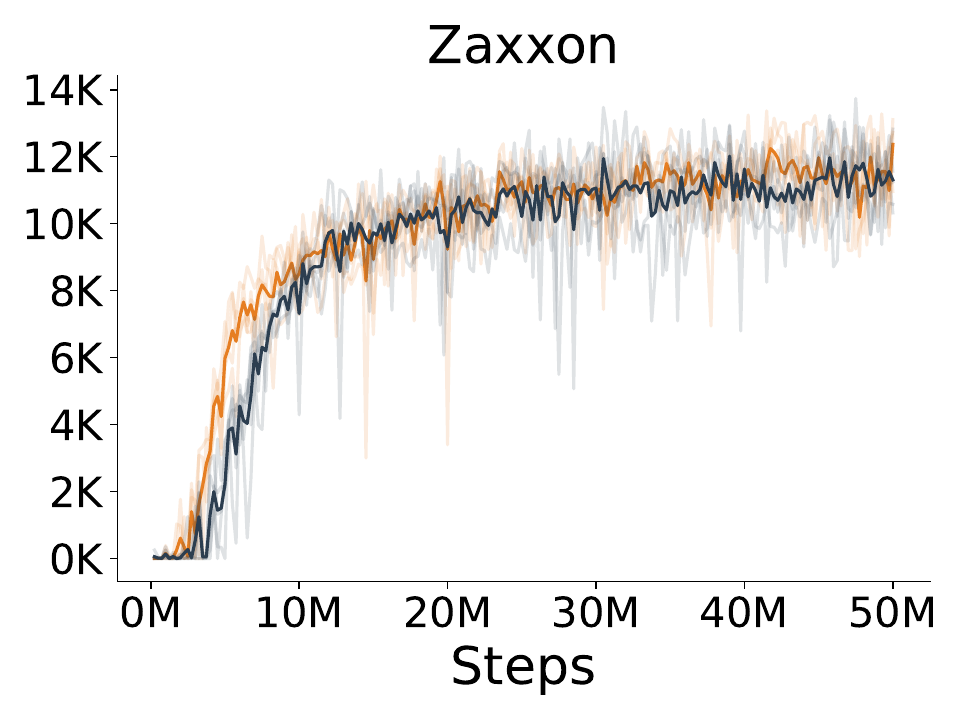} 
	\hspace{0.005\linewidth}
	\hspace{0.01\linewidth}\raisebox{2em}{\includegraphics[width=0.2\linewidth]{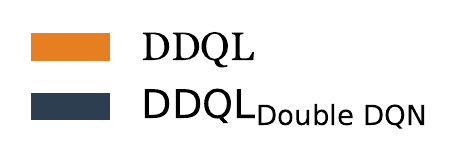}} 
	\caption{Performance of DDQL compared to $\text{DDQL}_{\text{Double DQN}}$. The algorithms perform at a similar level, with DDQL performing better in two environments.}
	\label{DHDQN_vs_DoubleDQN:Ablation11:Score}
\end{figure}

Figure~\ref{DHDQN_vs_DoubleDQN:Ablation11:Score} compares DDQL to $\text{DDQL}_{\text{Double DQN}}$.
The two algorithms generally perform on par with one another, though DDQL substantially outperforms $\text{DDQL}_{\text{Double DQN}}$ in \textsc{Phoenix} and appears to do reliably better in \textsc{DoubleDunk}.
DDQL may be more reliable in \textsc{BattleZone}, but it is difficult to draw any conclusions as performance is similar between the two algorithms if we exclude the single bad seed for $\text{DDQL}_{\text{Double DQN}}$.

\begin{figure}[h!]
        \centering
    	\includegraphics[width=0.21\linewidth]{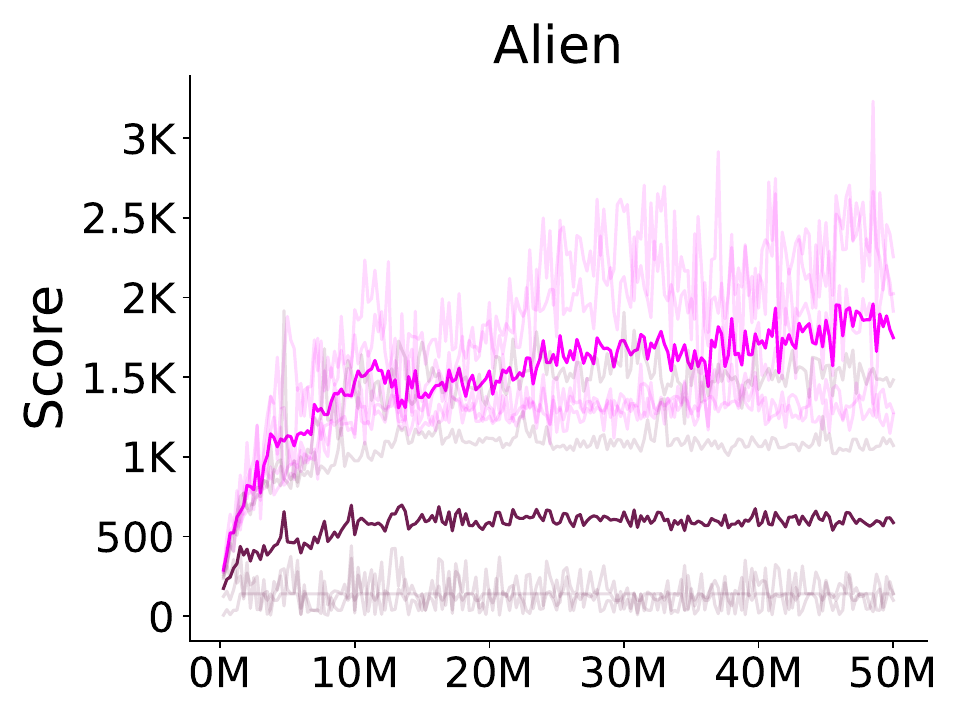} 
	\hspace{0.005\linewidth}
	\includegraphics[width=0.21\linewidth]{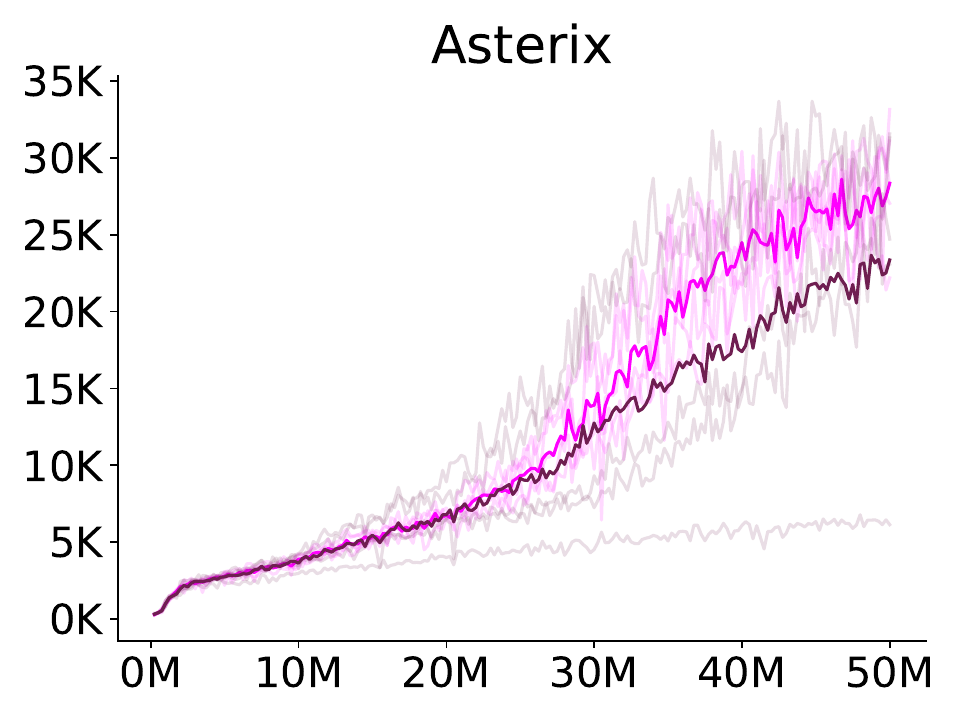} 
	\hspace{0.005\linewidth}
	\includegraphics[width=0.21\linewidth]{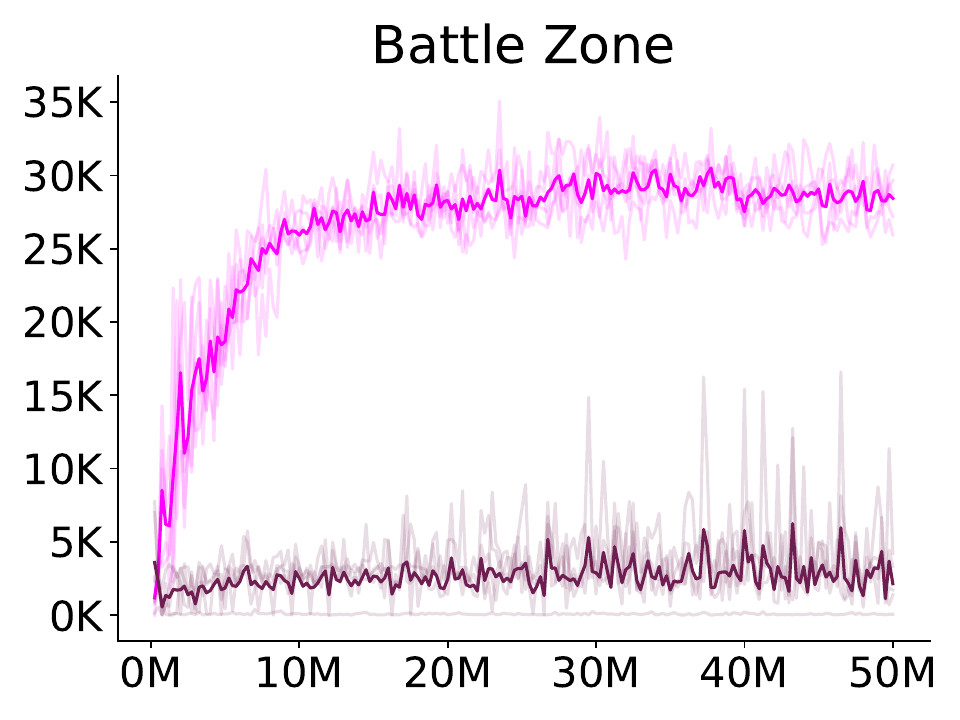} 
	\hspace{0.005\linewidth}
	\includegraphics[width=0.21\linewidth]{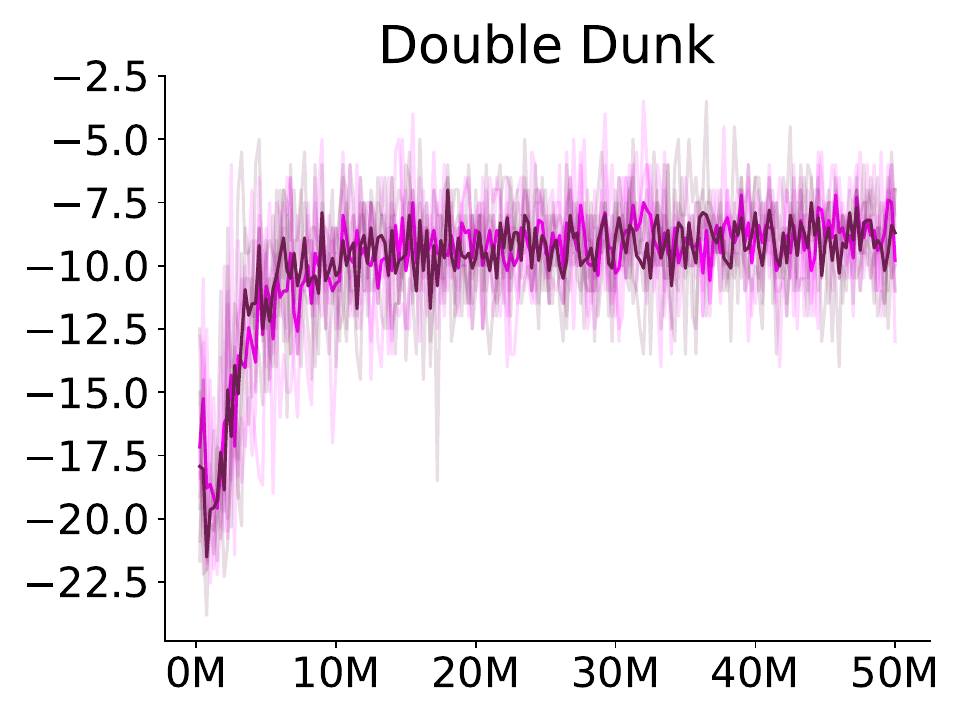} 
	\hspace{0.005\linewidth}
	\includegraphics[width=0.21\linewidth]{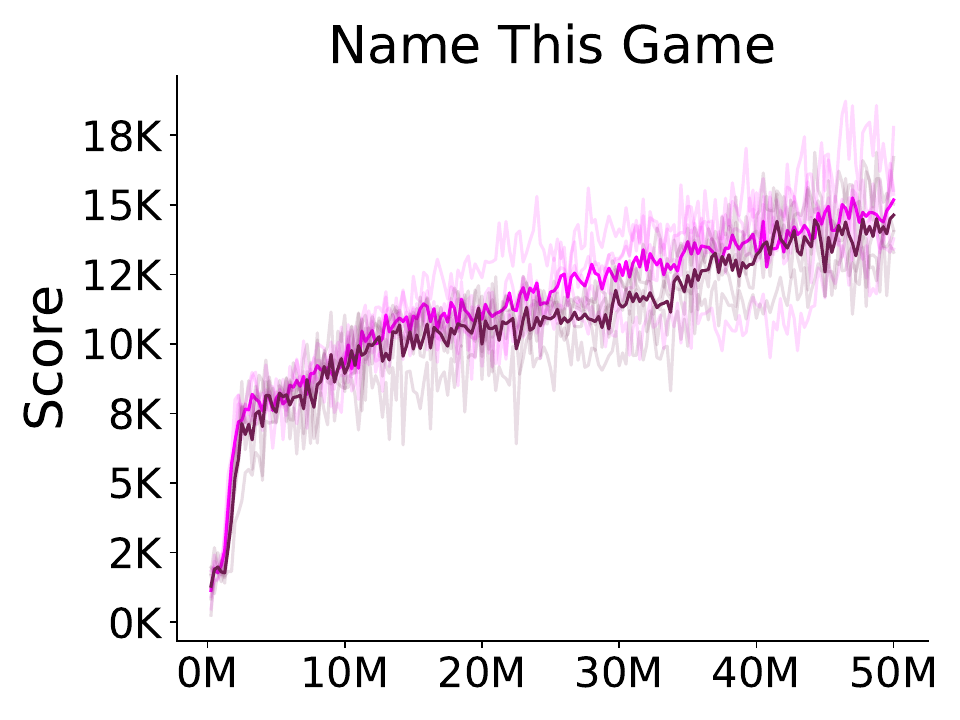} 
	\includegraphics[width=0.21\linewidth]{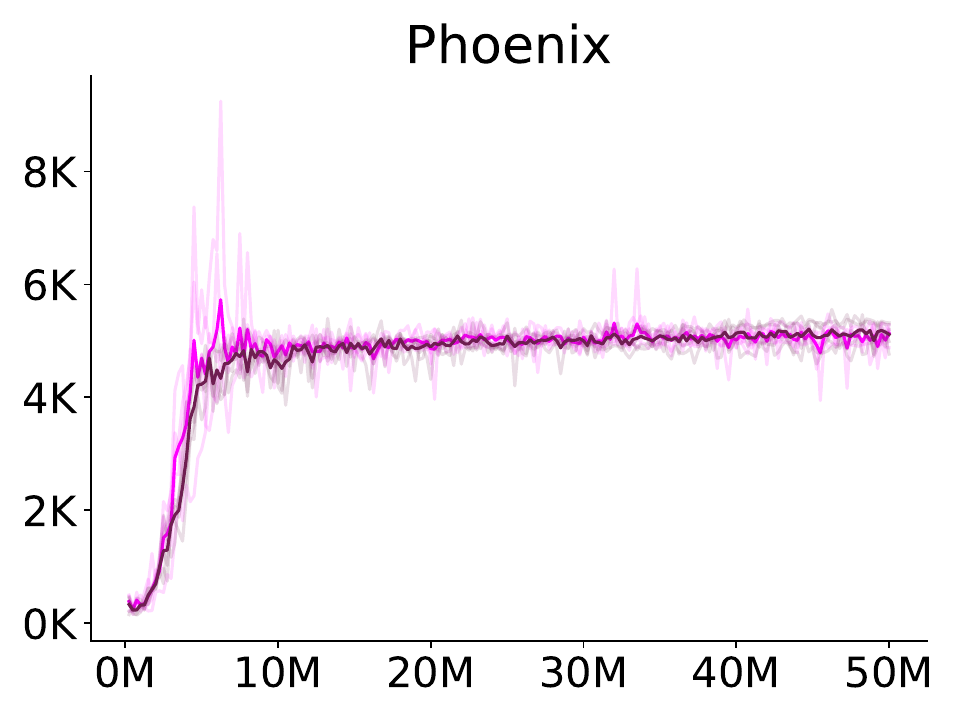} 
	\hspace{0.005\linewidth}
	\includegraphics[width=0.21\linewidth]{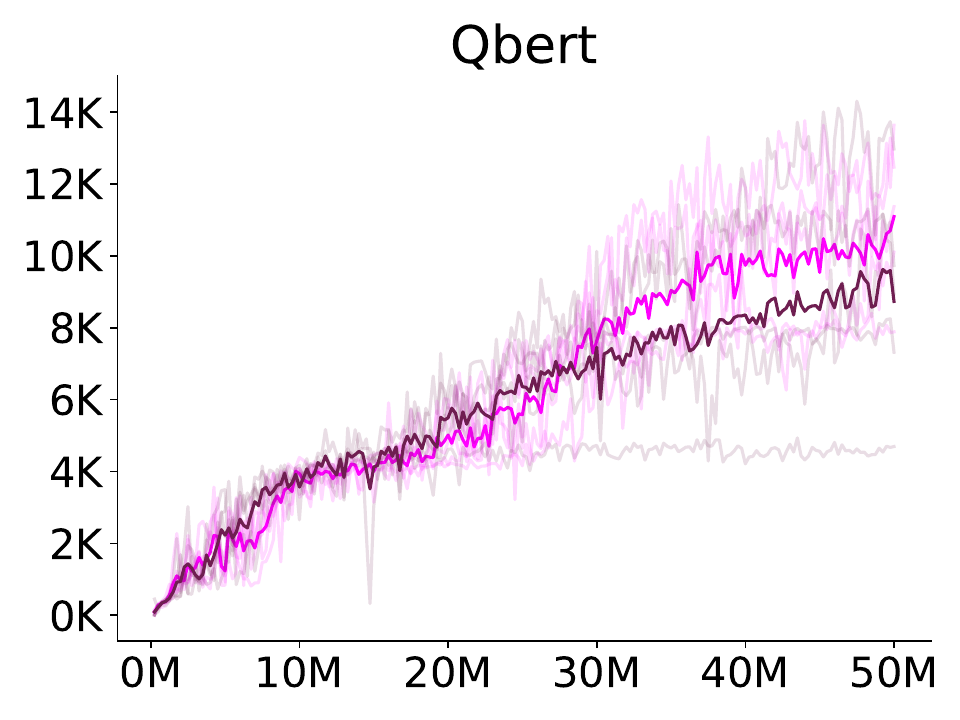} 
	\hspace{0.005\linewidth}
	\includegraphics[width=0.21\linewidth]{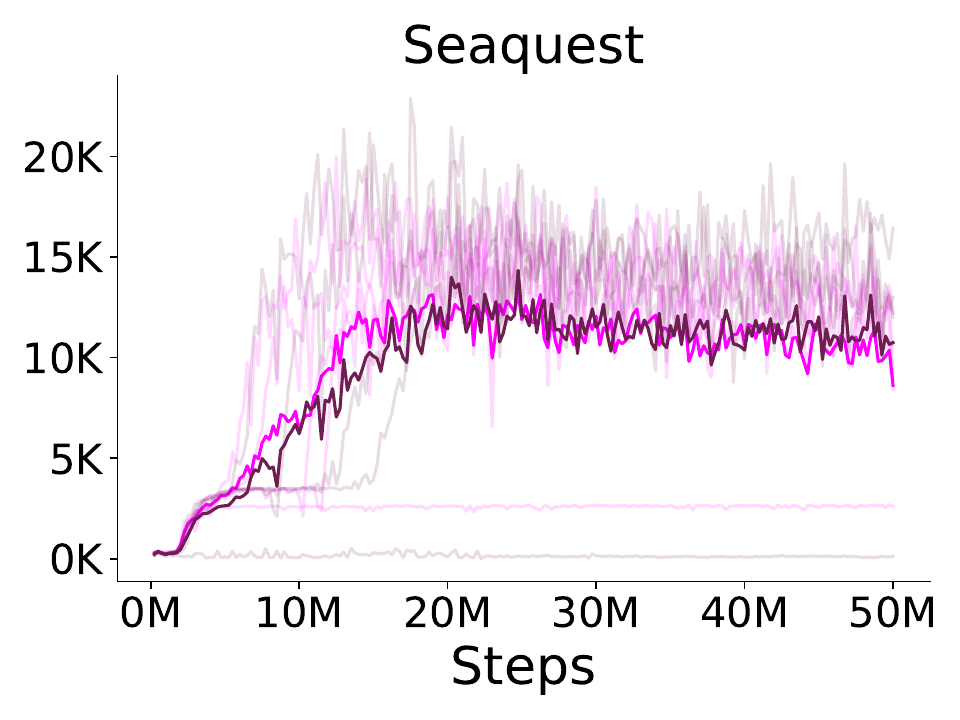} 
	\hspace{0.005\linewidth}
	\includegraphics[width=0.21\linewidth]{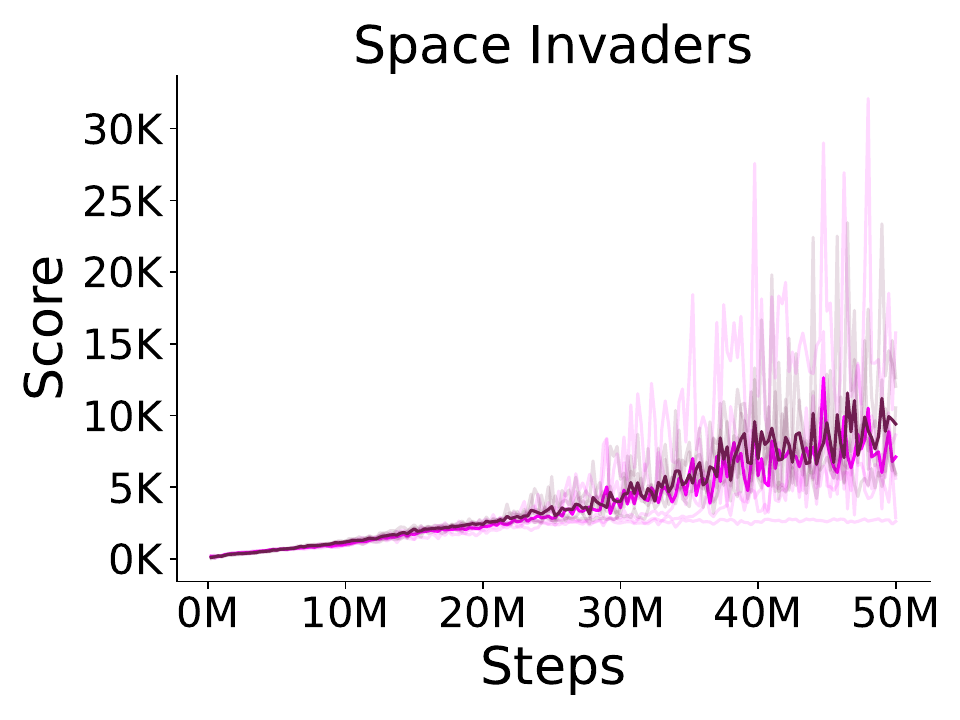} 
	\hspace{0.005\linewidth}
	\includegraphics[width=0.21\linewidth]{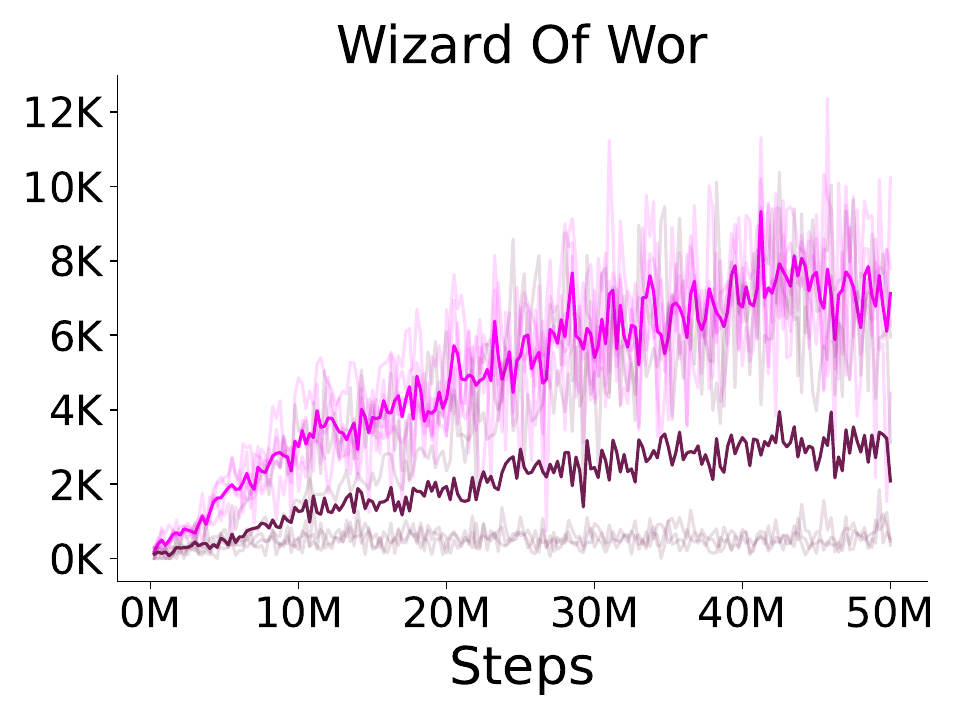} 
	\includegraphics[width=0.21\linewidth]{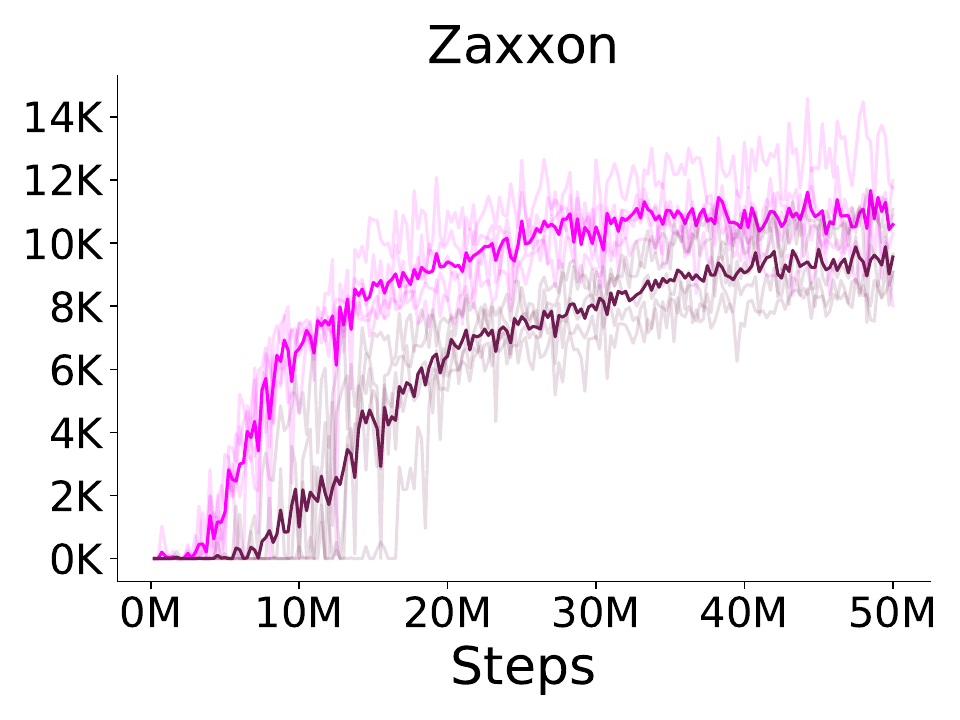} 
	\hspace{0.005\linewidth}
	\hspace{0.01\linewidth}\raisebox{2em}{\includegraphics[width=0.2\linewidth]{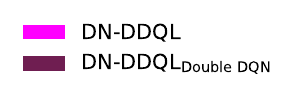}} 
	\caption{Performance of DN-DDQL compared to $\text{DN-DDQL}_\text{Double DQN}$. DN-DDQL performs more stably.}
	\label{DNDQN_vs_DoubleDQN:Ablation11:Score}
\end{figure}

Figure~\ref{DNDQN_vs_DoubleDQN:Ablation11:Score} compares DN-DDQL to $\text{DN-DDQL}_{\text{Double DQN}}$.
Again, it seems the stationarity of DQN-style updates is helpful.
In many environments, the performances are almost indistinguishable.
However, in environments like \textsc{Zaxxon}, \textsc{WizardOfWor}, \textsc{Alien}, and \textsc{Qbert}, DN-DDQL performs reliably better. 
In \textsc{BattleZone}, DN-DDQL is clearly better, with $\text{DN-DDQL}_{\text{Double DQN}}$ seemingly being unable to learn.
We conclude that for DN-DDQL, $\text{DN-DDQL}_{\text{DQN}}$ is more stable than $\text{DN-DDQL}_{\text{Double DQN}}$.

\clearpage

\clearpage
\section{Deep Double Q-learning: full results} \label{appendix:more_results}
This appendix does not include new results.
Rather, it expands the results shown in the body regarding our ablations, overestimation, and performance of Double DQN, DDQL, DM-DDQL, and DN-DDQL.

\subsection{Ablations}

\begin{figure}[h]
        \centering
    	\includegraphics[width=0.22\linewidth]{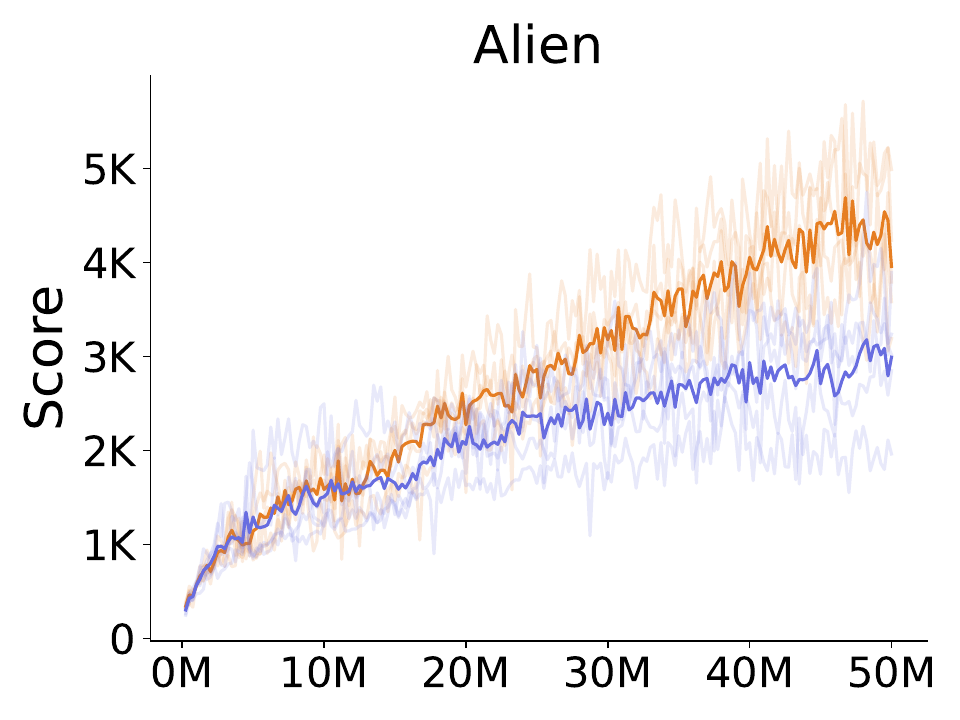} 
	\hspace{0.005\linewidth}
	\includegraphics[width=0.22\linewidth]{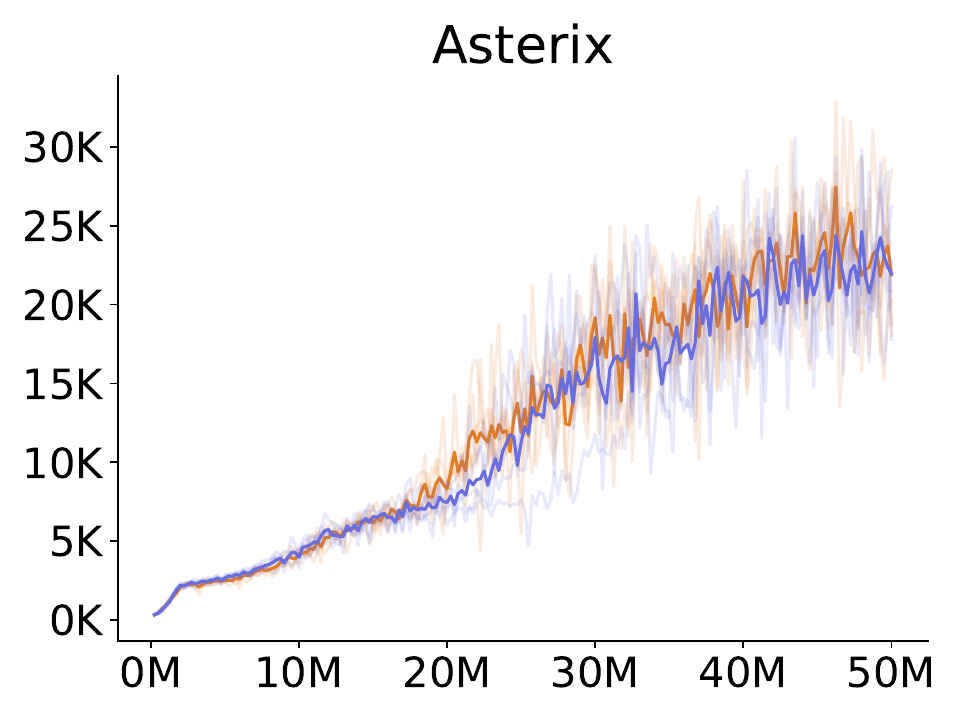} 
	\hspace{0.005\linewidth}
	\includegraphics[width=0.22\linewidth]{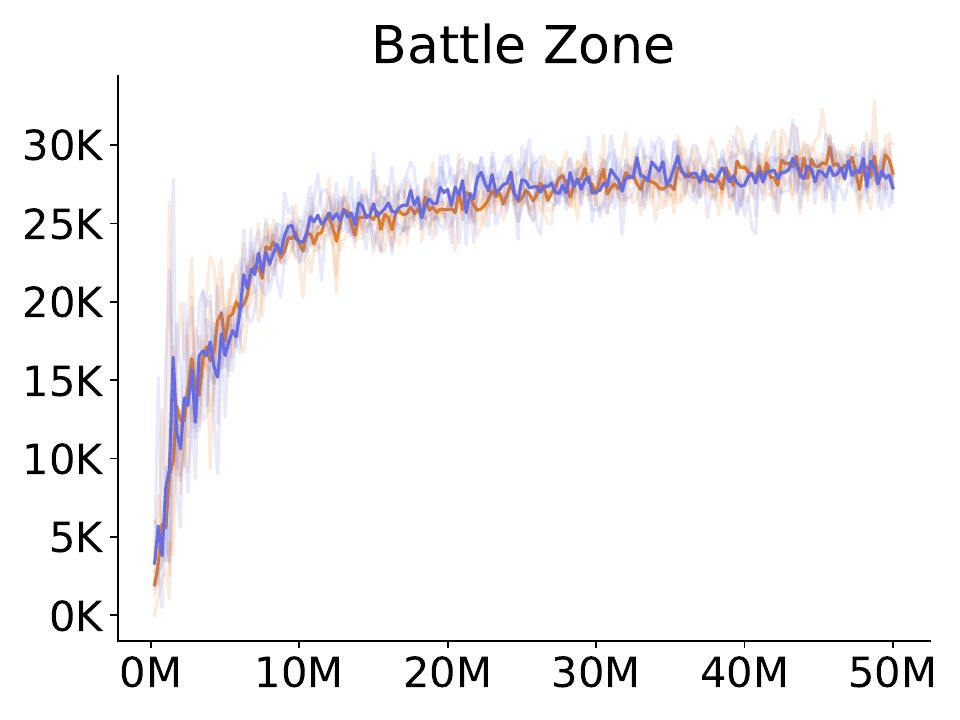} 
	\hspace{0.005\linewidth}
	\includegraphics[width=0.22\linewidth]{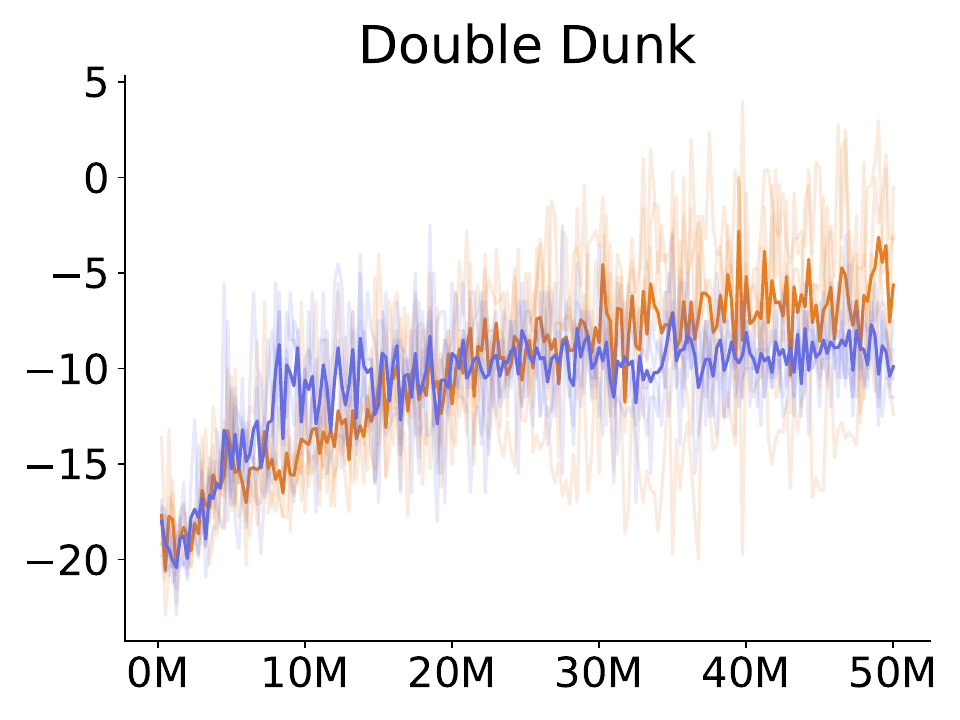} 
	\hspace{0.005\linewidth}
	\includegraphics[width=0.22\linewidth]{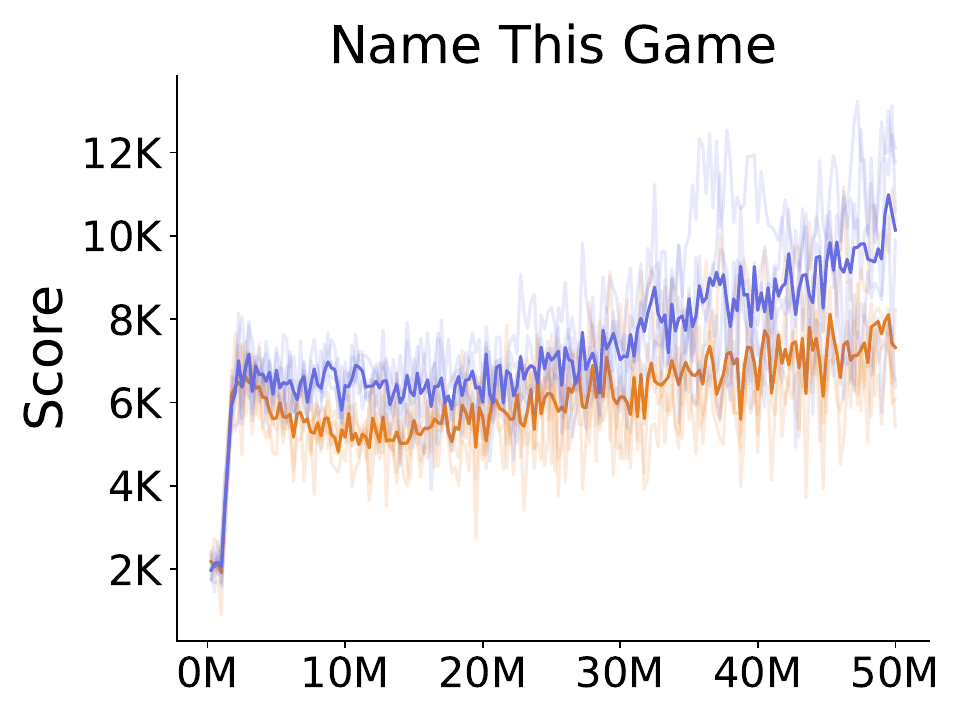}
	\includegraphics[width=0.22\linewidth]{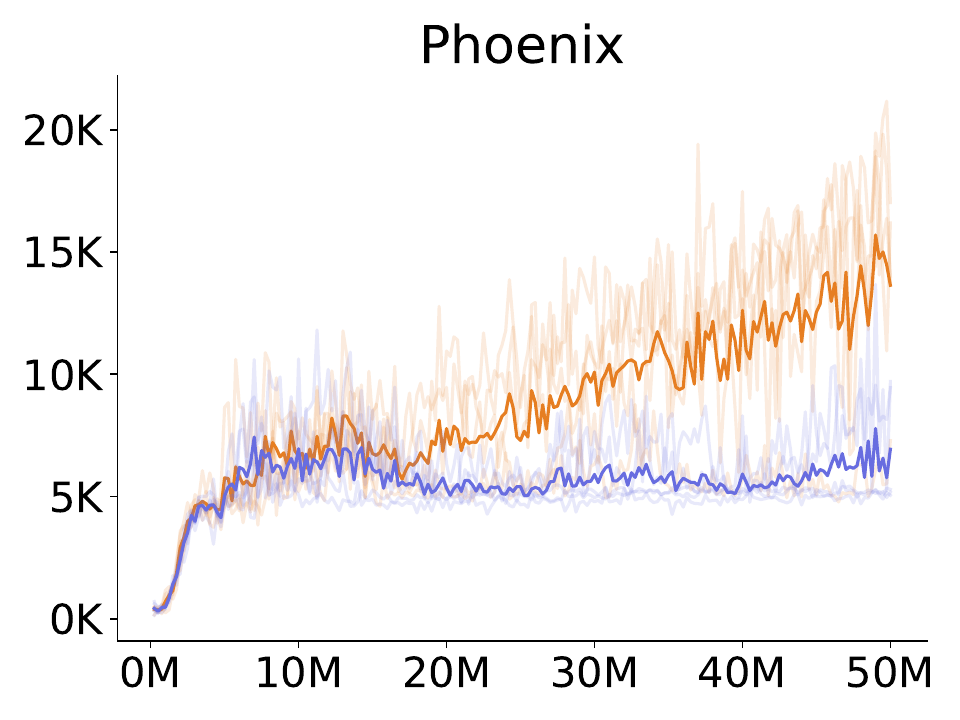} 
	\hspace{0.005\linewidth}
	\includegraphics[width=0.22\linewidth]{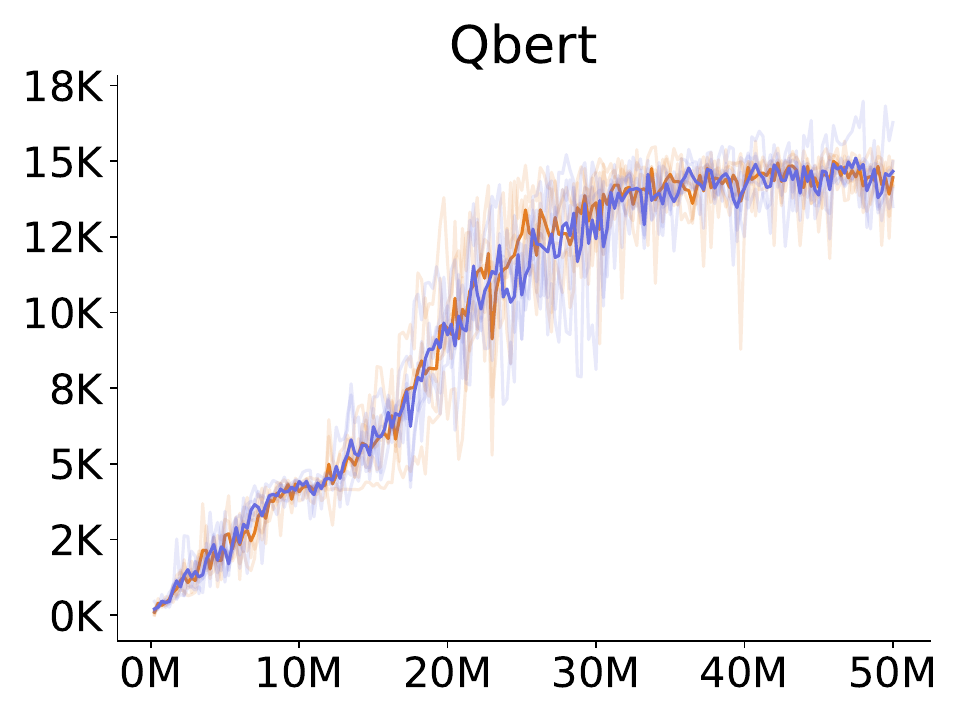} 
	\hspace{0.005\linewidth}
	\includegraphics[width=0.22\linewidth]{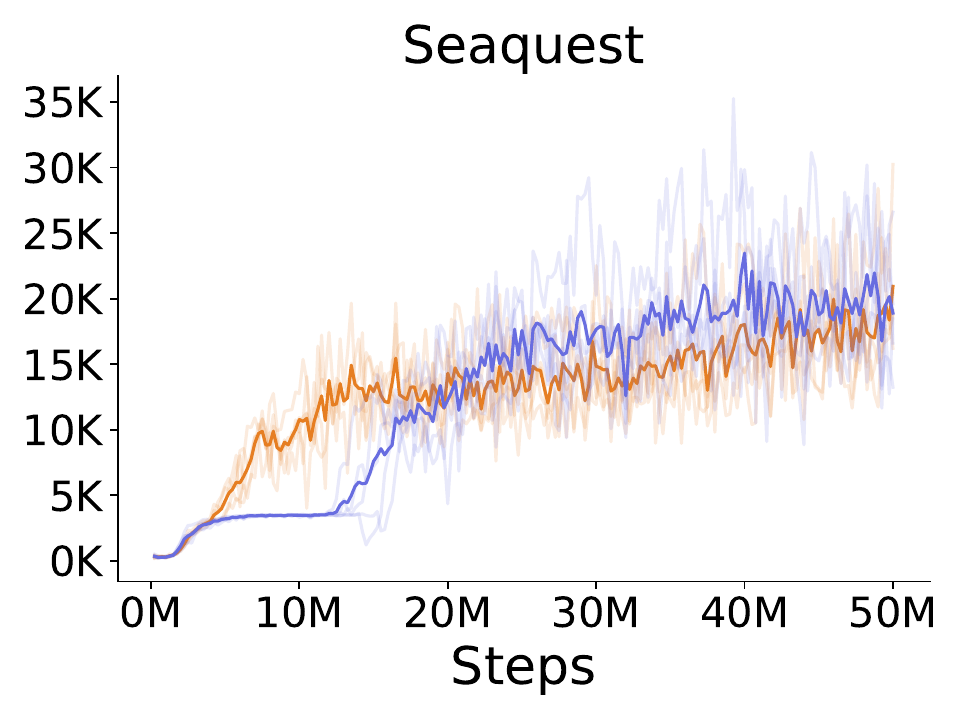} 
	\hspace{0.005\linewidth}
	\includegraphics[width=0.22\linewidth]{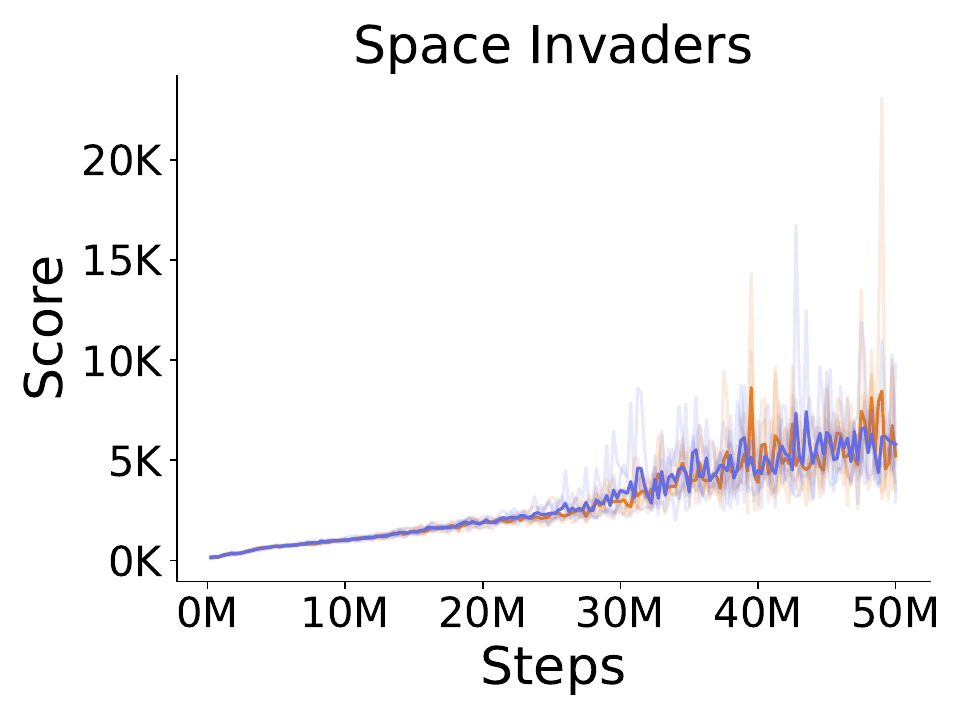} 
	\hspace{0.005\linewidth}
	\includegraphics[width=0.22\linewidth]{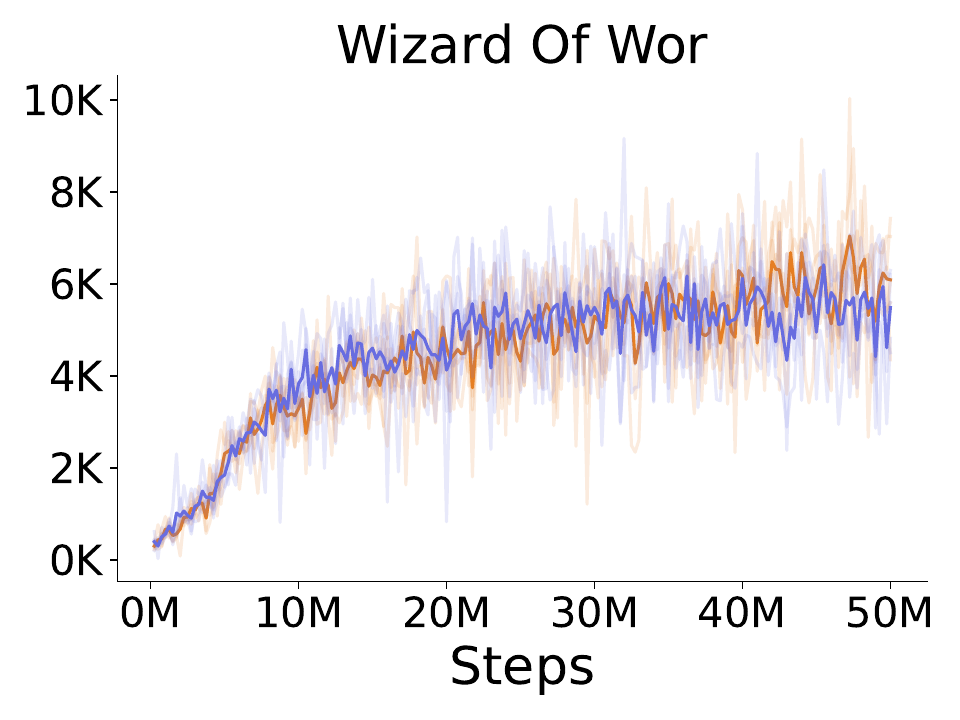} 
	\includegraphics[width=0.22\linewidth]{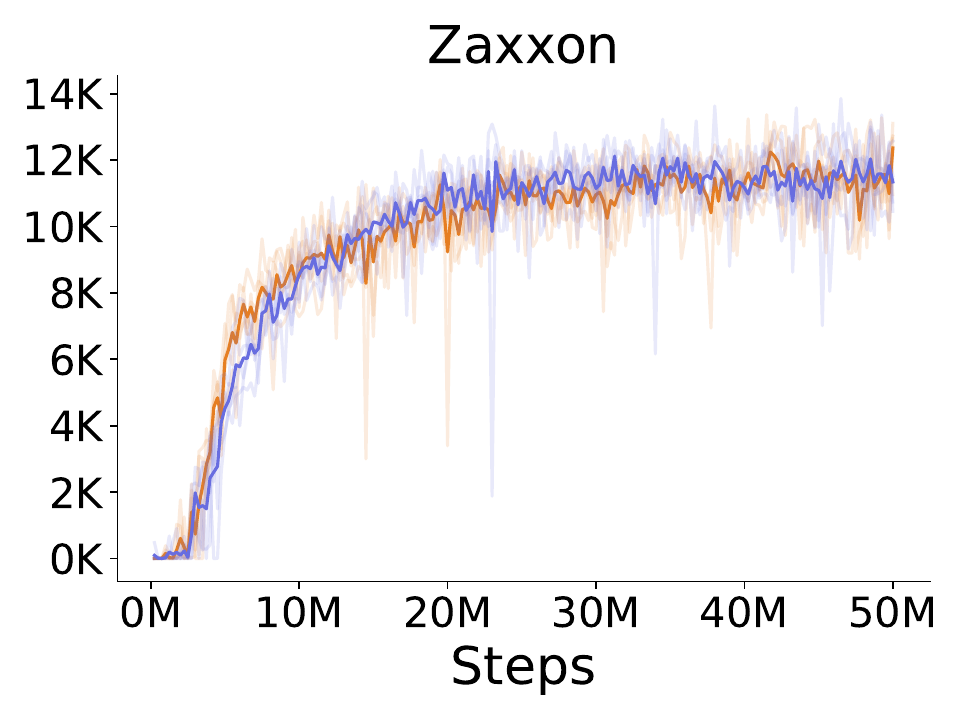} 
	\hspace{0.005\linewidth}
	\hspace{0.02\linewidth}\raisebox{2.6em}{\includegraphics[width=0.2\linewidth]{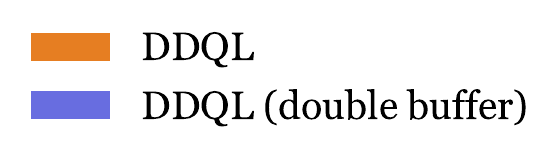}} 
	\caption{Performance of DDQL compared to DDQL (double buffer). The algorithms largely perform at a similar level, with a few exceptions.}
	\label{DatasetPartitioningDH:Ablation11:Score}
\end{figure}

\begin{figure}[h]
        \centering
    	\includegraphics[width=0.22\linewidth]{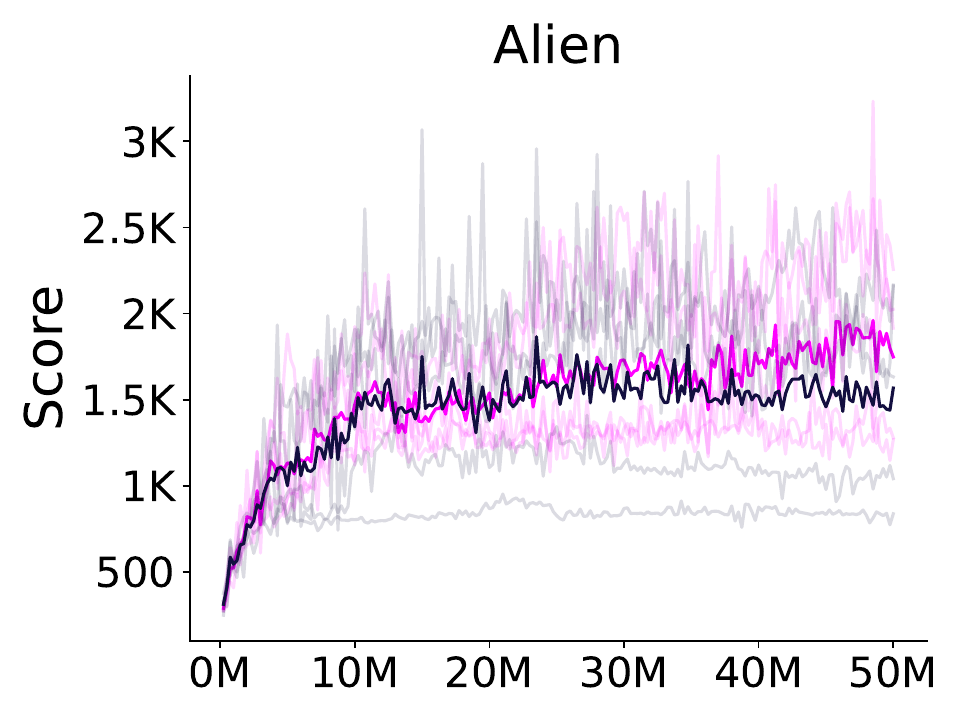} 
	\hspace{0.005\linewidth}
	\includegraphics[width=0.22\linewidth]{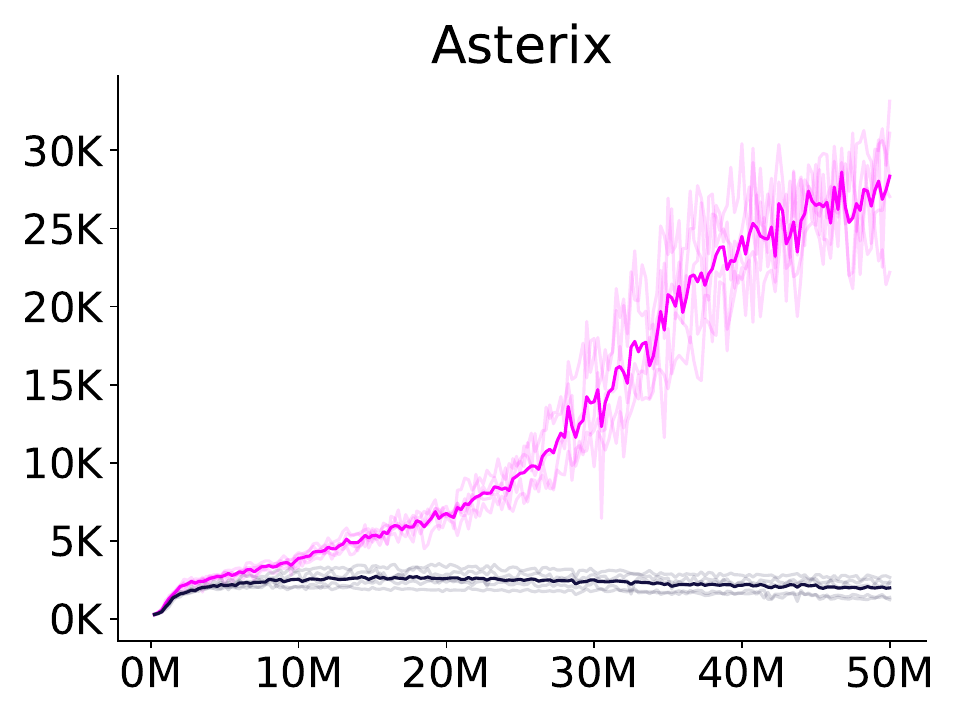} 
	\hspace{0.005\linewidth}
	\includegraphics[width=0.22\linewidth]{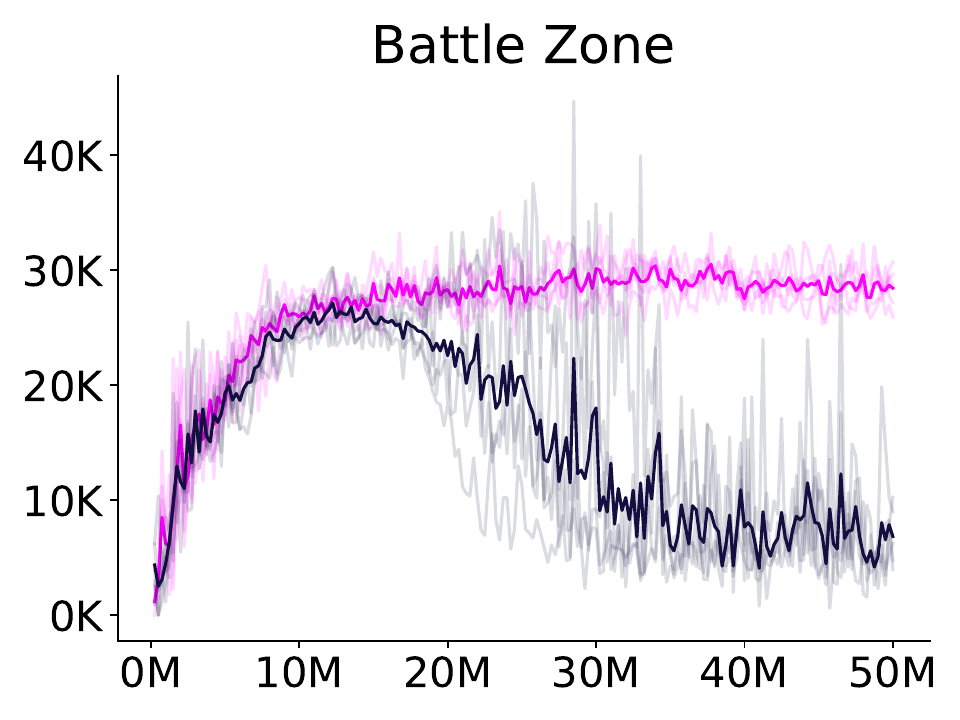} 
	\hspace{0.005\linewidth}
	\includegraphics[width=0.22\linewidth]{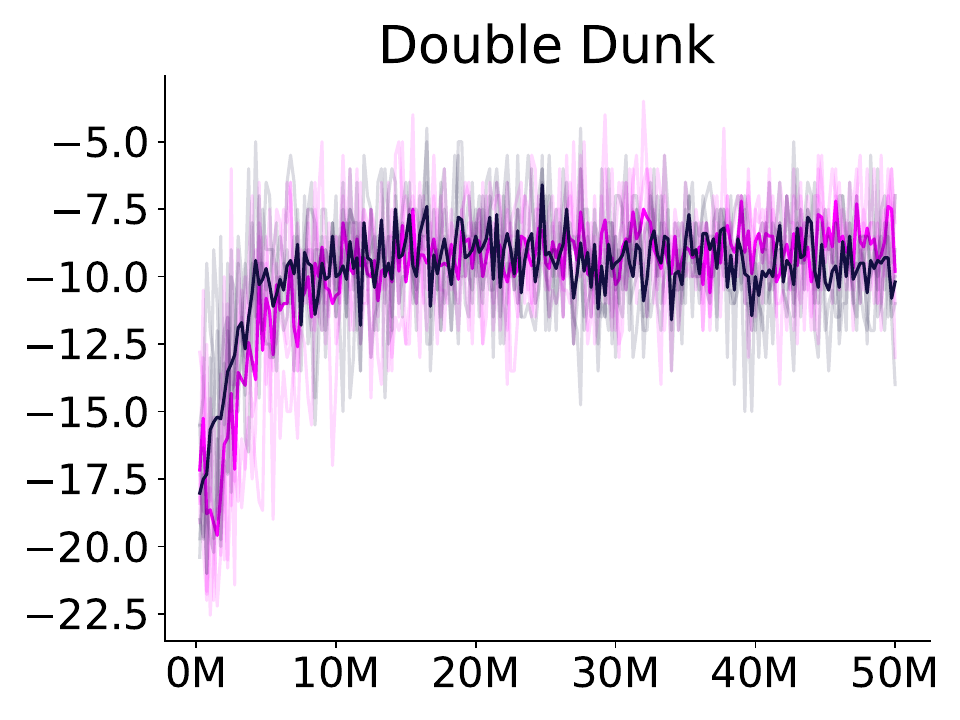} 
	\hspace{0.005\linewidth}
	\includegraphics[width=0.22\linewidth]{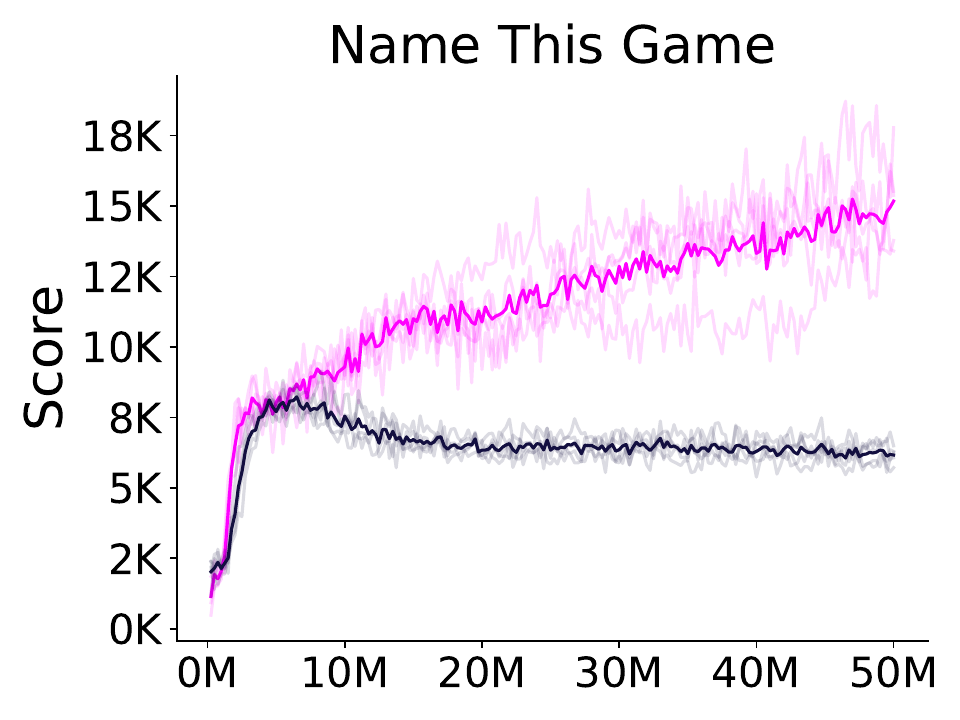} 
	\includegraphics[width=0.22\linewidth]{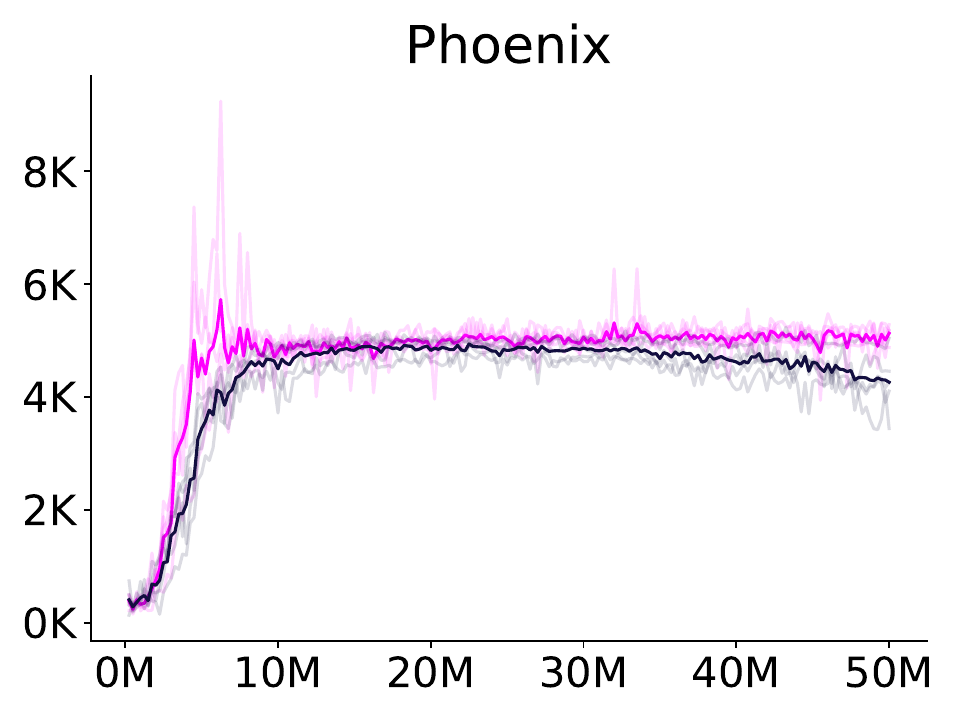} 
	\hspace{0.005\linewidth}
	\includegraphics[width=0.22\linewidth]{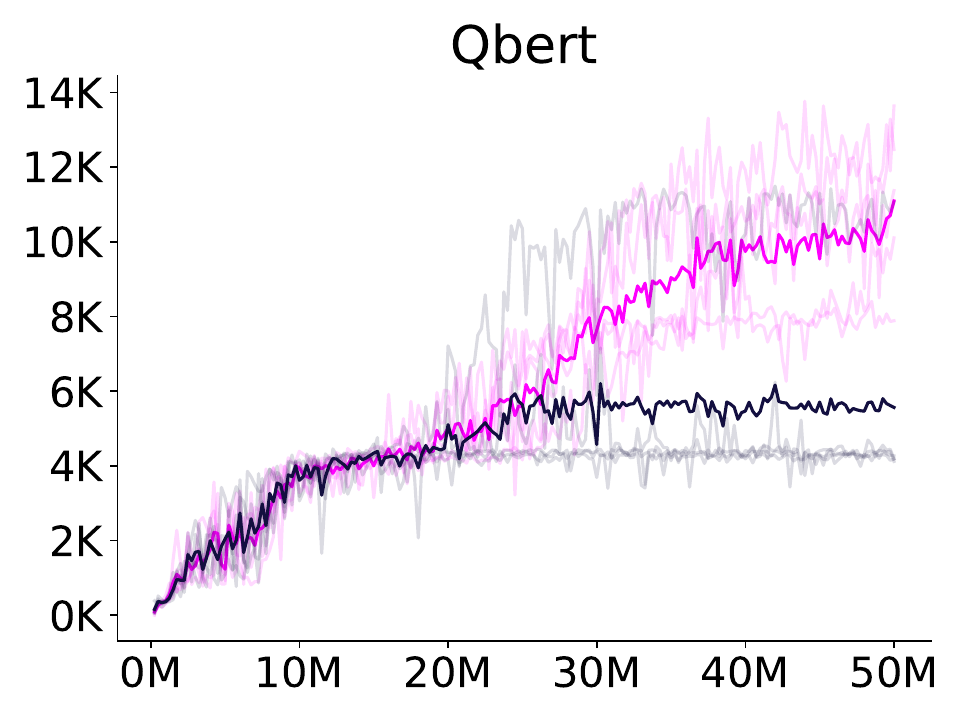} 
	\hspace{0.005\linewidth}
	\includegraphics[width=0.22\linewidth]{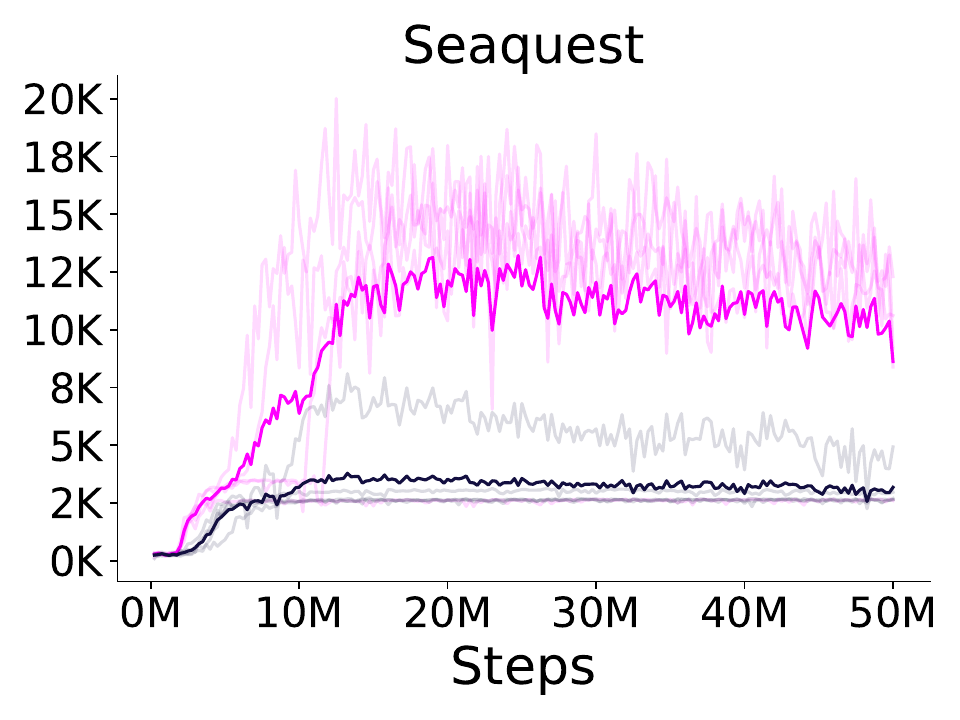} 
	\hspace{0.005\linewidth}
	\includegraphics[width=0.22\linewidth]{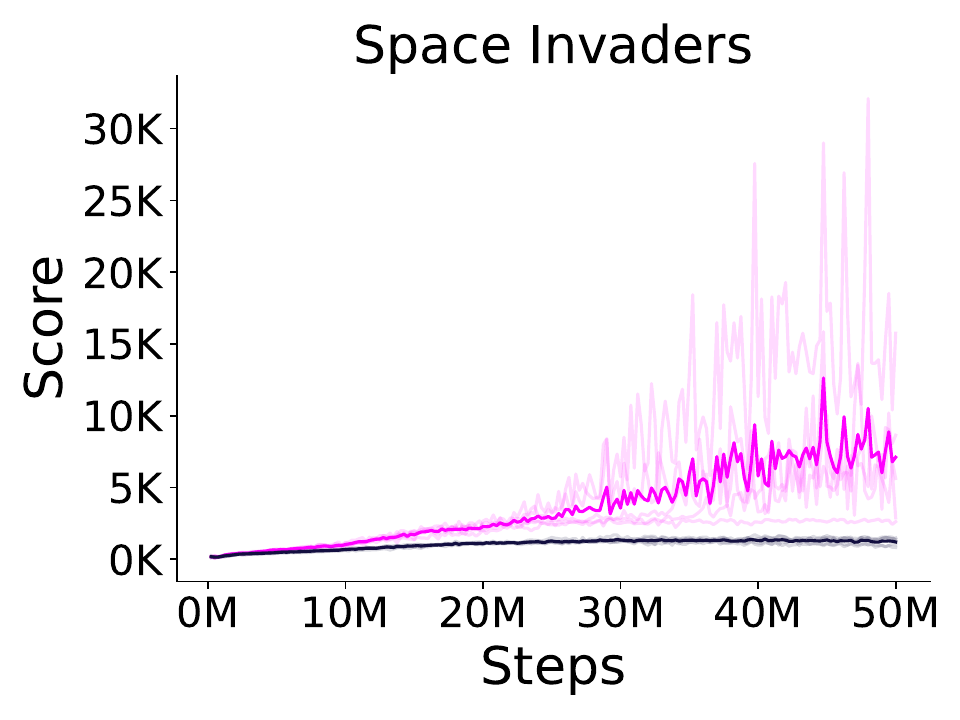} 
	\hspace{0.005\linewidth}
	\includegraphics[width=0.22\linewidth]{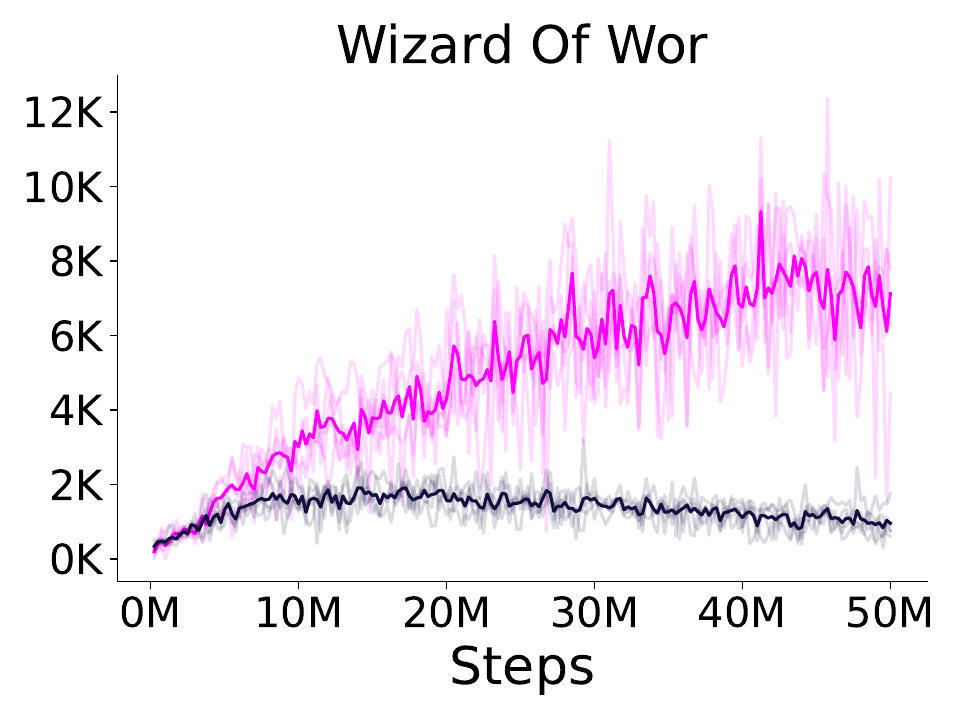} 
	\includegraphics[width=0.22\linewidth]{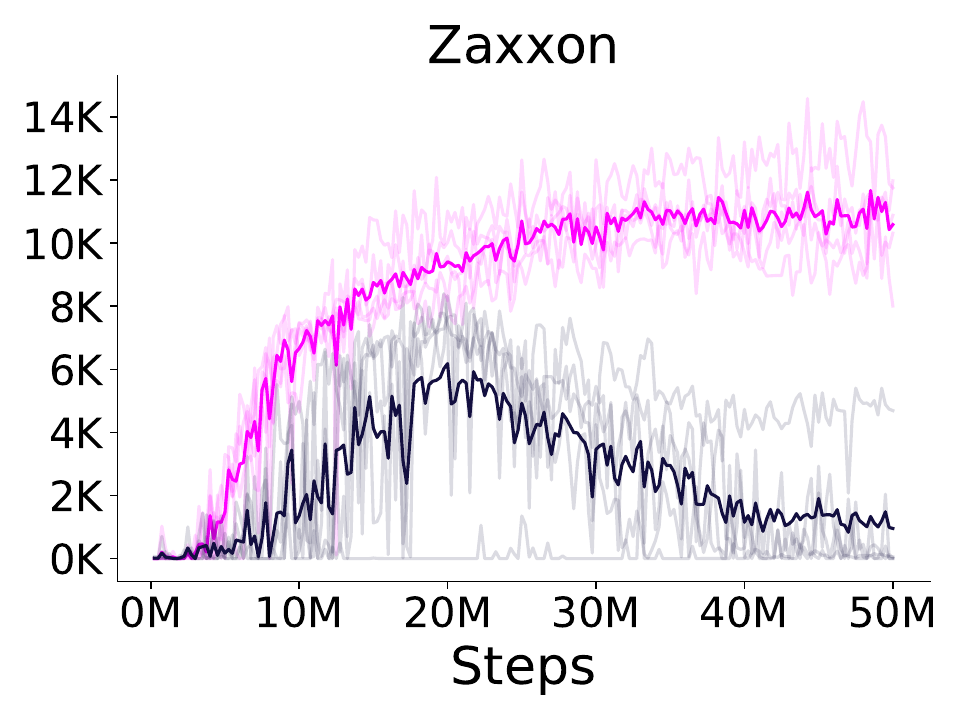} 
	\hspace{0.005\linewidth}
	\hspace{0.02\linewidth}\raisebox{2.6em}{\includegraphics[width=0.2\linewidth]{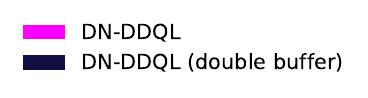}} 
	\caption{Performance of DN-DDQL compared to DN-DDQL (double buffer). DN-DDQL (double buffer) performs quite poorly.}
	\label{DatasetPartitioningDN:Ablation11:Score}
\end{figure}

\begin{figure}[h]
        \centering
    	\includegraphics[width=0.22\linewidth]{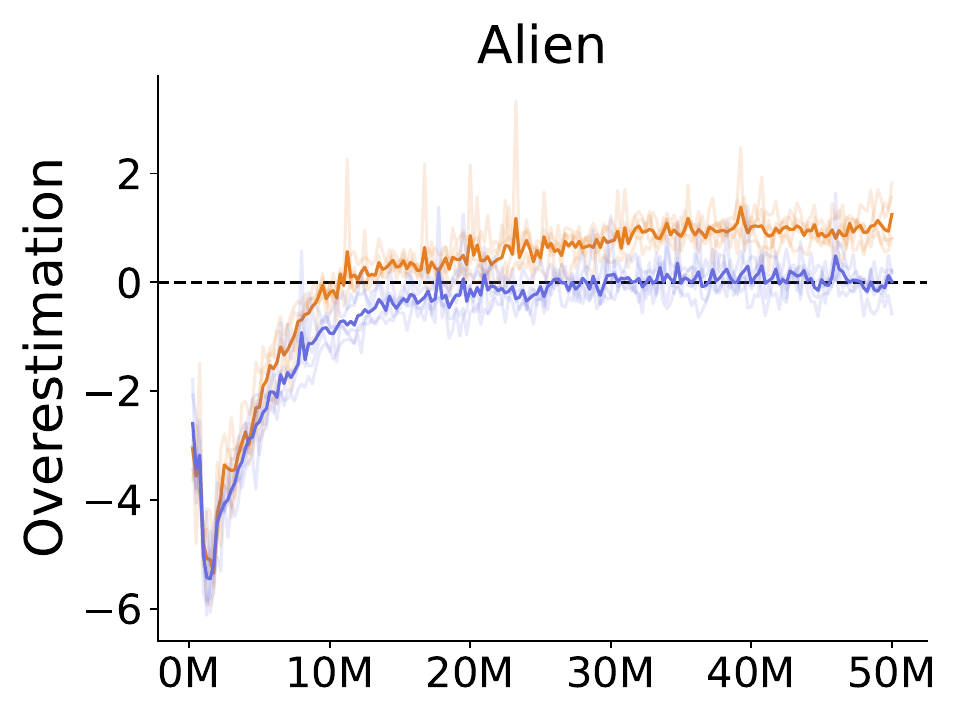} 
	\hspace{0.005\linewidth}
	\includegraphics[width=0.22\linewidth]{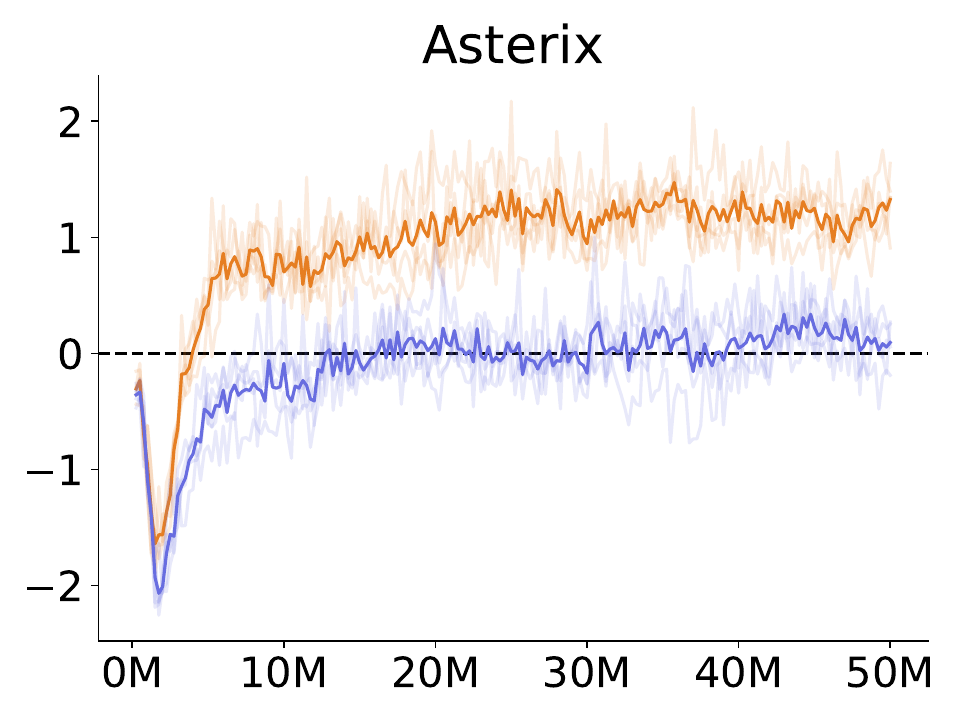} 
	\hspace{0.005\linewidth}
	\includegraphics[width=0.22\linewidth]{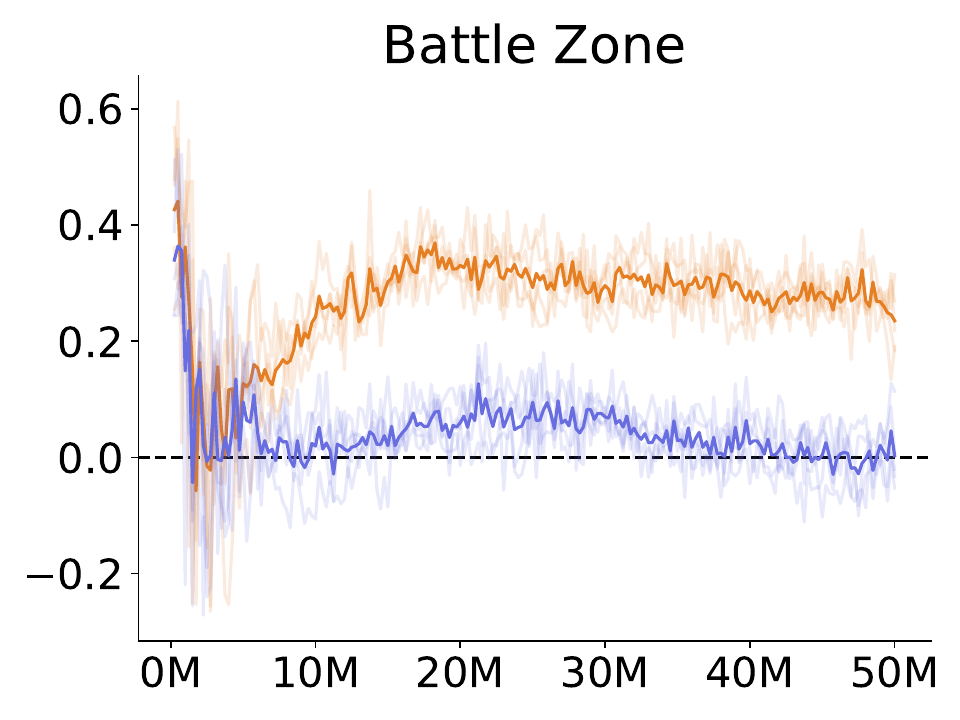} 
	\hspace{0.005\linewidth}
	\includegraphics[width=0.22\linewidth]{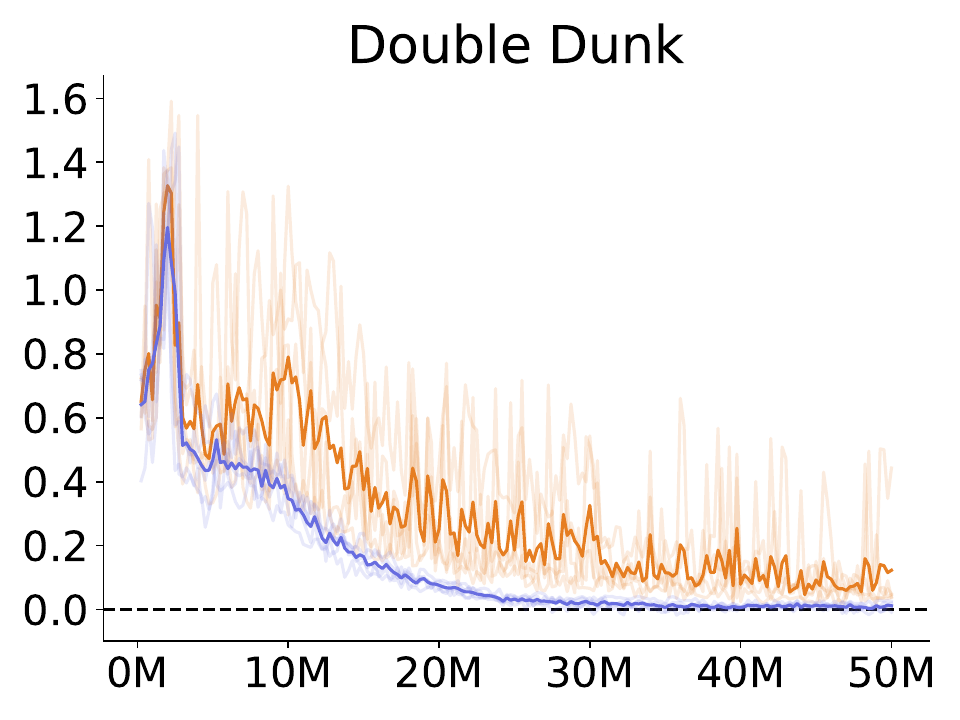} 
	\hspace{0.005\linewidth}
	\includegraphics[width=0.22\linewidth]{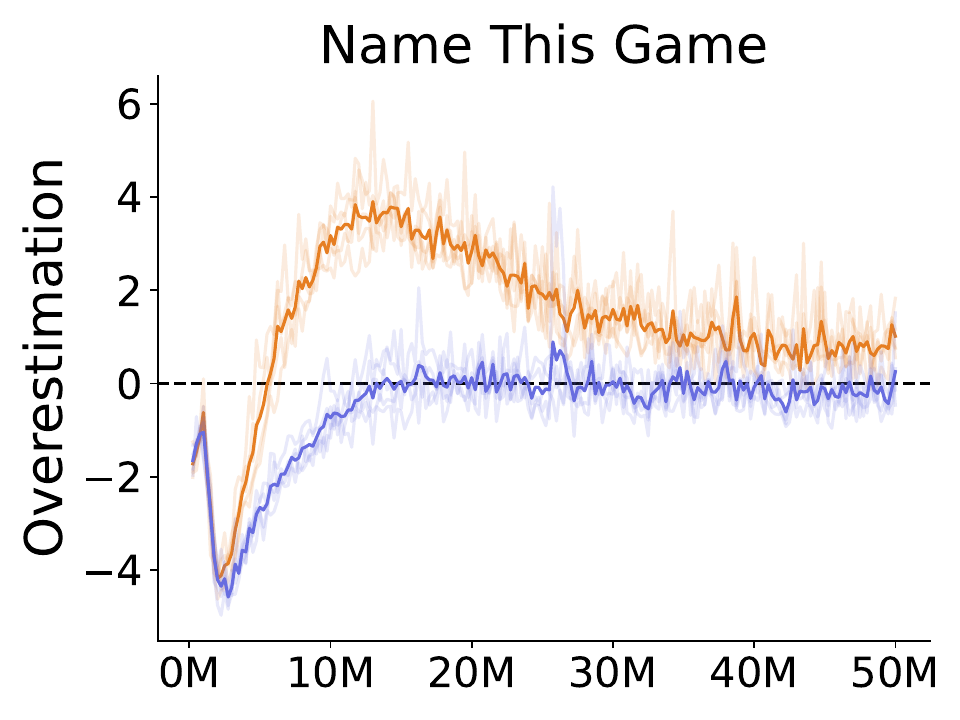} 
	\includegraphics[width=0.22\linewidth]{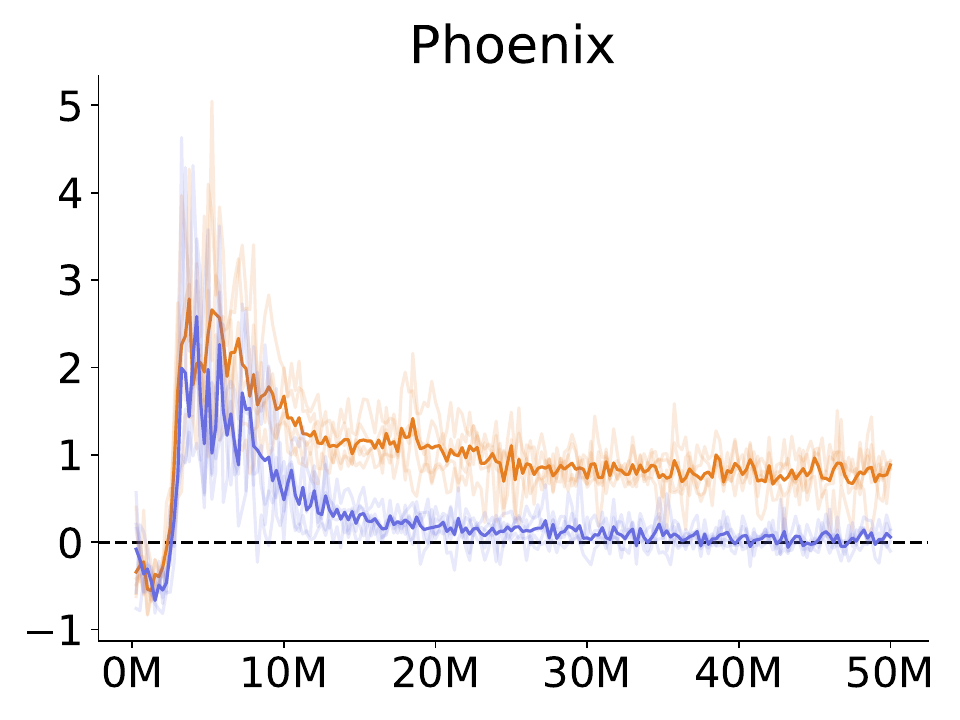} 
	\hspace{0.005\linewidth}
	\includegraphics[width=0.22\linewidth]{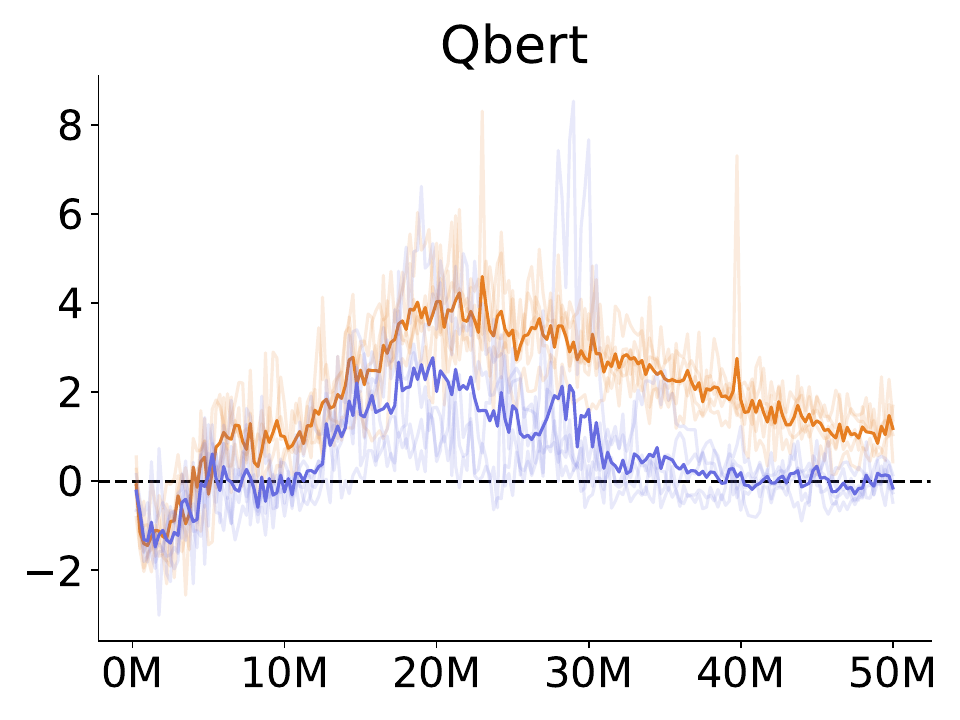} 
	\hspace{0.005\linewidth}
	\includegraphics[width=0.22\linewidth]{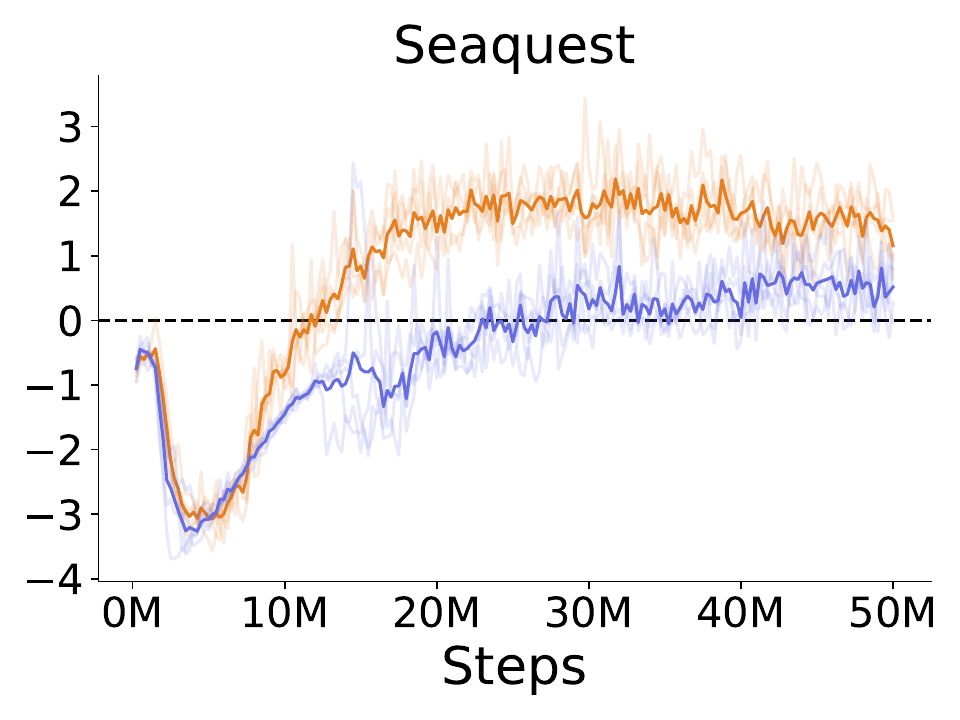} 
	\hspace{0.005\linewidth}
	\includegraphics[width=0.22\linewidth]{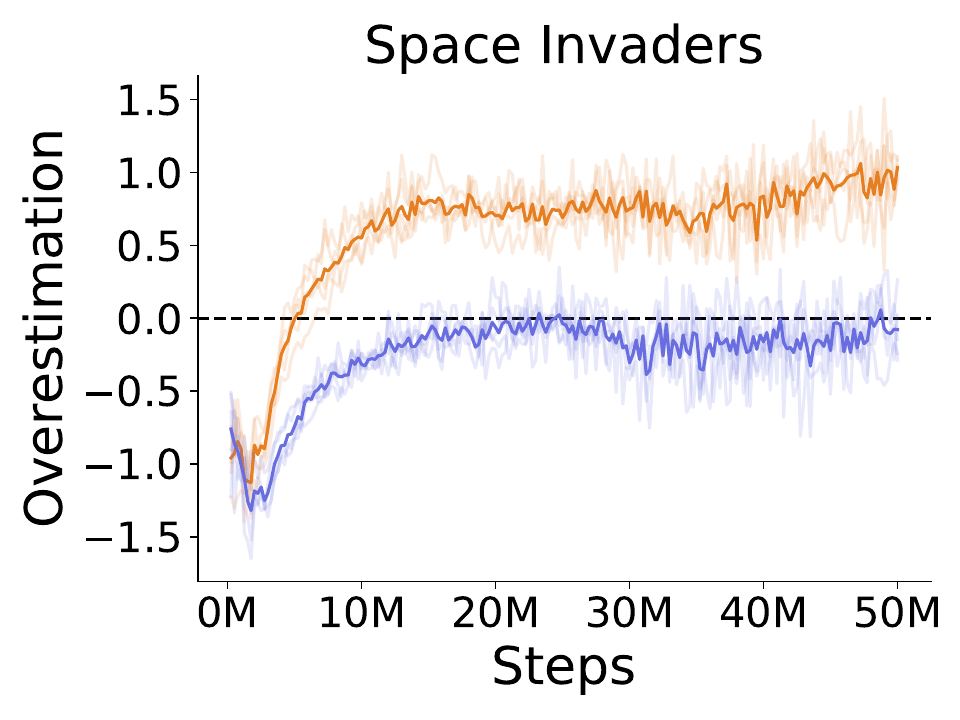} 
	\hspace{0.005\linewidth}
	\includegraphics[width=0.22\linewidth]{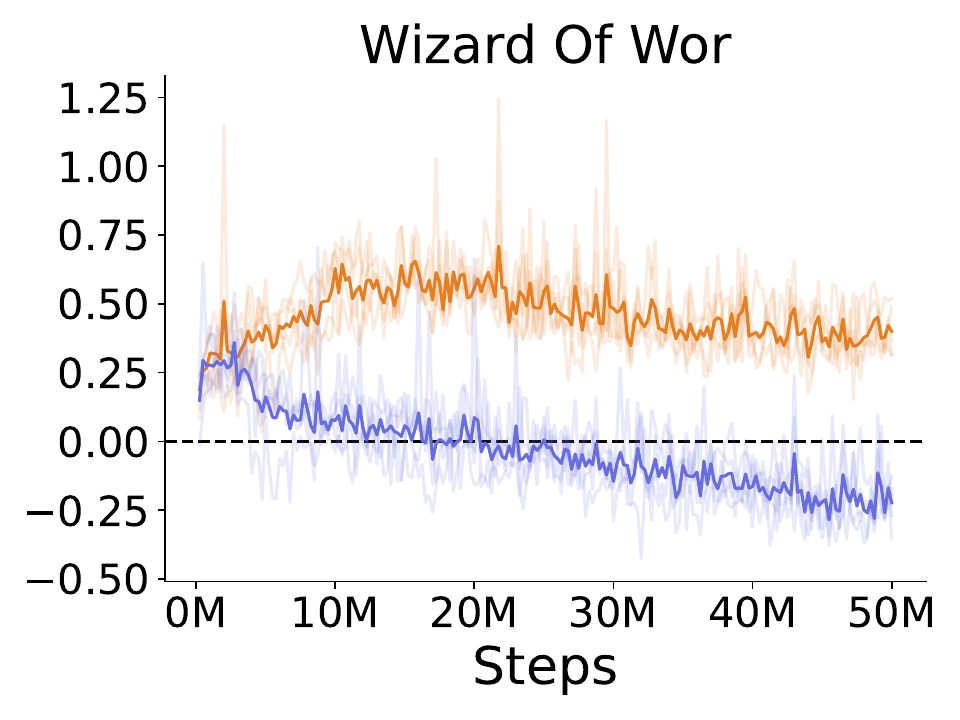} 
	\includegraphics[width=0.22\linewidth]{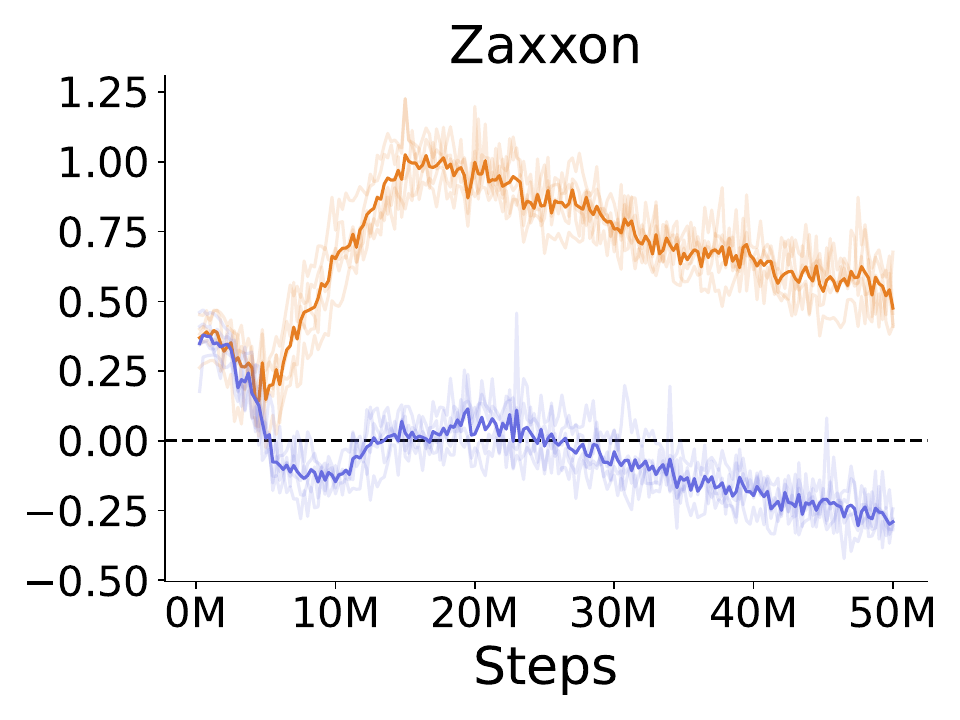} 
	\hspace{0.005\linewidth}
	\hspace{0.045\linewidth}\raisebox{10mm}[0pt][0pt]{\includegraphics[width=0.2\linewidth]{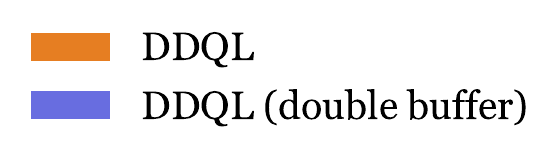}} 
	\caption{Overestimation of DDQL compared to DDQL (double buffer). The double buffer strategy reduces overestimation.}
	\label{DatasetPartitioningDH:Ablation11:Overestimation}
\end{figure}

\begin{figure}[h]
        \centering
    	\includegraphics[width=0.22\linewidth]{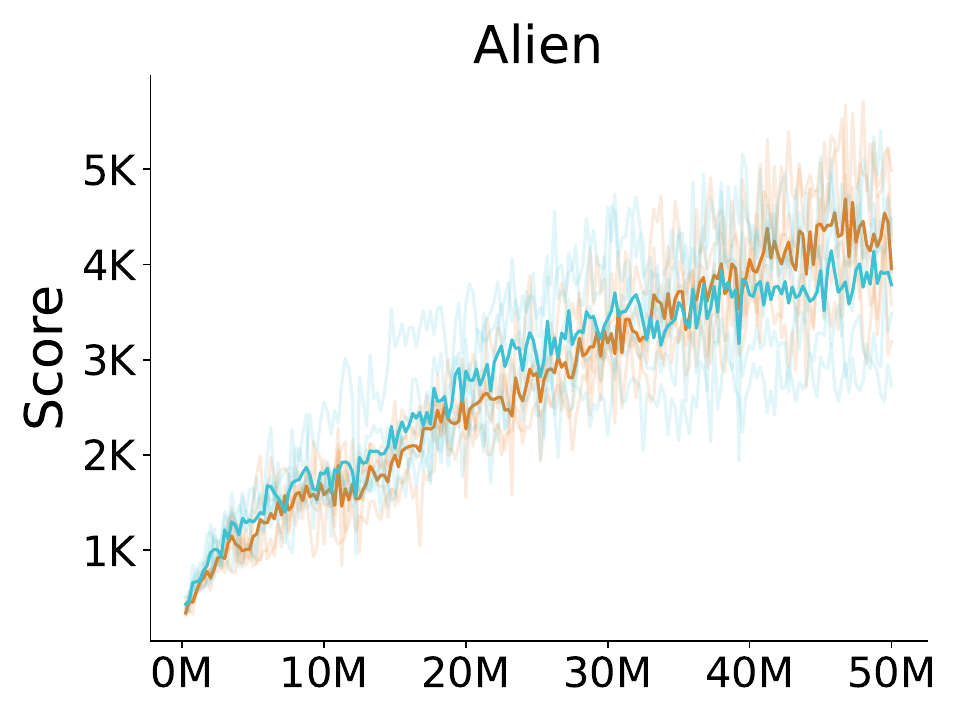} 
	\hspace{0.005\linewidth}
	\includegraphics[width=0.22\linewidth]{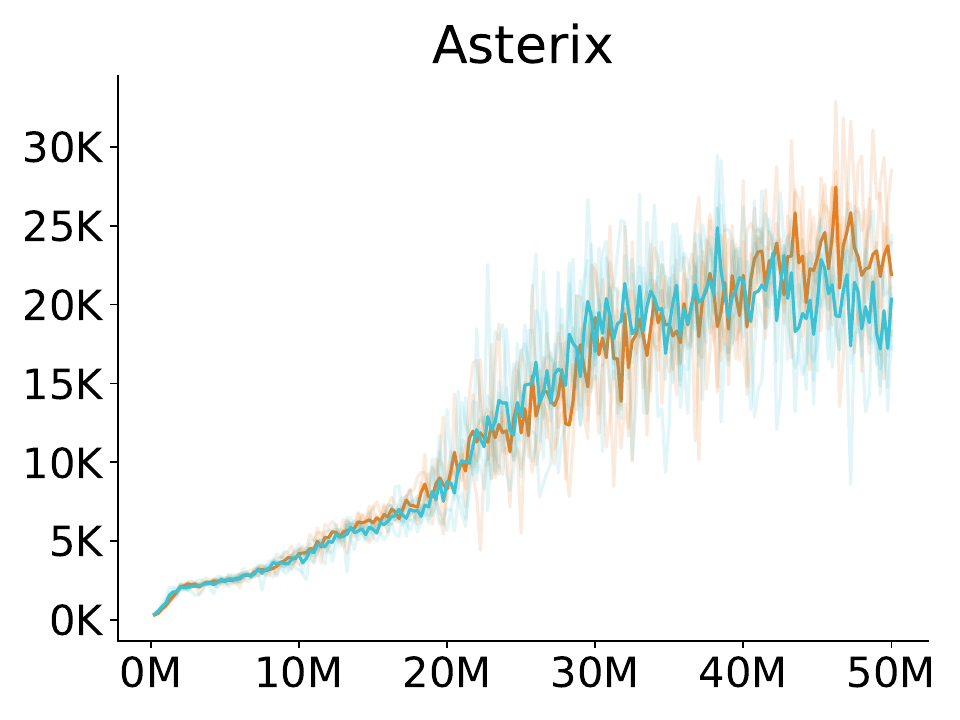} 
	\hspace{0.005\linewidth}
	\includegraphics[width=0.22\linewidth]{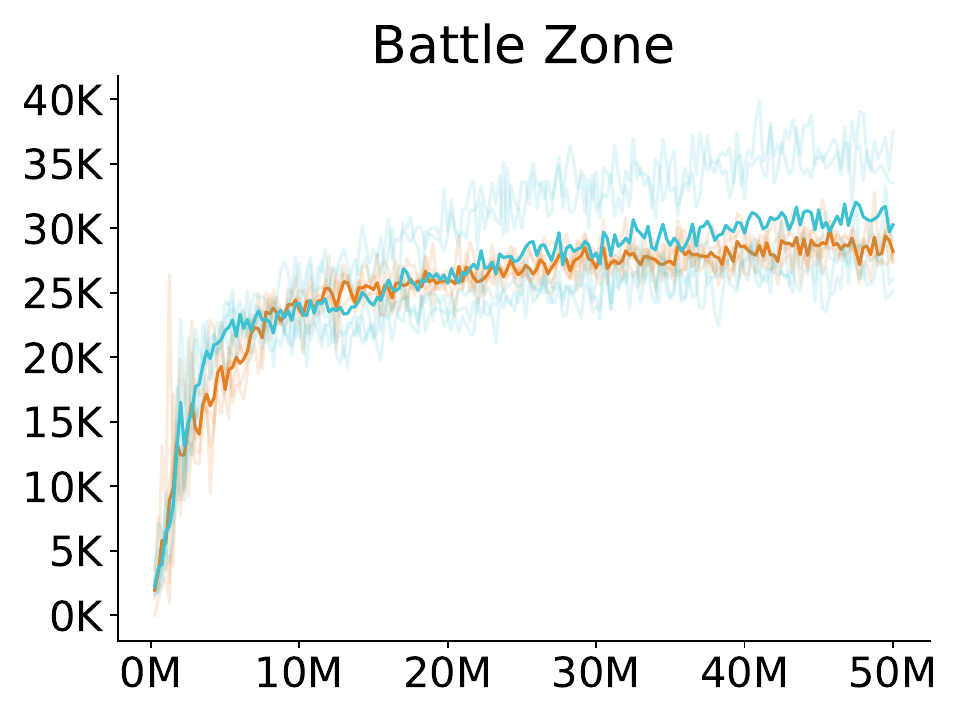} 
	\hspace{0.005\linewidth}
	\includegraphics[width=0.22\linewidth]{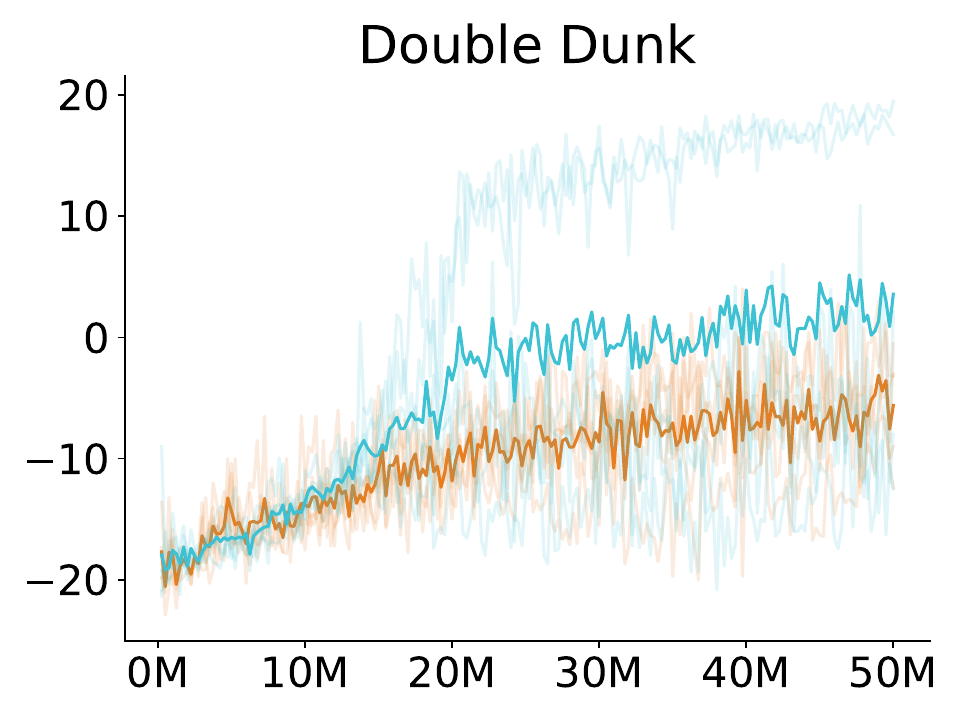} 
	\hspace{0.005\linewidth}
	\includegraphics[width=0.22\linewidth]{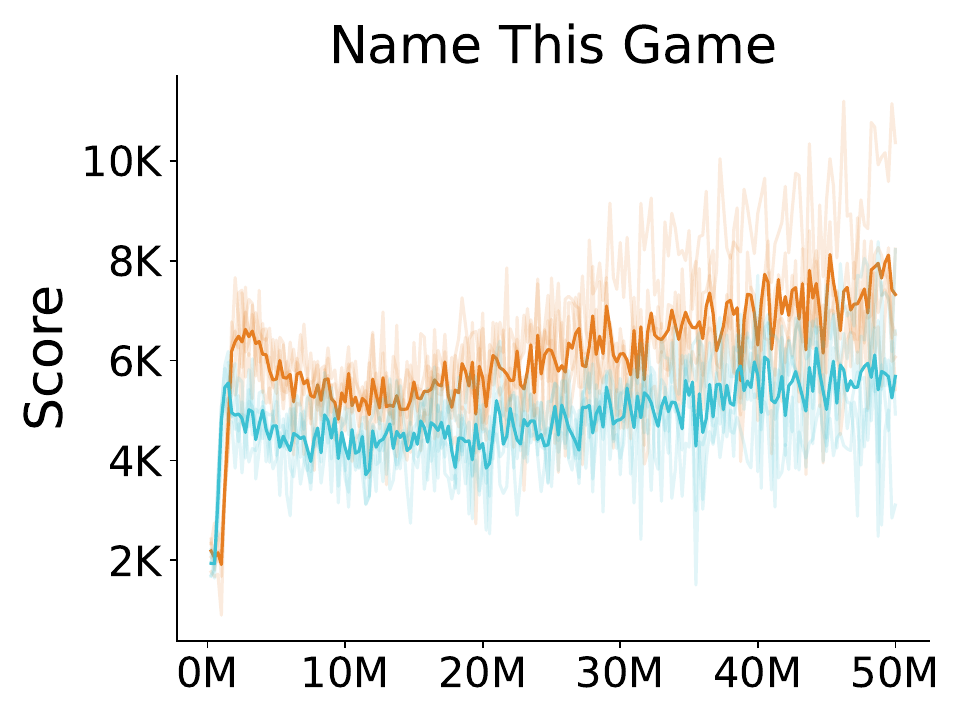} 
	\includegraphics[width=0.22\linewidth]{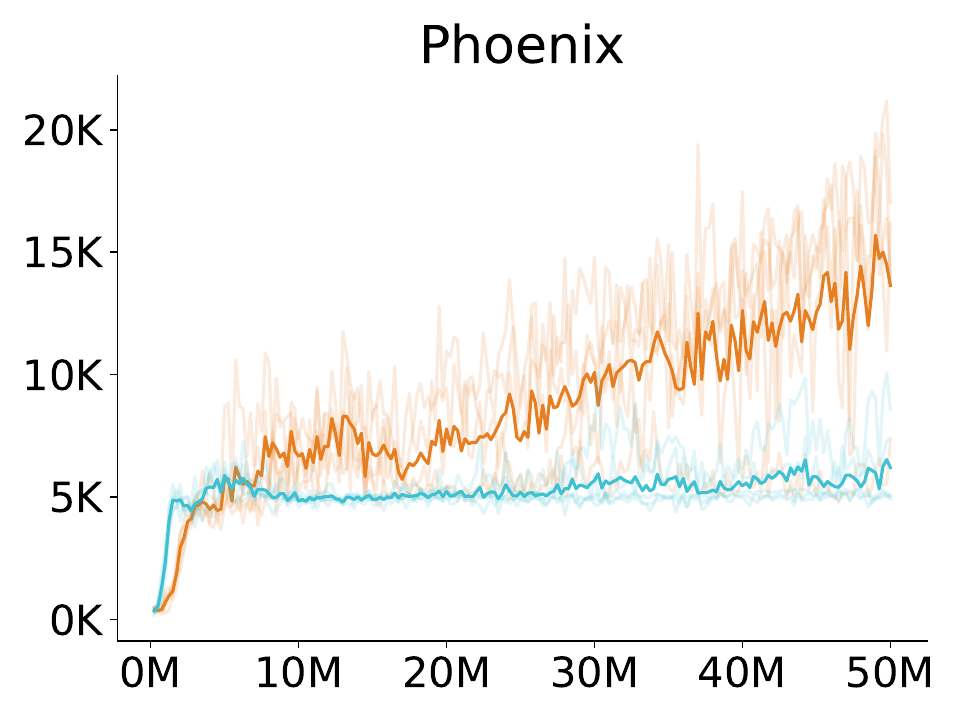} 
	\hspace{0.005\linewidth}
	\includegraphics[width=0.22\linewidth]{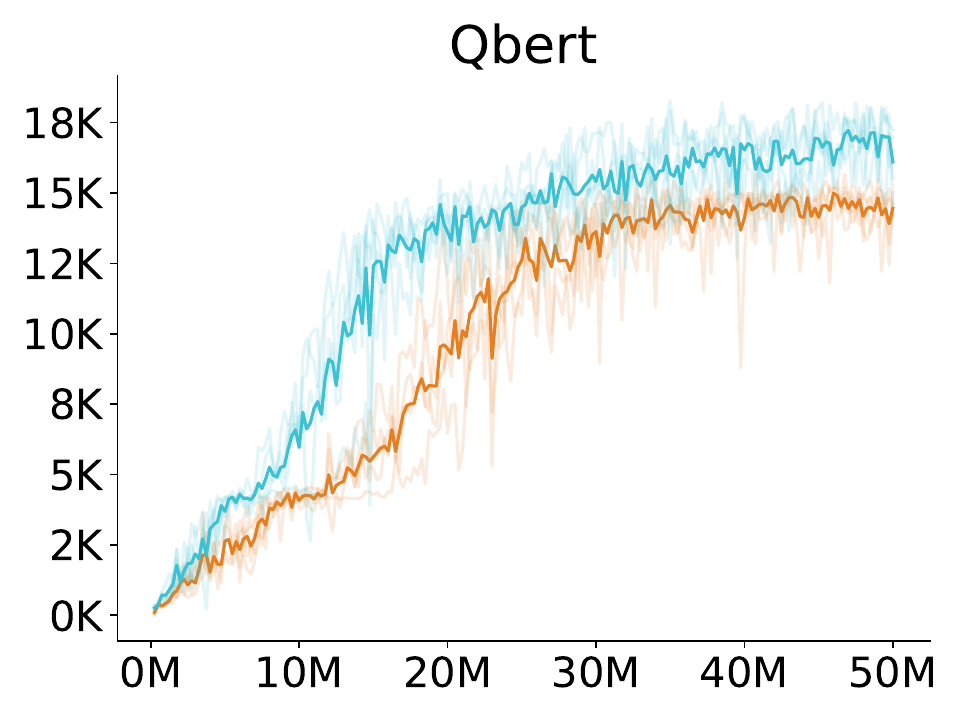} 
	\hspace{0.005\linewidth}
	\includegraphics[width=0.22\linewidth]{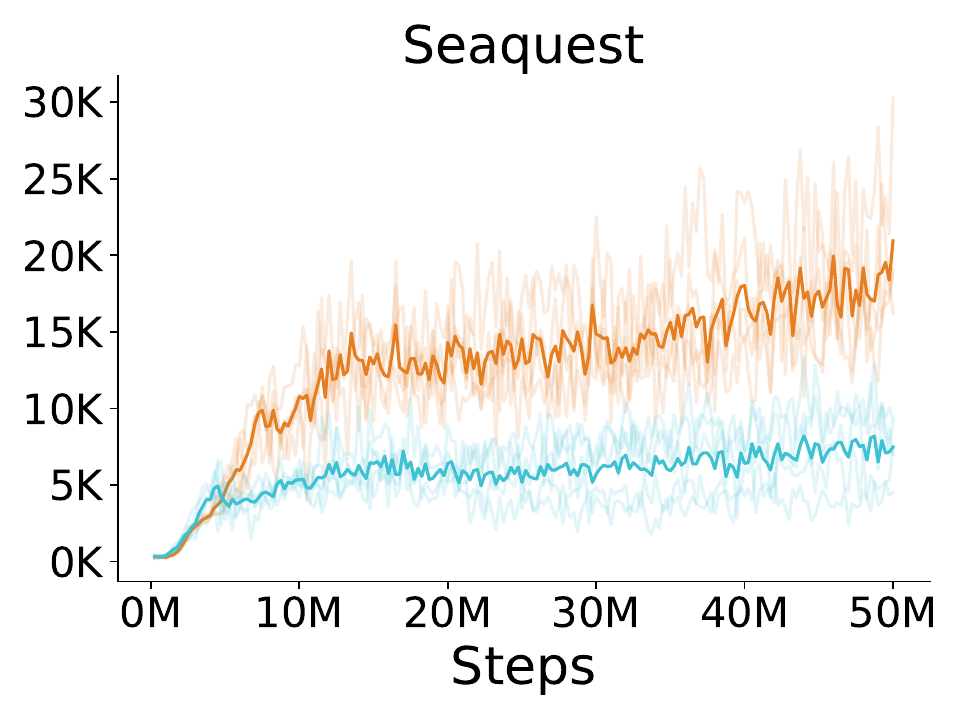} 
	\hspace{0.005\linewidth}
	\includegraphics[width=0.22\linewidth]{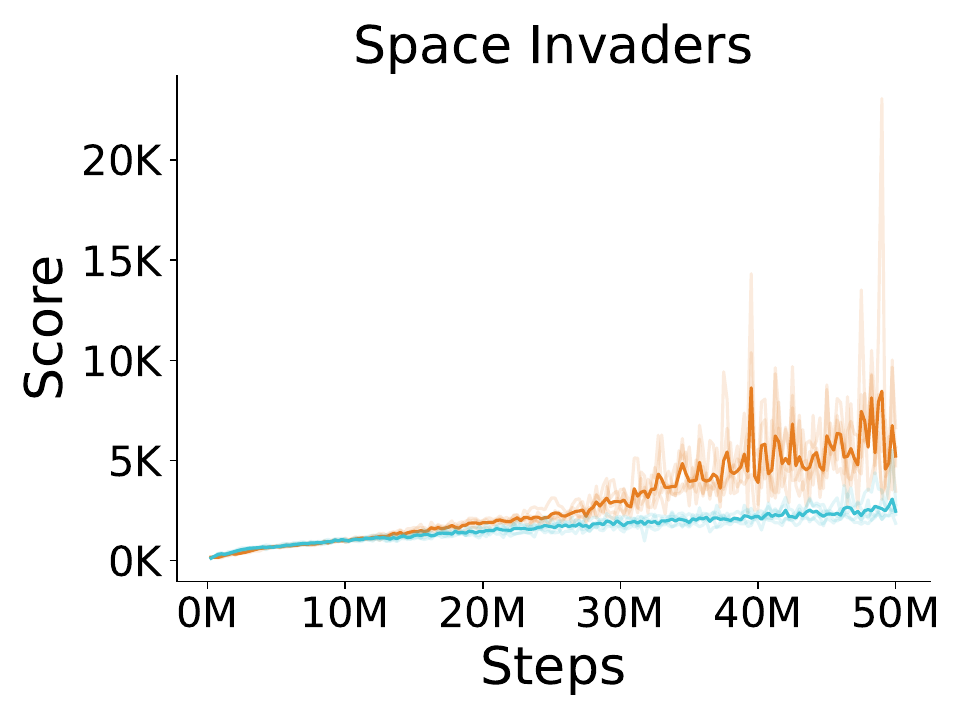} 
	\hspace{0.005\linewidth}
	\includegraphics[width=0.22\linewidth]{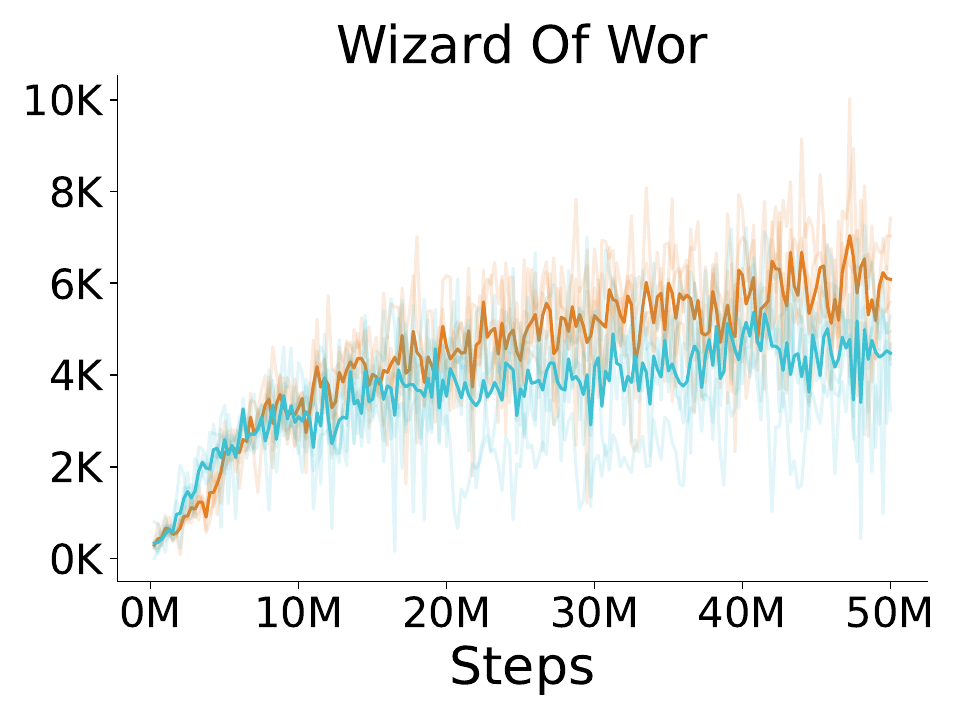} 
	\includegraphics[width=0.22\linewidth]{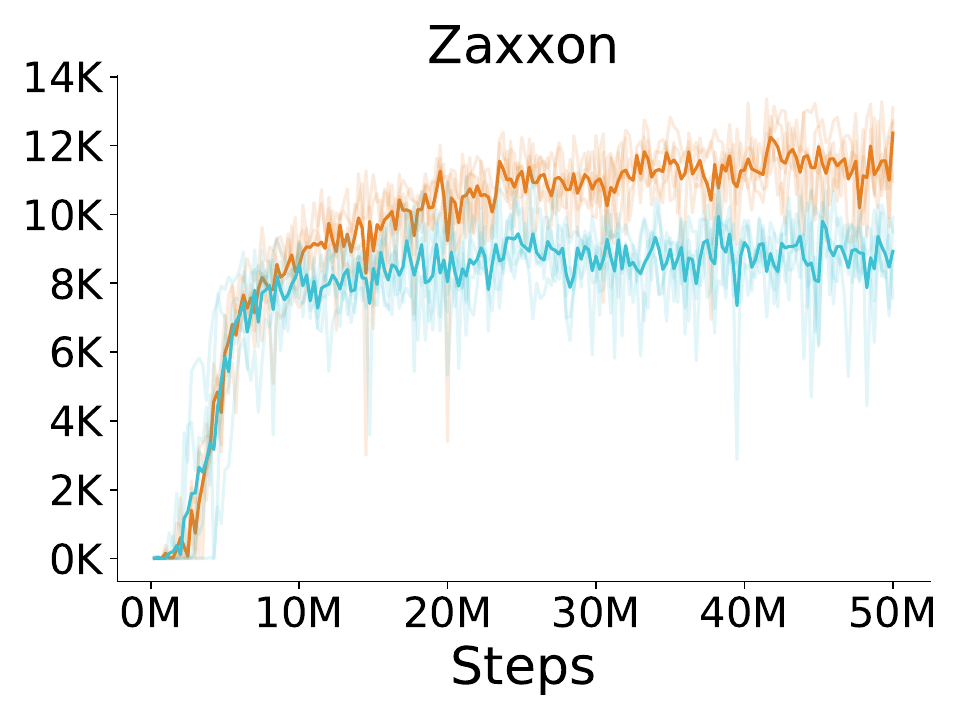} 
	\hspace{0.005\linewidth}
	\hspace{0.02\linewidth}\raisebox{1.8em}{\includegraphics[width=0.2\linewidth]{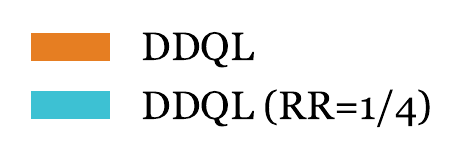}} 
    \caption{Performance of DDQL compared to $\text{DDQL} (\text{RR}=\nicefrac{1}{4})$. They perform similarly in some environments, but $\text{DDQL} (\text{RR}=\nicefrac{1}{4})$ is generally worse.}
    \label{ReplayRatioDH:Ablation11:Score}
\end{figure}

\begin{figure}[h]
        \centering
    	\includegraphics[width=0.22\linewidth]{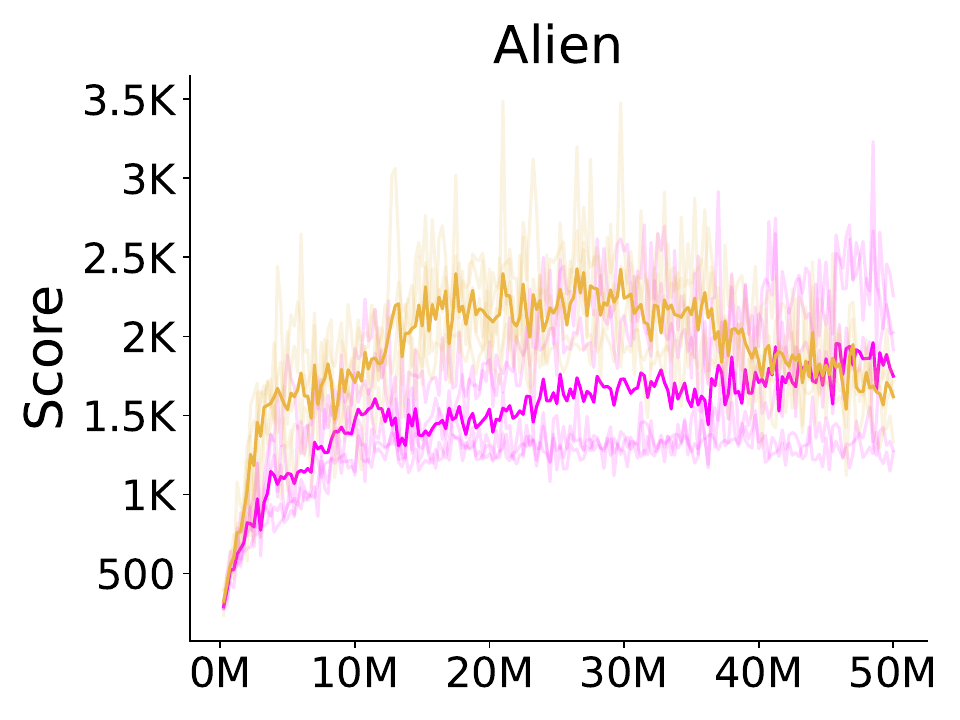} 
	\hspace{0.005\linewidth}
	\includegraphics[width=0.22\linewidth]{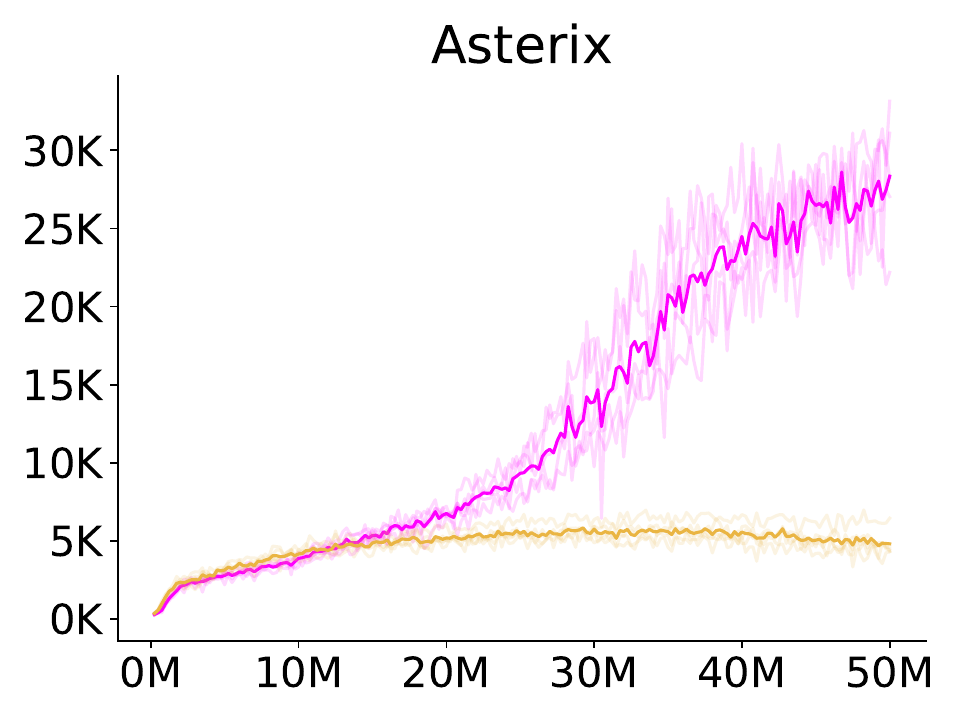} 
	\hspace{0.005\linewidth}
	\includegraphics[width=0.22\linewidth]{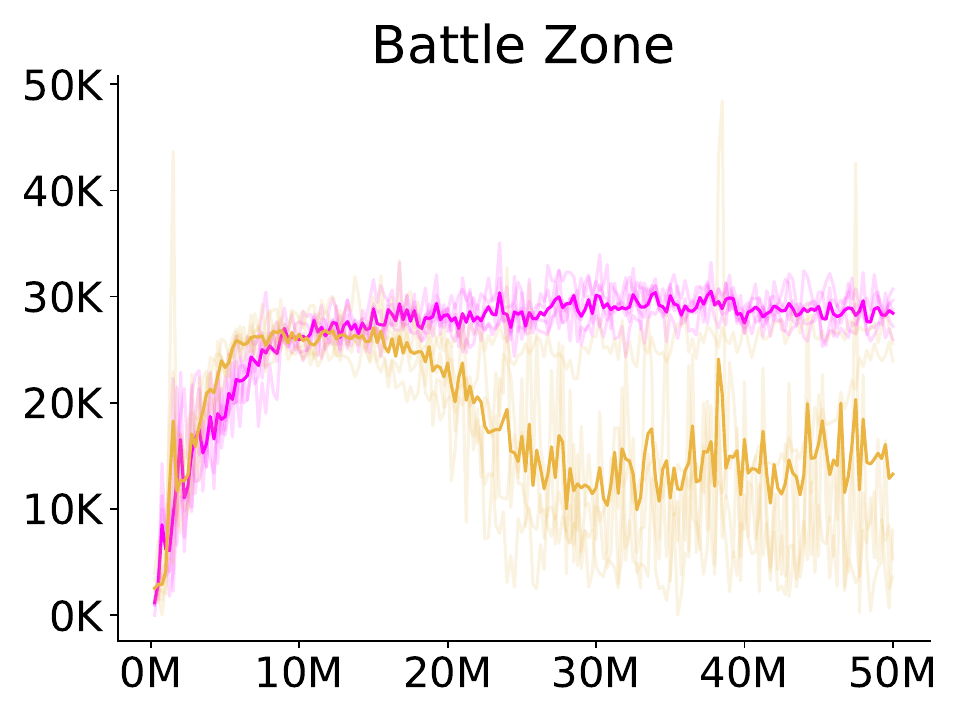} 
	\hspace{0.005\linewidth}
	\includegraphics[width=0.22\linewidth]{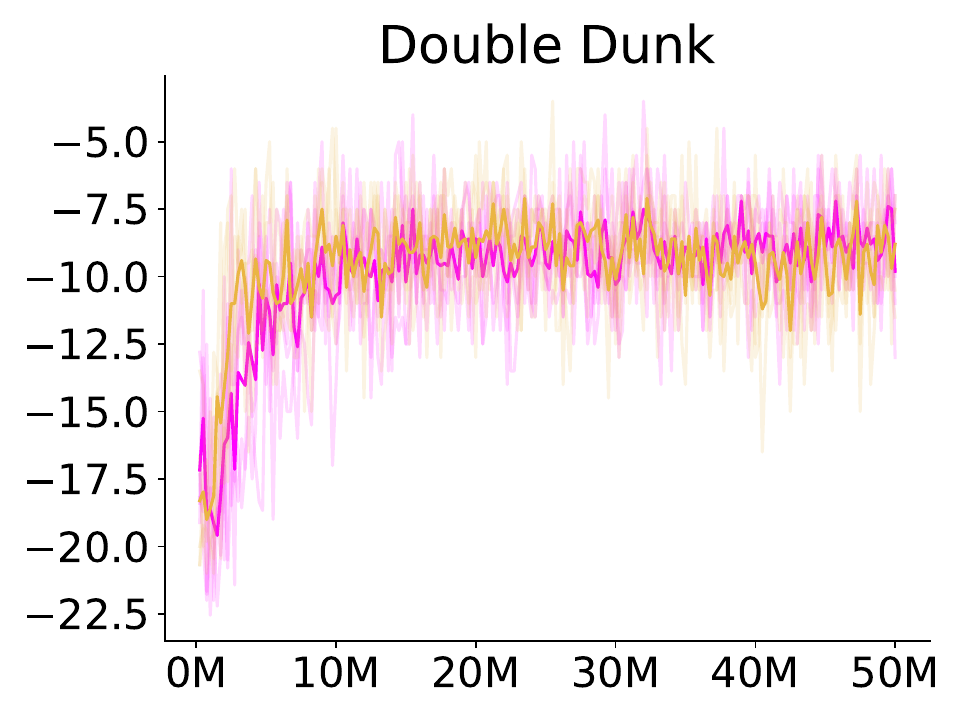} 
	\hspace{0.005\linewidth}
	\includegraphics[width=0.22\linewidth]{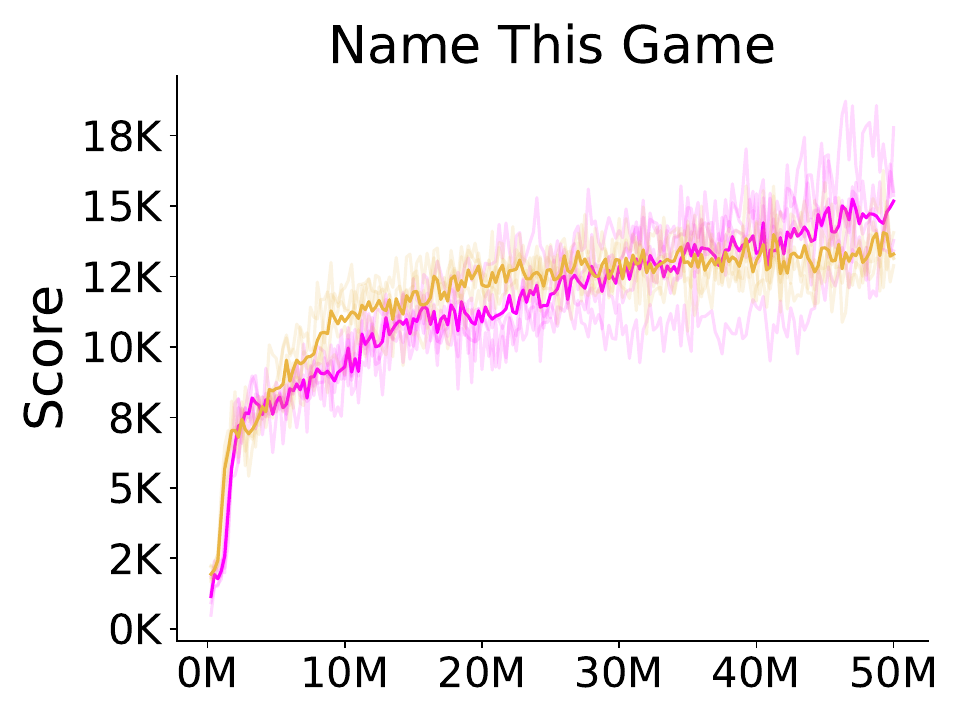} 
	\includegraphics[width=0.22\linewidth]{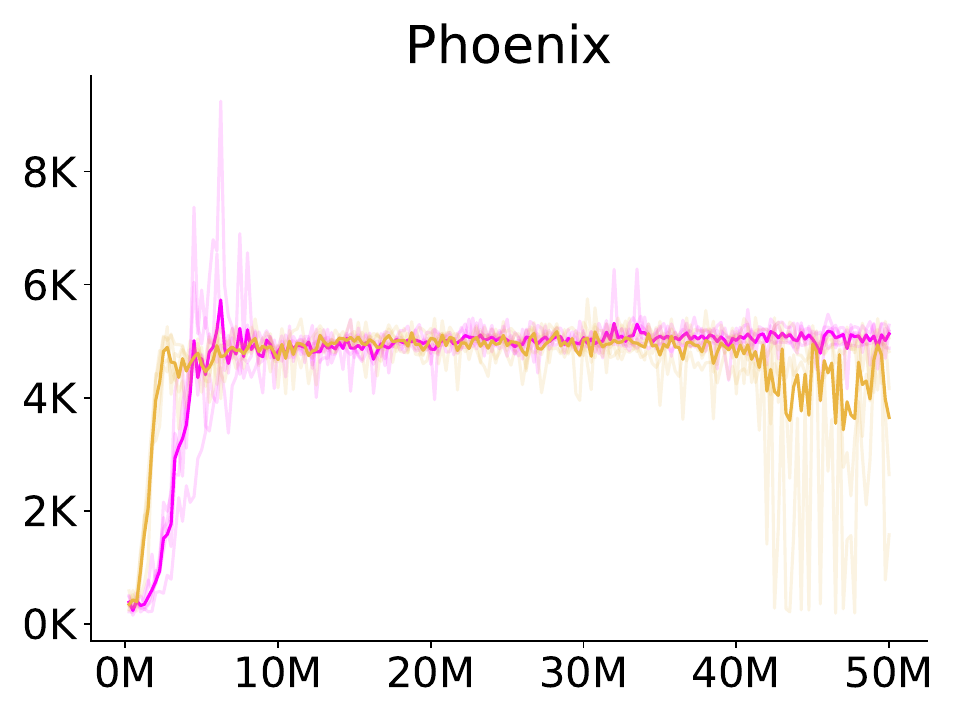} 
	\hspace{0.005\linewidth}
	\includegraphics[width=0.22\linewidth]{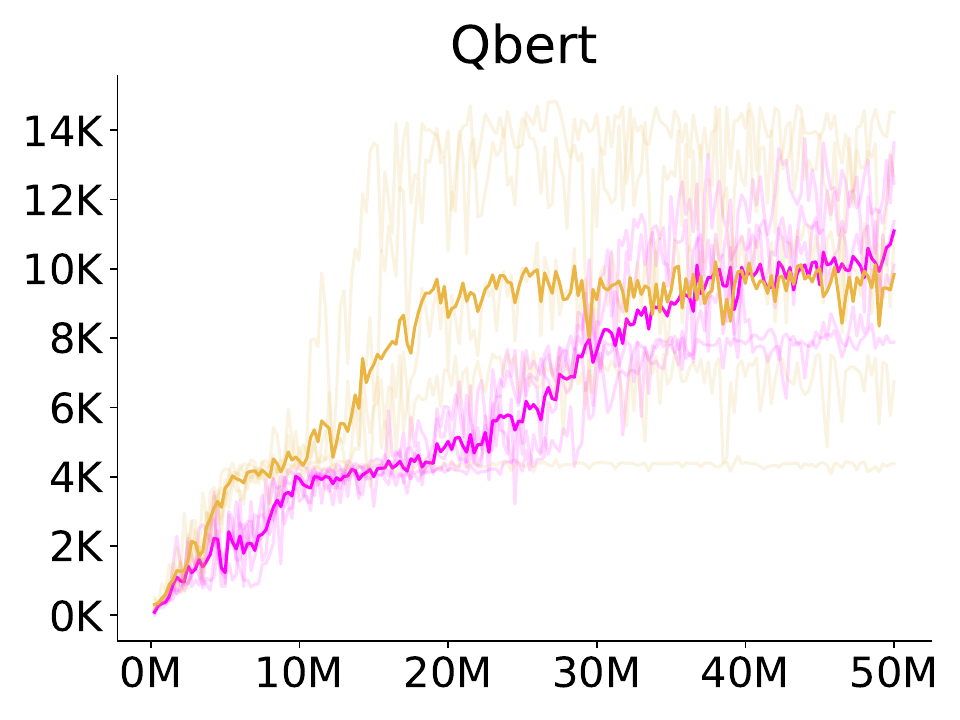} 
	\hspace{0.005\linewidth}
	\includegraphics[width=0.22\linewidth]{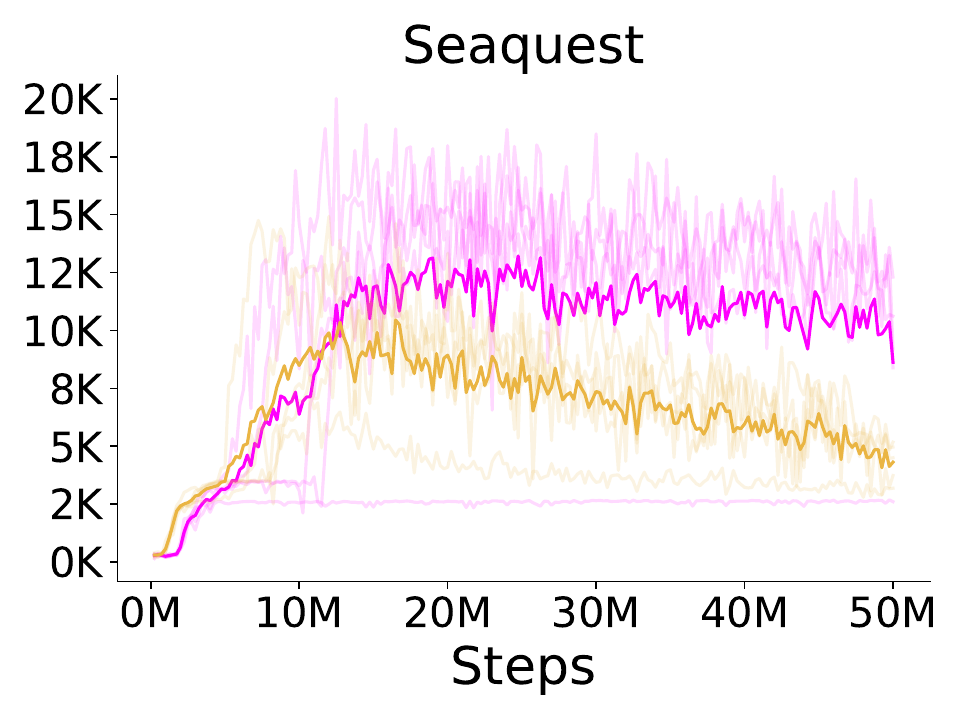} 
	\hspace{0.005\linewidth}
	\includegraphics[width=0.22\linewidth]{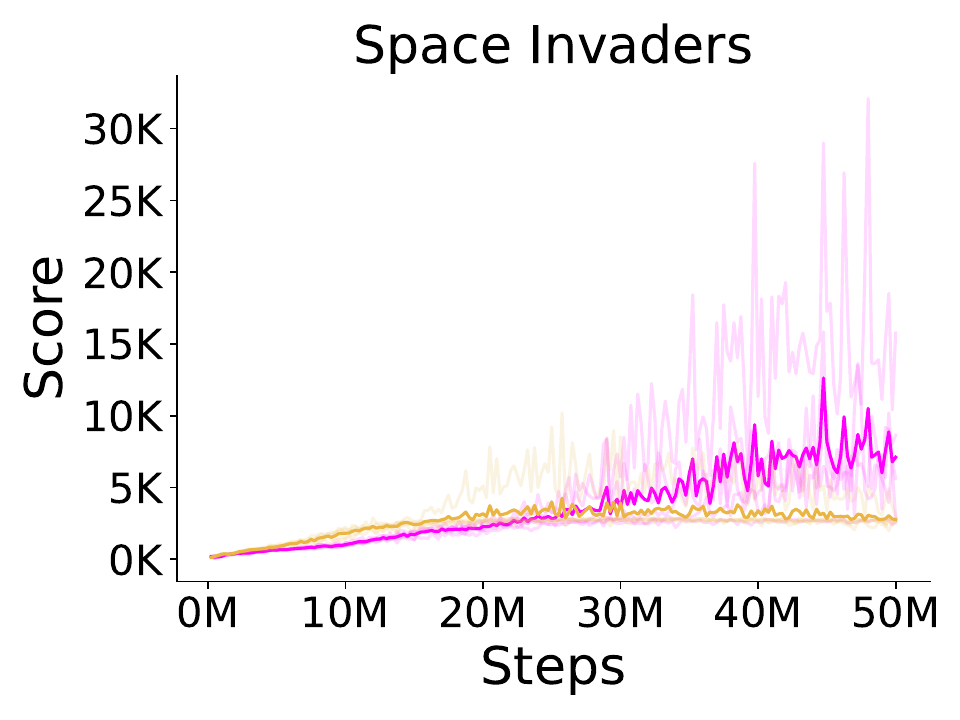} 
	\hspace{0.005\linewidth}
	\includegraphics[width=0.22\linewidth]{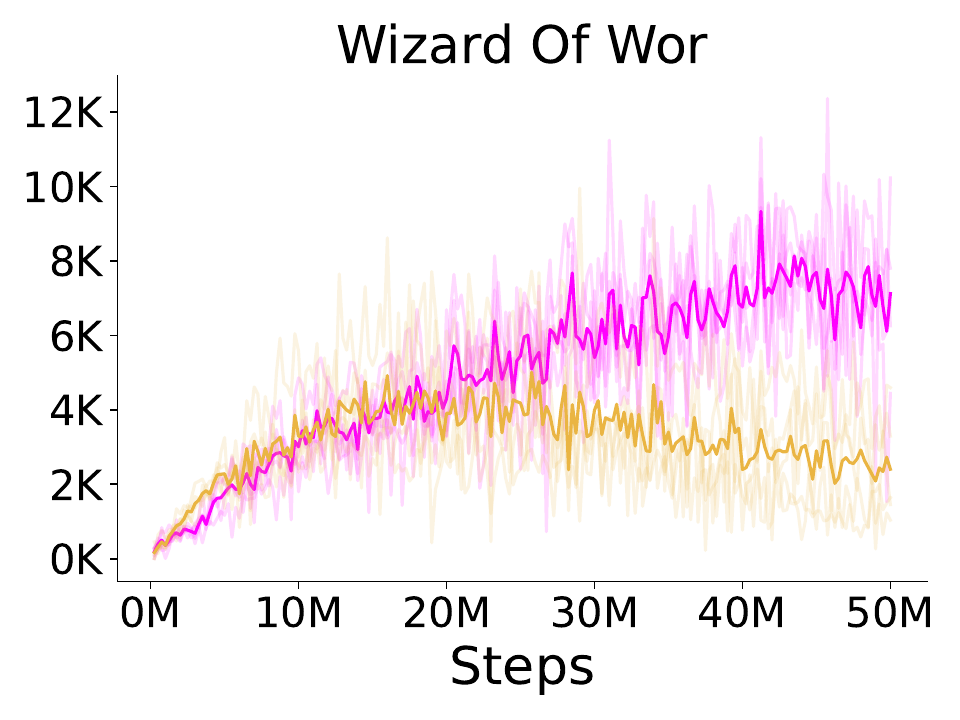} 
	\includegraphics[width=0.22\linewidth]{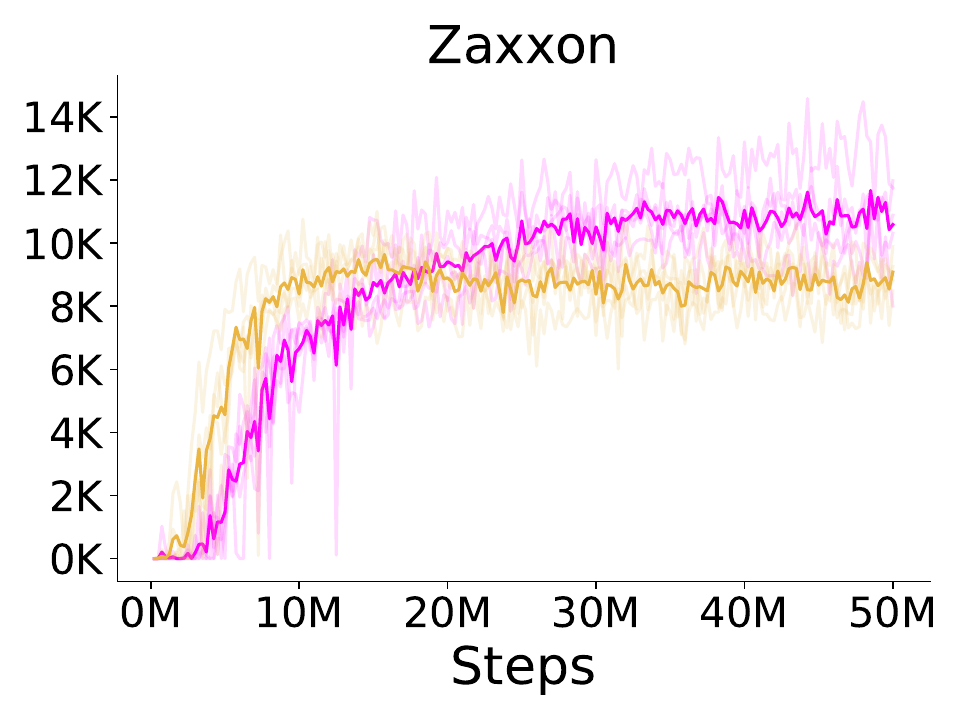} 
	\hspace{0.005\linewidth}
	\hspace{0.01\linewidth}\raisebox{2em}{\includegraphics[width=0.2\linewidth]{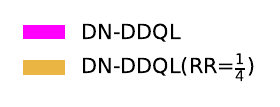}} 
	\caption{Performance of DN-DDQL compared to $\text{DN-DDQL} (\text{RR}=\nicefrac{1}{4})$. $\text{DN-DDQL} (\text{RR}=\nicefrac{1}{4})$ is generally worse.}
	\label{ReplayRatioDN:Ablation11:Score}
\end{figure}

\begin{figure}[h!]
        \centering
    	\includegraphics[width=0.22\linewidth]{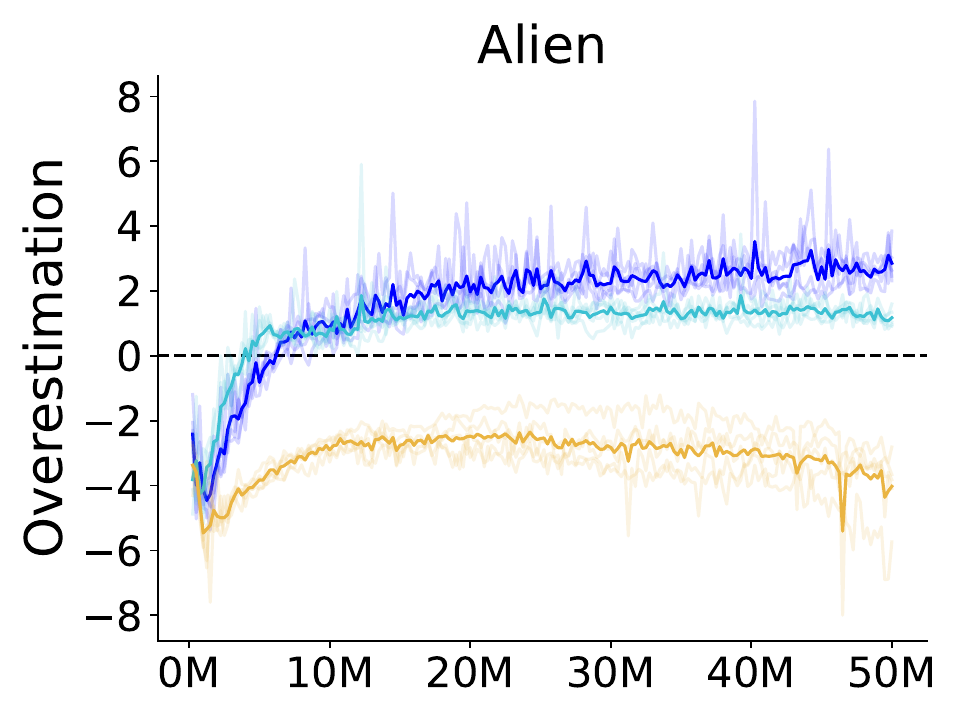} 
	\hspace{0.005\linewidth}
	\includegraphics[width=0.22\linewidth]{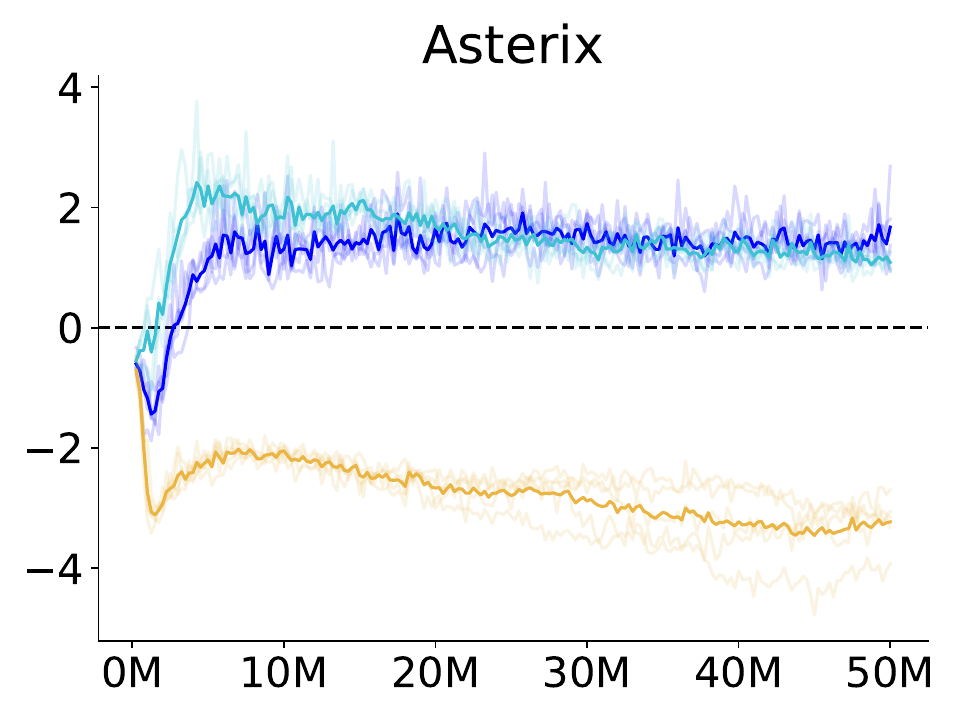} 
	\hspace{0.005\linewidth}
	\includegraphics[width=0.22\linewidth]{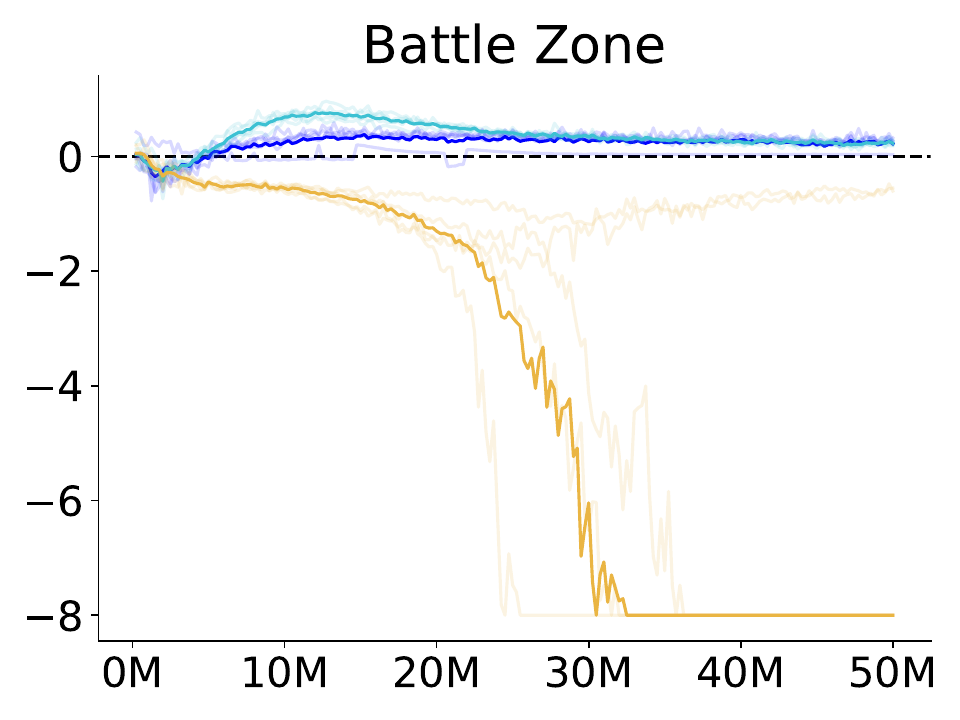} 
	\hspace{0.005\linewidth}
	\includegraphics[width=0.22\linewidth]{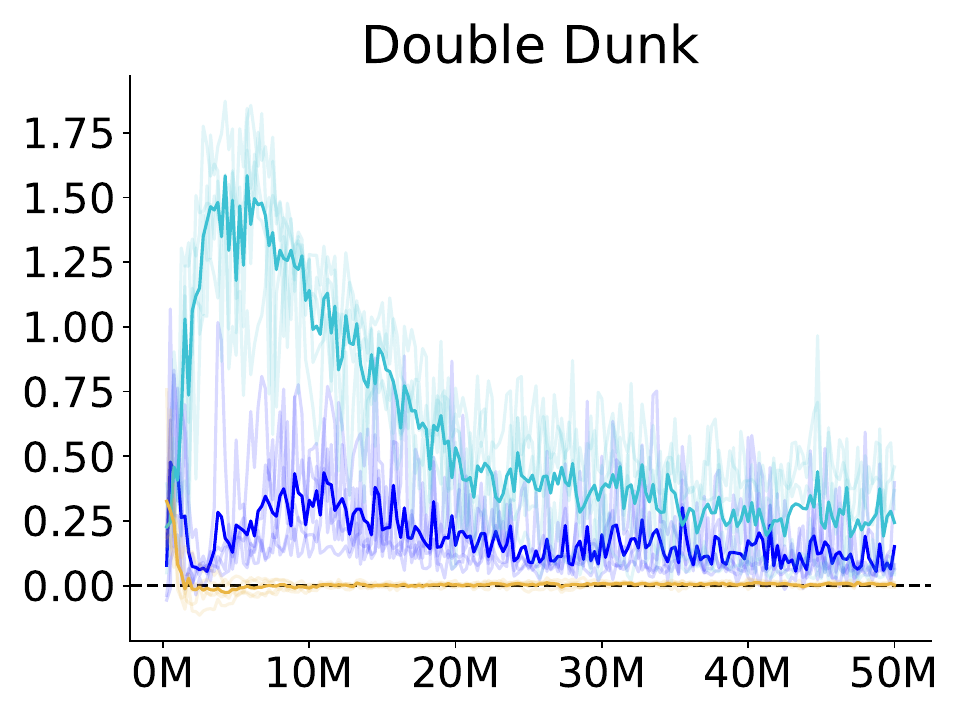} 
	\hspace{0.005\linewidth}
	\includegraphics[width=0.22\linewidth]{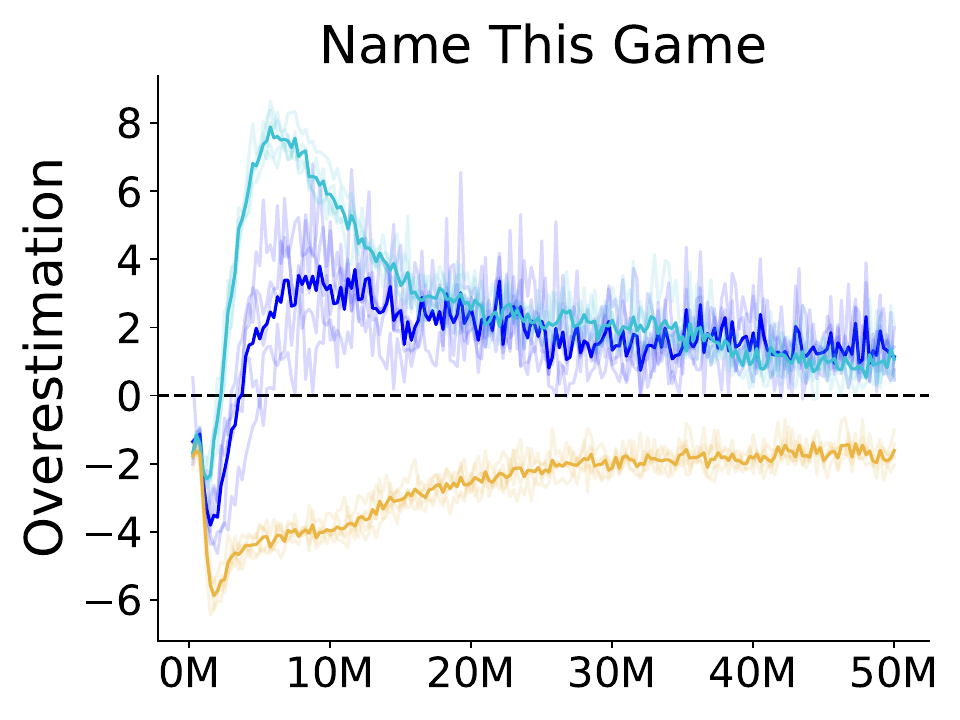} 
	\includegraphics[width=0.22\linewidth]{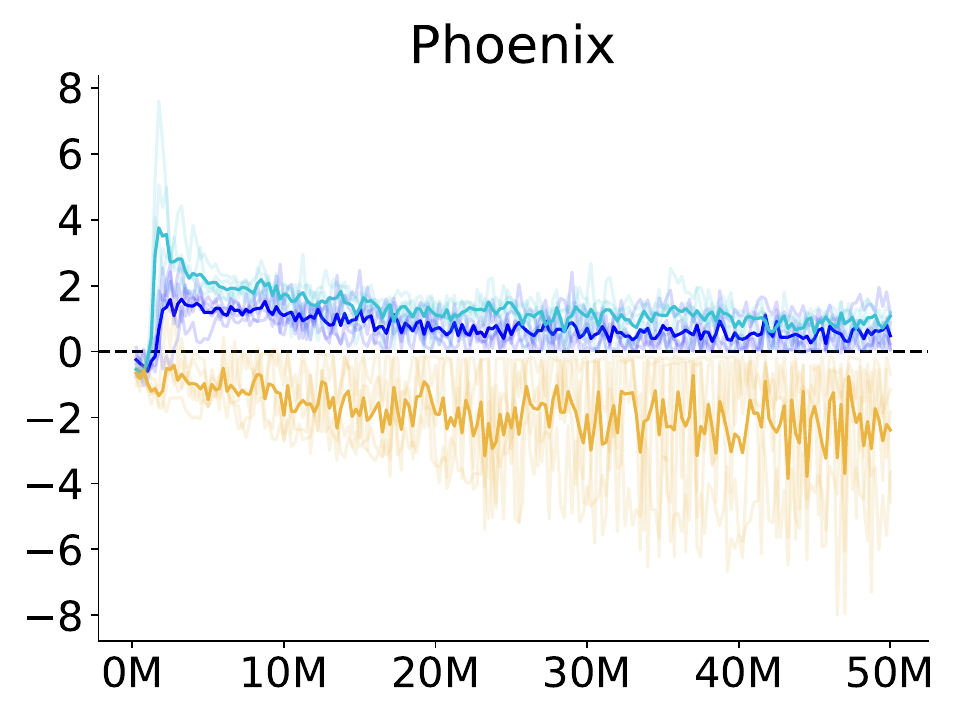} 
	\hspace{0.005\linewidth}
	\includegraphics[width=0.22\linewidth]{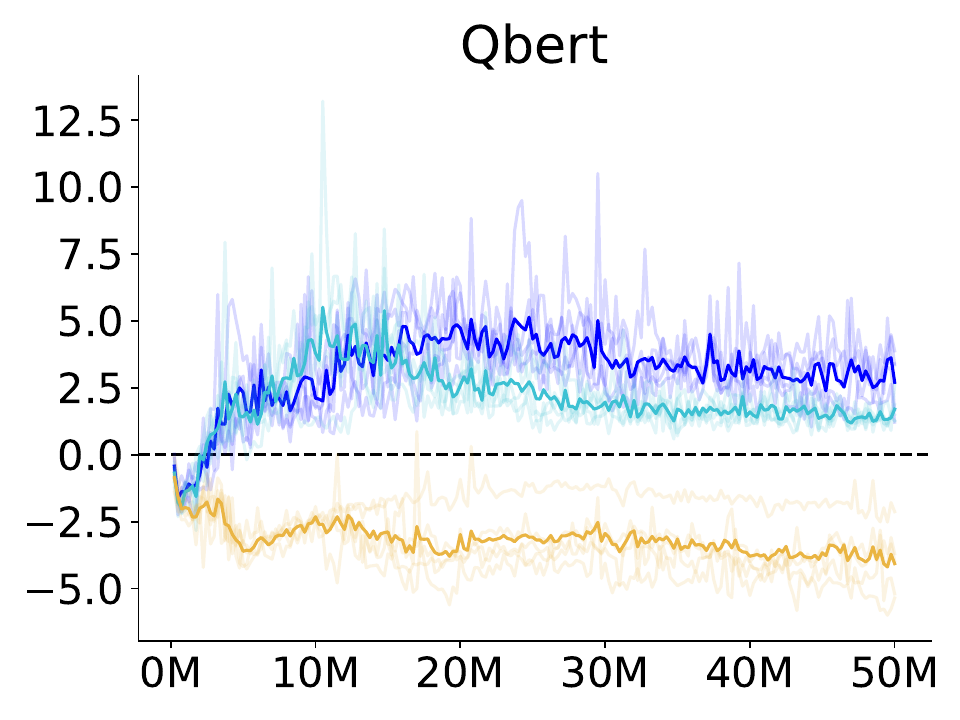} 
	\hspace{0.005\linewidth}
	\includegraphics[width=0.22\linewidth]{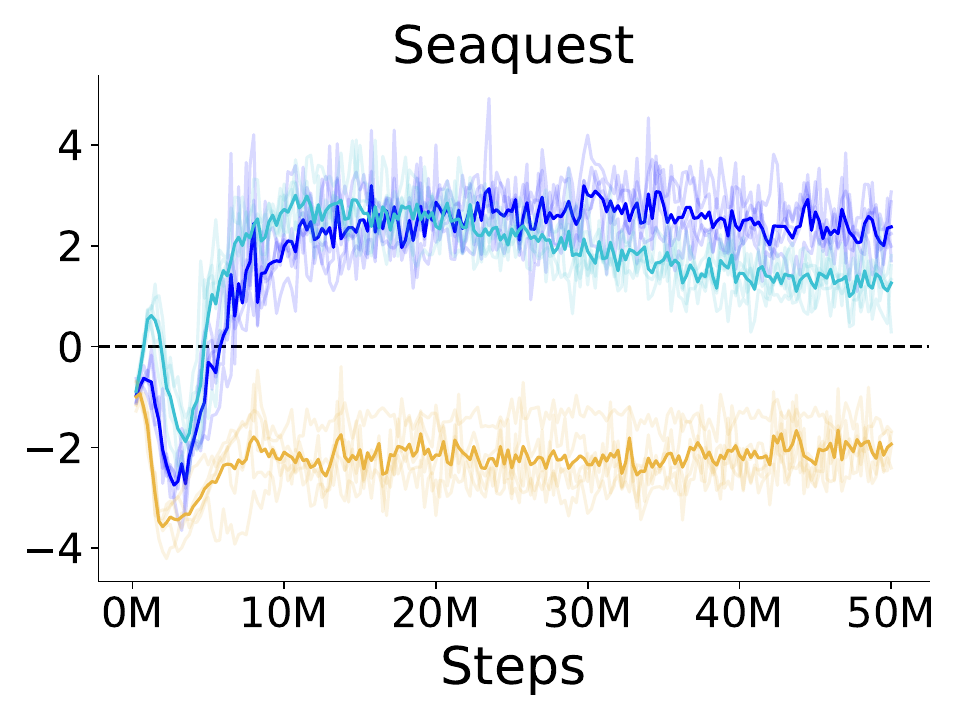} 
	\hspace{0.005\linewidth}
	\includegraphics[width=0.22\linewidth]{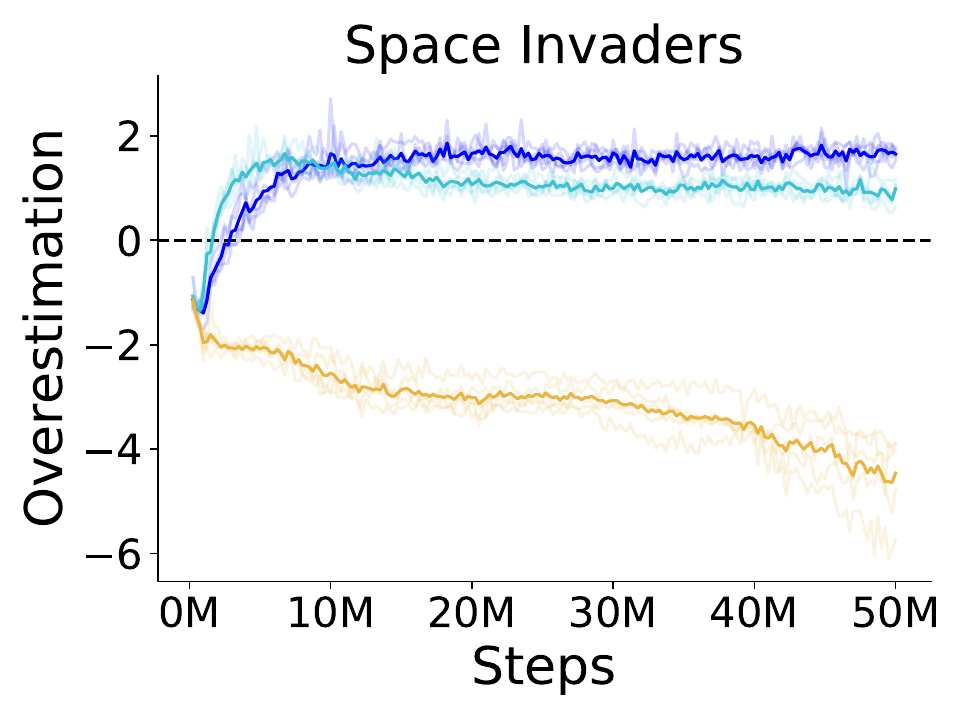} 
	\hspace{0.005\linewidth}
	\includegraphics[width=0.22\linewidth]{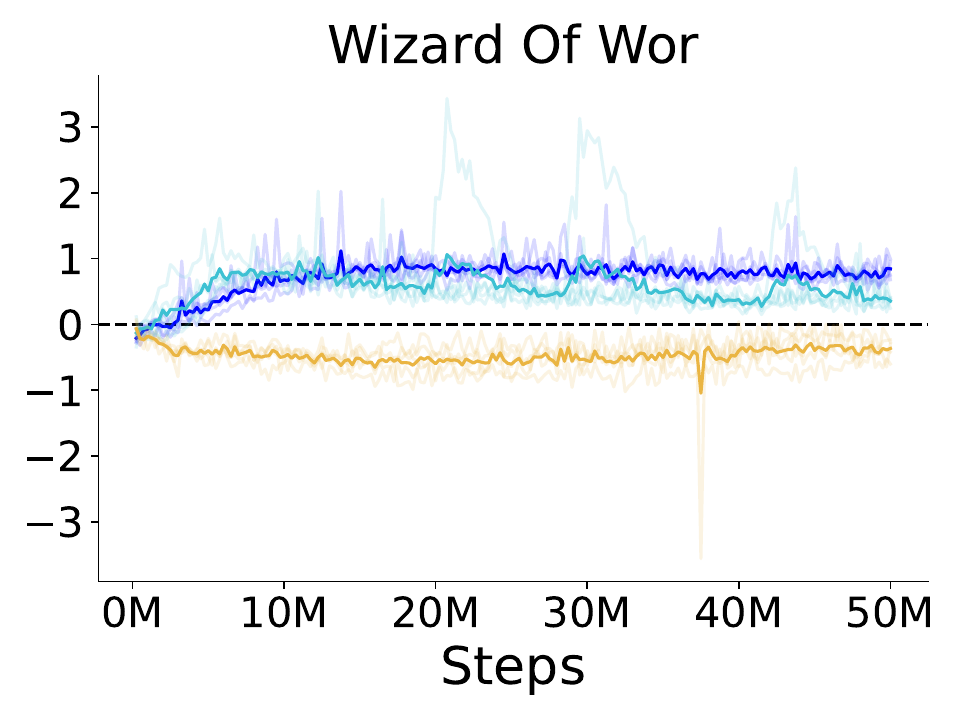} 
	\includegraphics[width=0.22\linewidth]{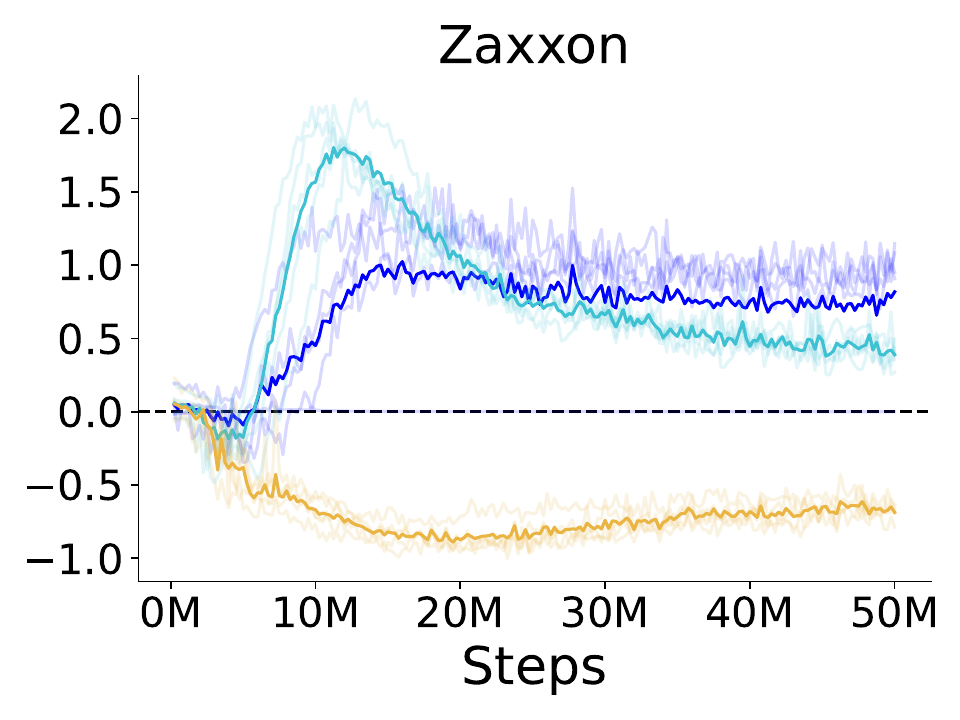} 
	\hspace{0.005\linewidth}
	\hspace{0.01\linewidth}\raisebox{3mm}[0pt][0pt]{\includegraphics[width=0.2\linewidth]{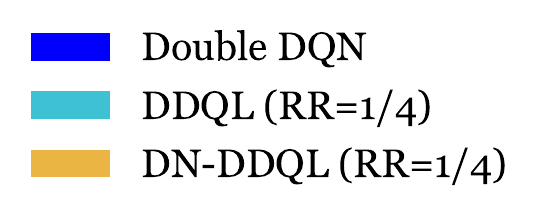}} 
	\caption{Overestimation of $\text{DDQL} (\text{RR}=\nicefrac{1}{4})$, $\text{DN-DDQL} (\text{RR}=\nicefrac{1}{4})$, and Double DQN.
    Overestimation is clipped at $-8$ due to divergence in \textsc{BattleZone}. The DDQL variants continue to reduce overestimation even with double the replay ratio.}
	\label{ReplayRatioDDQL:Ablation11:Overestimation}
\end{figure}

This subsection contains the full curves corresponding to the ablations in Section~\ref{understanding_ddql}.
Figure~\ref{DatasetPartitioningDH:Ablation11:Score} shows the performance curves during training for DDQL's dataset partitioning experiments.
Figure~\ref{DatasetPartitioningDN:Ablation11:Score} shows the performance curves during training for DN-DDQL's dataset partitioning experiments.
Figure~\ref{DatasetPartitioningDH:Ablation11:Overestimation} shows the overestimation curves during training for DDQL's dataset partitioning experiments.
Figure~\ref{ReplayRatioDH:Ablation11:Score} shows the performance curves during training for DDQL's replay ratio experiments.
Figure~\ref{ReplayRatioDN:Ablation11:Score} shows the performance curves during training for DN-DDQL's replay ratio experiments.
Figure~\ref{ReplayRatioDDQL:Ablation11:Overestimation} shows the overestimation curves during training for DDQL's replay ratio experiments. 

\FloatBarrier

\subsection{Ablation: Double DQN with a lower replay ratio}

\begin{figure}[h]
    \begin{center}
    \includegraphics[width=0.5\textwidth, valign=m]{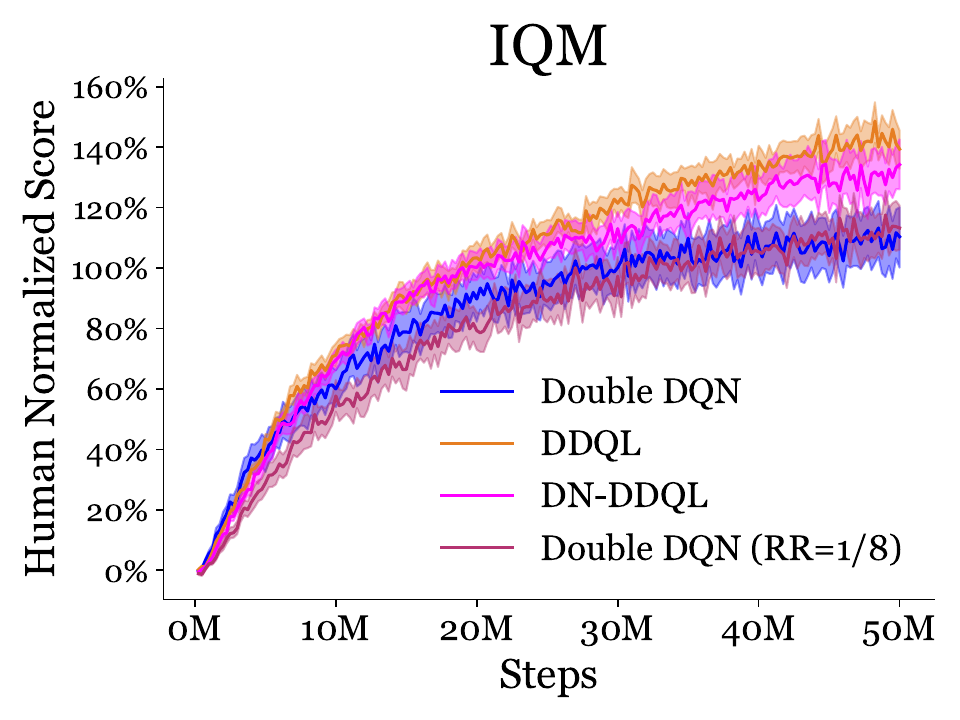}
    \end{center}
    \caption{
         Interquartile mean of the HNS throughout training on 57 Atari environments. The shaded region represents a 95\% stratified bootstrap confidence interval. $\text{Double DQN} (\text{RR}=\nicefrac{1}{8})$ is run for three seeds per game and the  other algorithms are run for five seeds per game.
    }
    \label{fig:half_rr_perf}
\end{figure}
To understand whether the performance improvements of DDQL come from using a lower replay ratio, we ran Double DQN for three seeds on the 57 games with a replay ratio of $\nicefrac{1}{8}$.
\ShortFigref{fig:half_rr_perf} shows that Double DQN using a replay ratio of $\nicefrac{1}{8}$ performs similarly to Double DQN.
Importantly, both variants of Double DQN perform worse than DDQL, indicating that it is not solely DDQL's lower replay ratio that causes improved performance.

\subsection{Atari-57 overestimation results}

\begin{figure}[p]
        \centering
    	\includegraphics[width=0.22\linewidth]{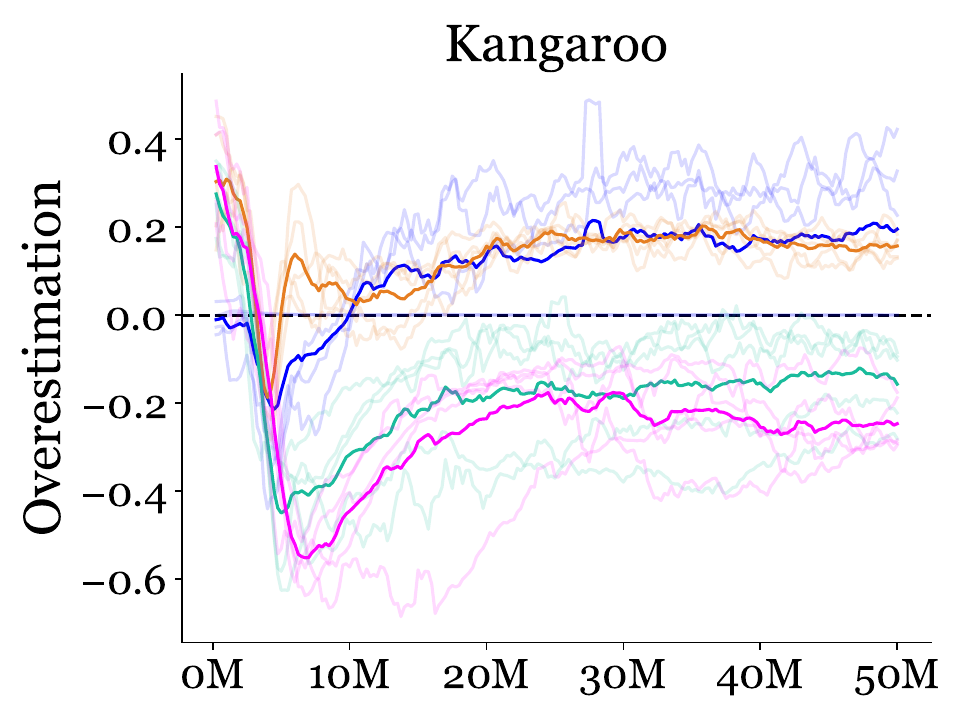} 
	\includegraphics[width=0.22\linewidth]{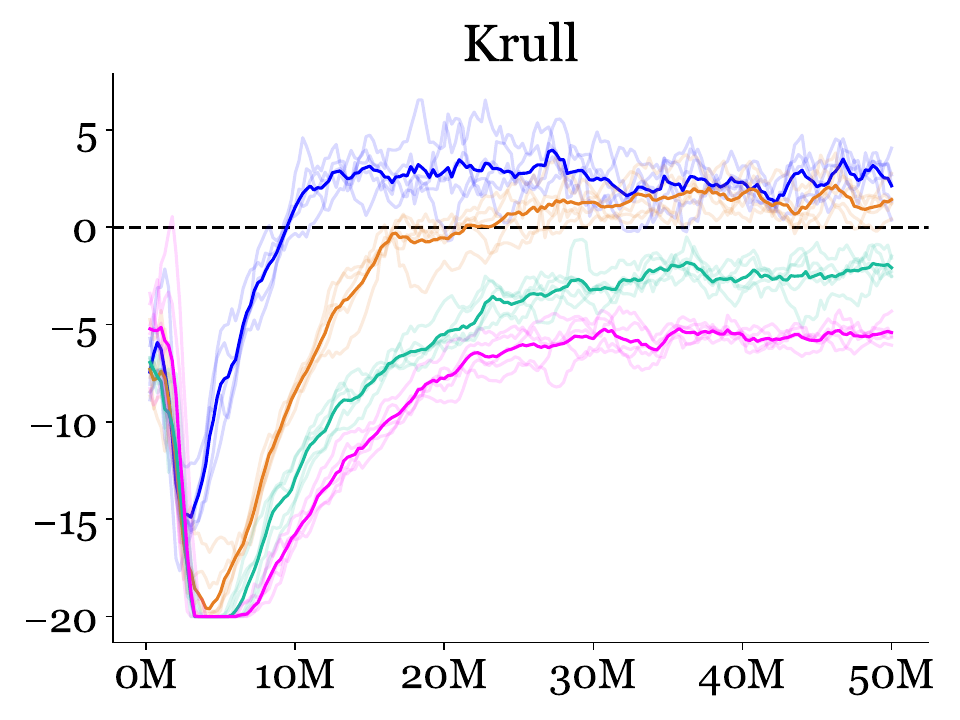} 
	\includegraphics[width=0.22\linewidth]{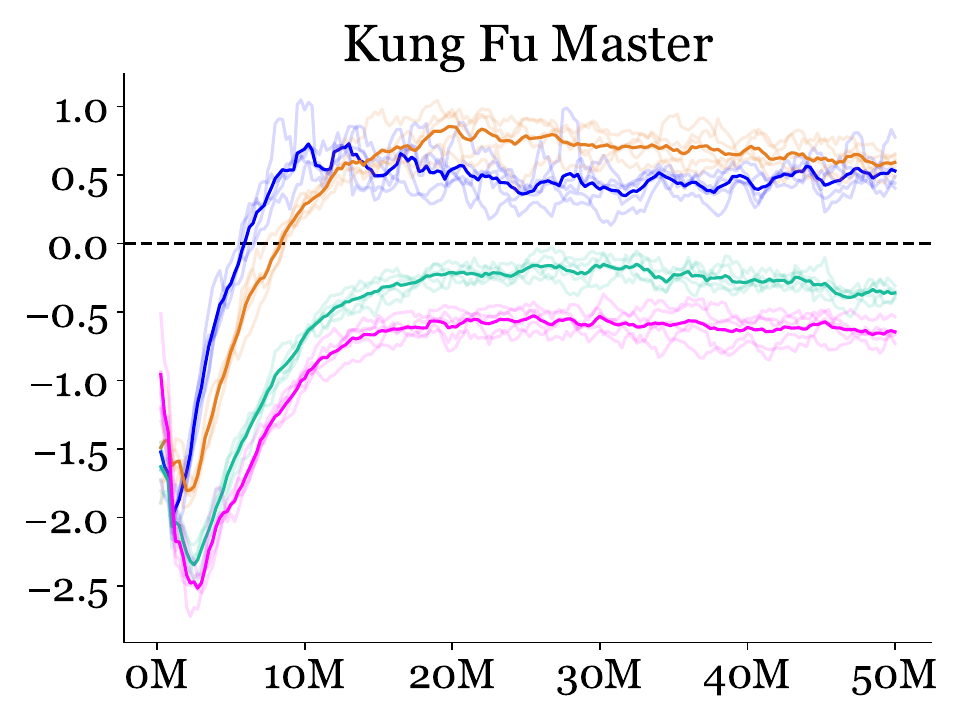} 
	\includegraphics[width=0.22\linewidth]{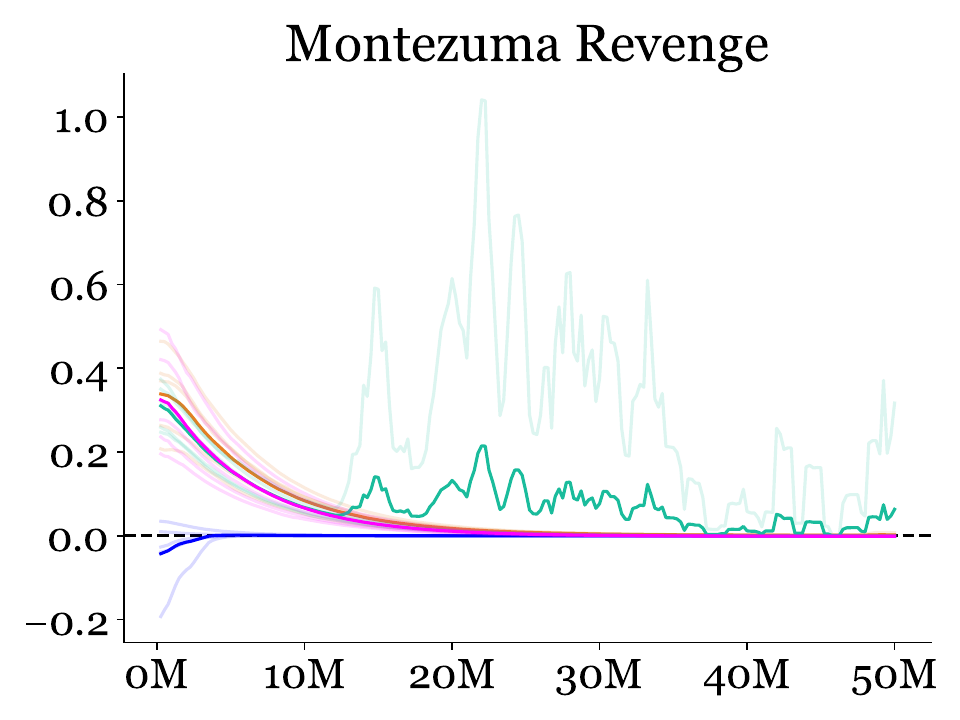} 
	\includegraphics[width=0.22\linewidth]{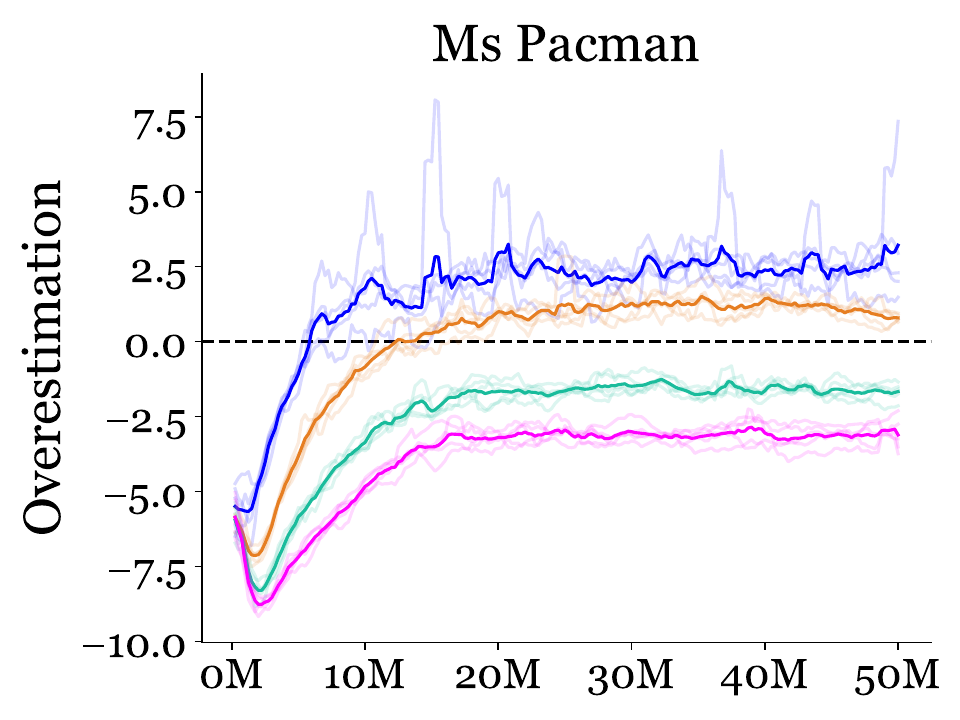} 
	\includegraphics[width=0.22\linewidth]{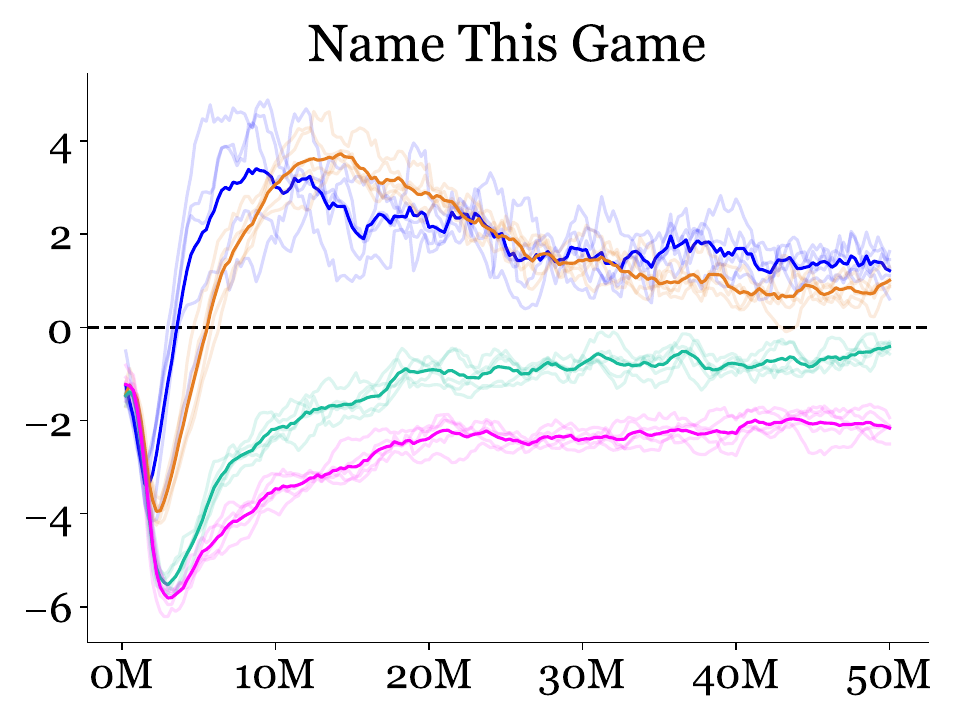} 
	\includegraphics[width=0.22\linewidth]{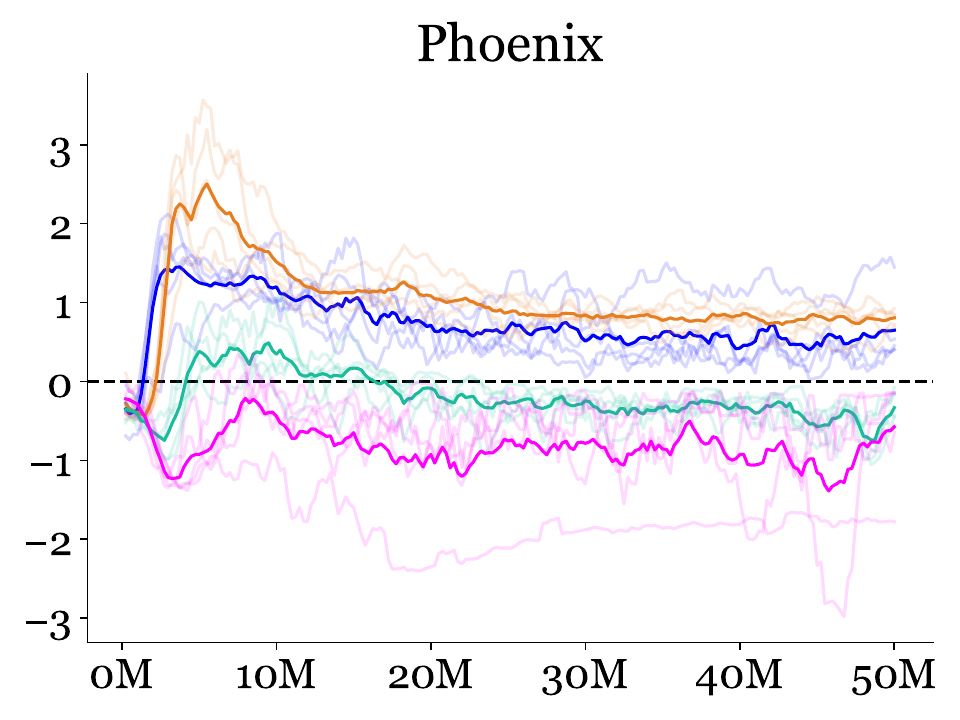} 
	\includegraphics[width=0.22\linewidth]{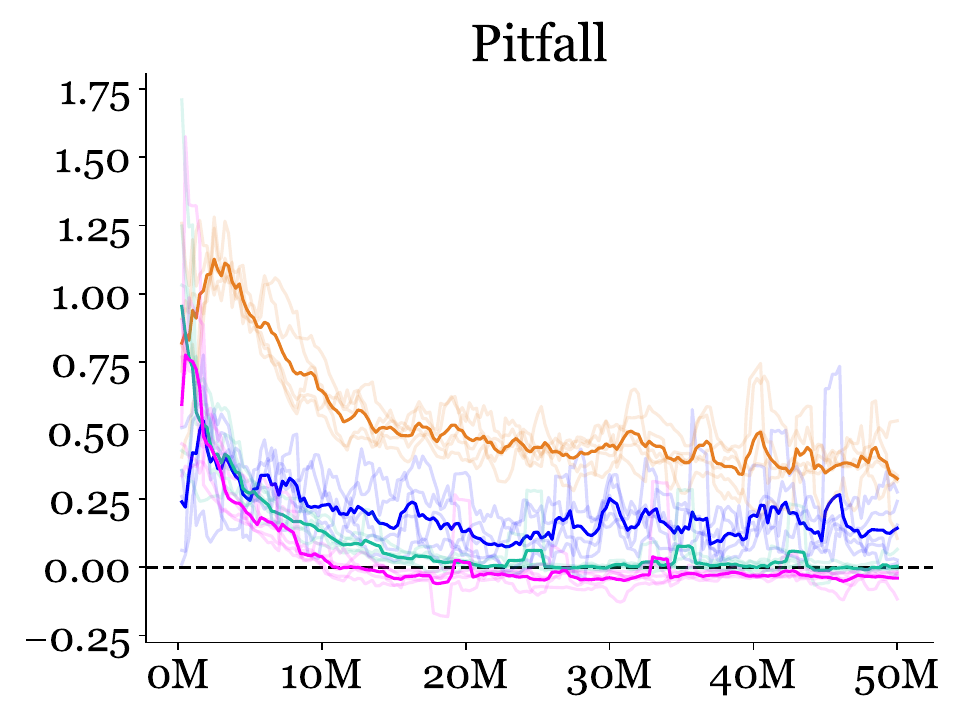} 
	\includegraphics[width=0.22\linewidth]{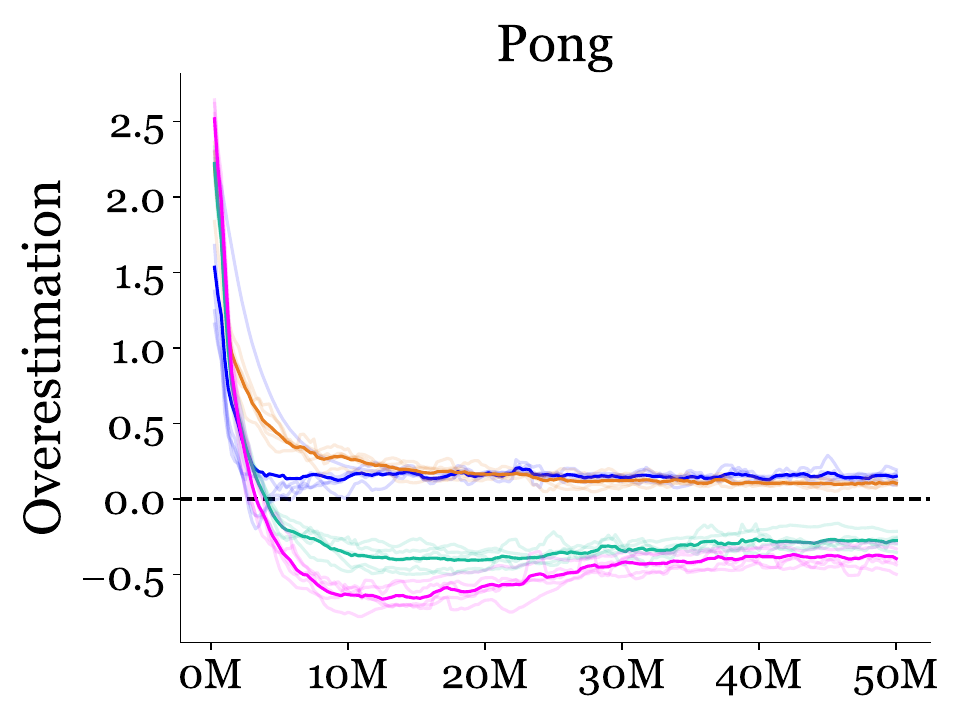} 
	\includegraphics[width=0.22\linewidth]{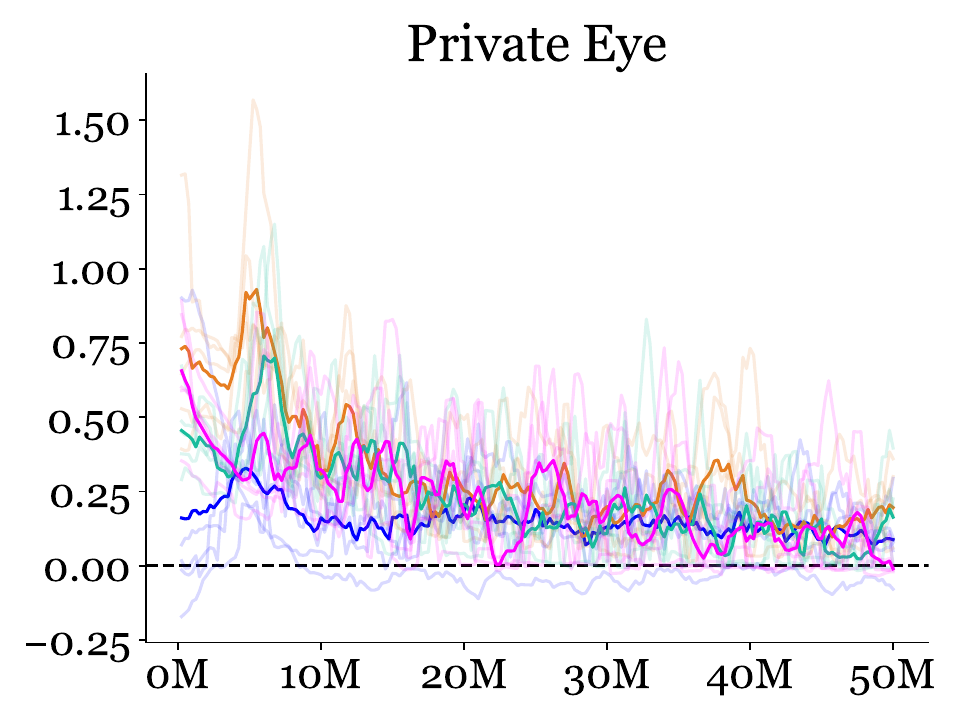} 
	\includegraphics[width=0.22\linewidth]{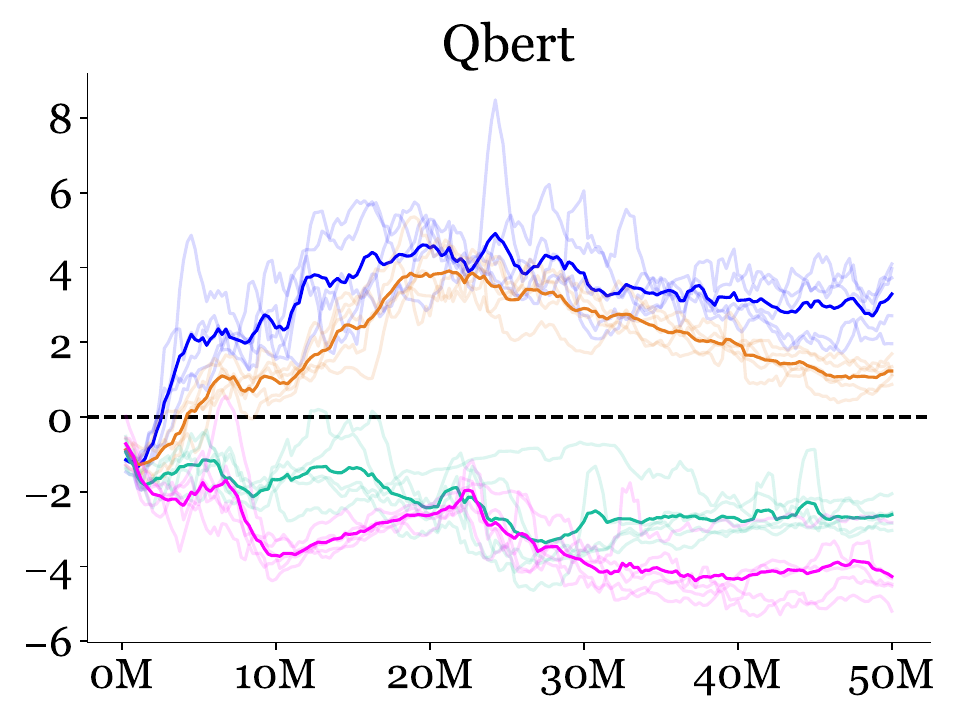} 
	\includegraphics[width=0.22\linewidth]{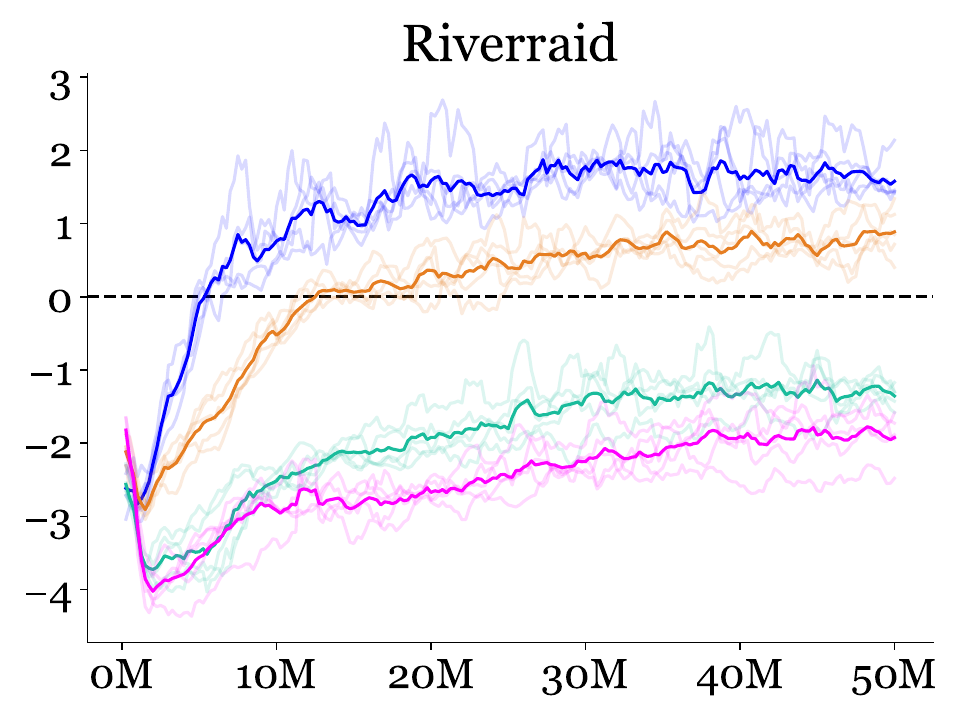} 
	\includegraphics[width=0.22\linewidth]{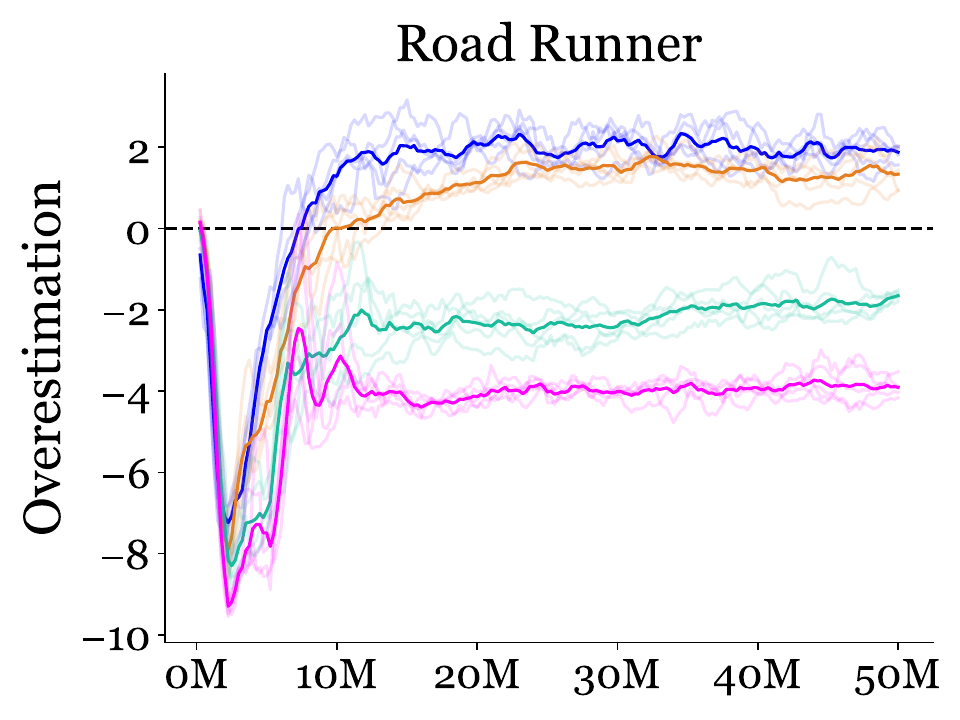} 
	\includegraphics[width=0.22\linewidth]{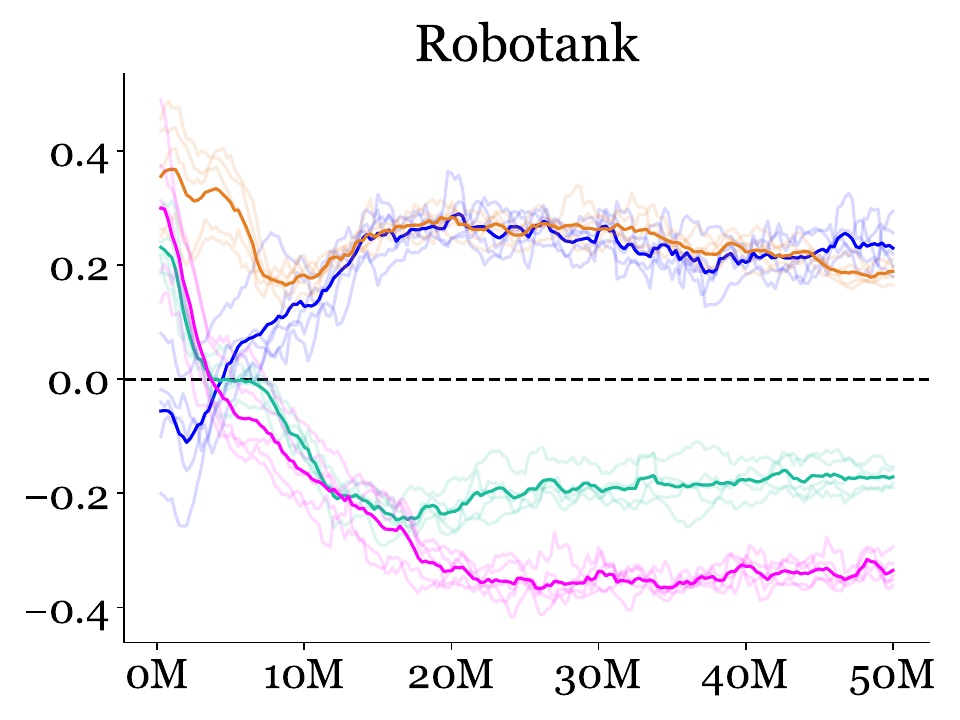} 
	\includegraphics[width=0.22\linewidth]{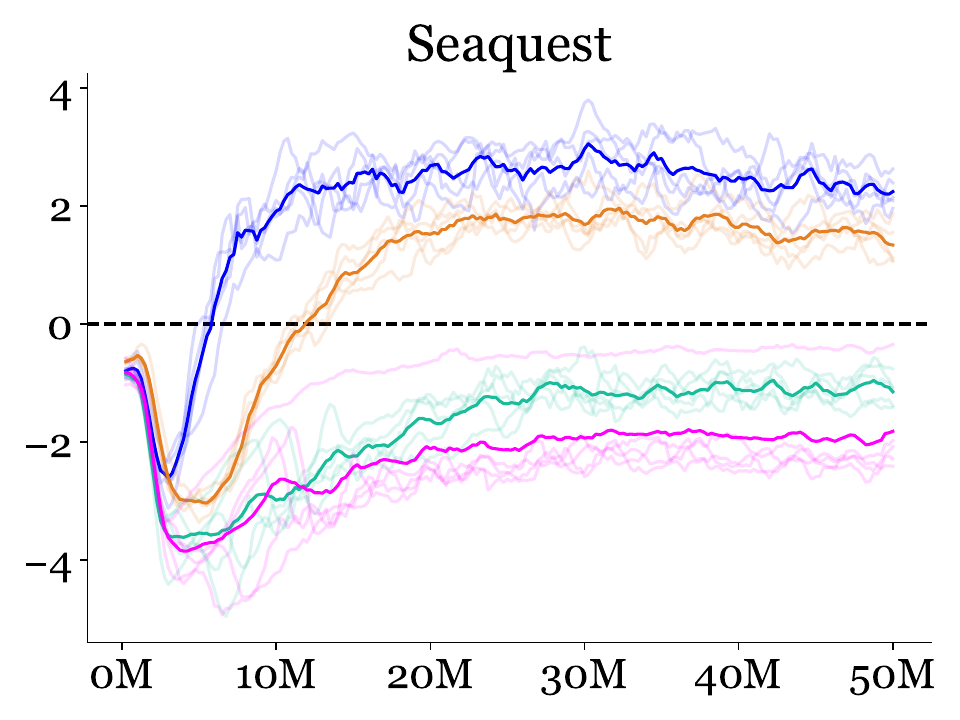} 
	\includegraphics[width=0.22\linewidth]{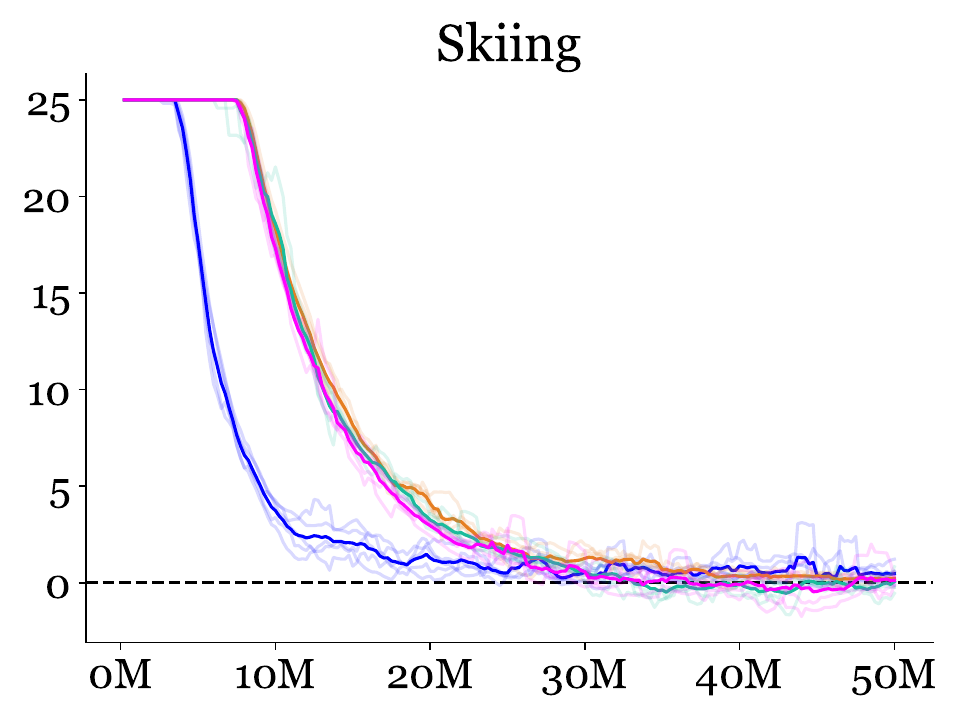} 
	\includegraphics[width=0.22\linewidth]{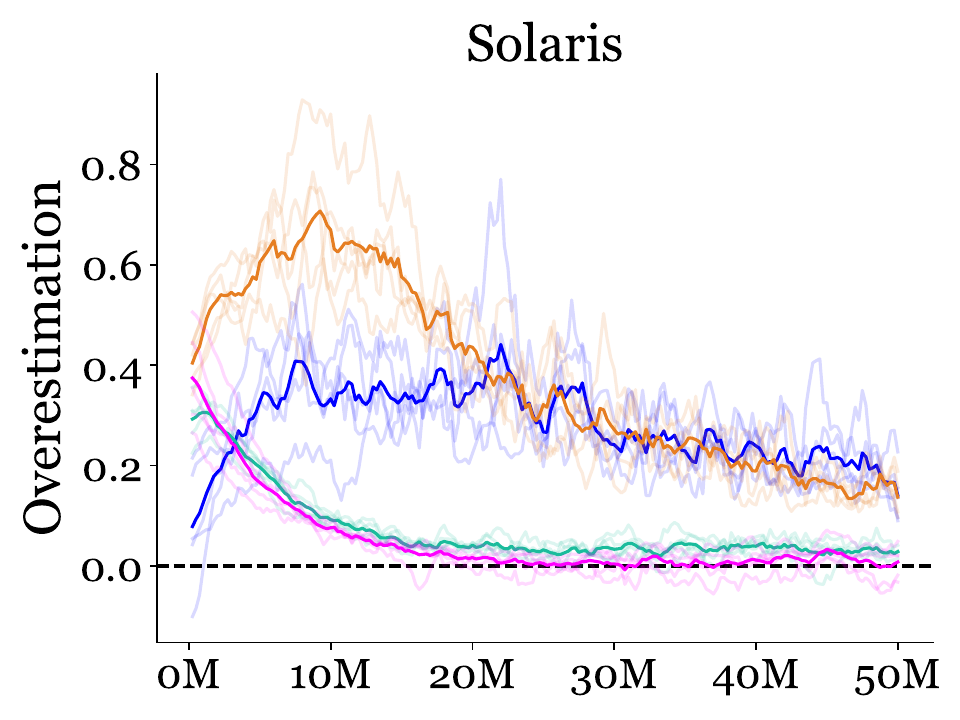} 
	\includegraphics[width=0.22\linewidth]{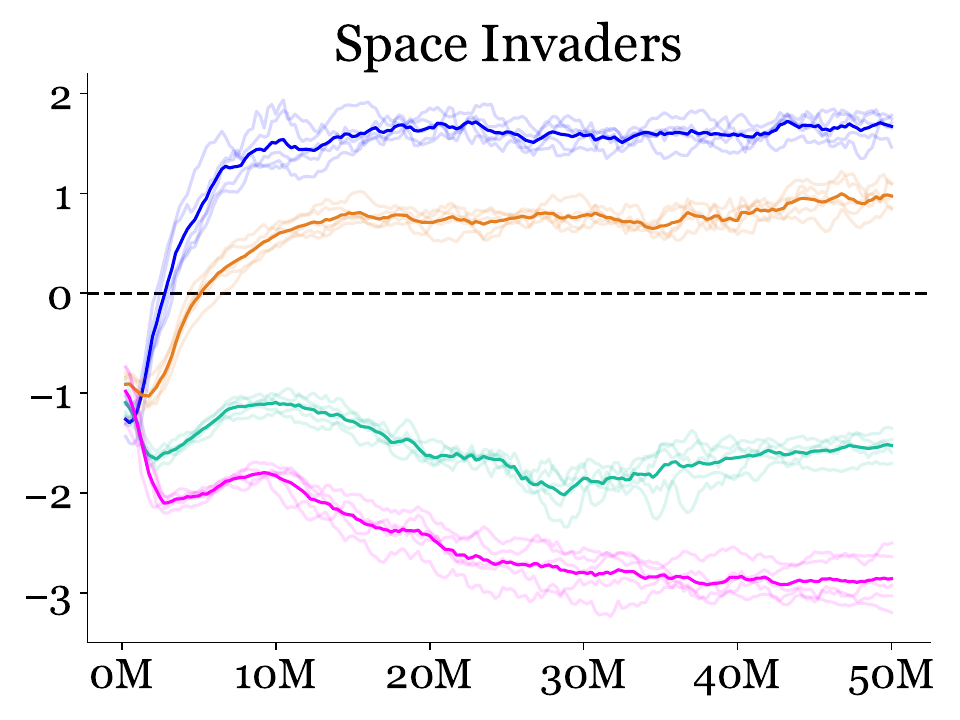} 
	\includegraphics[width=0.22\linewidth]{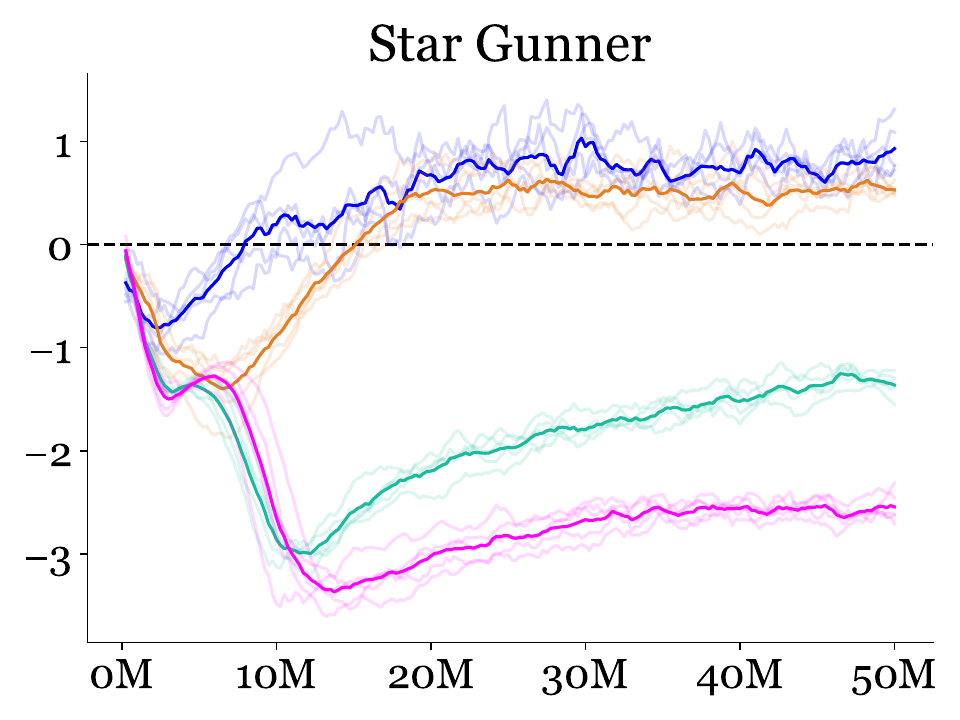} 
	\includegraphics[width=0.22\linewidth]{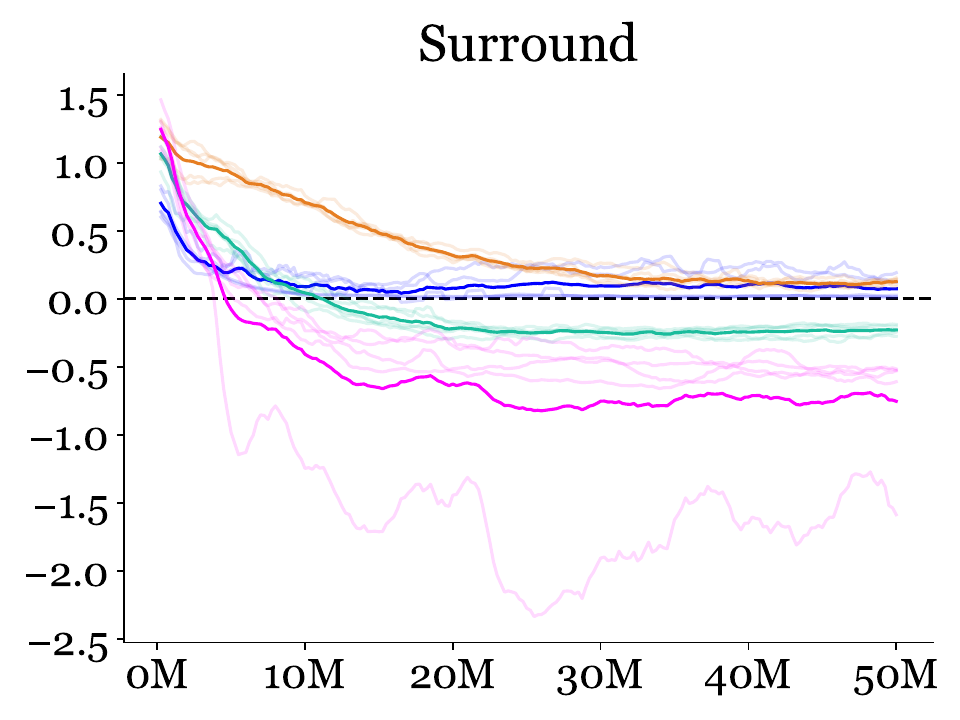} 
	\includegraphics[width=0.22\linewidth]{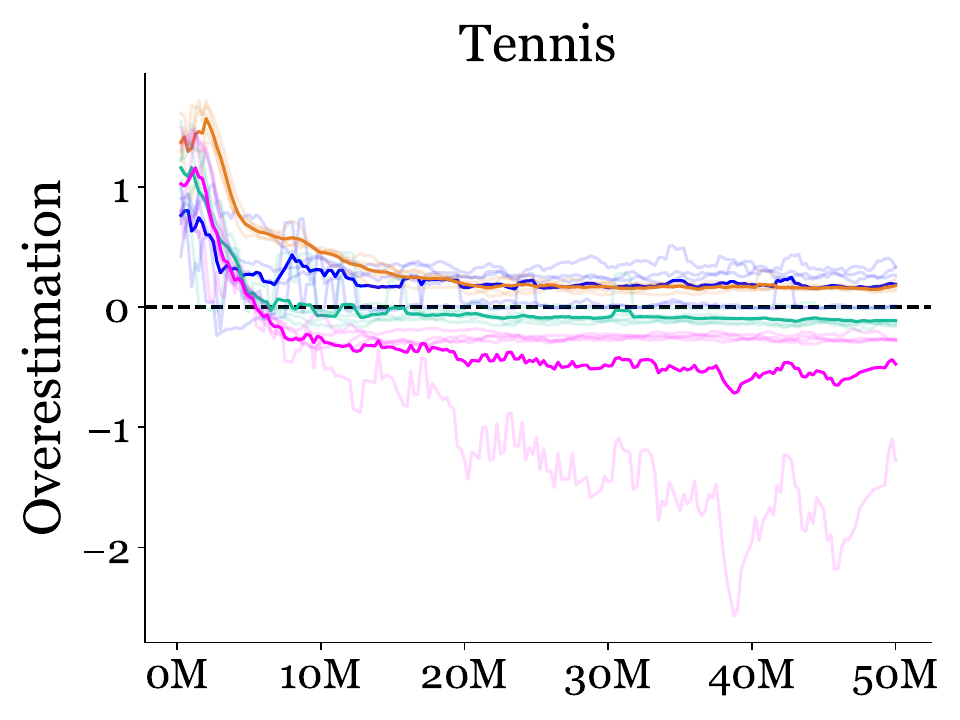} 
	\includegraphics[width=0.22\linewidth]{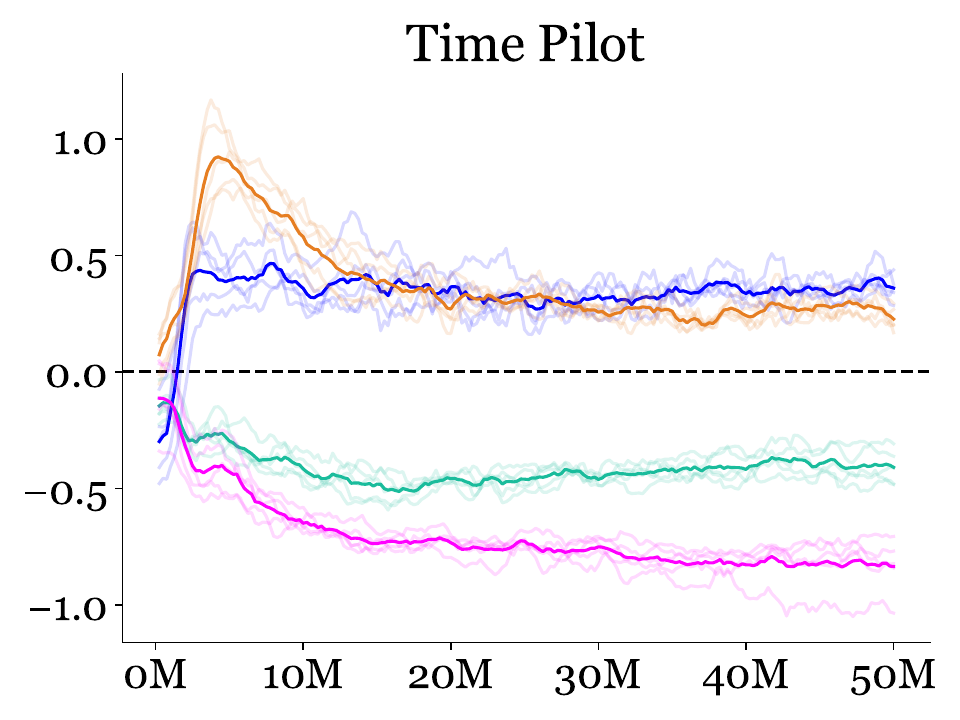} 
	\includegraphics[width=0.22\linewidth]{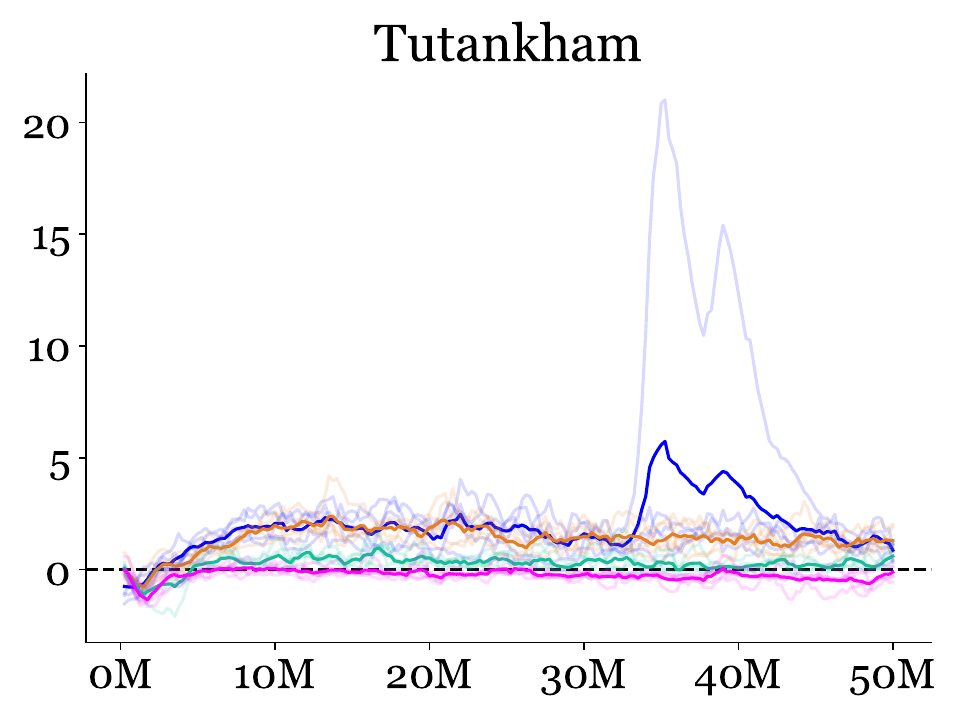} 
	\includegraphics[width=0.22\linewidth]{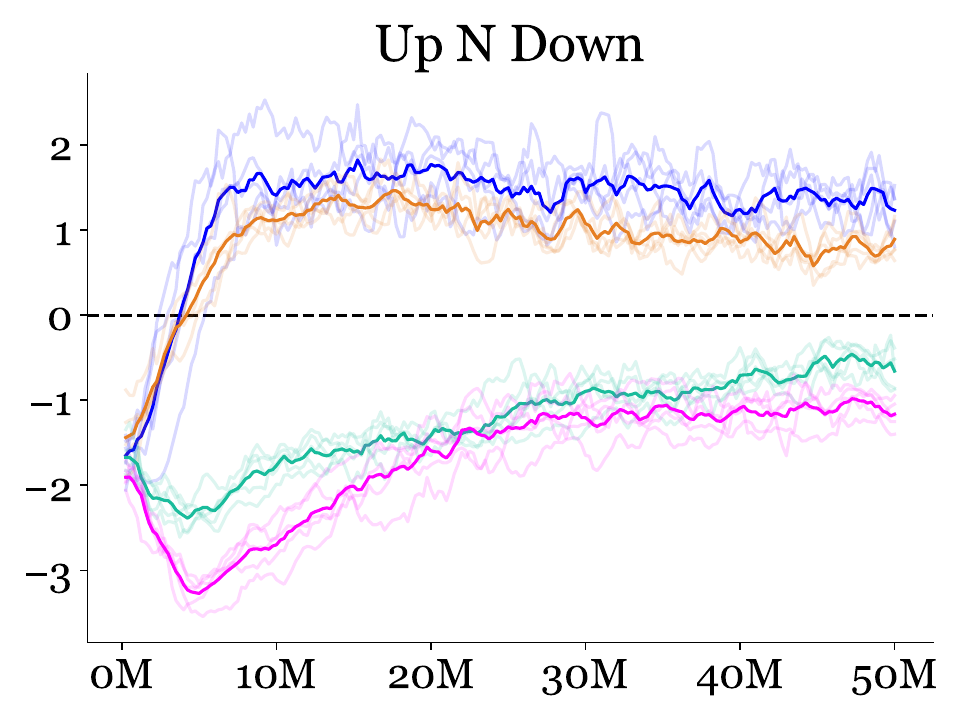} 
	\includegraphics[width=0.22\linewidth]{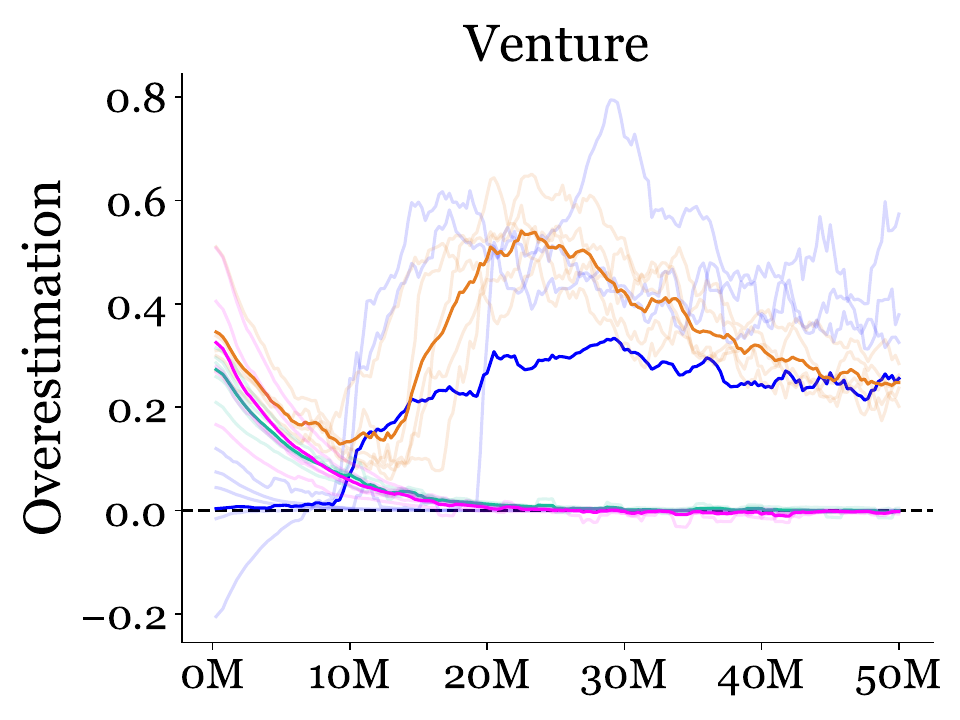} 
	\includegraphics[width=0.22\linewidth]{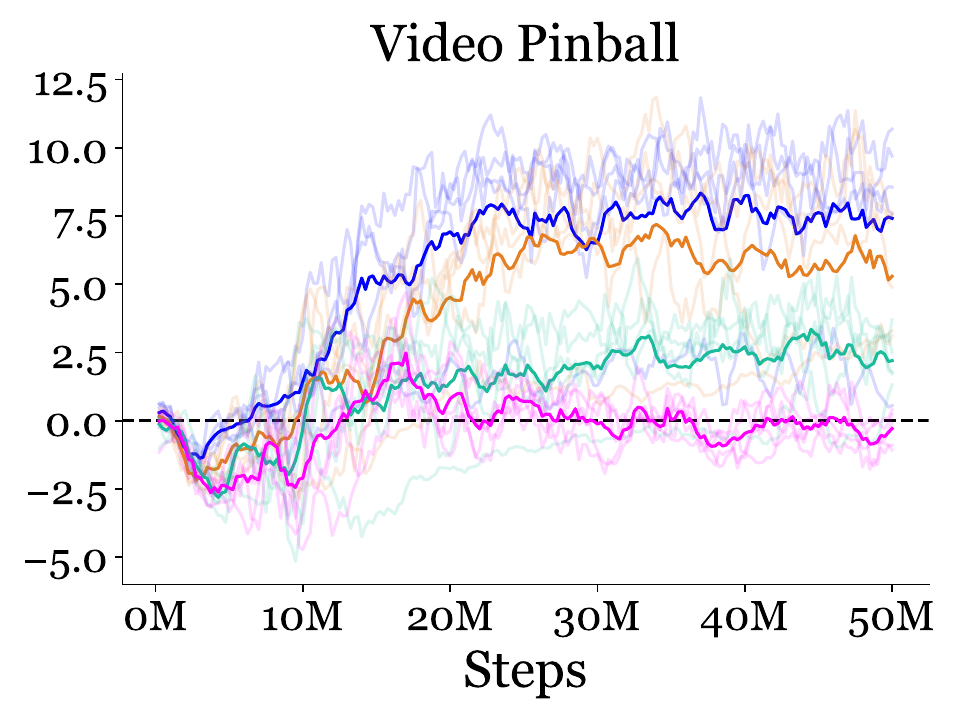} 
	\includegraphics[width=0.22\linewidth]{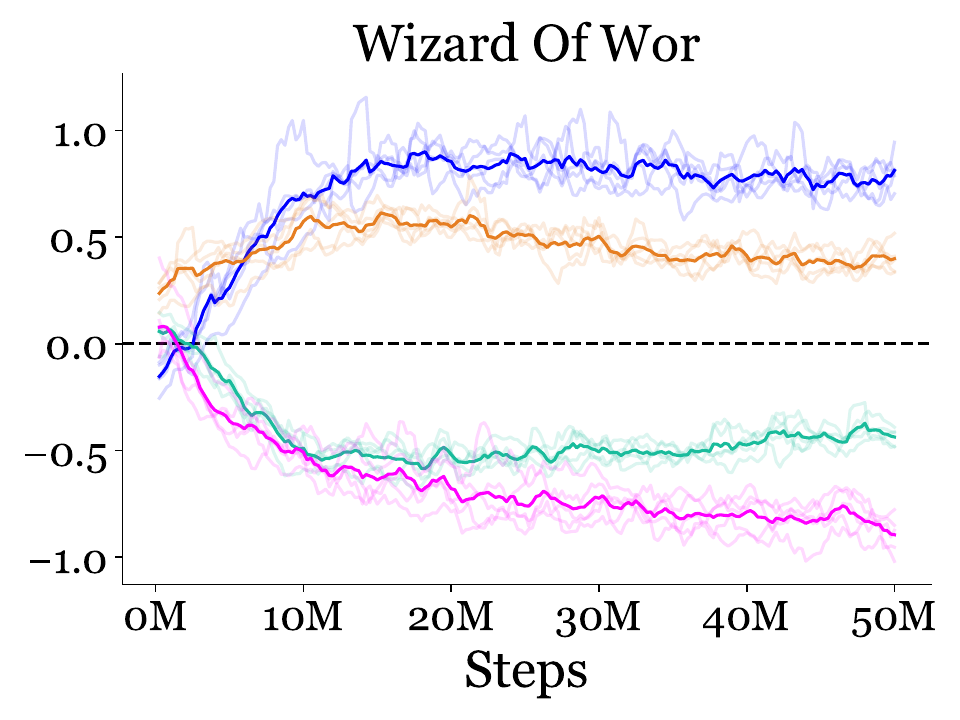} 
	\includegraphics[width=0.22\linewidth]{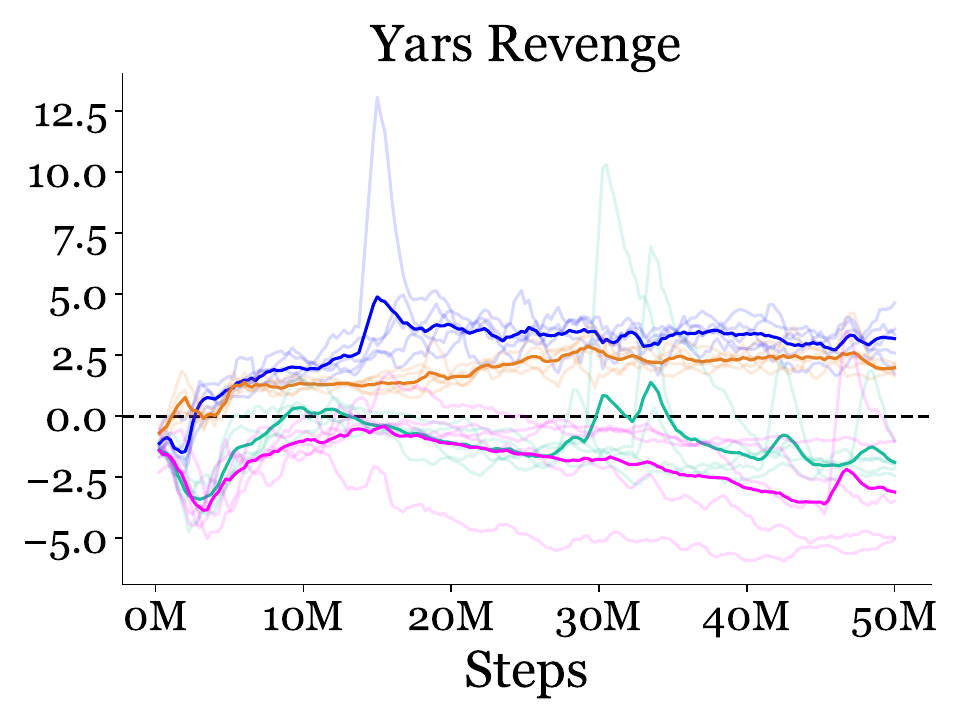} 
	\includegraphics[width=0.22\linewidth]{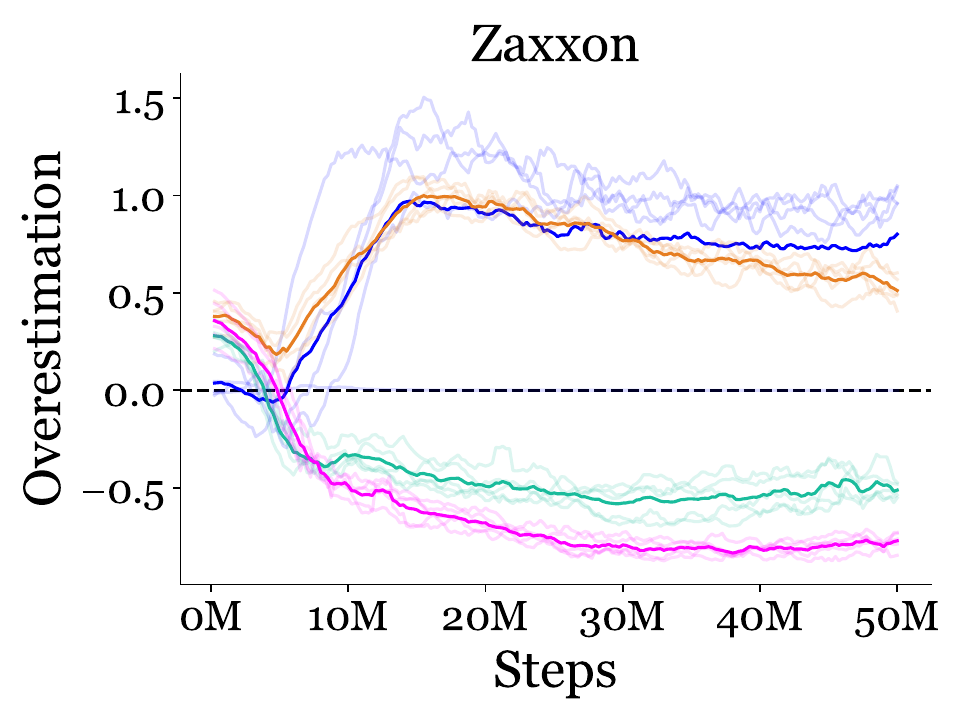} 
	\hspace{0.02\linewidth}\raisebox{3mm}{\includegraphics[width=0.65\linewidth]{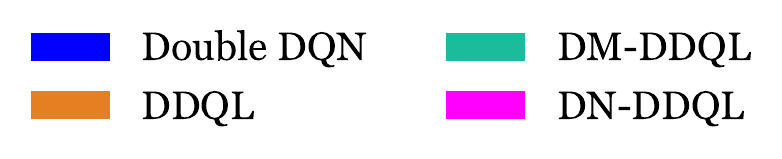}} 
	\caption{Overestimation over 50M timesteps for 57 Atari 2600 games. For readability, overestimation is bounded below at $-20$ and above at 25. Plots are smoothed with a moving average of five. The semitransparent curves are the individual seeds.}
	\label{Atari57:Overestimation:page_2}
\end{figure}

Figure~\ref{Atari57:Overestimation:page_2} depicts the overestimation measured throughout training across all environments and seeds.

\clearpage
\begin{figure}[p]
        \centering
    	\includegraphics[width=0.22\linewidth]{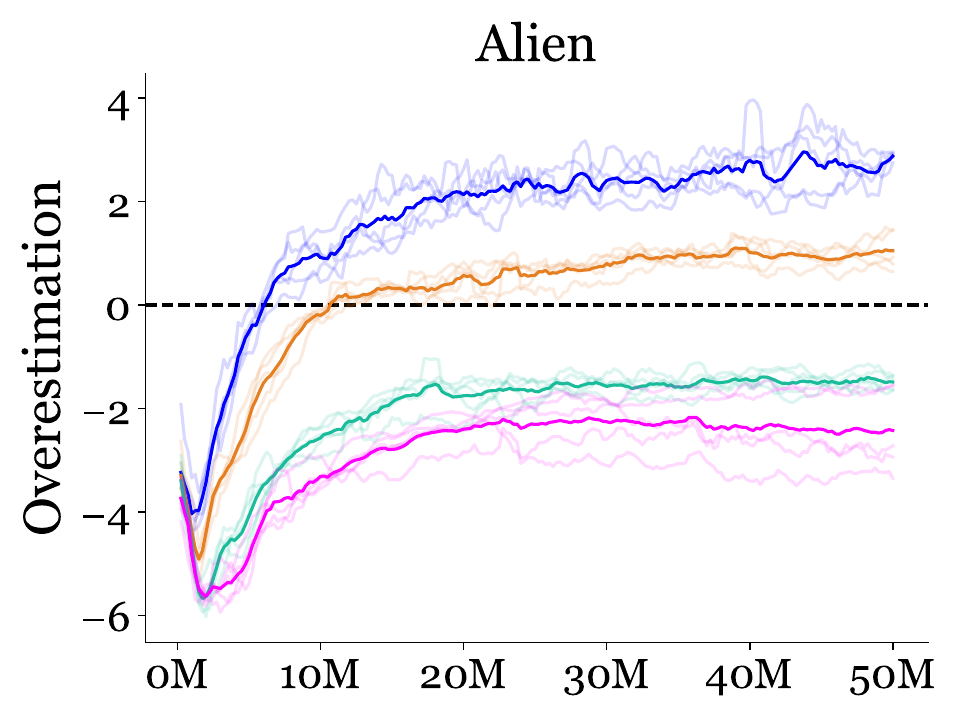} 
	\includegraphics[width=0.22\linewidth]{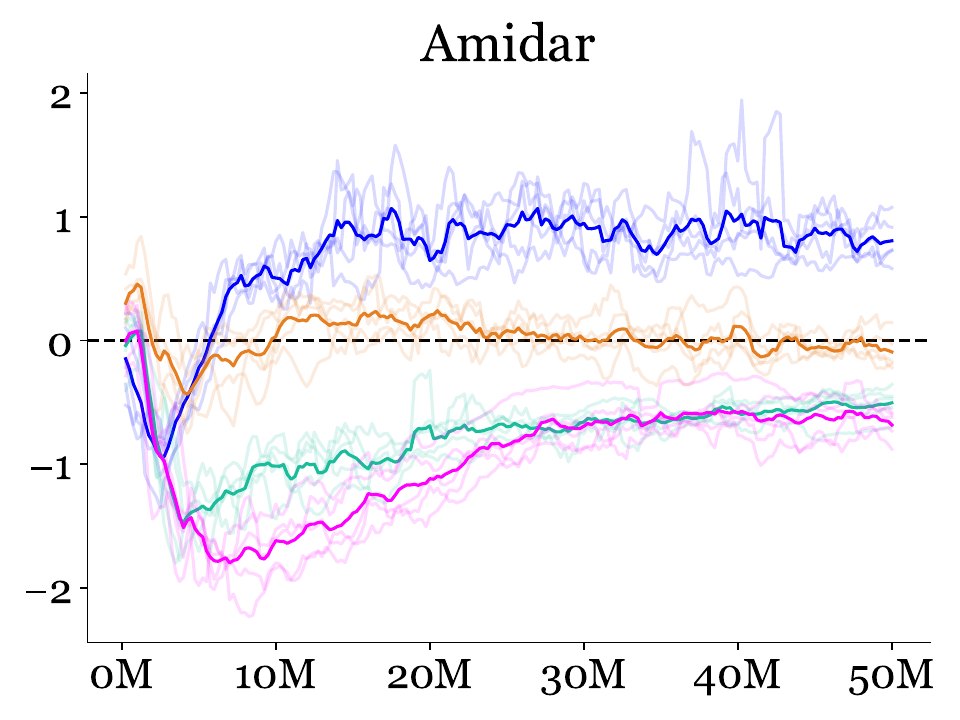} 
	\includegraphics[width=0.22\linewidth]{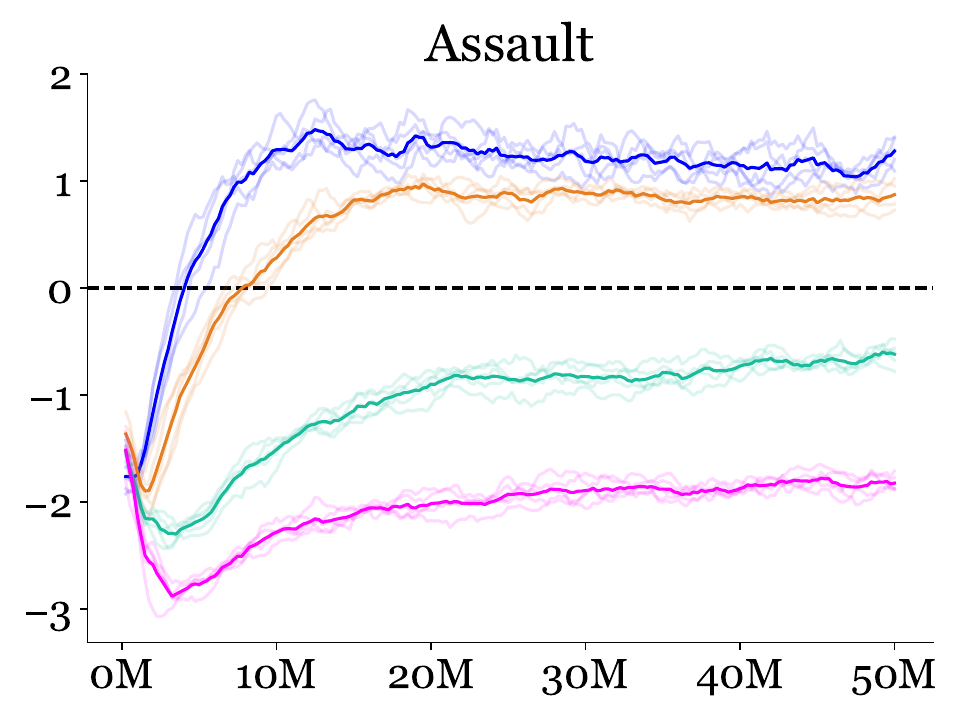} 
	\includegraphics[width=0.22\linewidth]{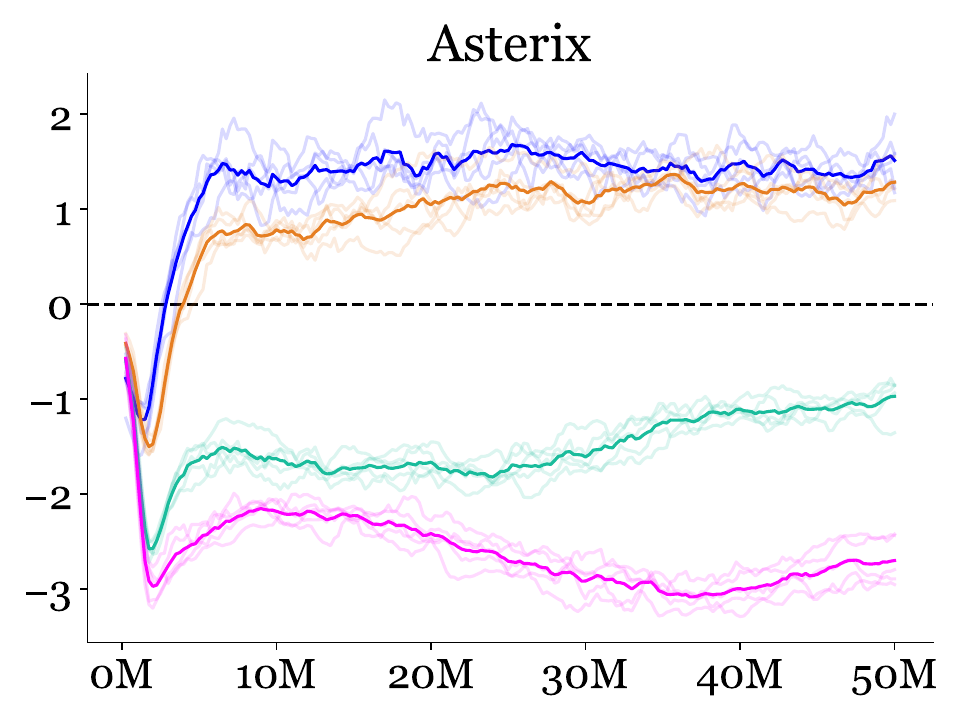} 
	\includegraphics[width=0.22\linewidth]{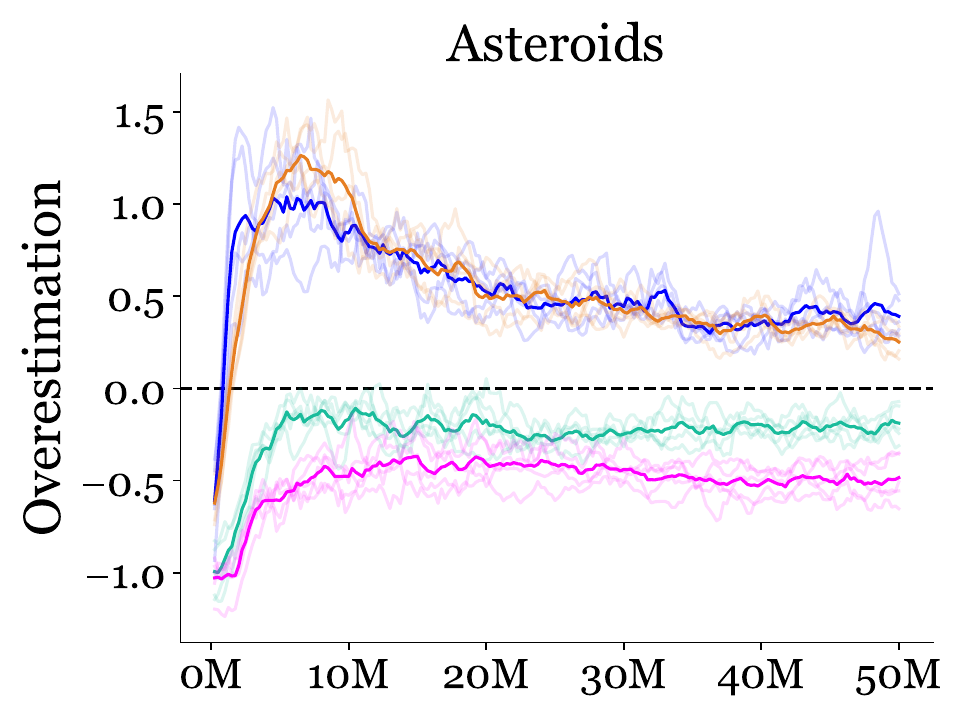} 
	\includegraphics[width=0.22\linewidth]{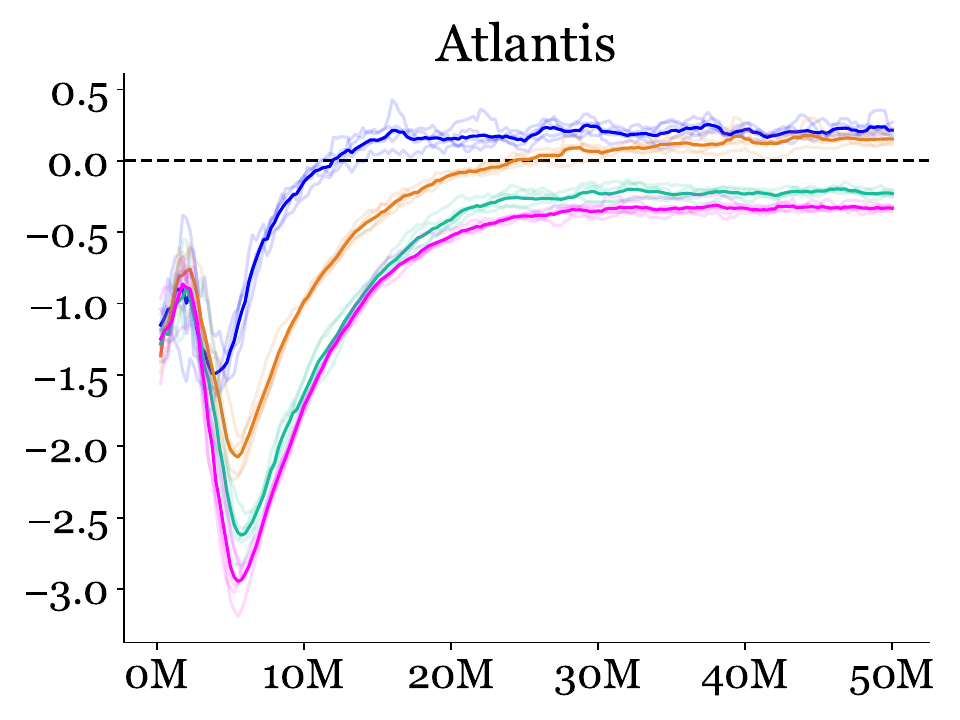} 
	\includegraphics[width=0.22\linewidth]{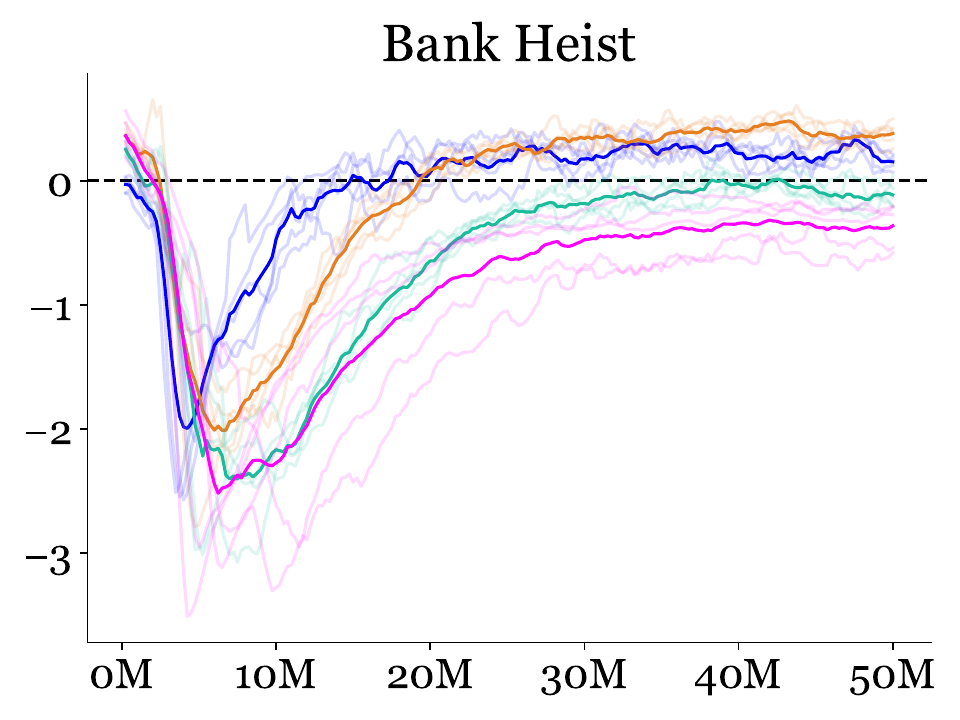} 
	\includegraphics[width=0.22\linewidth]{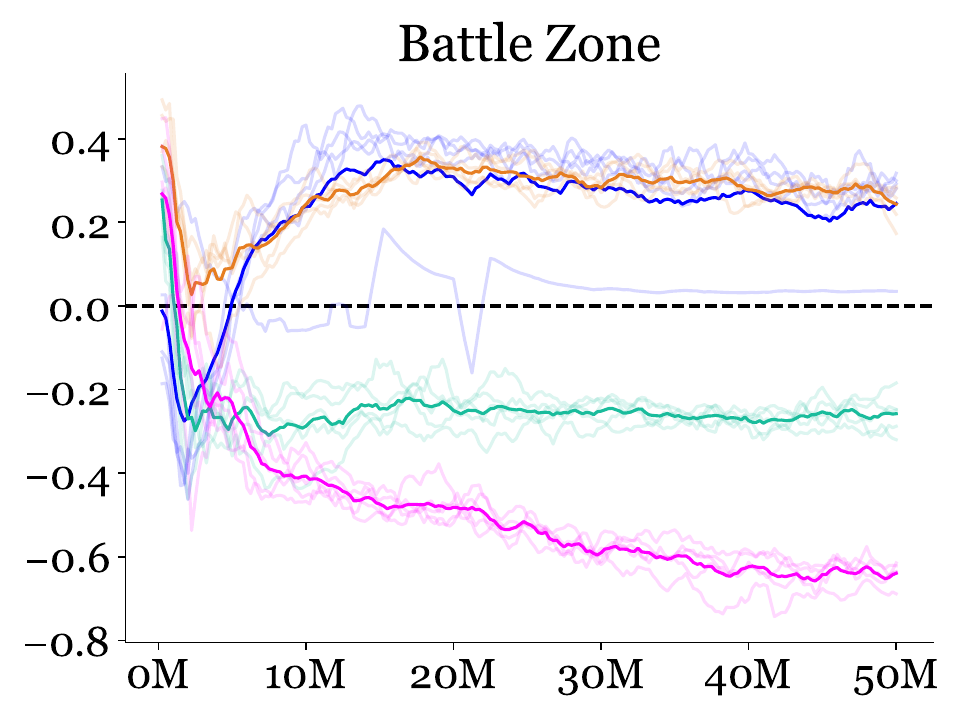} 
	\includegraphics[width=0.22\linewidth]{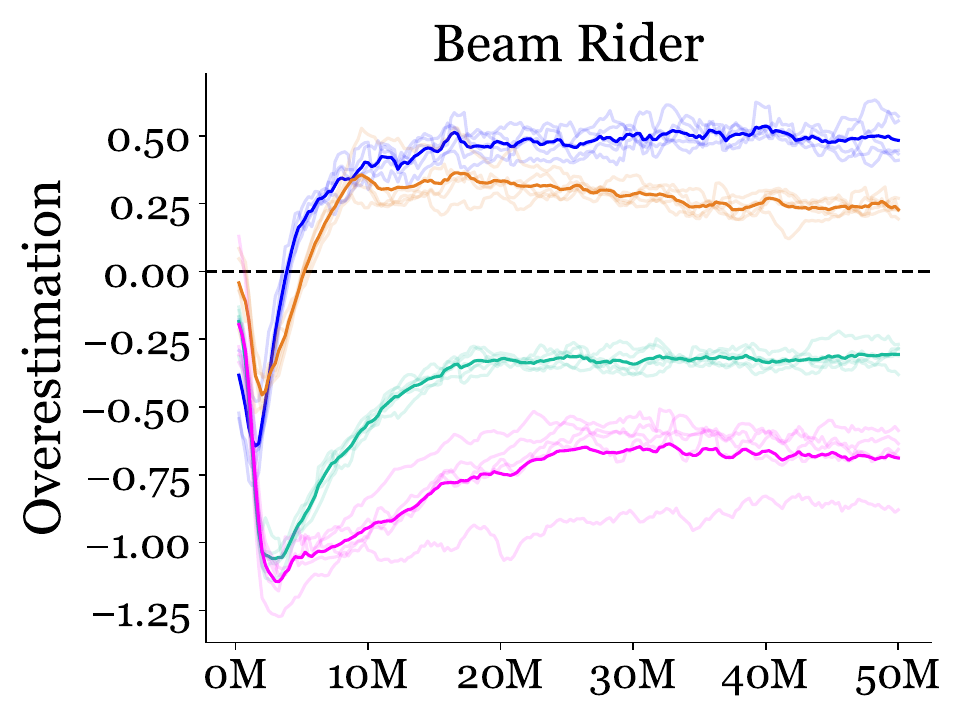} 
	\includegraphics[width=0.22\linewidth]{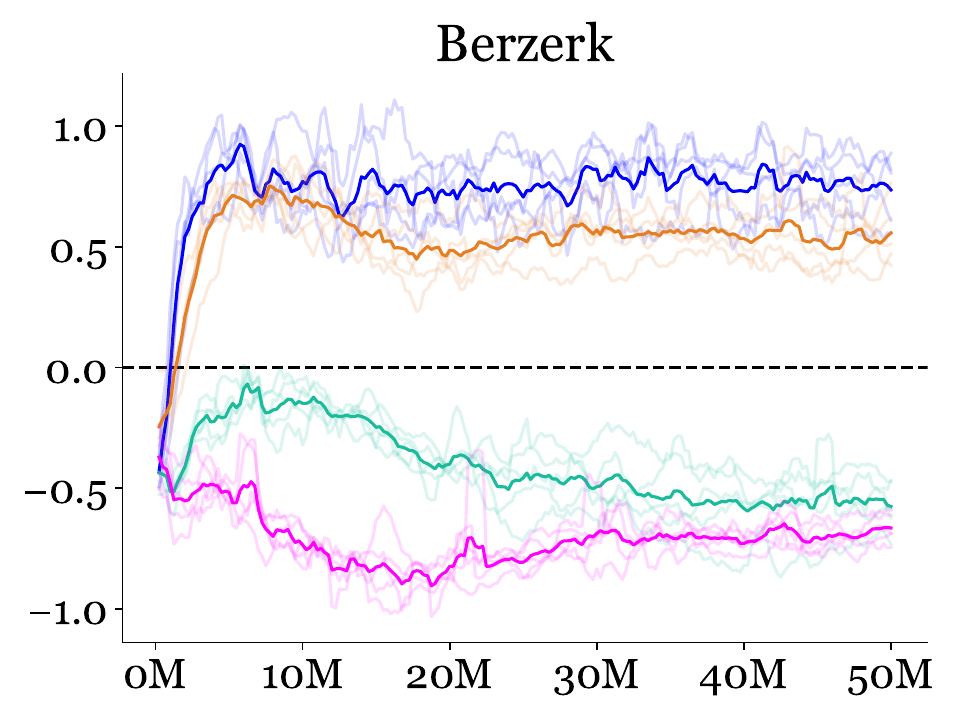} 
	\includegraphics[width=0.22\linewidth]{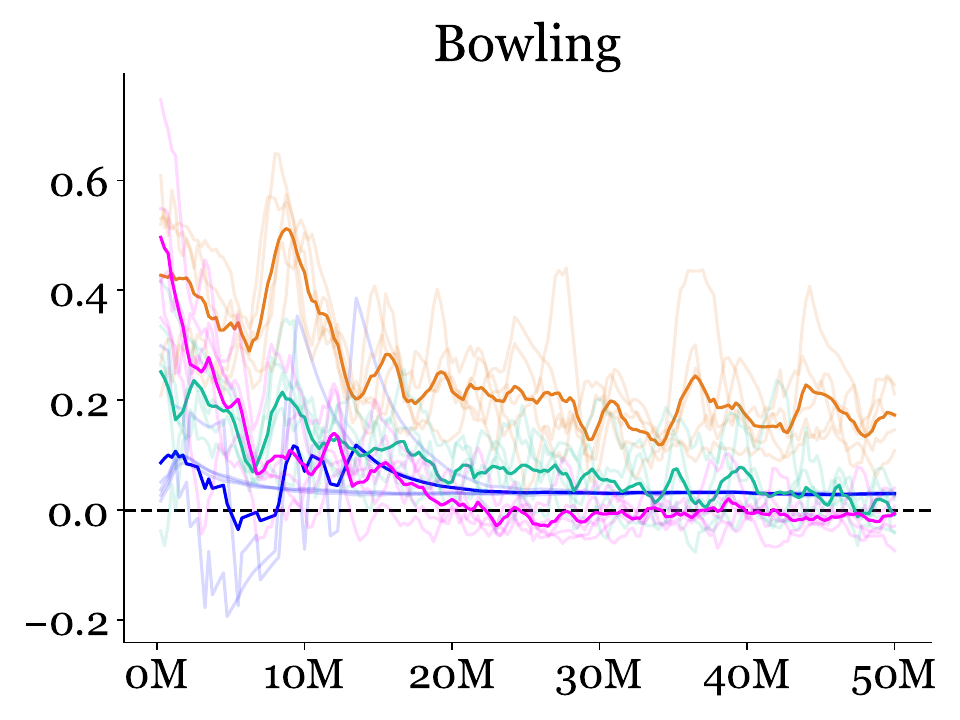} 
	\includegraphics[width=0.22\linewidth]{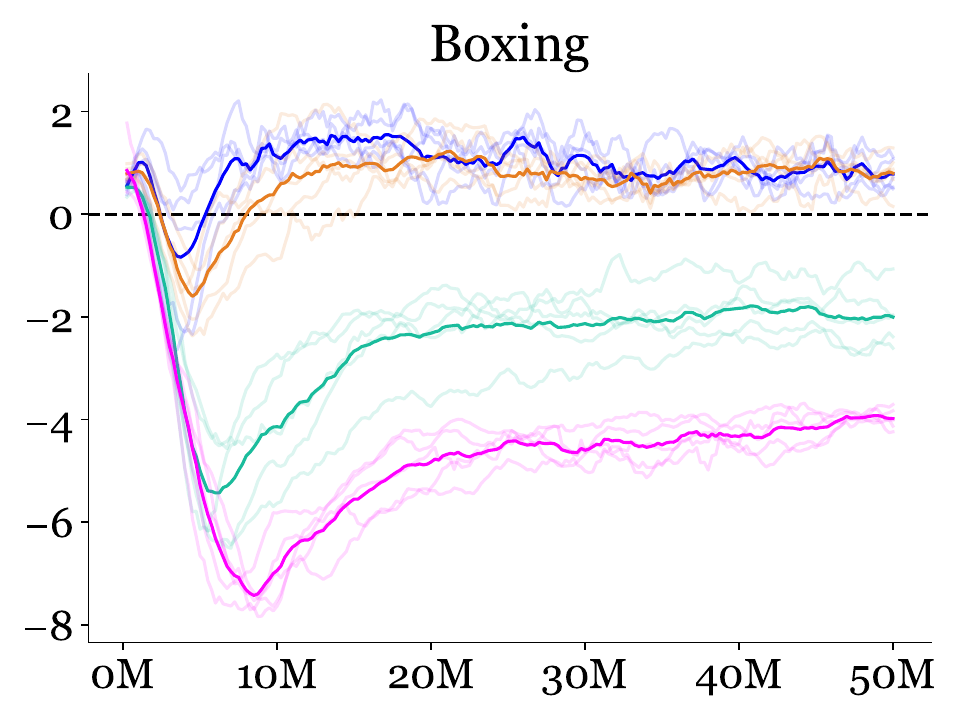} 
	\includegraphics[width=0.22\linewidth]{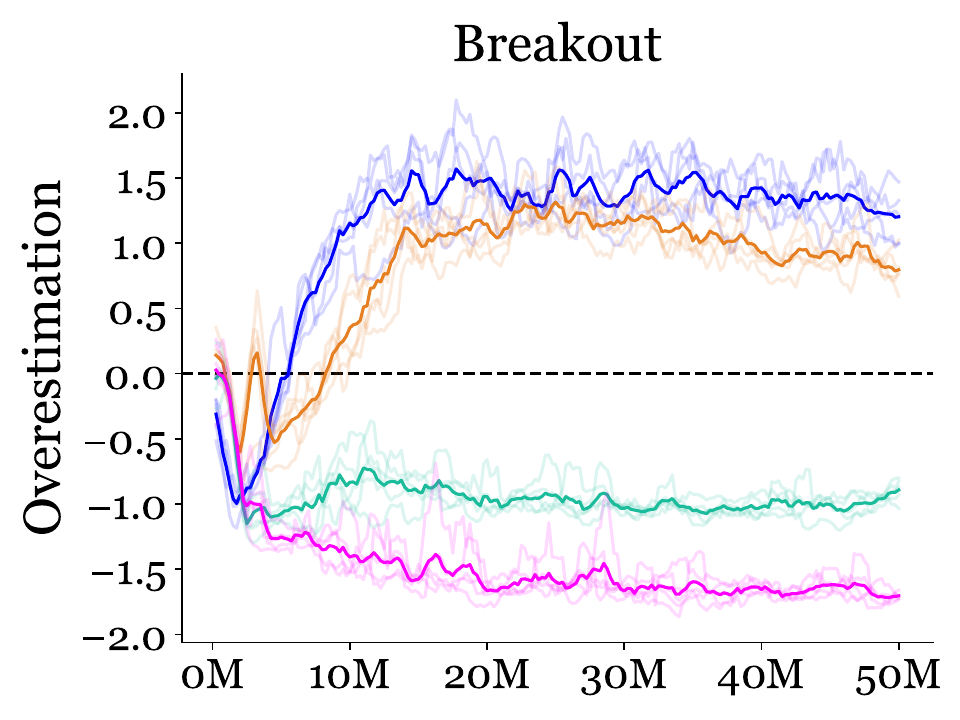} 
	\includegraphics[width=0.22\linewidth]{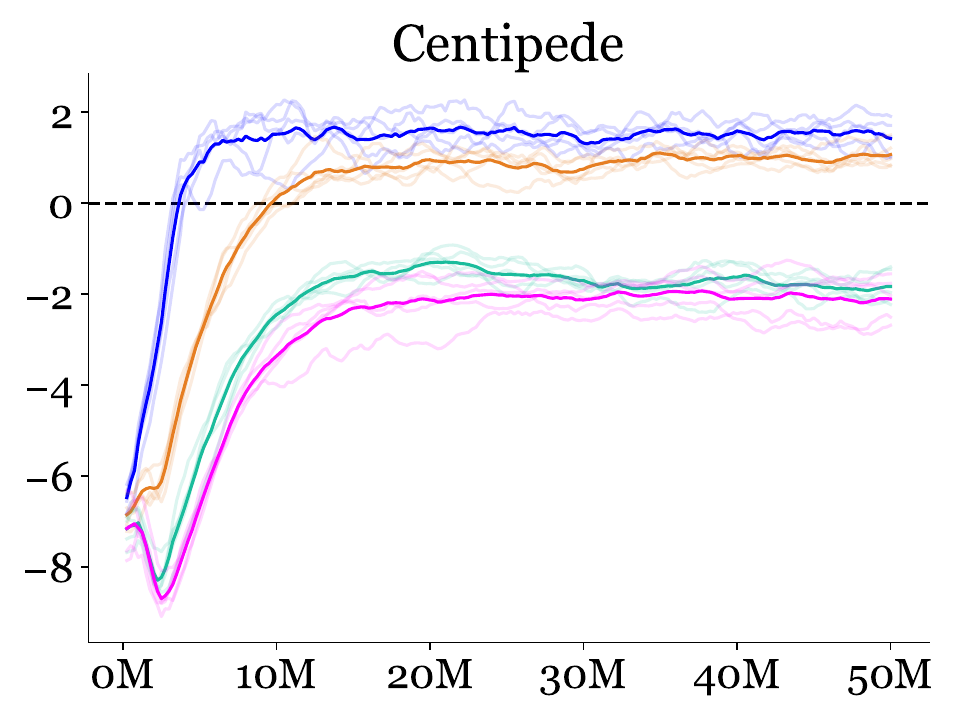} 
	\includegraphics[width=0.22\linewidth]{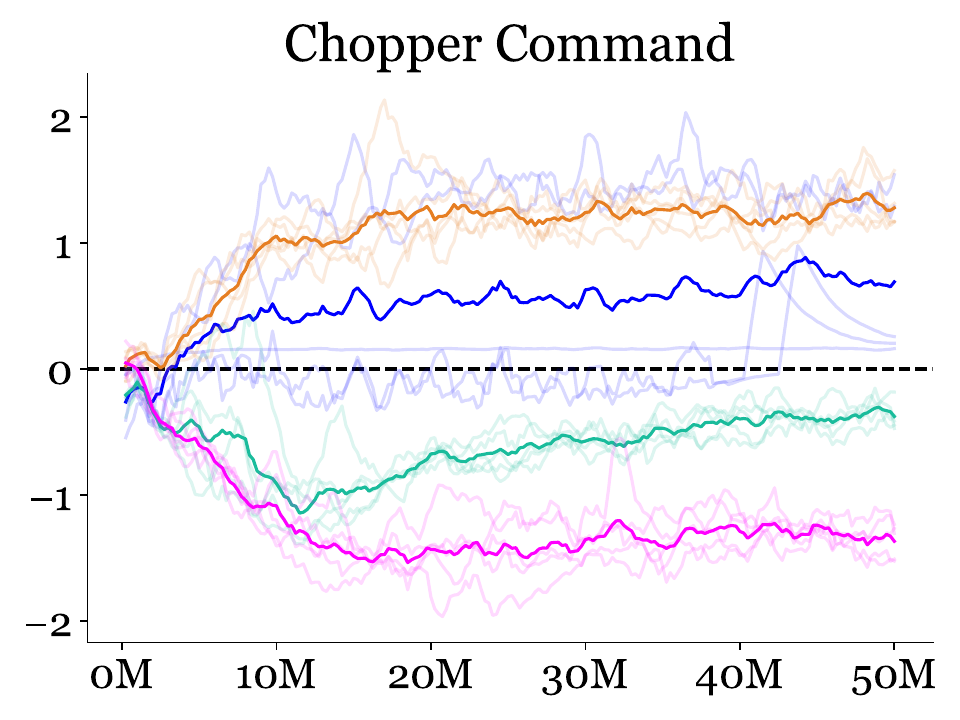} 
	\includegraphics[width=0.22\linewidth]{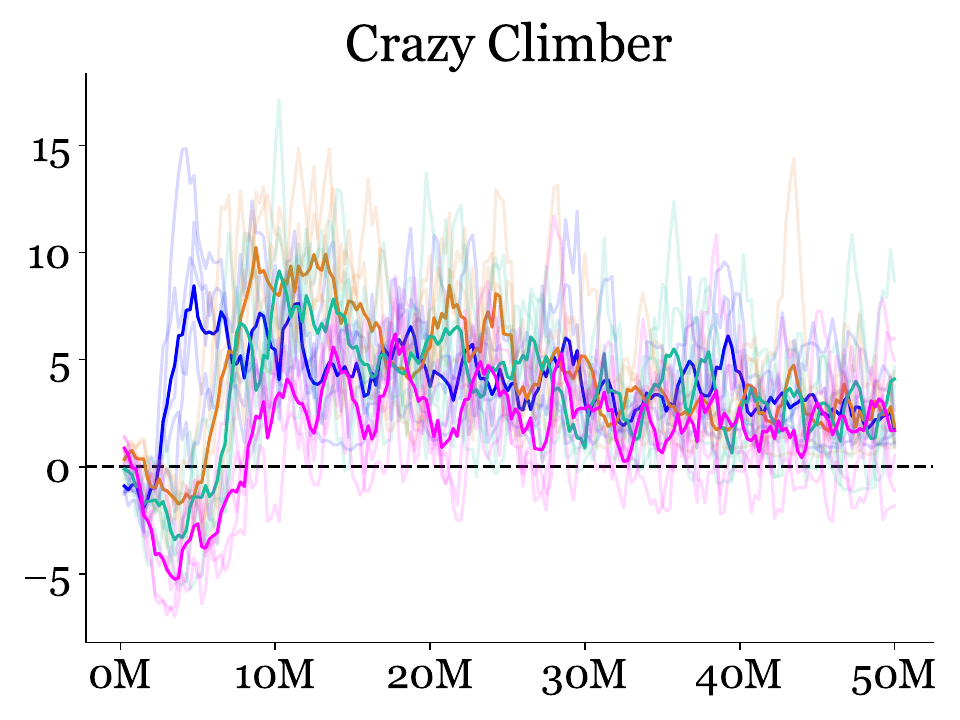} 
	\includegraphics[width=0.22\linewidth]{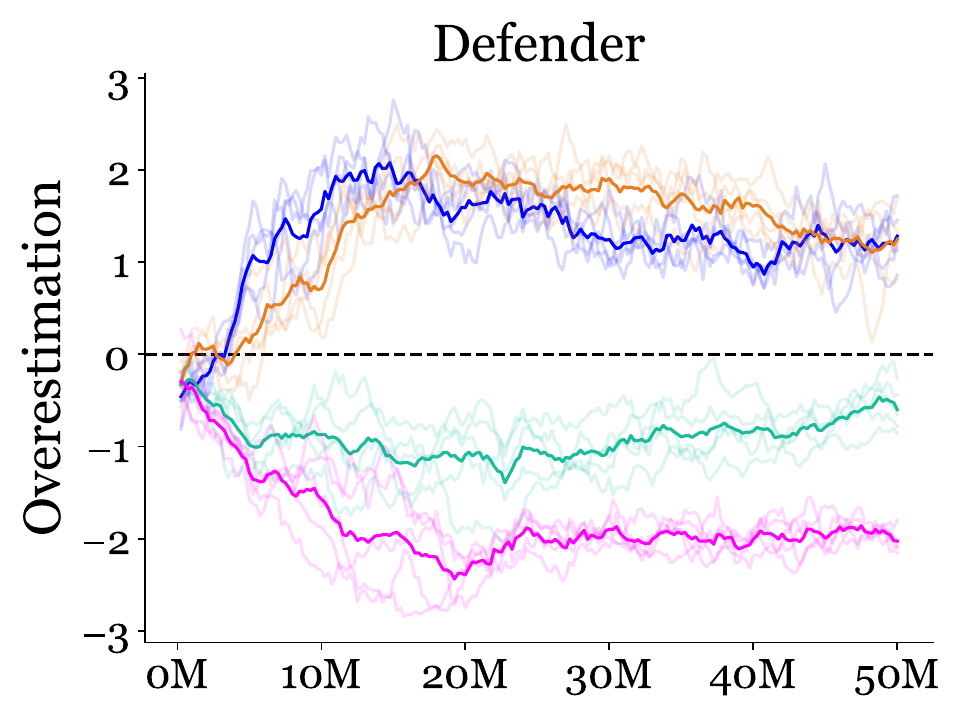} 
	\includegraphics[width=0.22\linewidth]{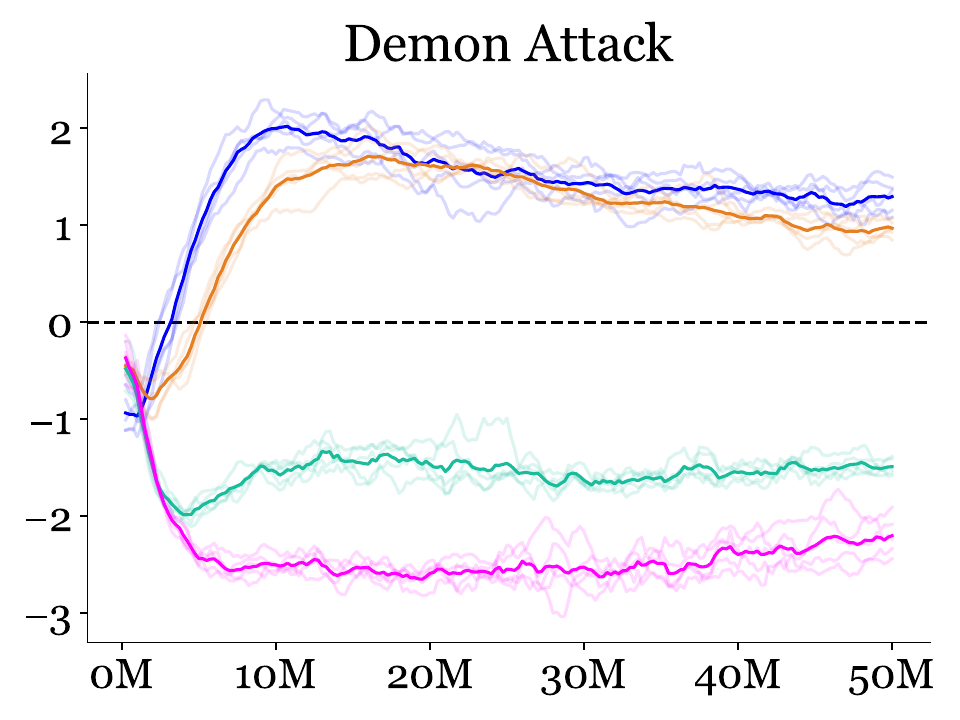} 
	\includegraphics[width=0.22\linewidth]{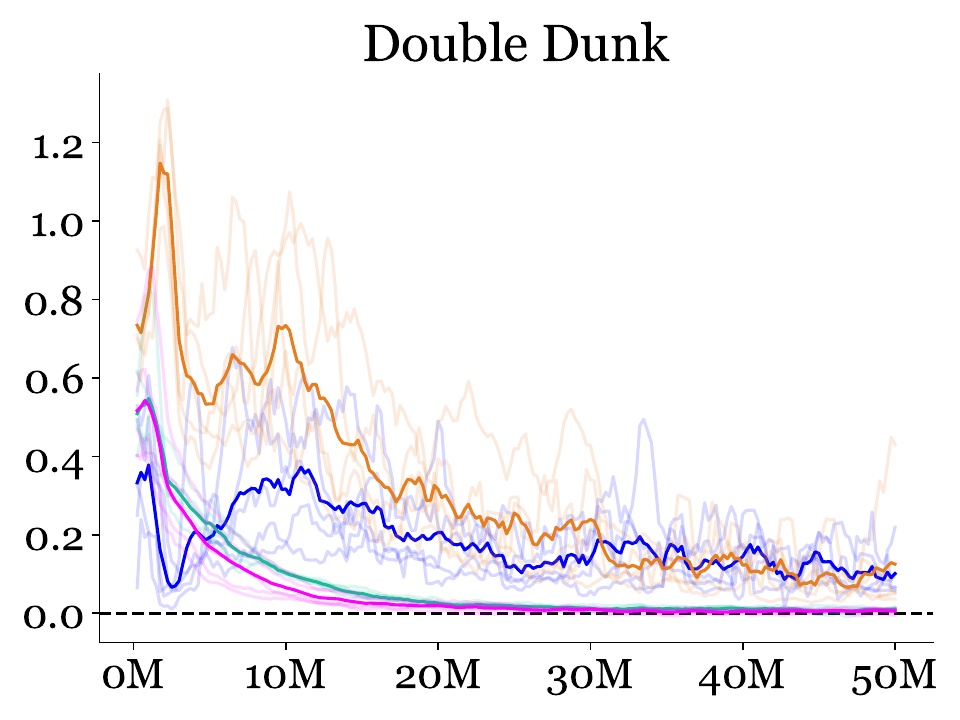} 
	\includegraphics[width=0.22\linewidth]{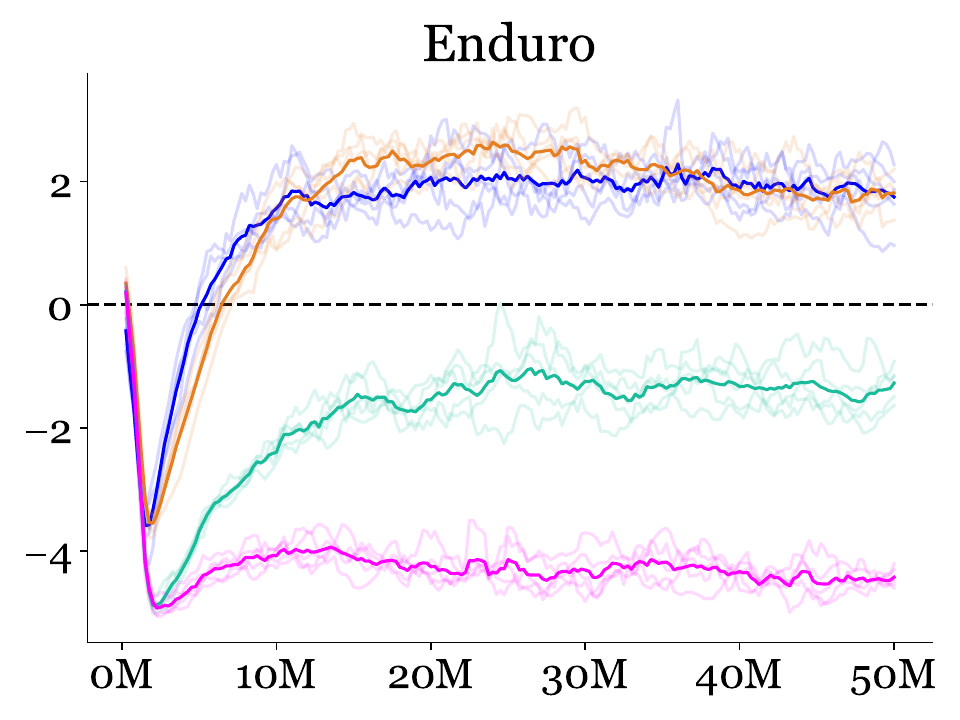} 
	\includegraphics[width=0.22\linewidth]{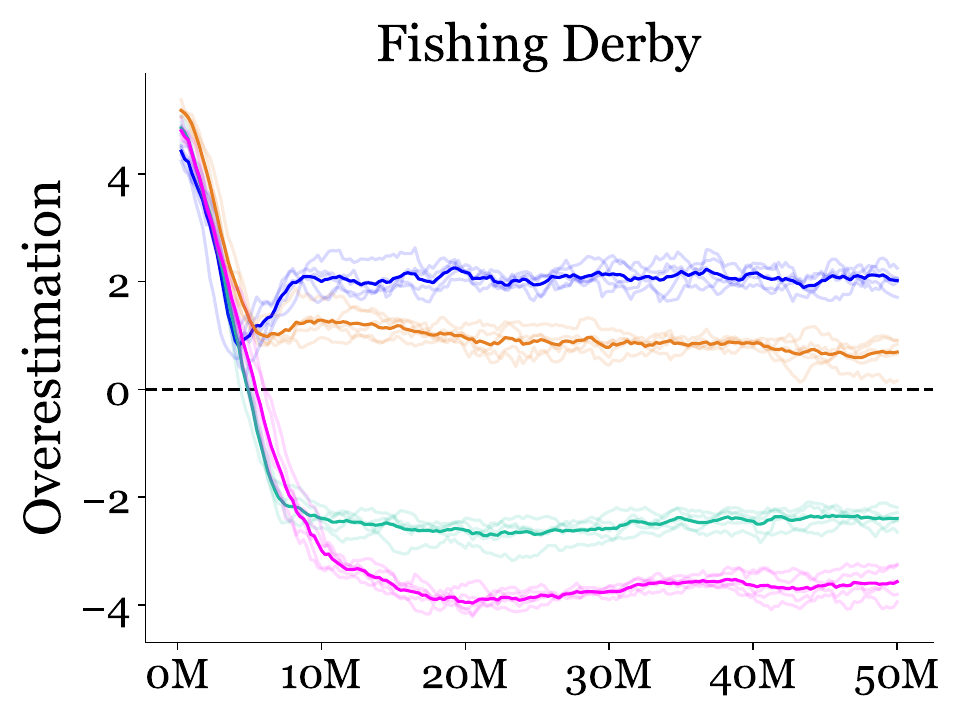} 
	\includegraphics[width=0.22\linewidth]{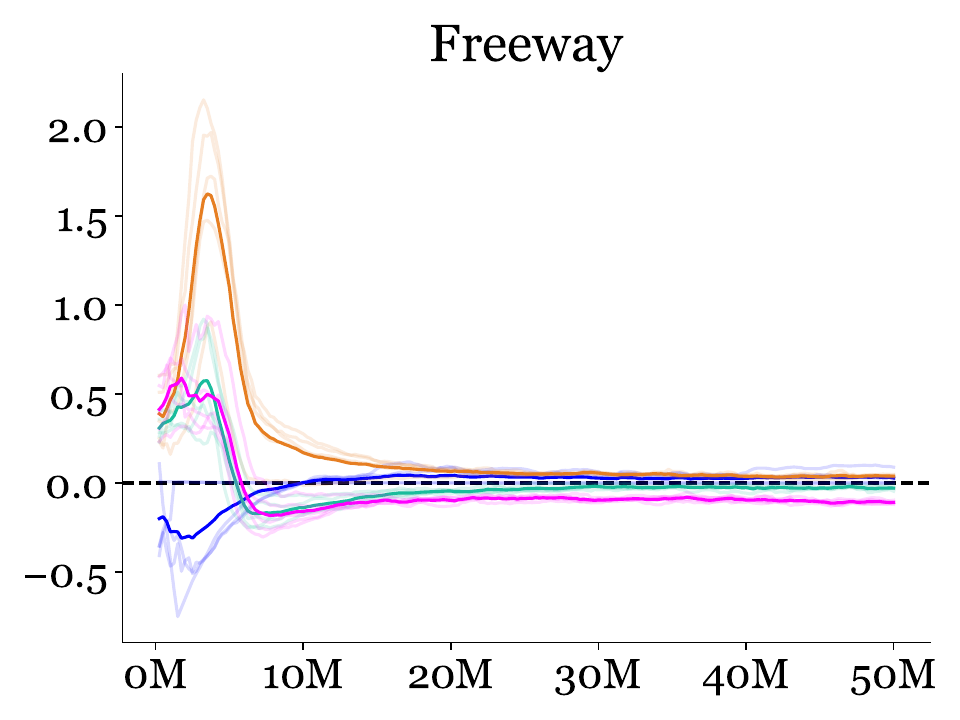} 
	\includegraphics[width=0.22\linewidth]{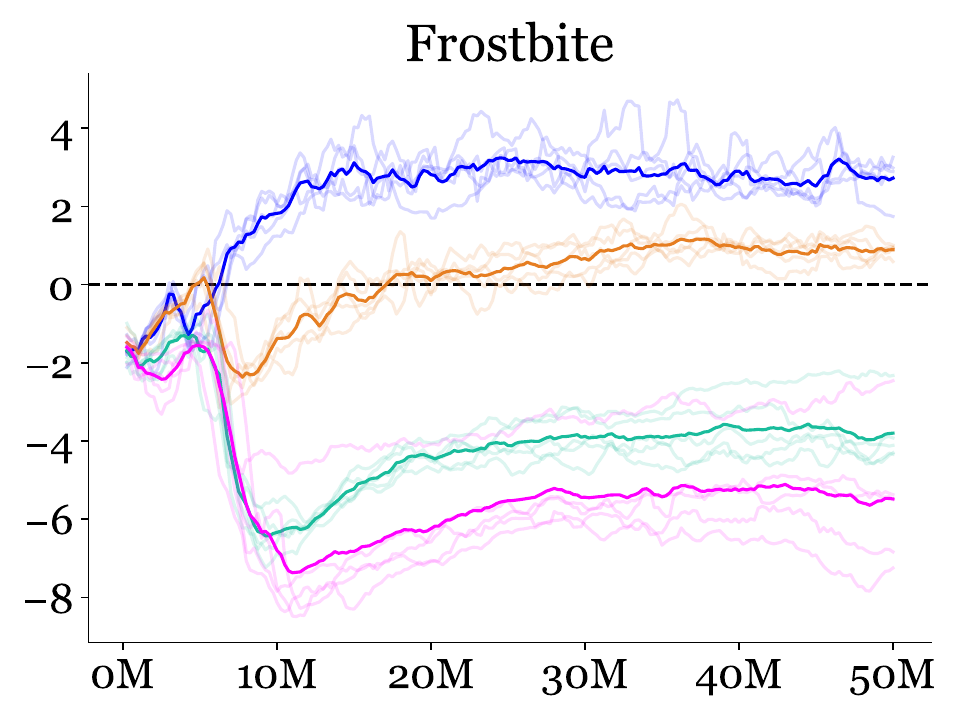} 
	\includegraphics[width=0.22\linewidth]{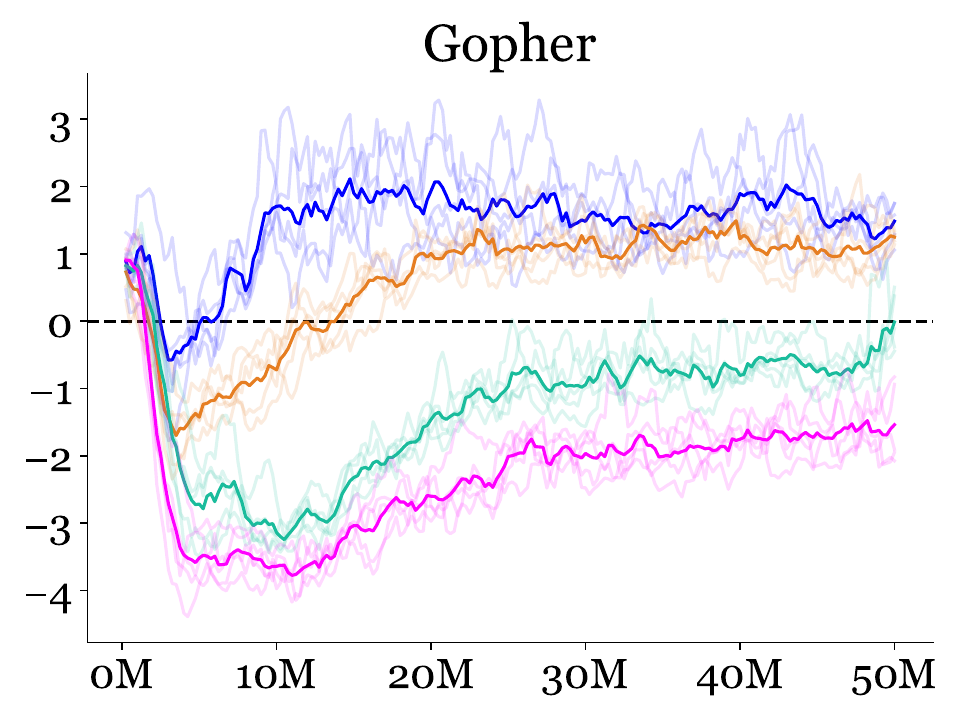} 
	\includegraphics[width=0.22\linewidth]{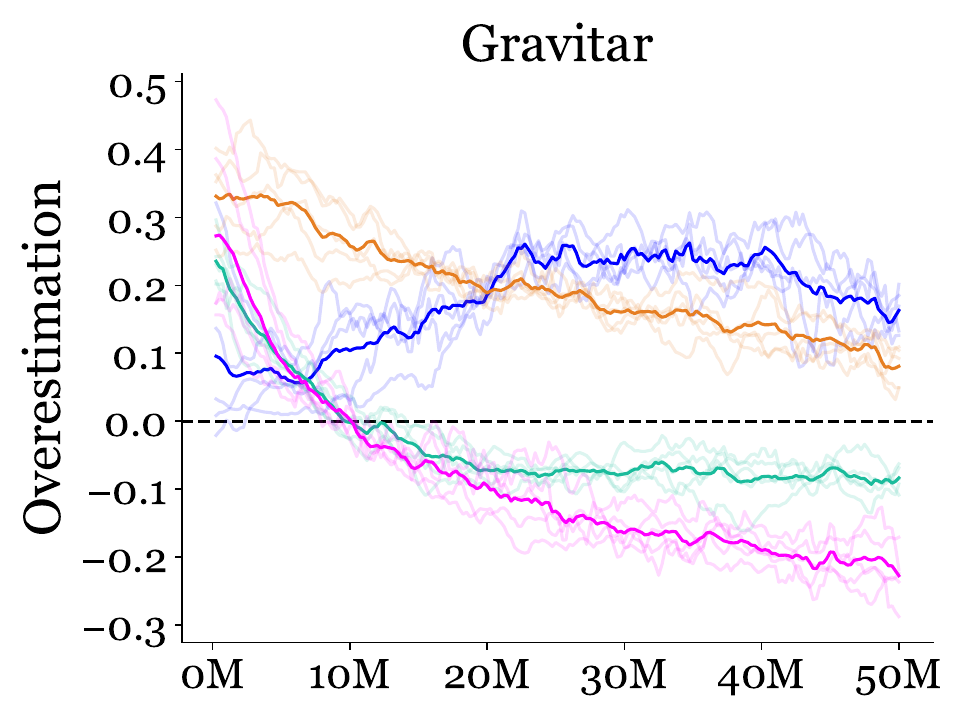} 
	\includegraphics[width=0.22\linewidth]{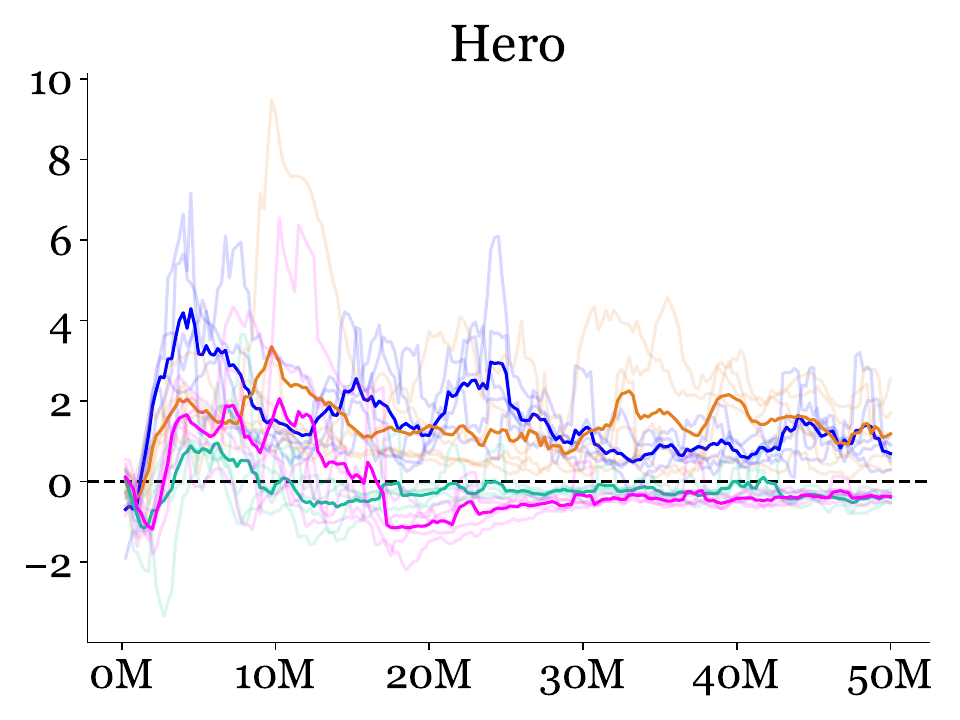} 
	\includegraphics[width=0.22\linewidth]{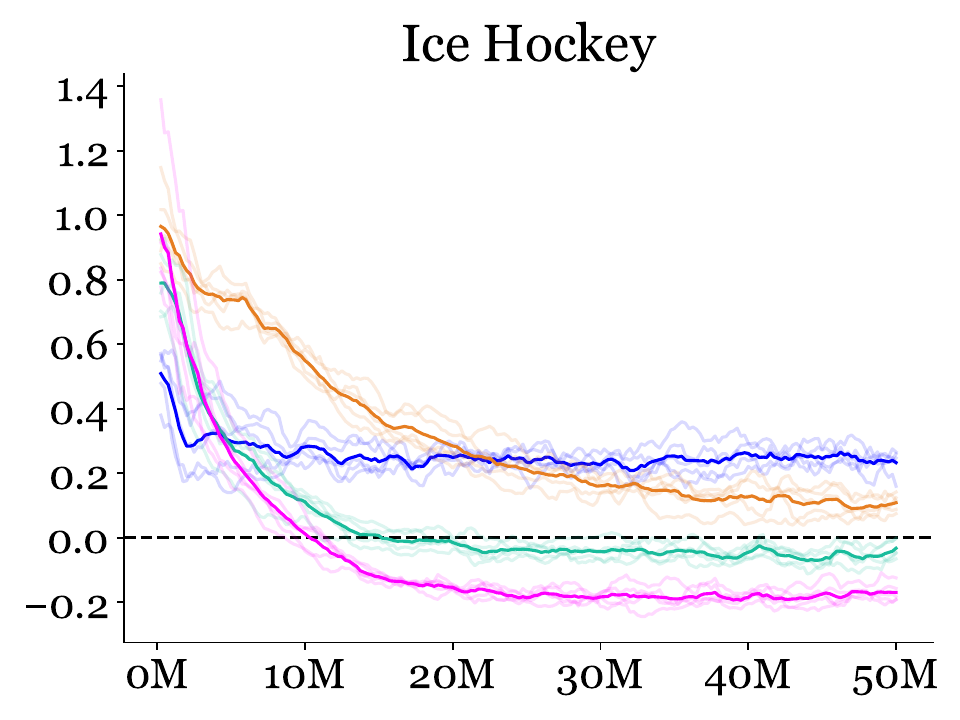} 
	\includegraphics[width=0.22\linewidth]{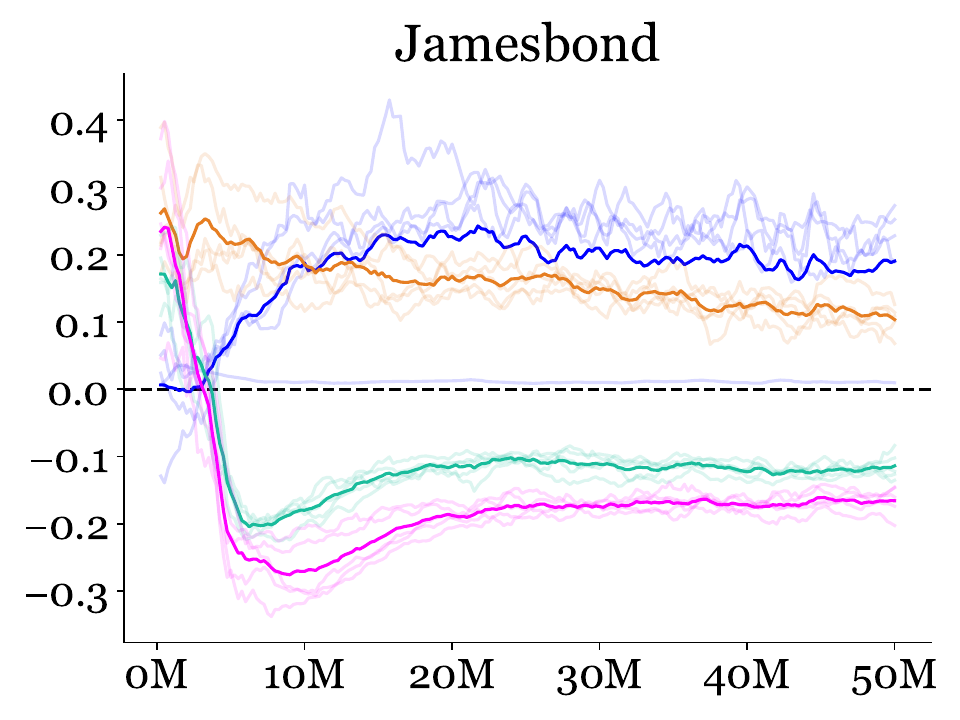} 
\end{figure}

\clearpage
\subsection{Atari-57 Performance}

\begin{figure}[t]
    \centering
    \includegraphics[width=0.95\textwidth, valign=m]{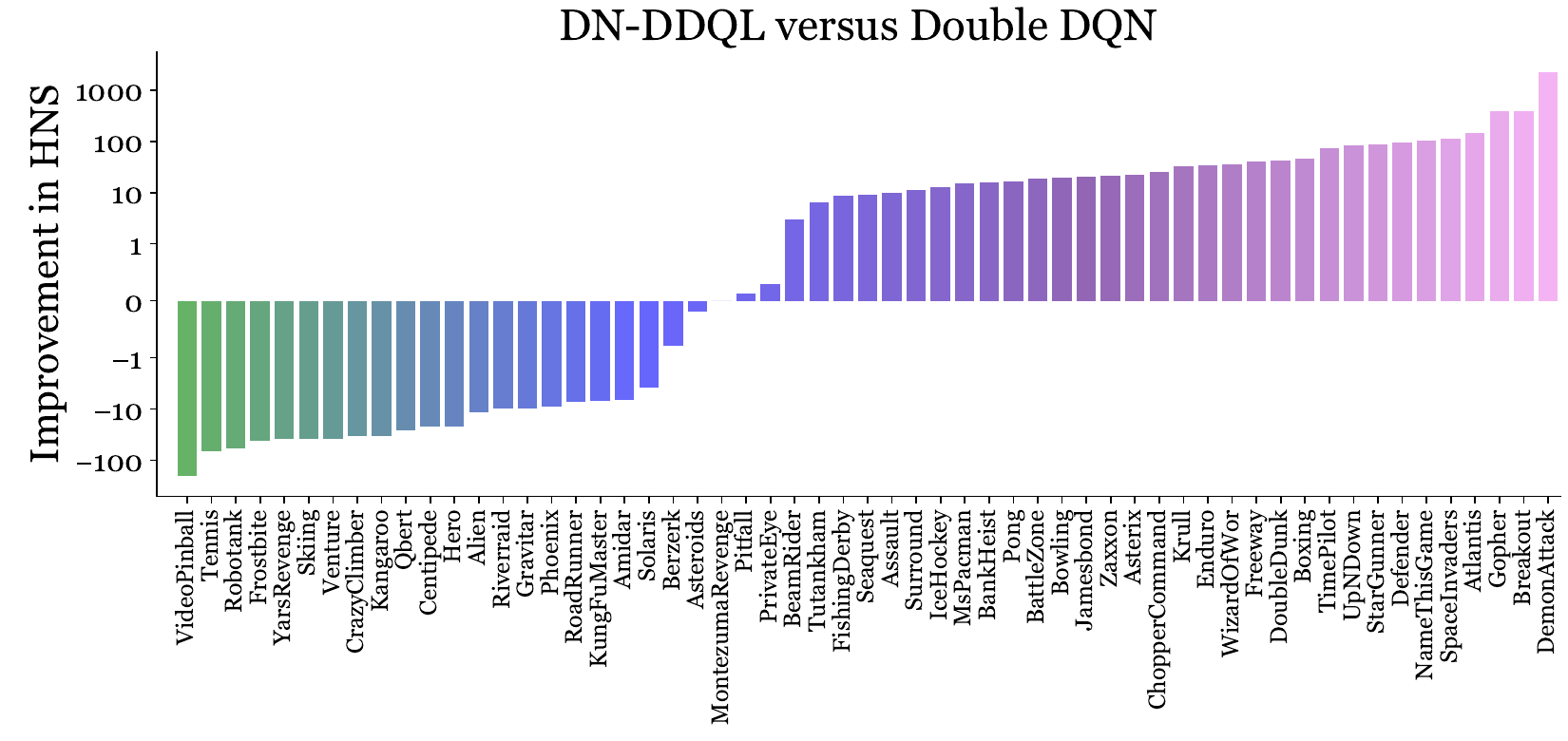}
        \caption{\textbf{Per-game performance improvements of DN-DDQL.} Per-game improvements in HNS of DN-DDQL over Double DQN in each of 57 games, calculated as the average area under the curve across 5 seeds. The y-axis is log-scale.}
    \label{fig:per_game_dn_ddql_perf}
\end{figure}

\begin{figure}[p]
        \centering
    	\includegraphics[width=0.22\linewidth]{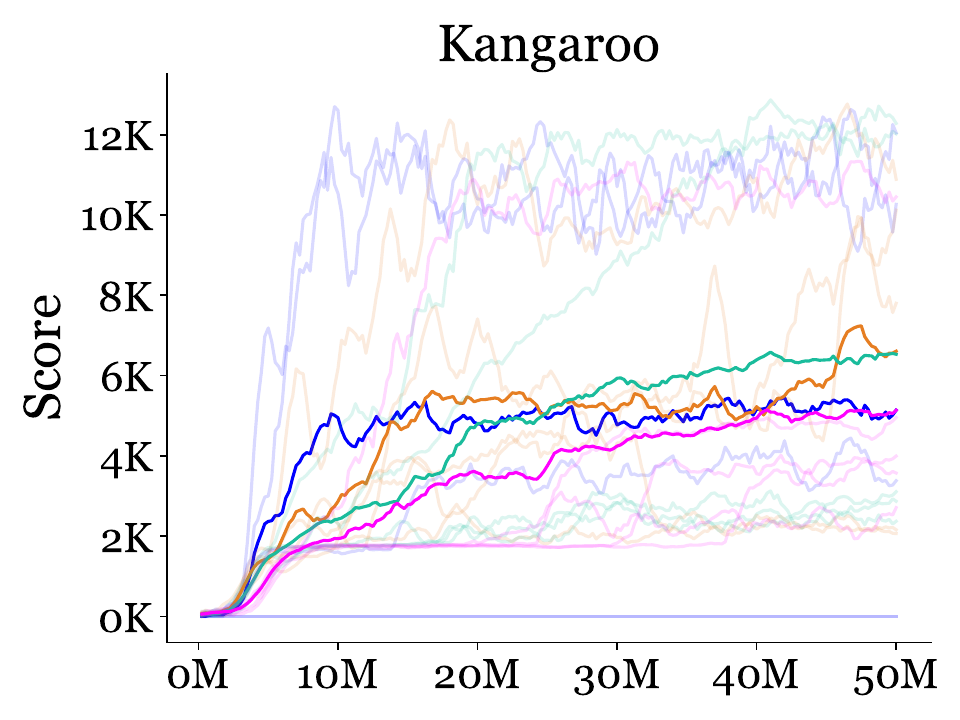} 
	\includegraphics[width=0.22\linewidth]{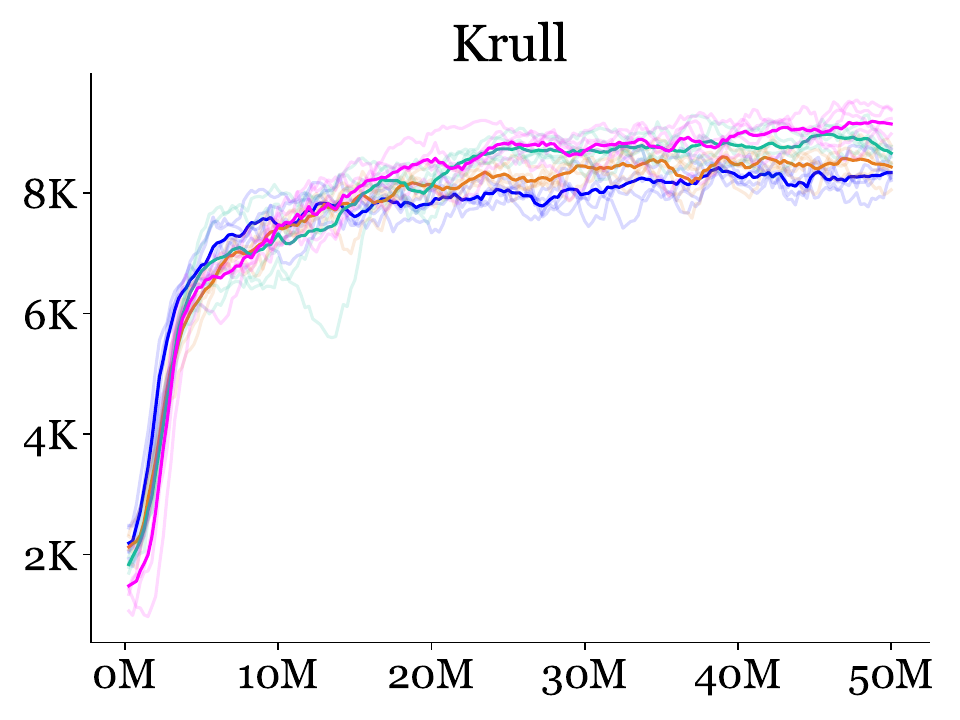} 
	\includegraphics[width=0.22\linewidth]{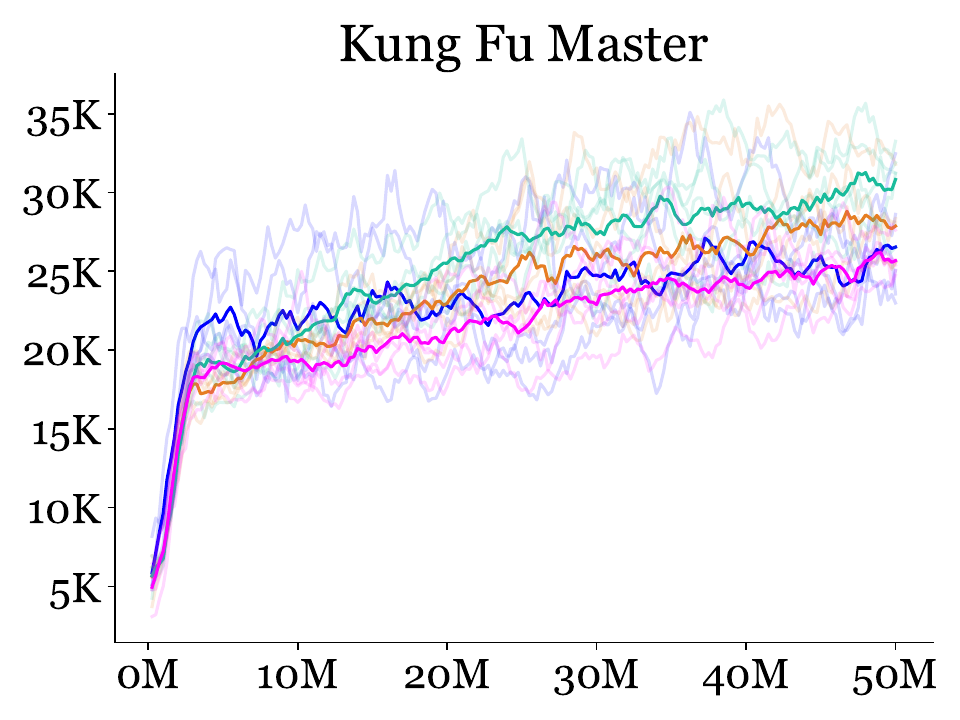} 
	\includegraphics[width=0.22\linewidth]{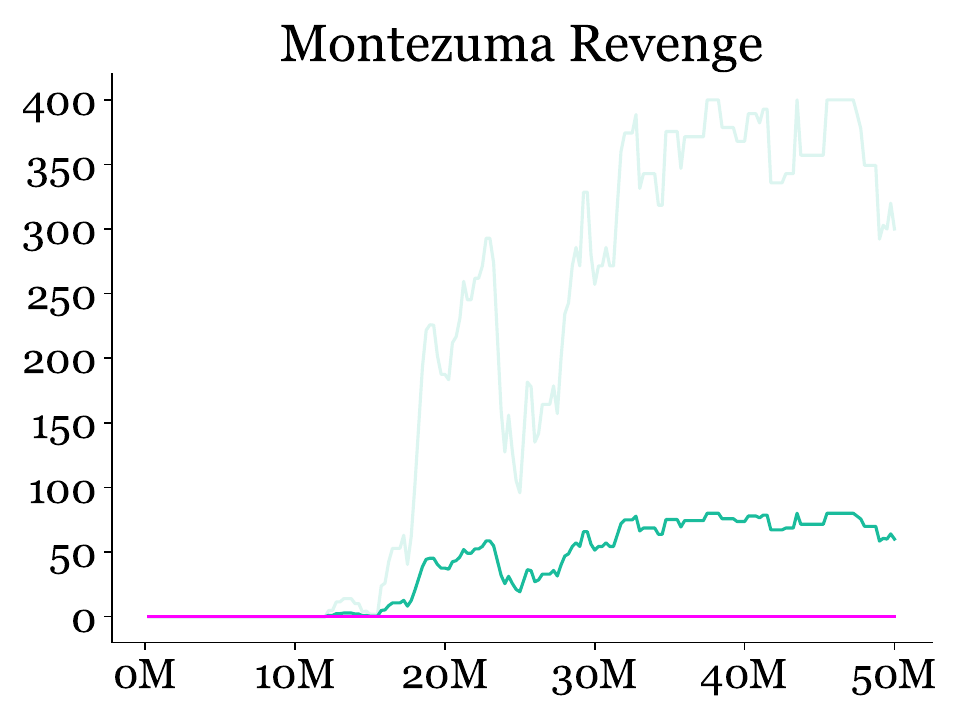} 
	\includegraphics[width=0.22\linewidth]{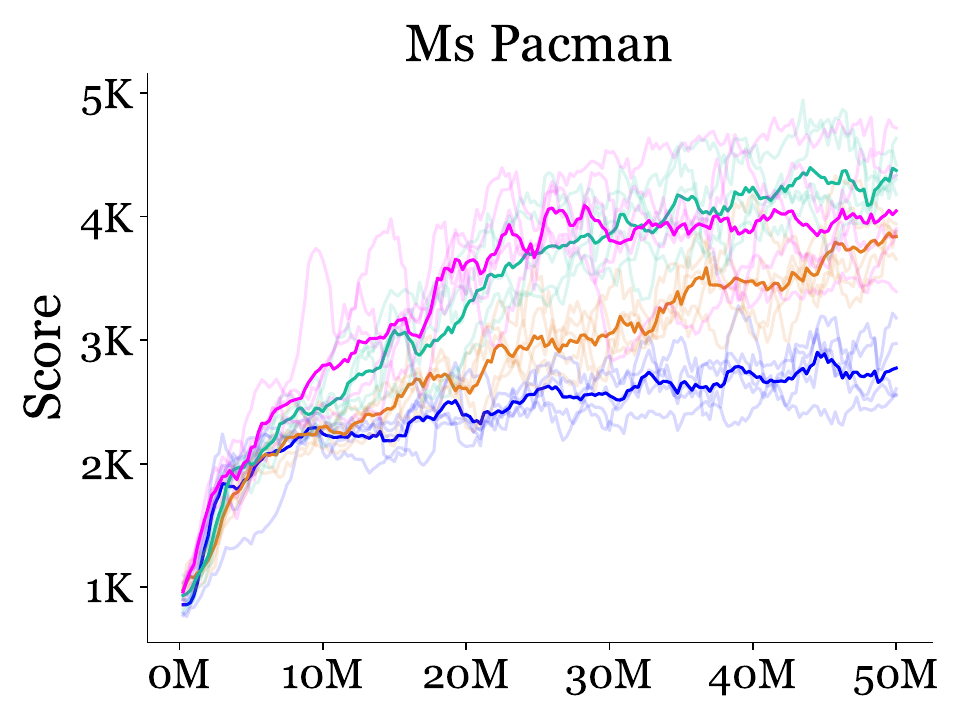} 
	\includegraphics[width=0.22\linewidth]{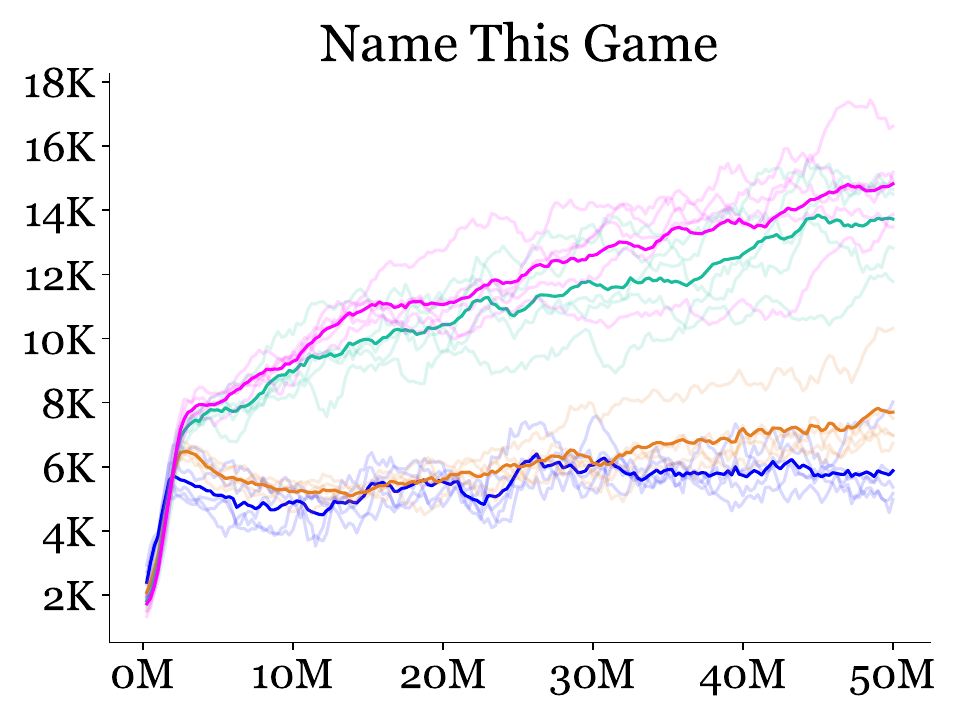} 
	\includegraphics[width=0.22\linewidth]{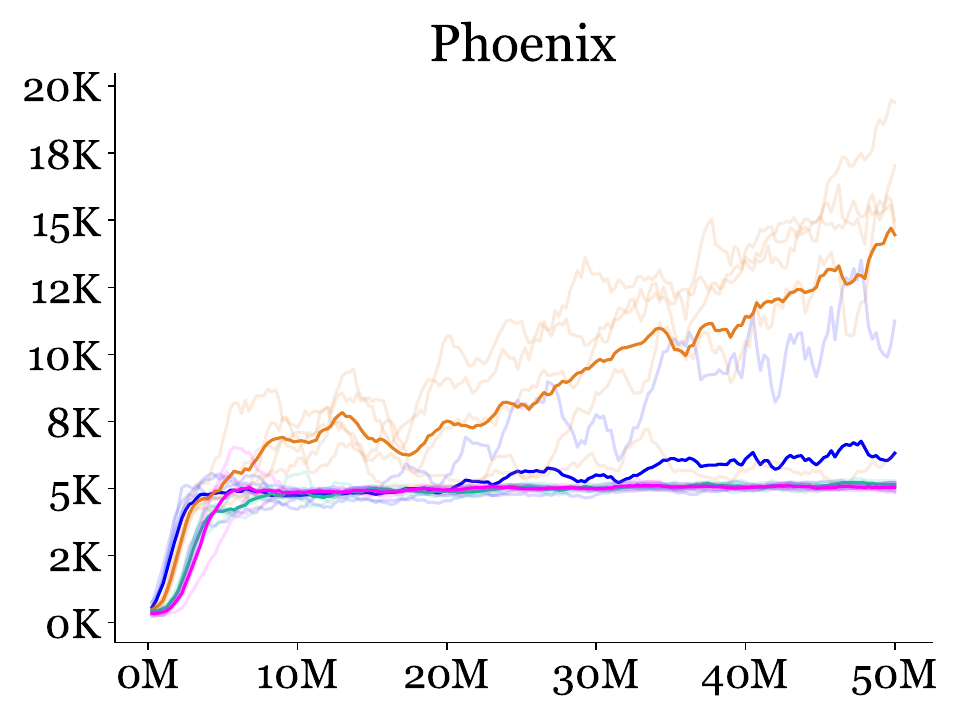} 
	\includegraphics[width=0.22\linewidth]{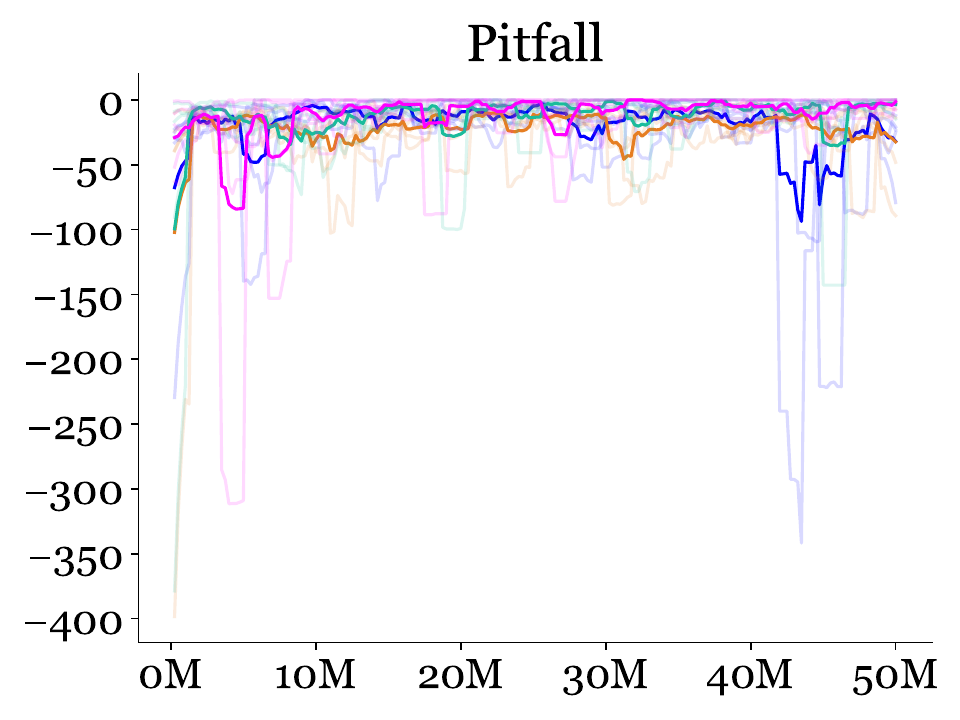} 
	\includegraphics[width=0.22\linewidth]{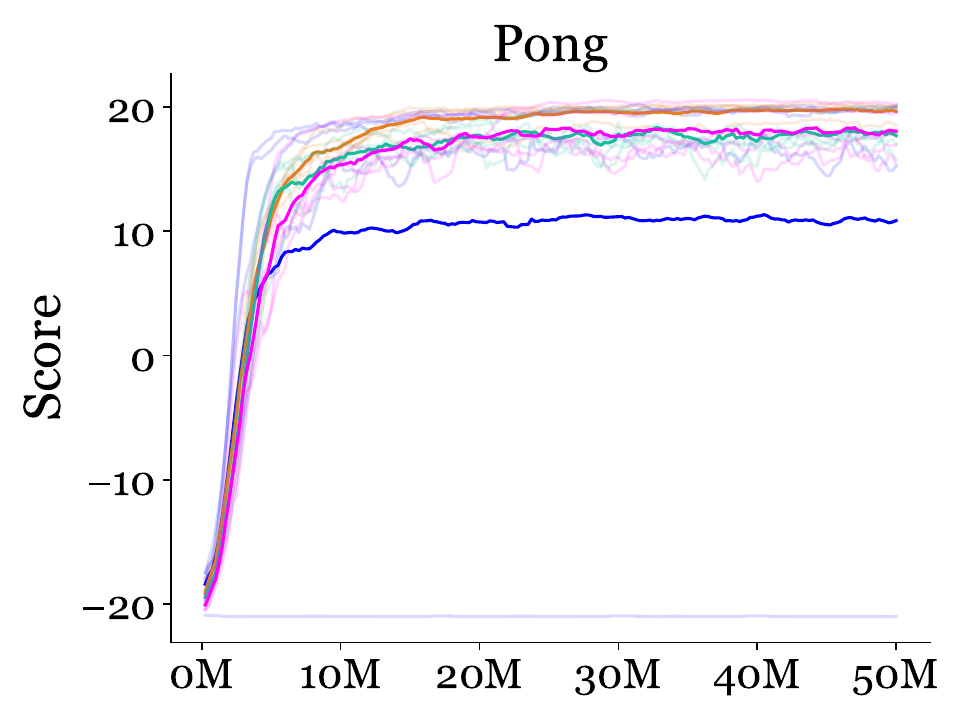} 
	\includegraphics[width=0.22\linewidth]{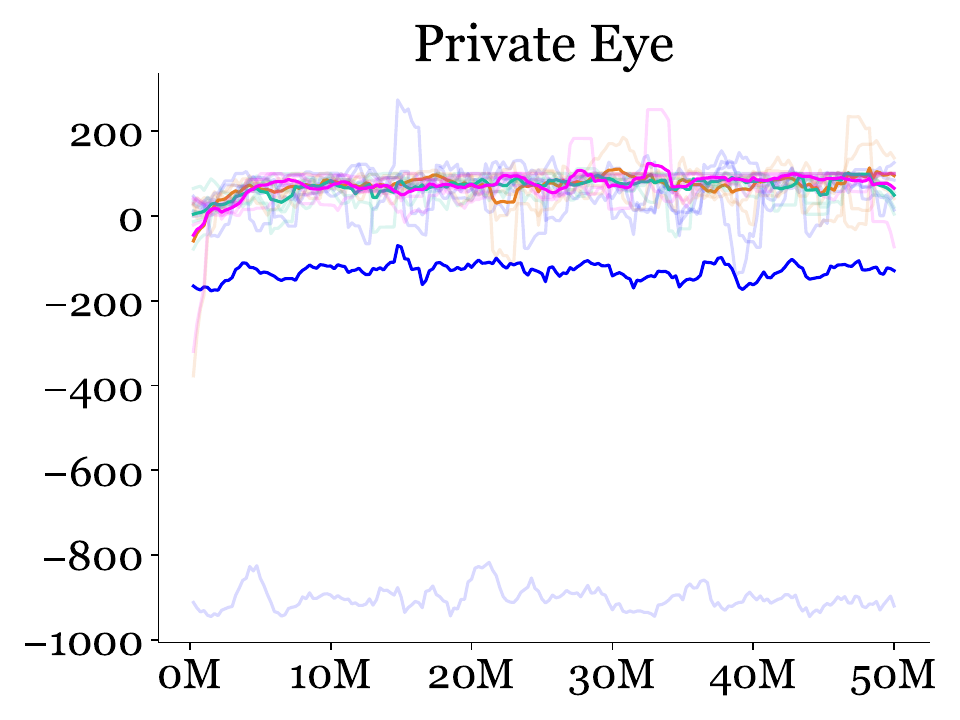} 
	\includegraphics[width=0.22\linewidth]{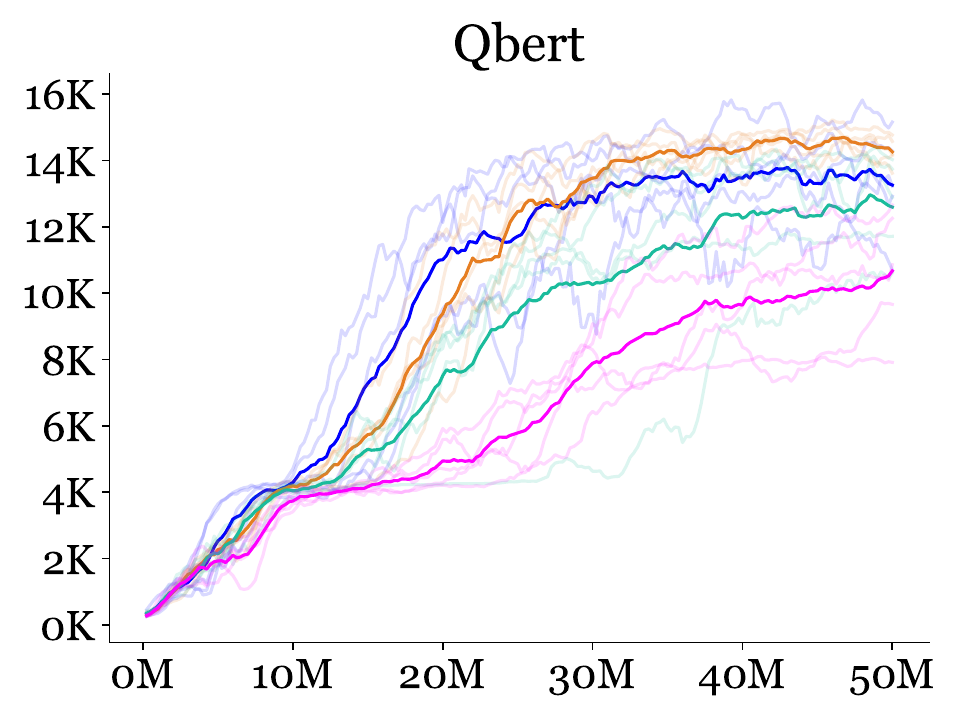} 
	\includegraphics[width=0.22\linewidth]{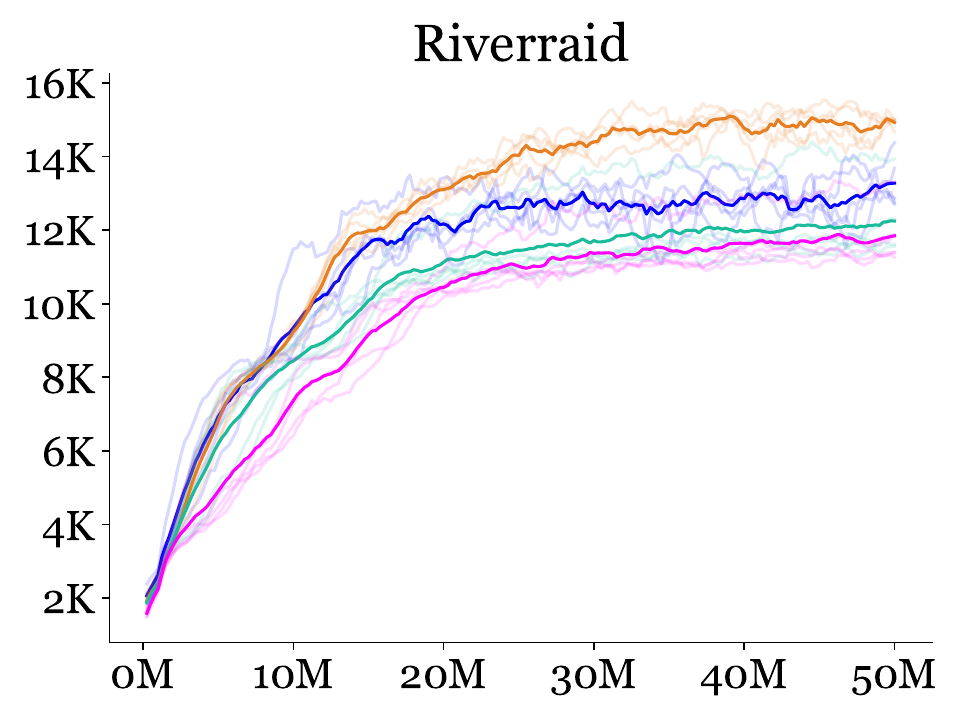} 
	\includegraphics[width=0.22\linewidth]{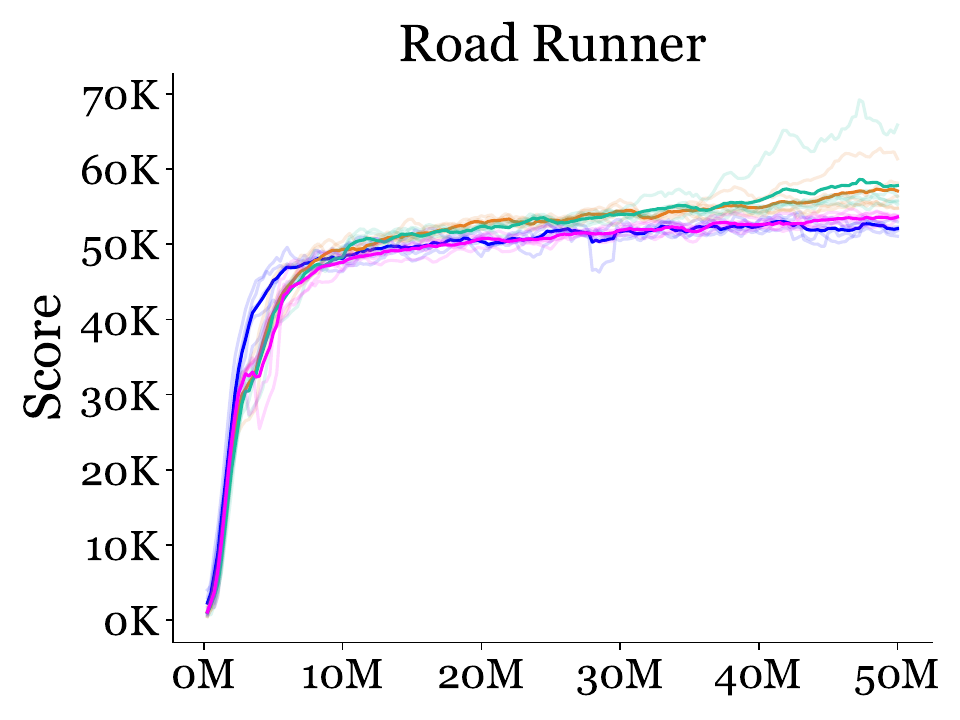} 
	\includegraphics[width=0.22\linewidth]{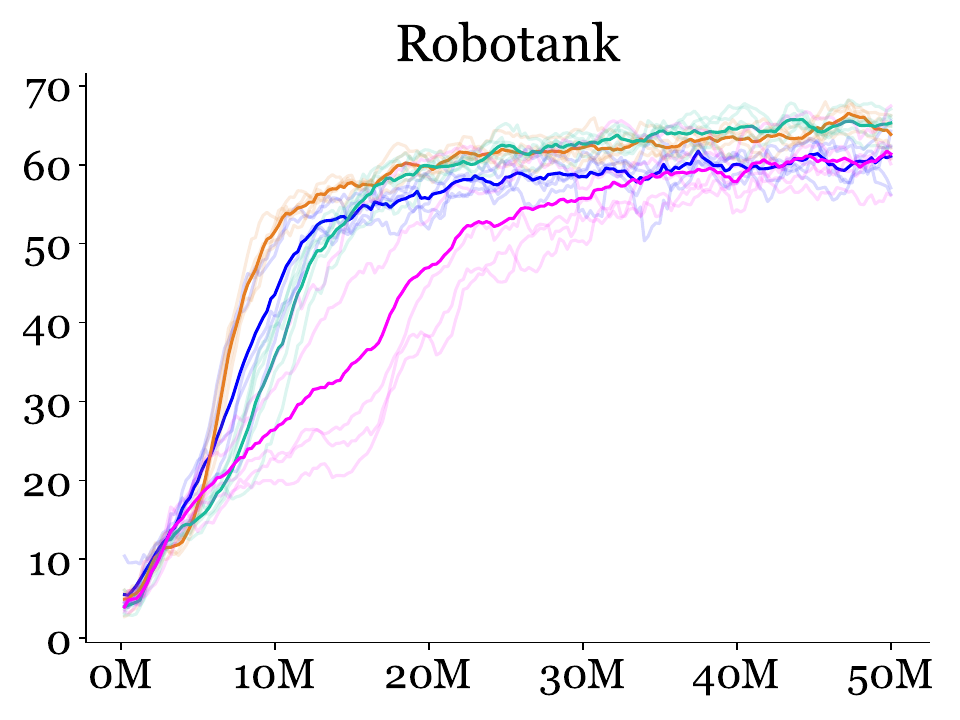} 
	\includegraphics[width=0.22\linewidth]{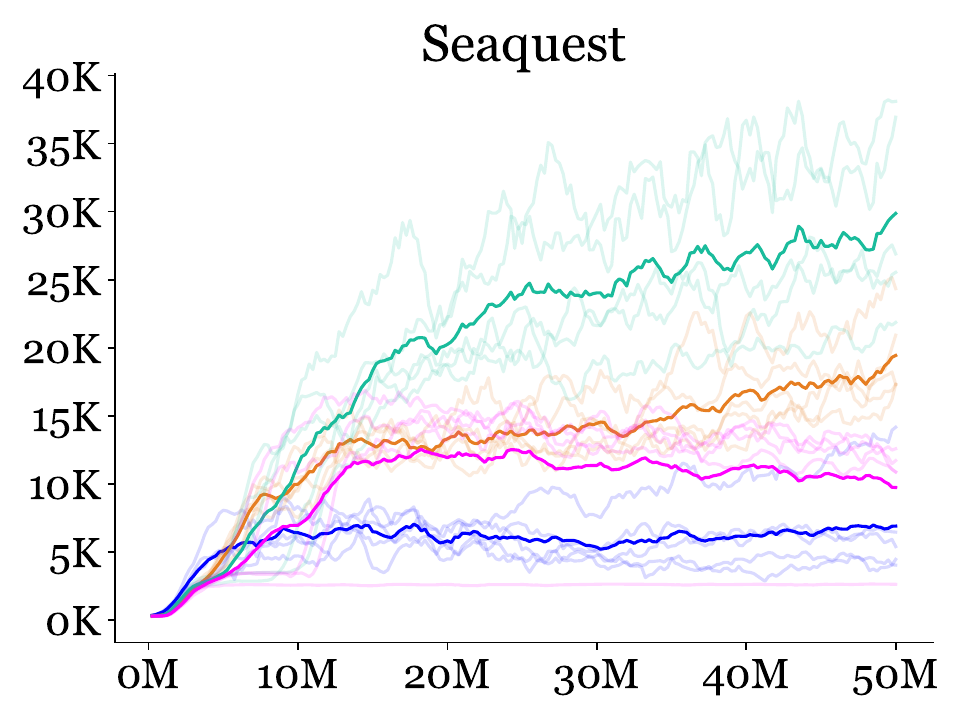} 
	\includegraphics[width=0.22\linewidth]{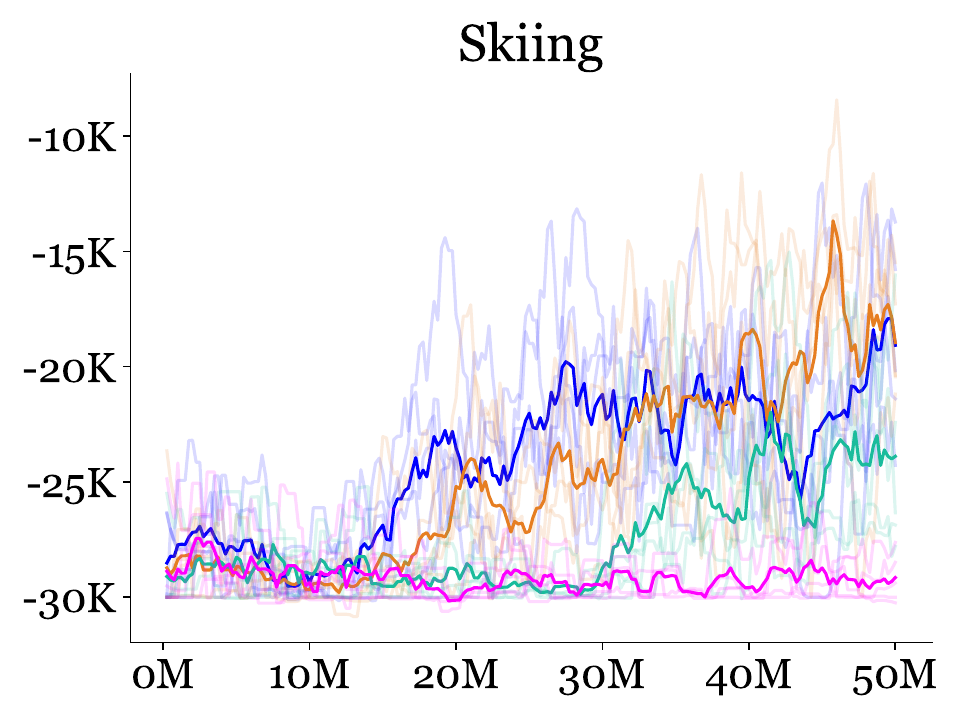} 
	\includegraphics[width=0.22\linewidth]{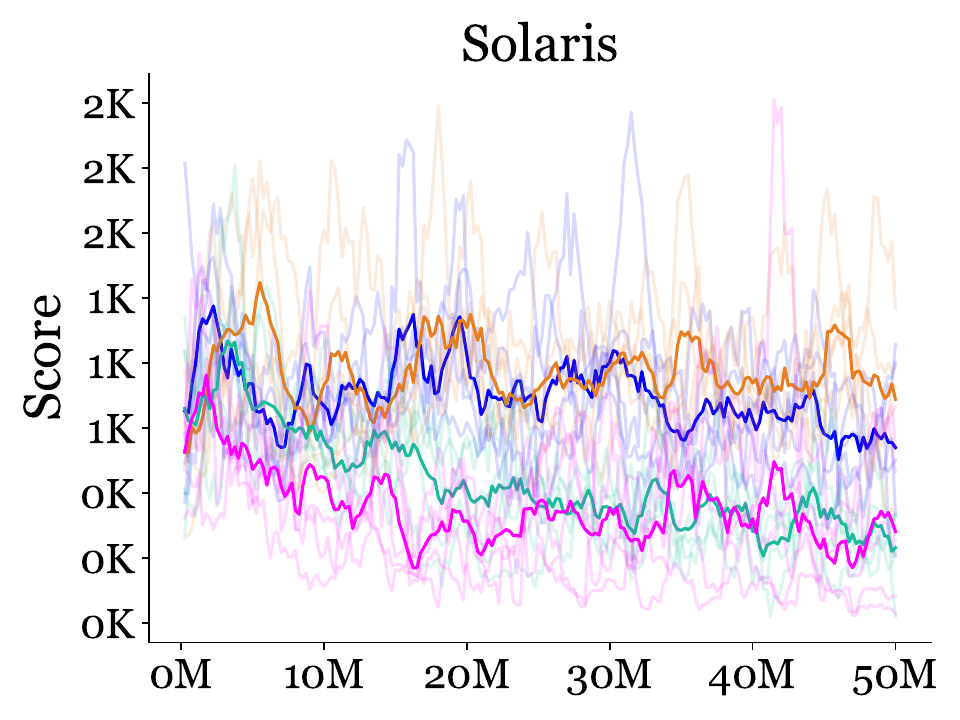} 
	\includegraphics[width=0.22\linewidth]{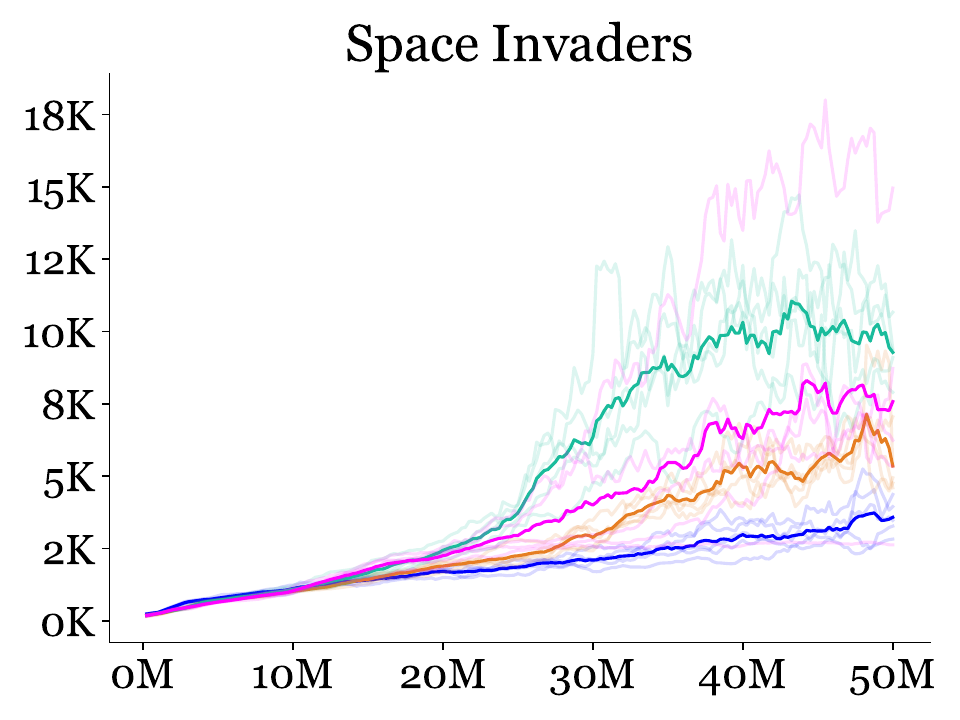} 
	\includegraphics[width=0.22\linewidth]{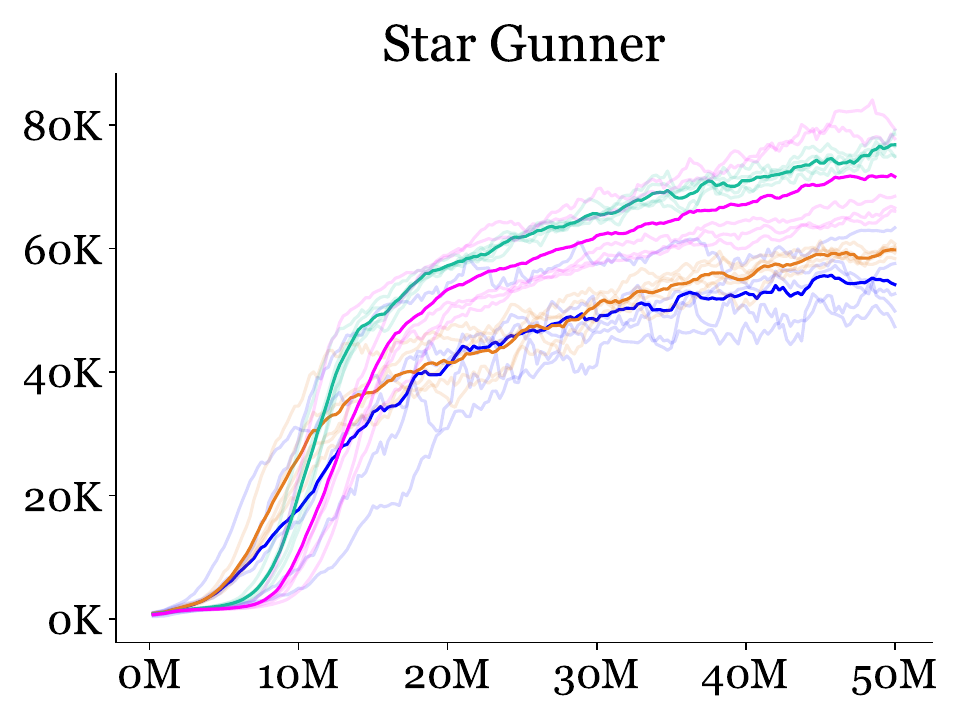} 
	\includegraphics[width=0.22\linewidth]{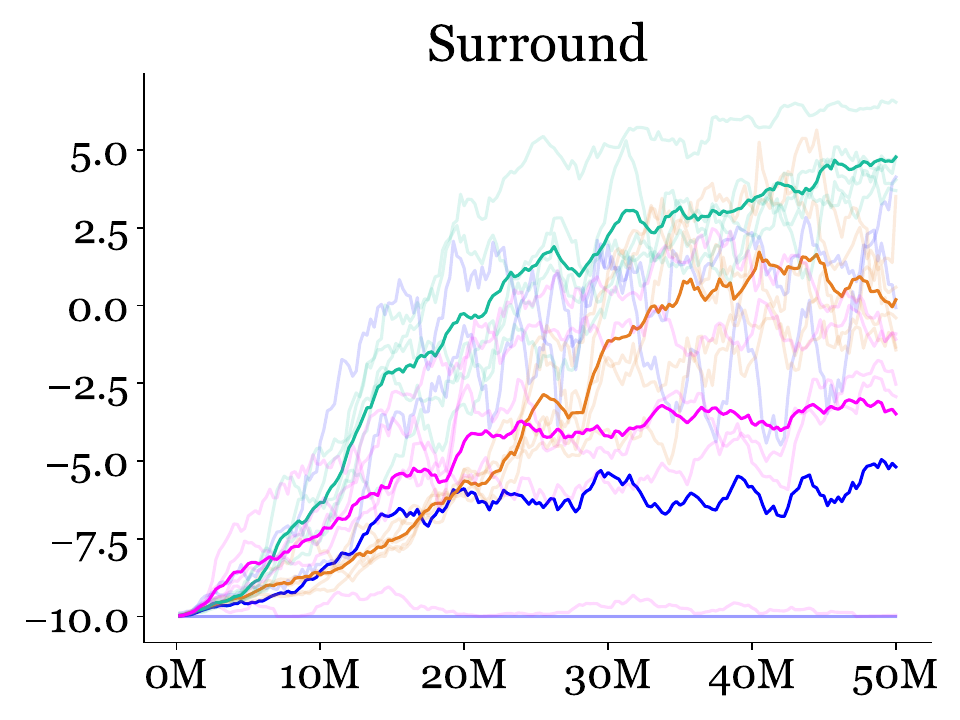} 
	\includegraphics[width=0.22\linewidth]{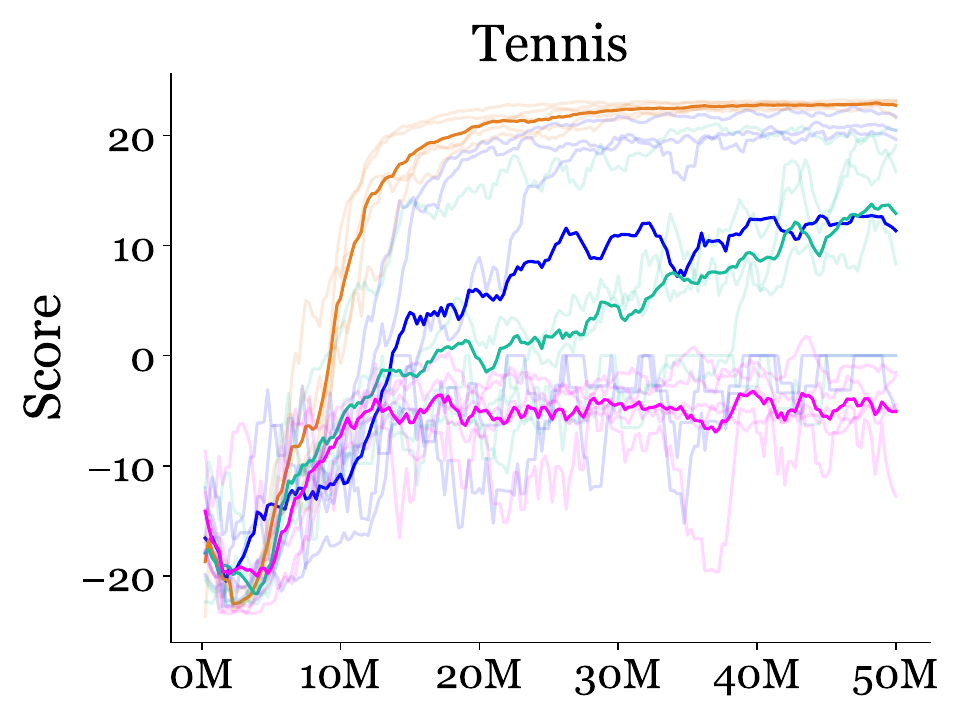} 
	\includegraphics[width=0.22\linewidth]{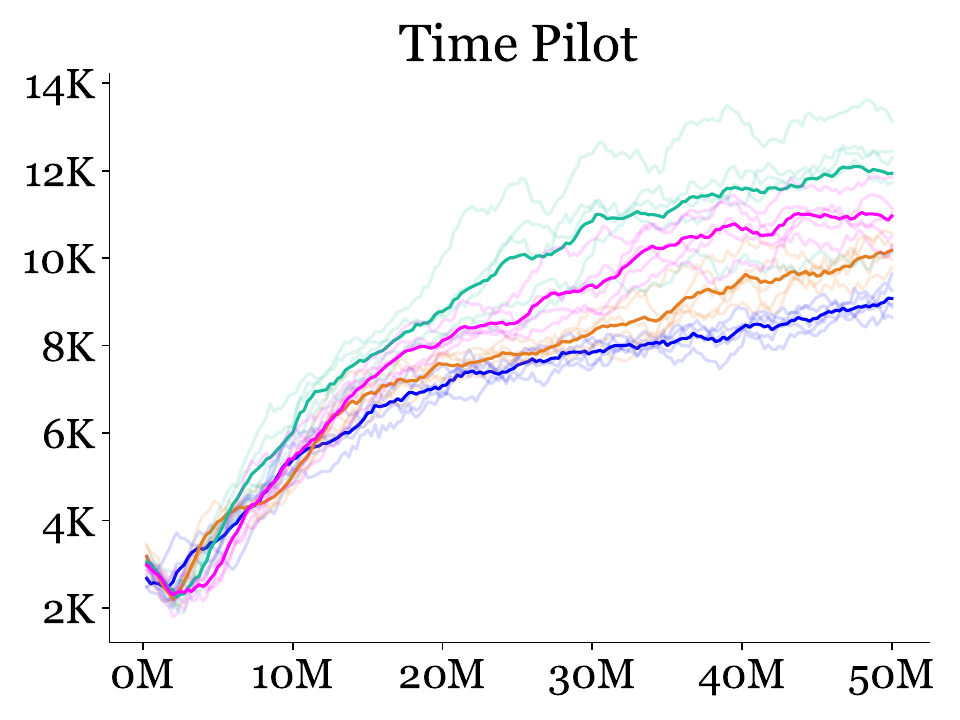} 
	\includegraphics[width=0.22\linewidth]{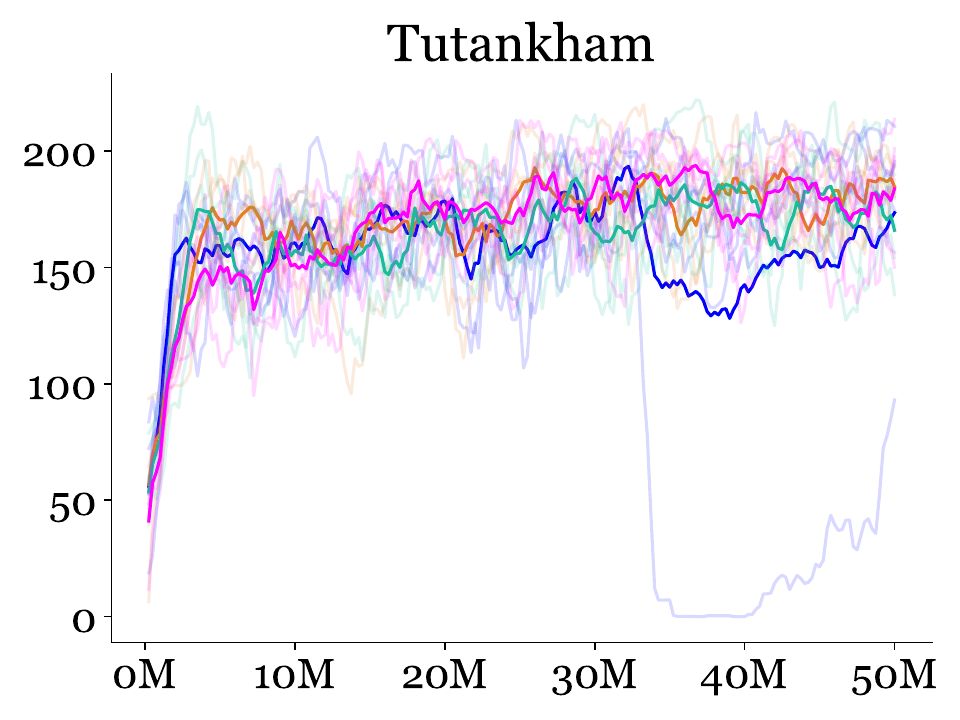} 
	\includegraphics[width=0.22\linewidth]{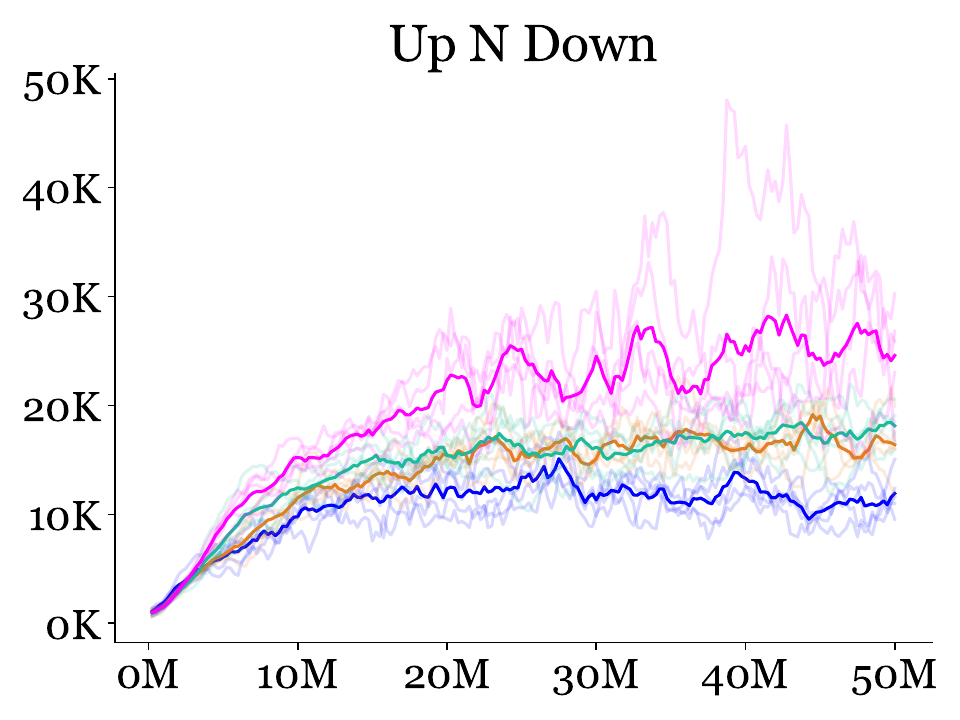} 
	\includegraphics[width=0.22\linewidth]{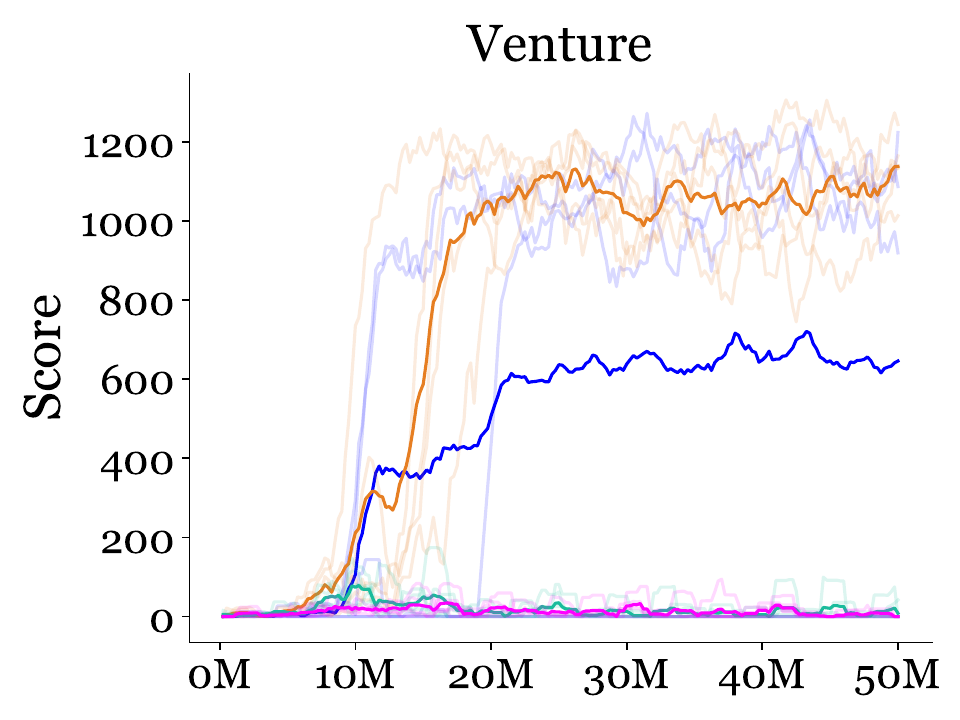} 
	\includegraphics[width=0.22\linewidth]{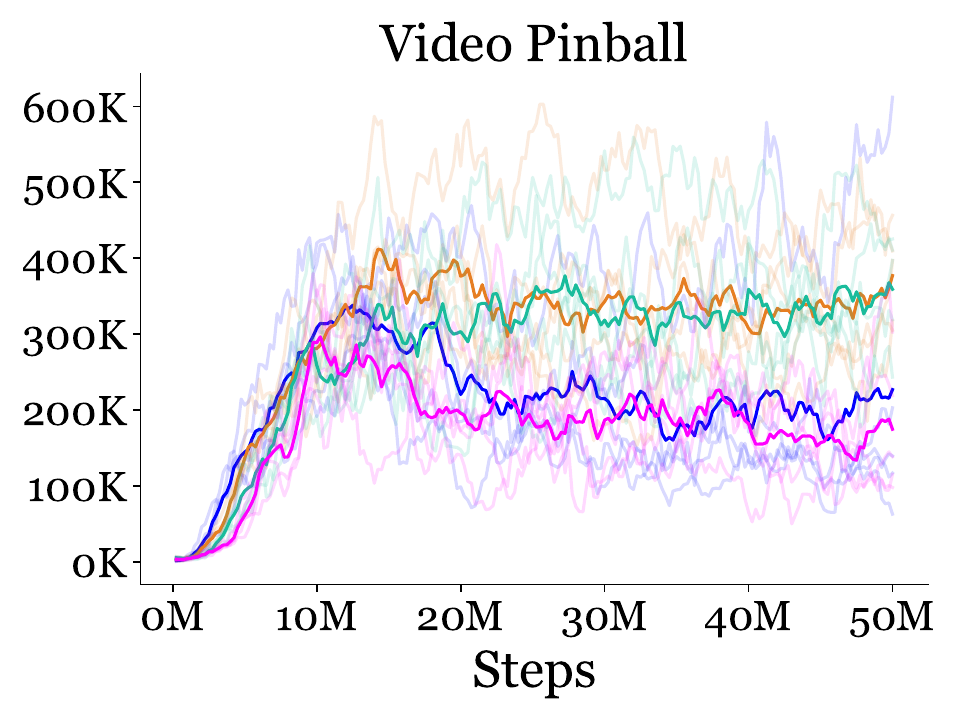} 
	\includegraphics[width=0.22\linewidth]{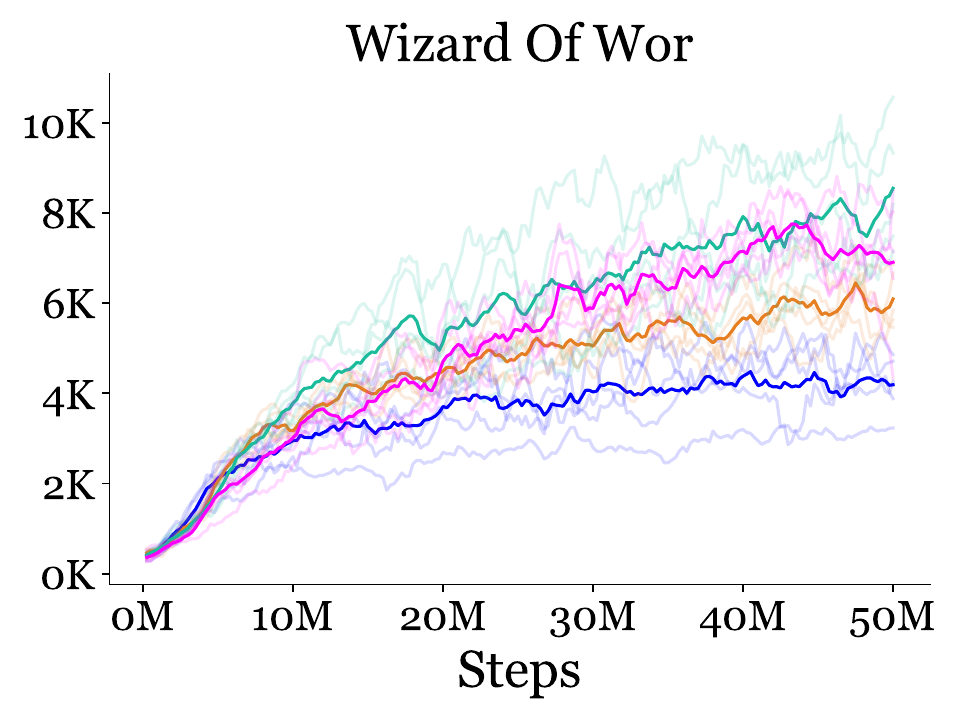} 
	\includegraphics[width=0.22\linewidth]{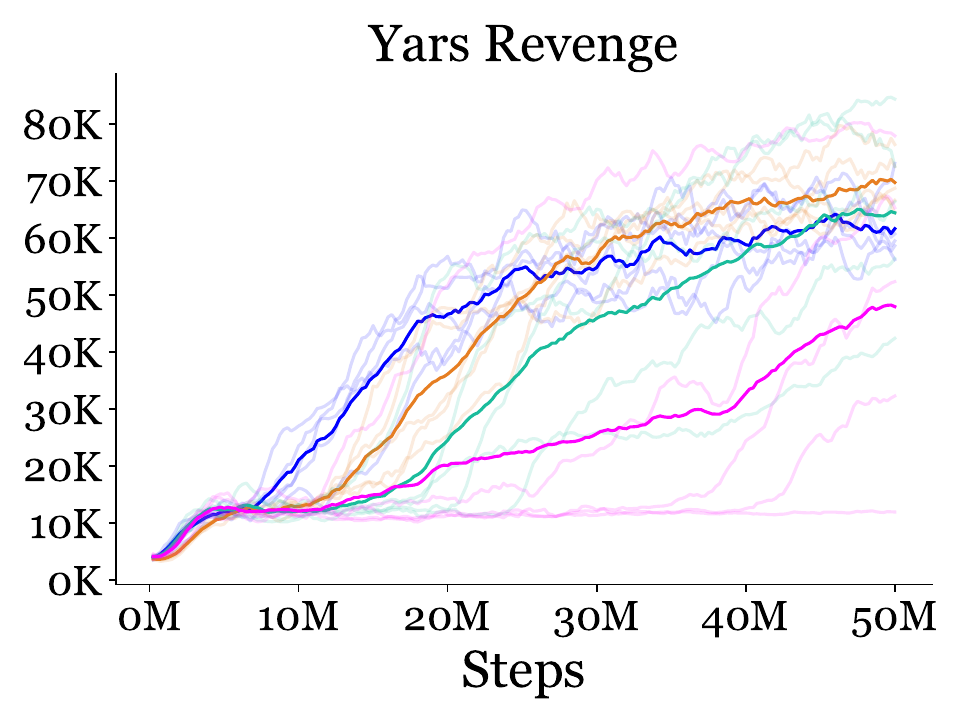} 
	\includegraphics[width=0.22\linewidth]{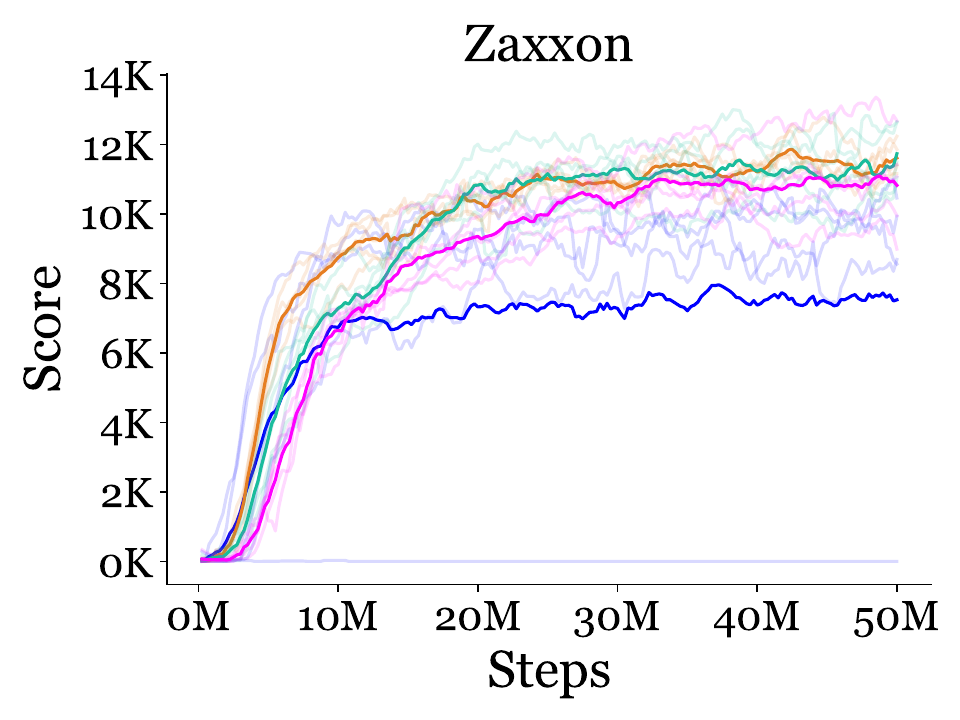} 
	\hspace{0.006\linewidth}
    \raisebox{3.5mm}[0pt][0pt]{
    {\includegraphics[width=0.62\linewidth]
    {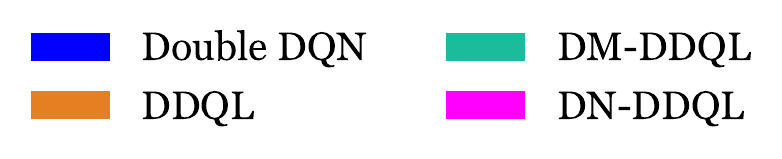}}} 
	\caption{Scores across 50M timesteps across 57 Atari 2600 games. Plots are smoothed with a moving average of seven. The semitransparent curves are the individual seeds.}
	\label{Atari57:Score:page_2}
\end{figure}

Figure~\ref{Atari57:Score:page_2} depicts the performance curves throughout training for Double DQN, DDQL, DM-DDQL, and DN-DDQL, across all games and seeds.
These curves depict the raw game scores instead of the human-normalized scores.
These curves depict the data used to produce 
Figure~\ref{fig:combined_hns_results}.
Figure~\ref{fig:per_game_dn_ddql_perf} depicts the per-game improvements of DN-DDQL over Double DQN.

Table~\ref{table_results} reports the mean score across the final three evaluations for each of our four algorithms in all 57 games for five seeds.
Note that this differs from standard reporting protocols, which often use the best-performing checkpoint and re-evaluate it.
Prior evaluation protocols should be expected to give better results than using the final evaluation~\citep{pfrl}.
Moreover, it should also be noted that we are using sticky actions, with the full action set, and game-over termination. 
As of this writing, most research in the sticky action setting still uses game-specific knowledge through the minimal action set, which should make the problem easier.

\clearpage

\begin{figure}[p]
        \centering
    	\includegraphics[width=0.22\linewidth]{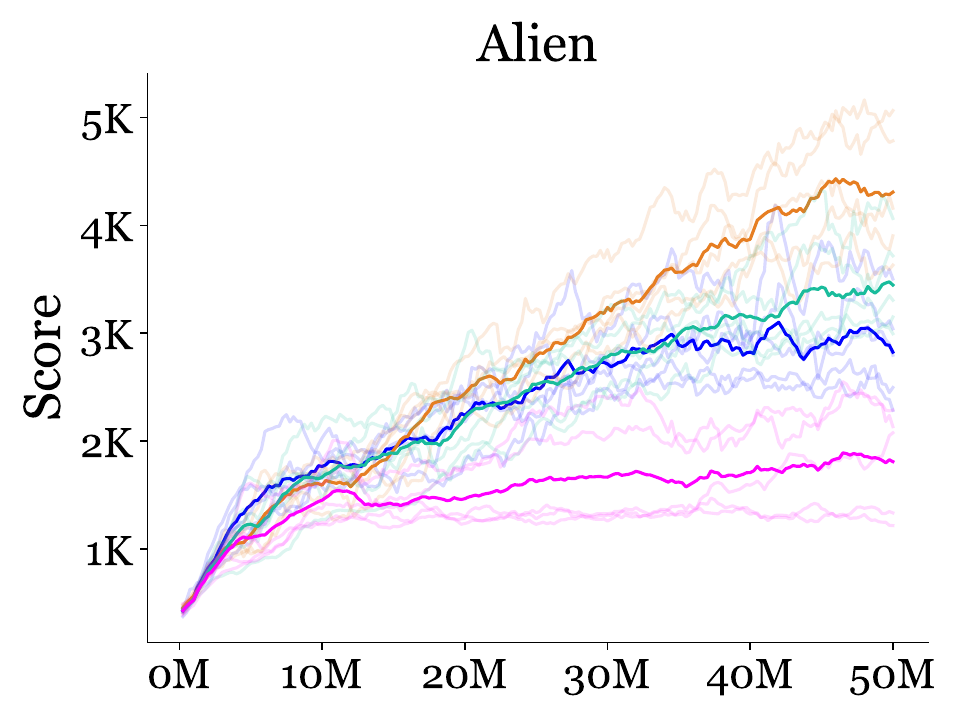} 
	\includegraphics[width=0.22\linewidth]{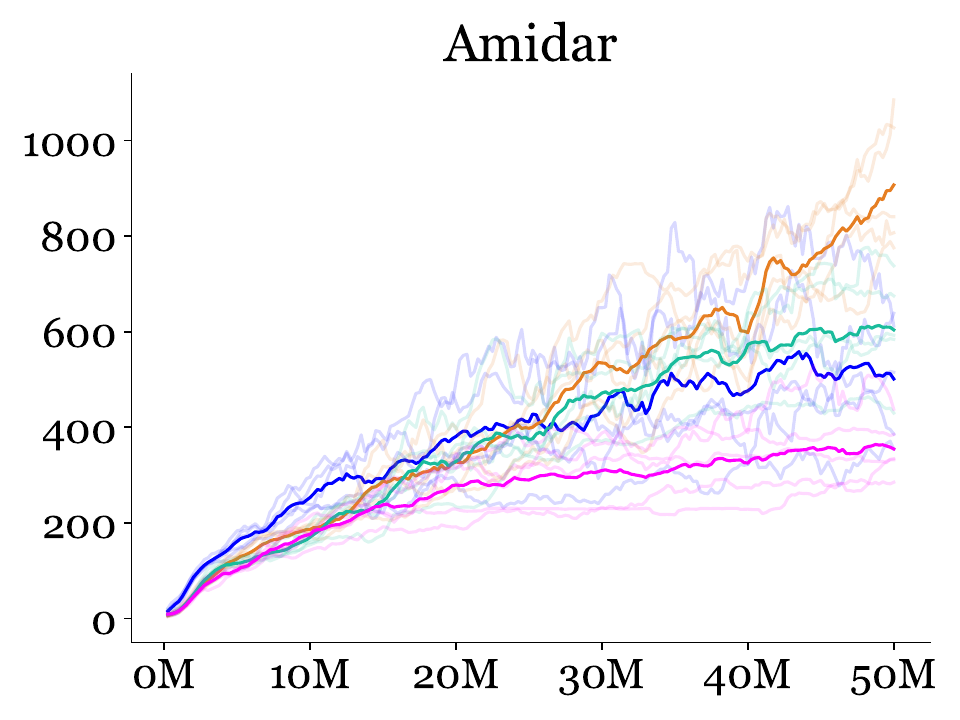} 
	\includegraphics[width=0.22\linewidth]{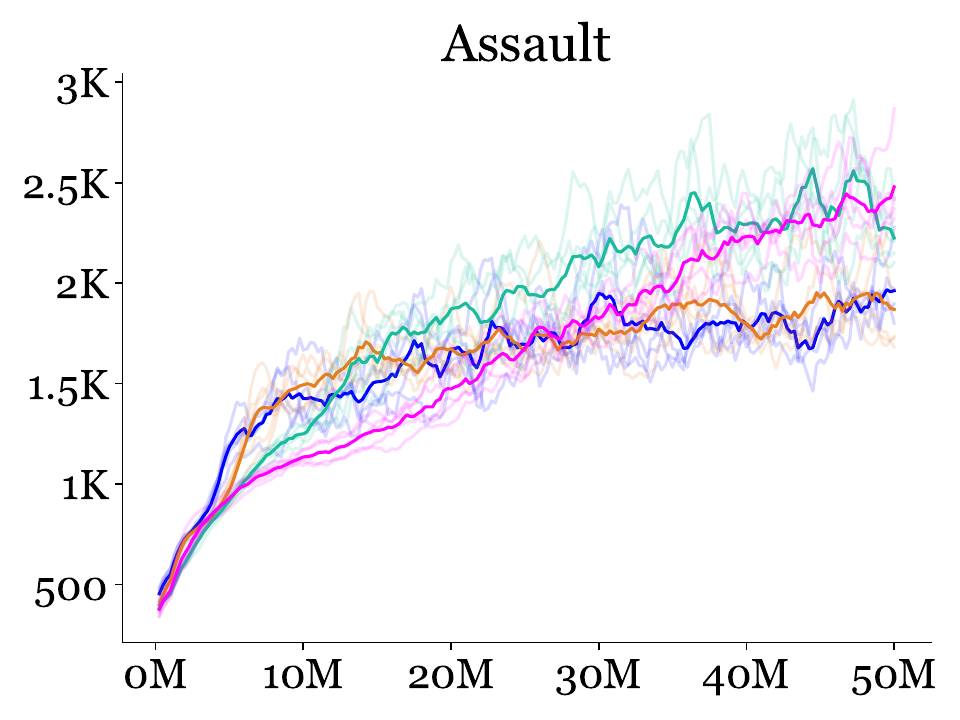} 
	\includegraphics[width=0.22\linewidth]{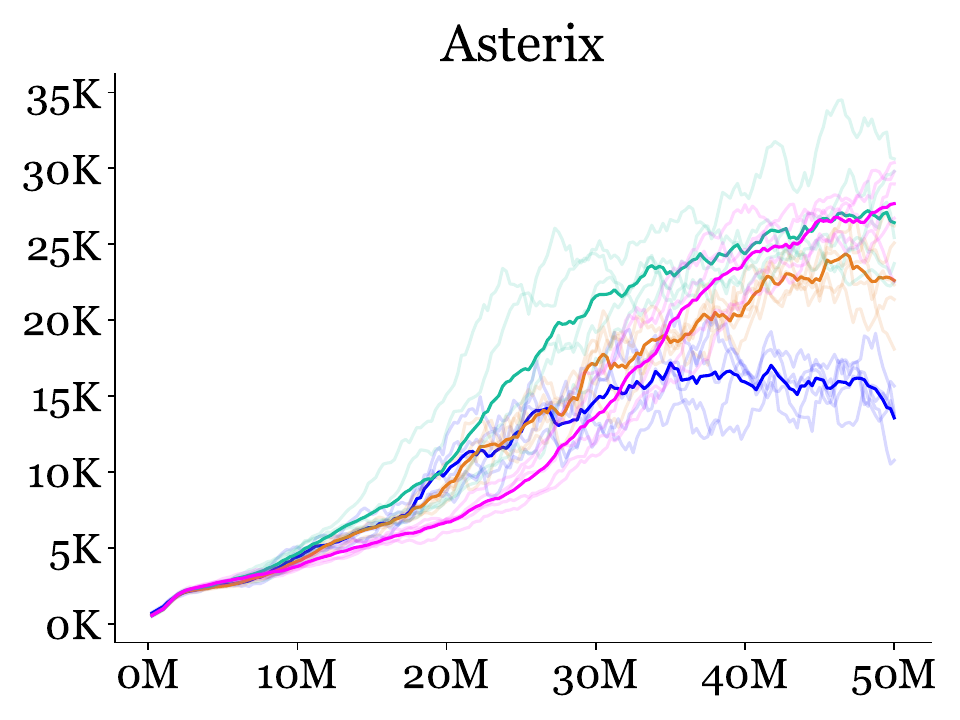} 
	\includegraphics[width=0.22\linewidth]{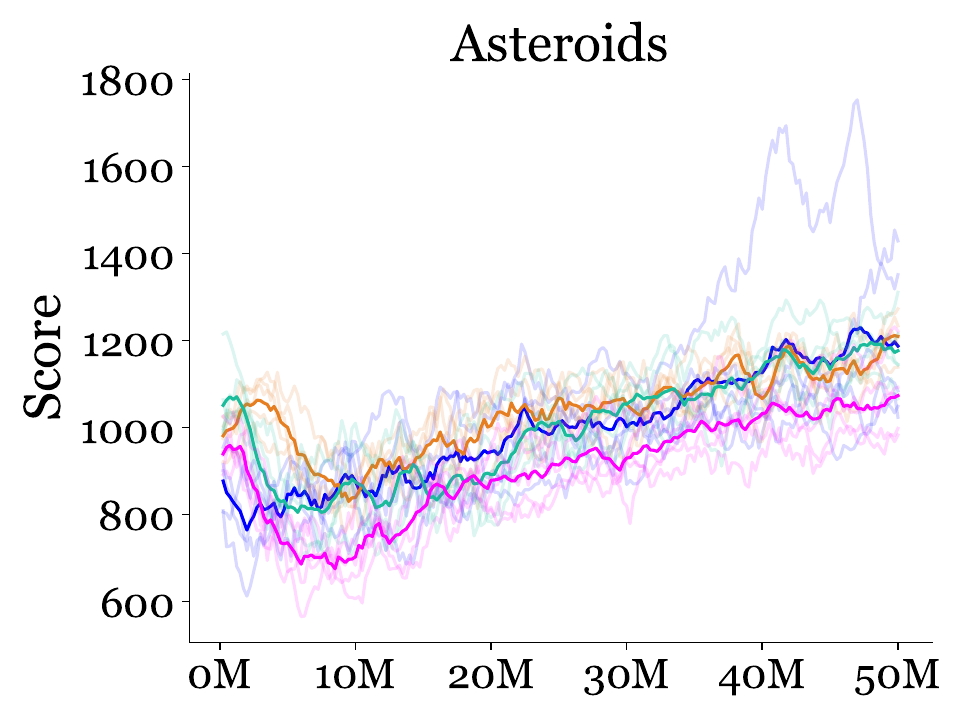} 
	\includegraphics[width=0.22\linewidth]{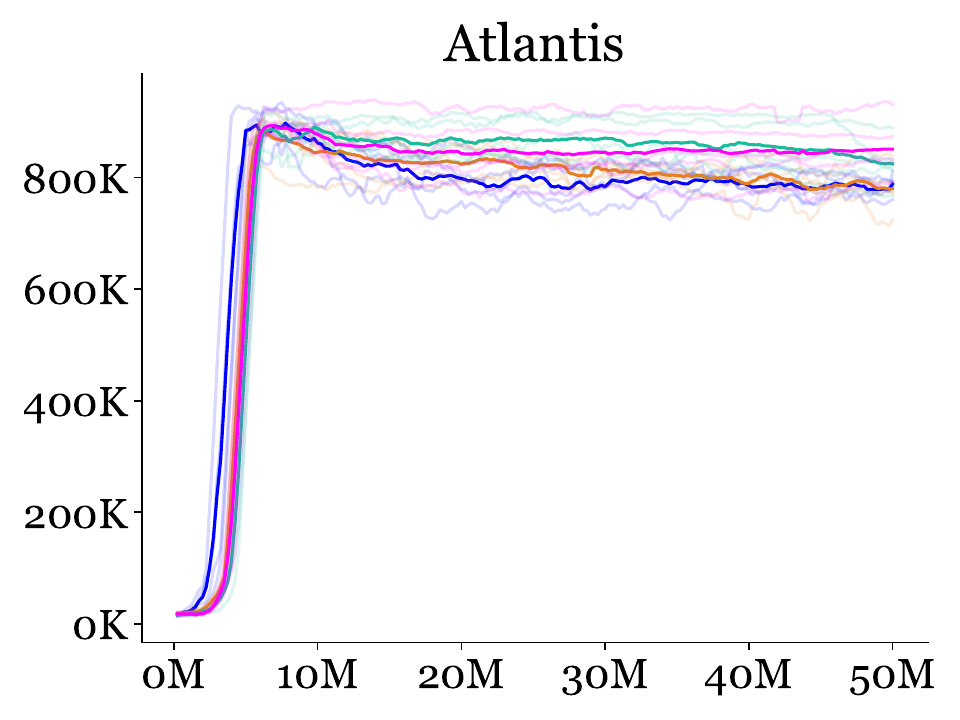} 
	\includegraphics[width=0.22\linewidth]{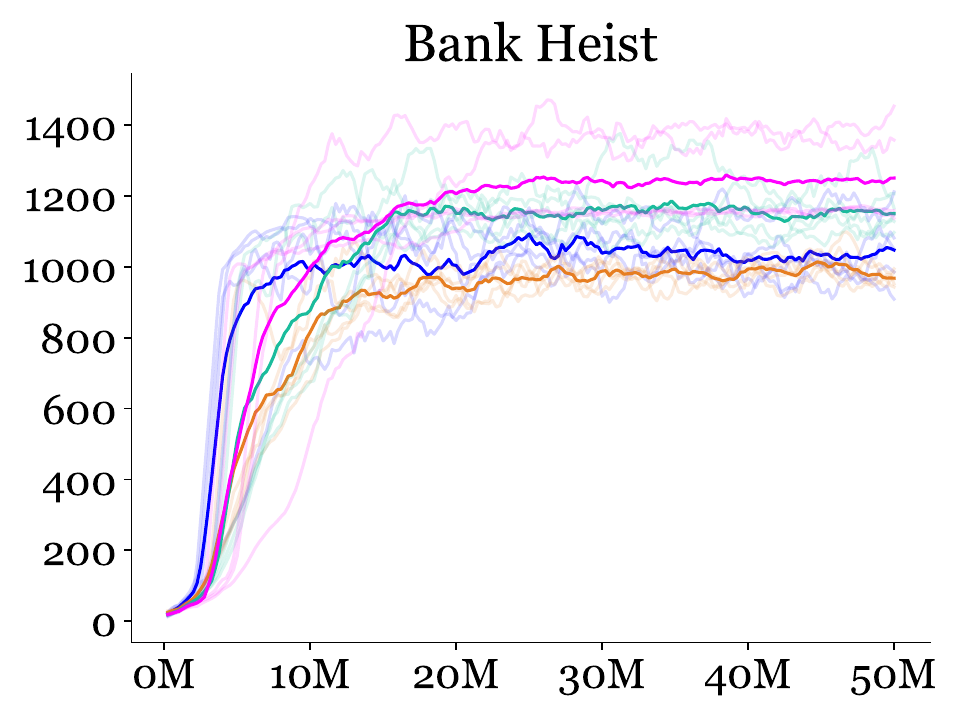} 
	\includegraphics[width=0.22\linewidth]{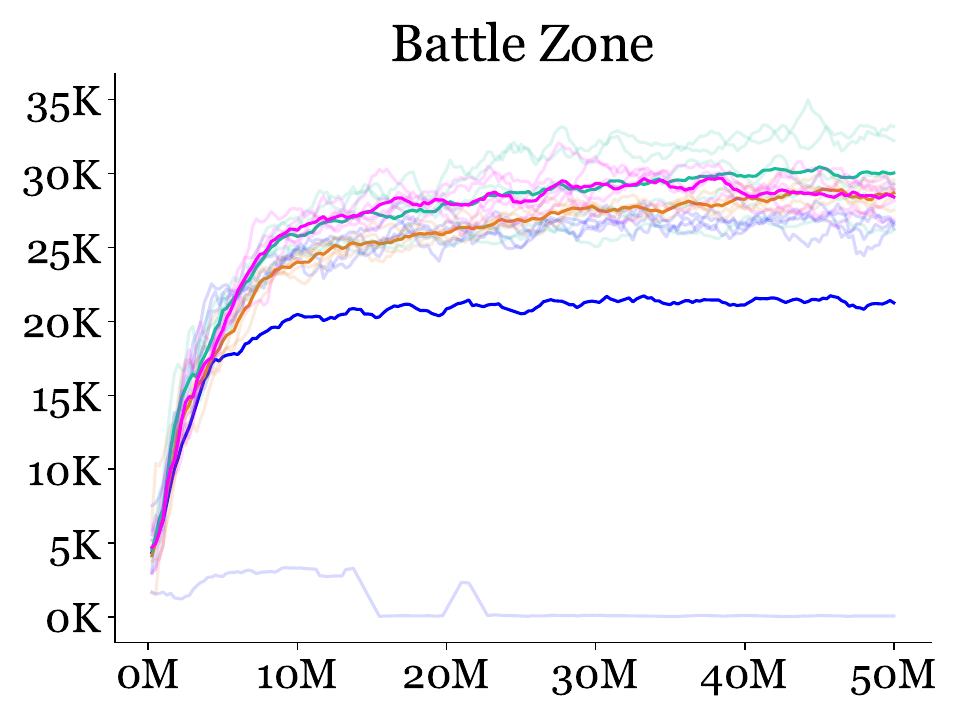} 
	\includegraphics[width=0.22\linewidth]{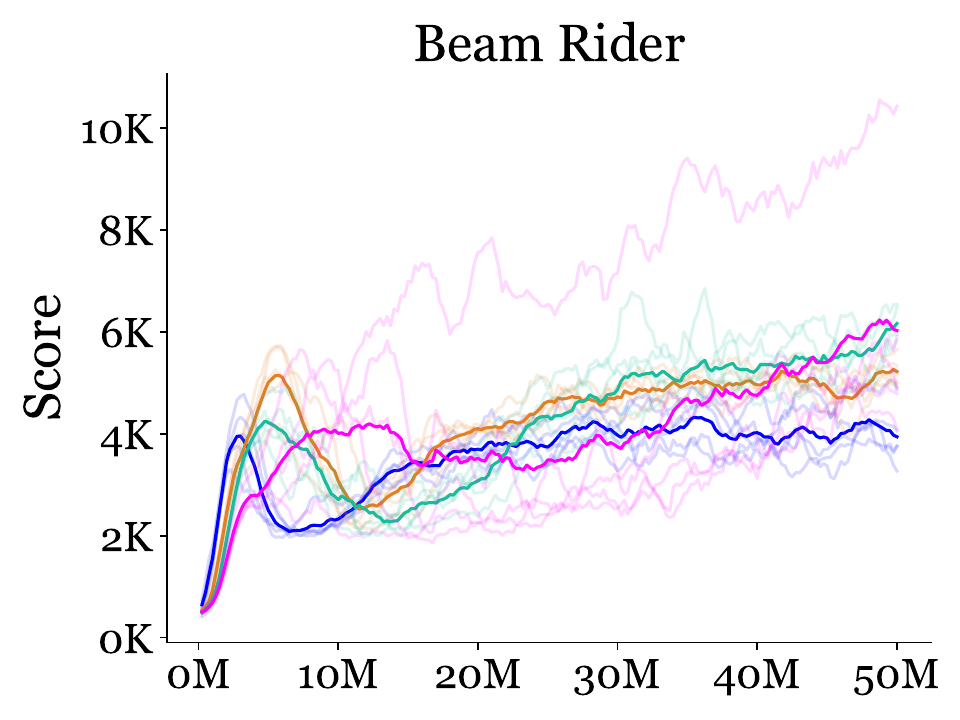} 
	\includegraphics[width=0.22\linewidth]{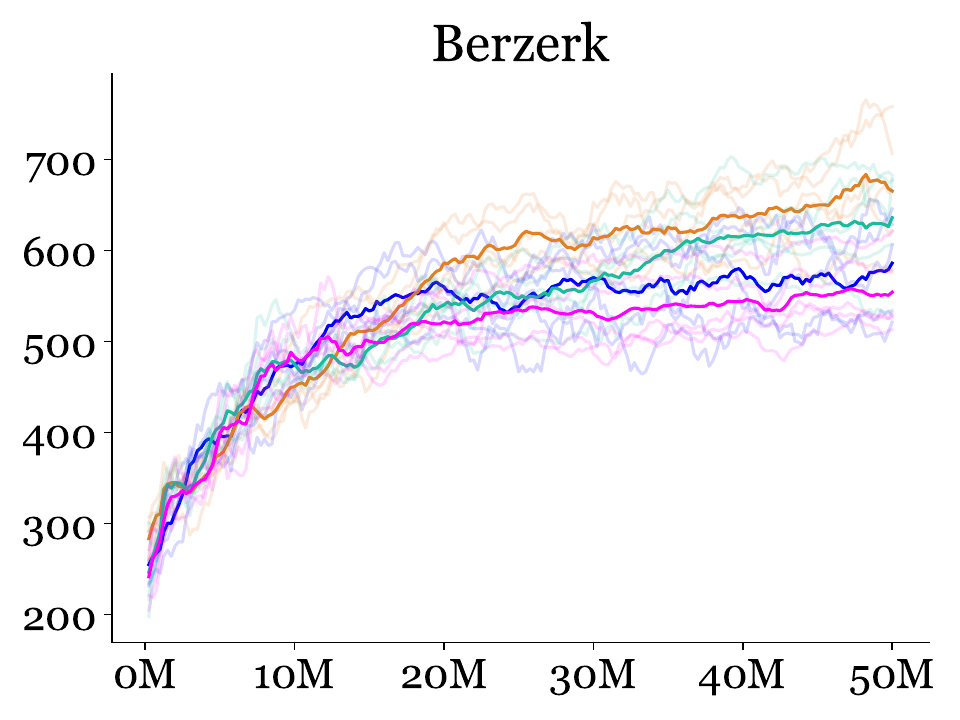} 
	\includegraphics[width=0.22\linewidth]{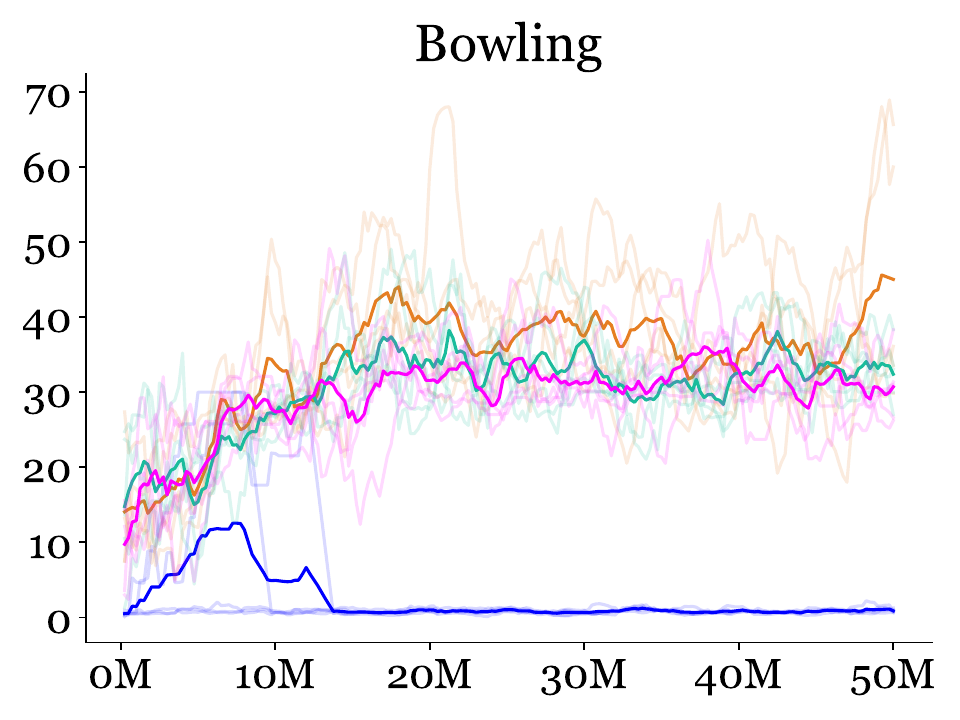} 
	\includegraphics[width=0.22\linewidth]{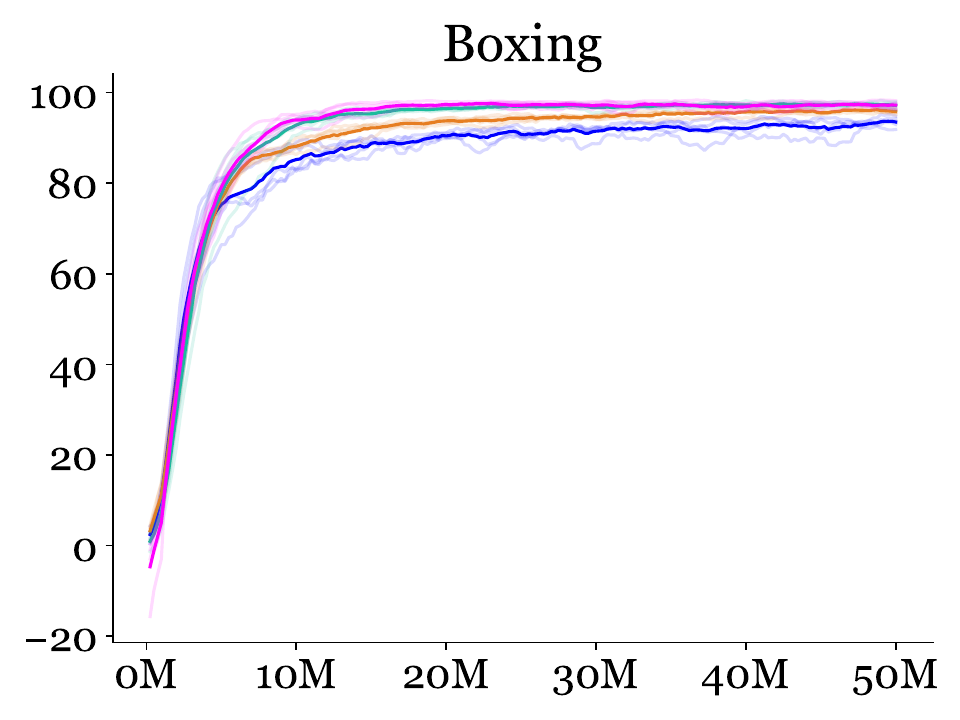} 
	\includegraphics[width=0.22\linewidth]{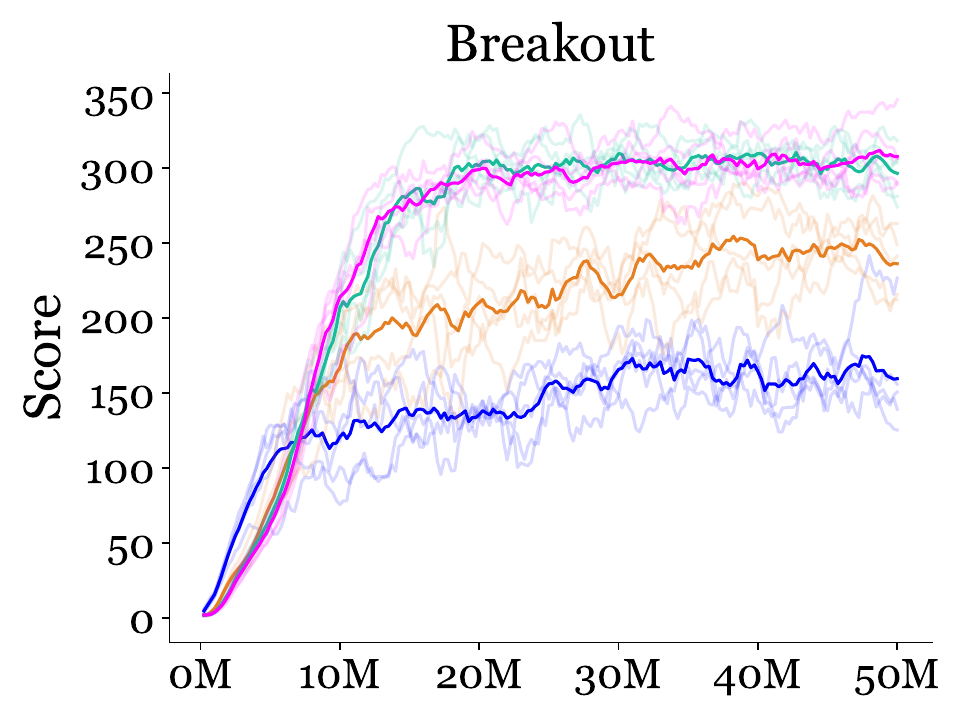} 
	\includegraphics[width=0.22\linewidth]{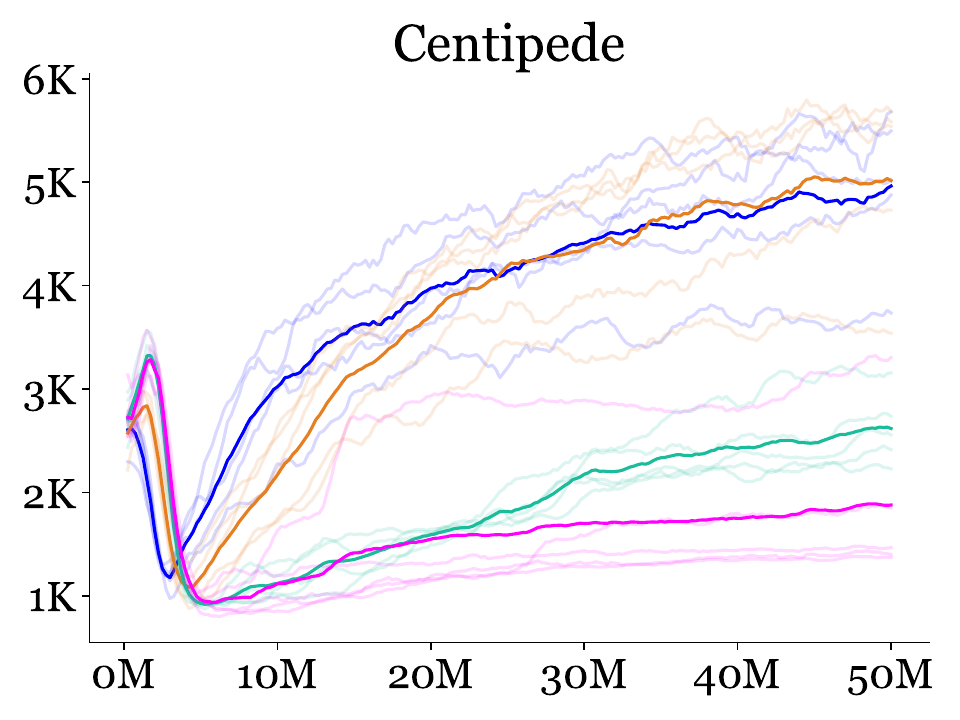} 
	\includegraphics[width=0.22\linewidth]{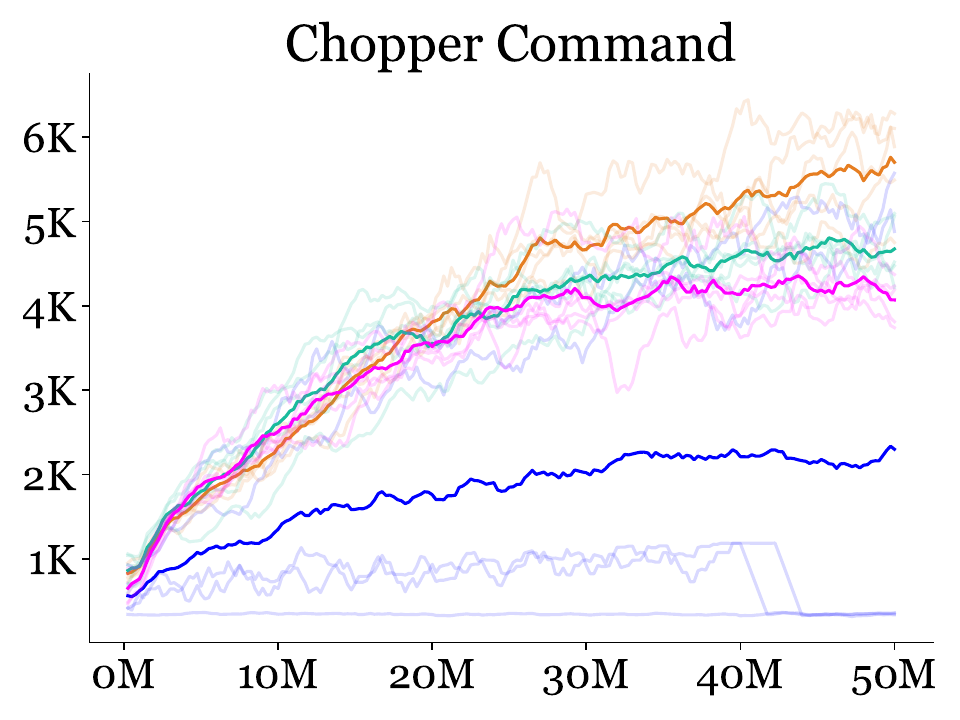} 
	\includegraphics[width=0.22\linewidth]{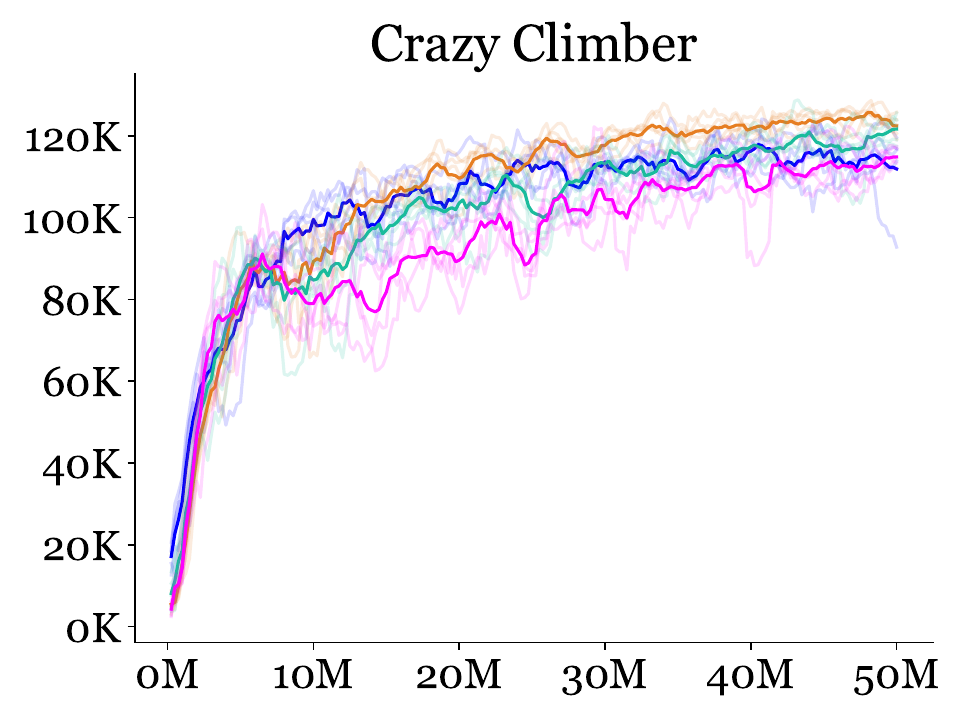} 
	\includegraphics[width=0.22\linewidth]{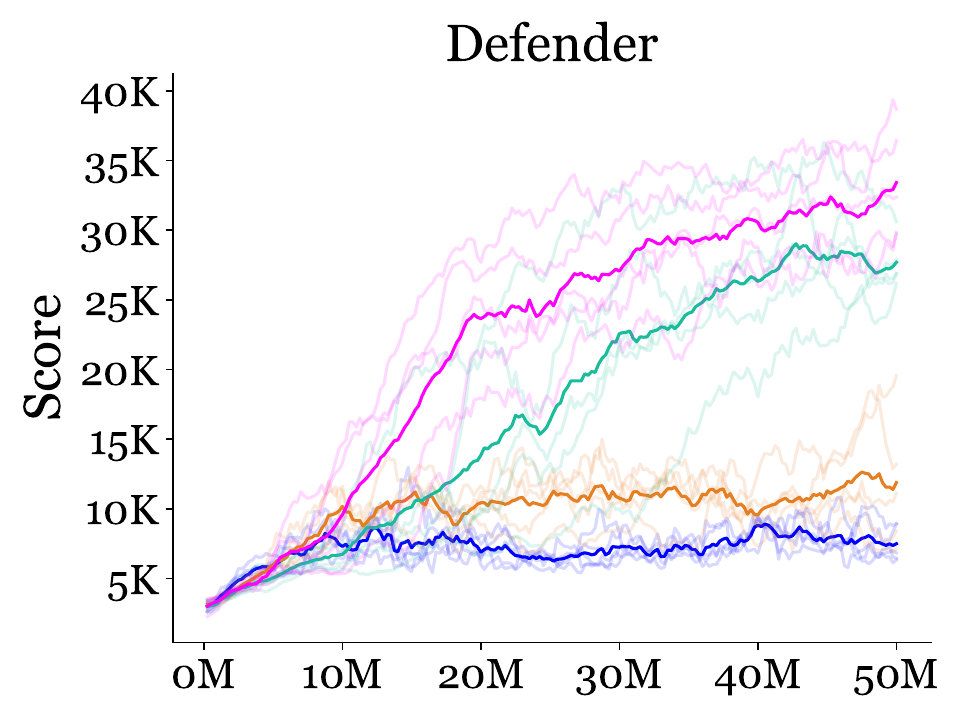} 
	\includegraphics[width=0.22\linewidth]{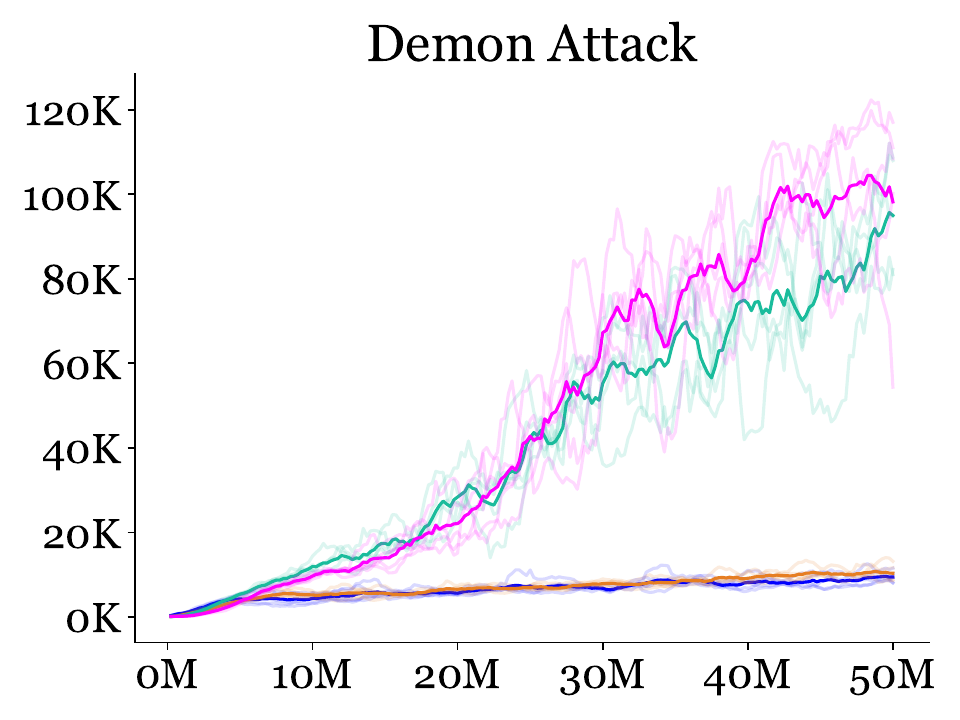} 
	\includegraphics[width=0.22\linewidth]{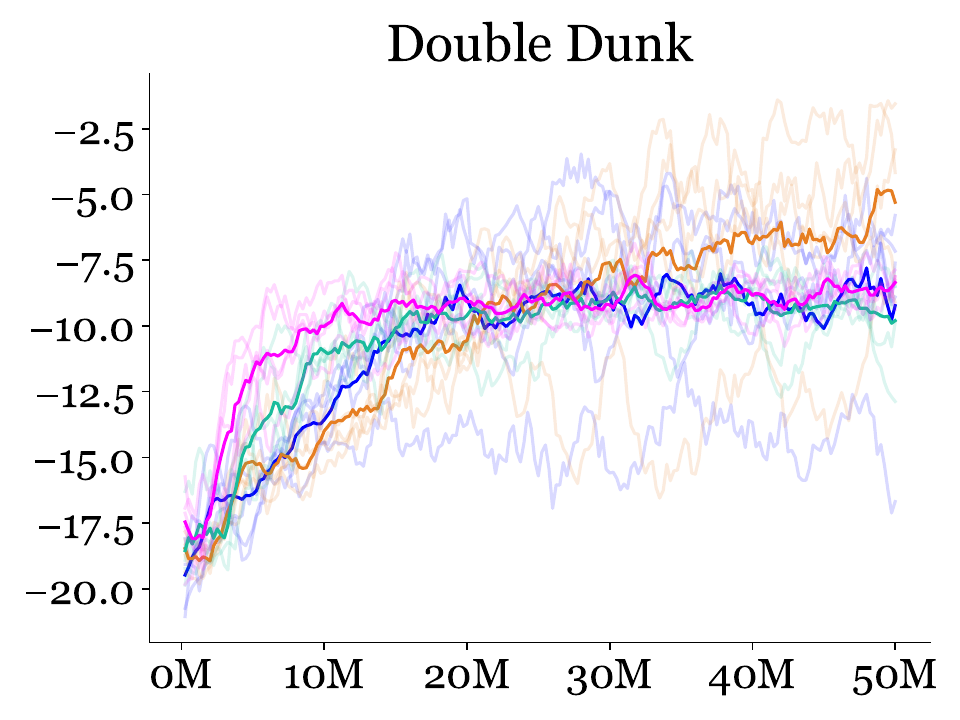} 
	\includegraphics[width=0.22\linewidth]{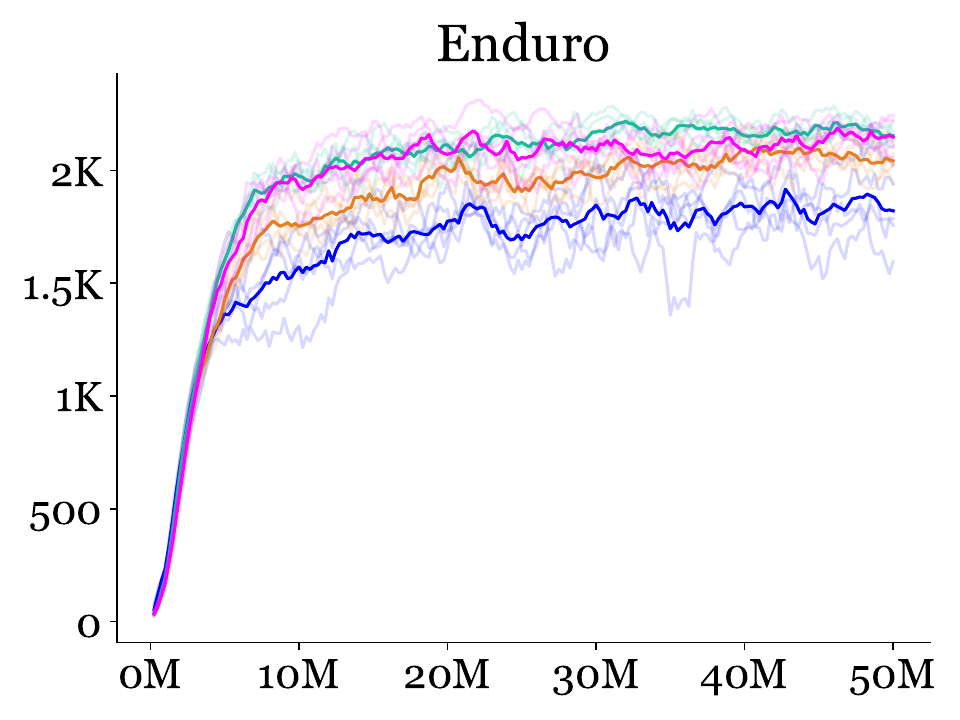} 
	\includegraphics[width=0.22\linewidth]{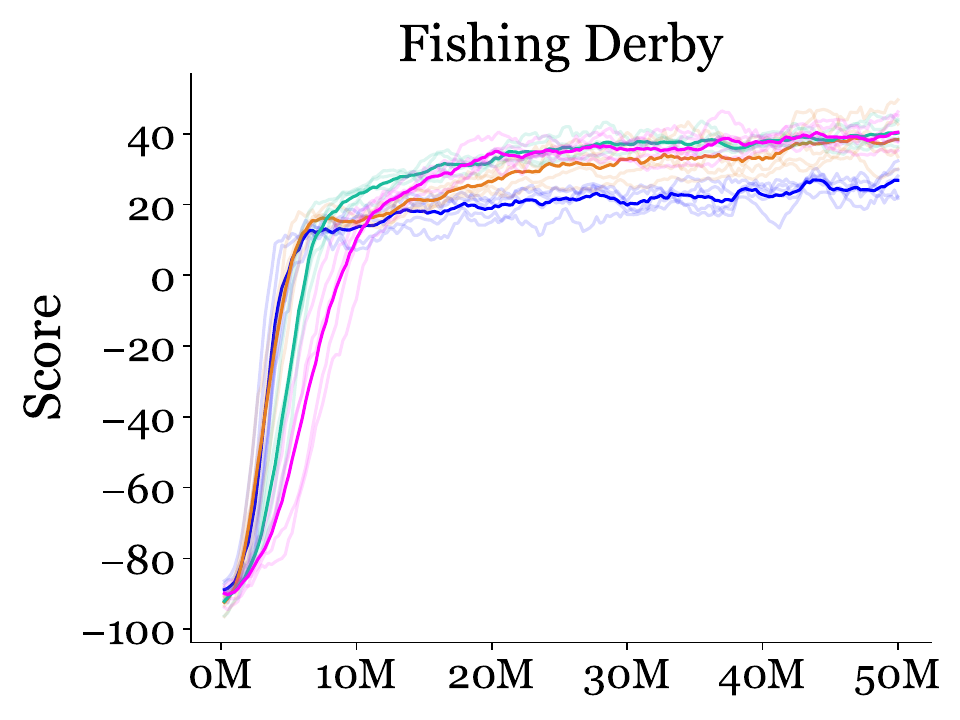} 
	\includegraphics[width=0.22\linewidth]{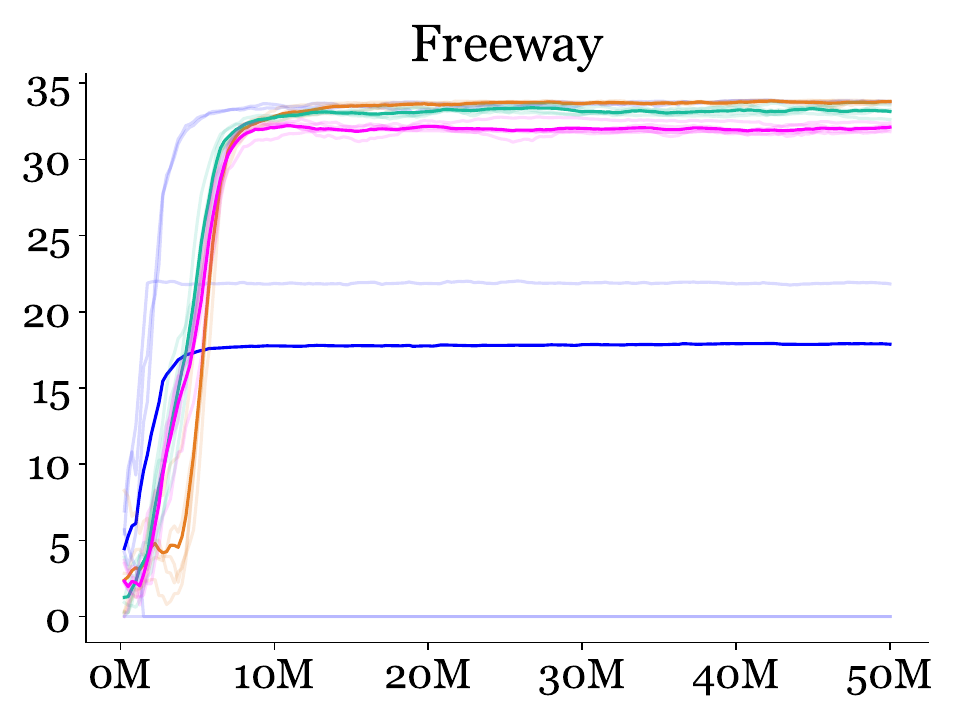} 
	\includegraphics[width=0.22\linewidth]{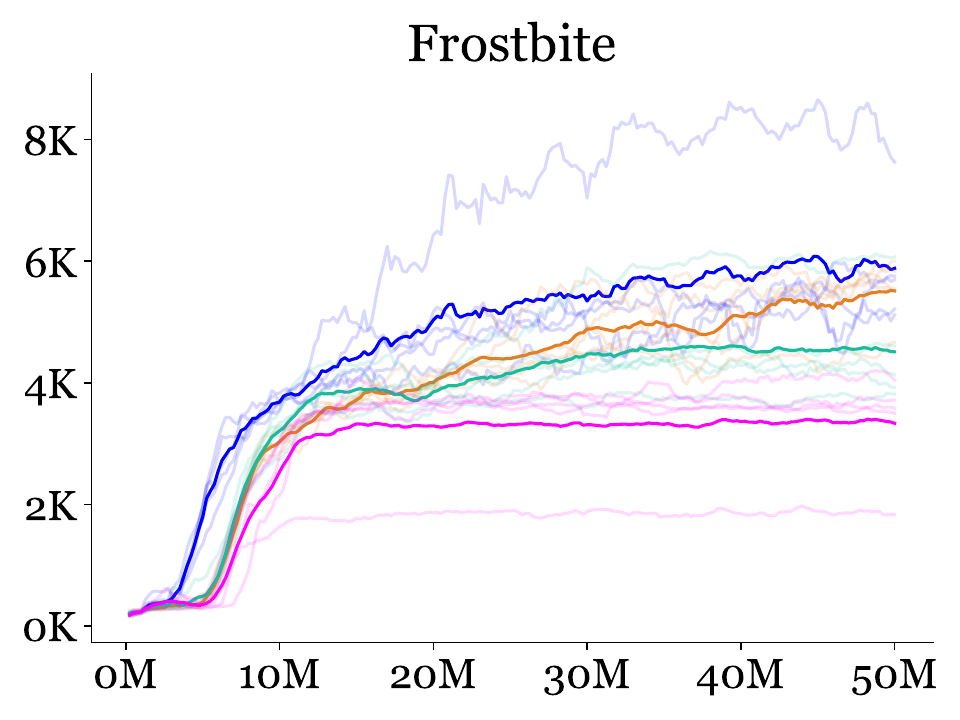} 
	\includegraphics[width=0.22\linewidth]{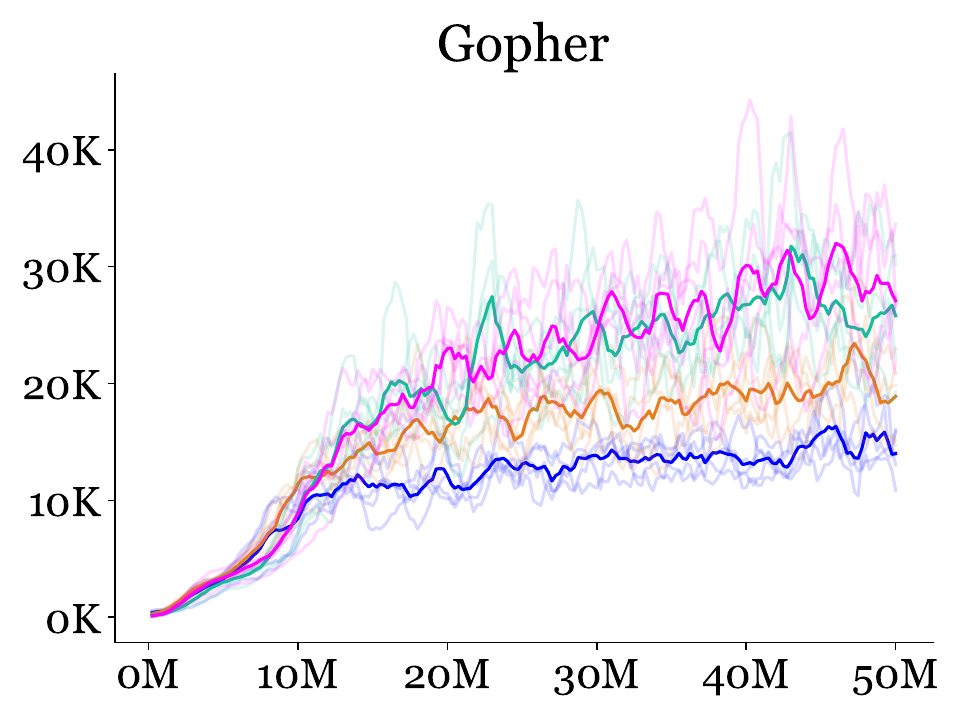} 
	\includegraphics[width=0.22\linewidth]{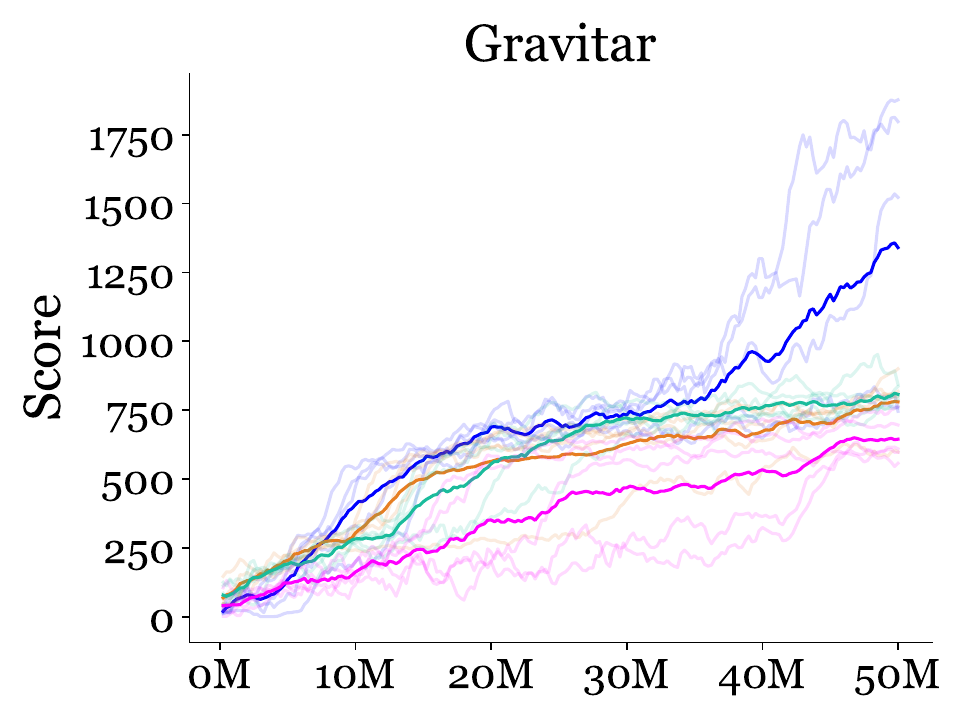} 
	\includegraphics[width=0.22\linewidth]{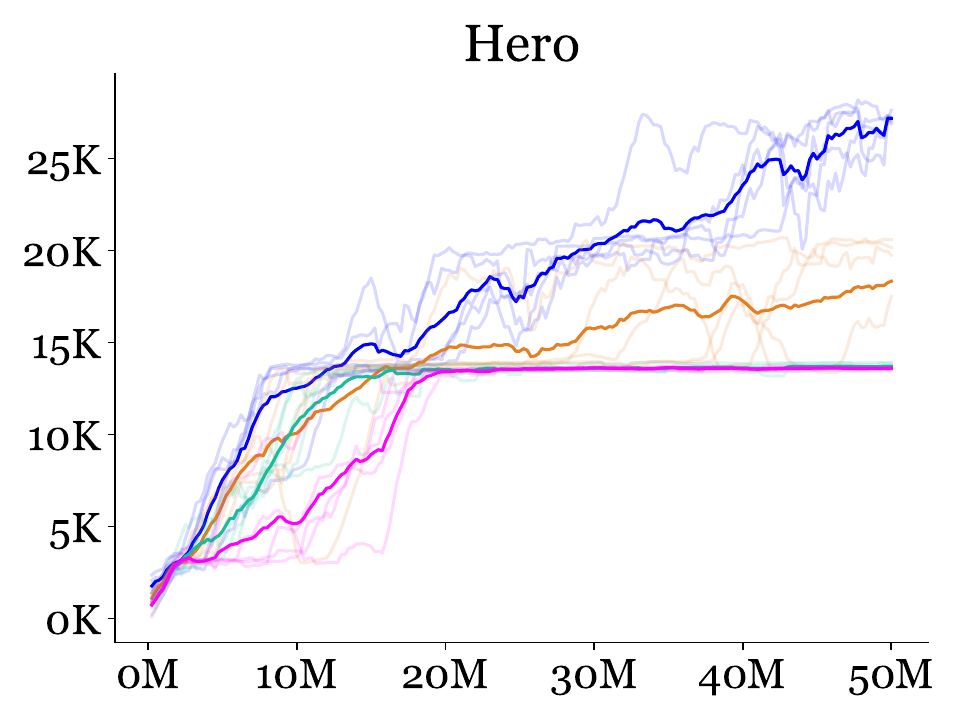} 
	\includegraphics[width=0.22\linewidth]{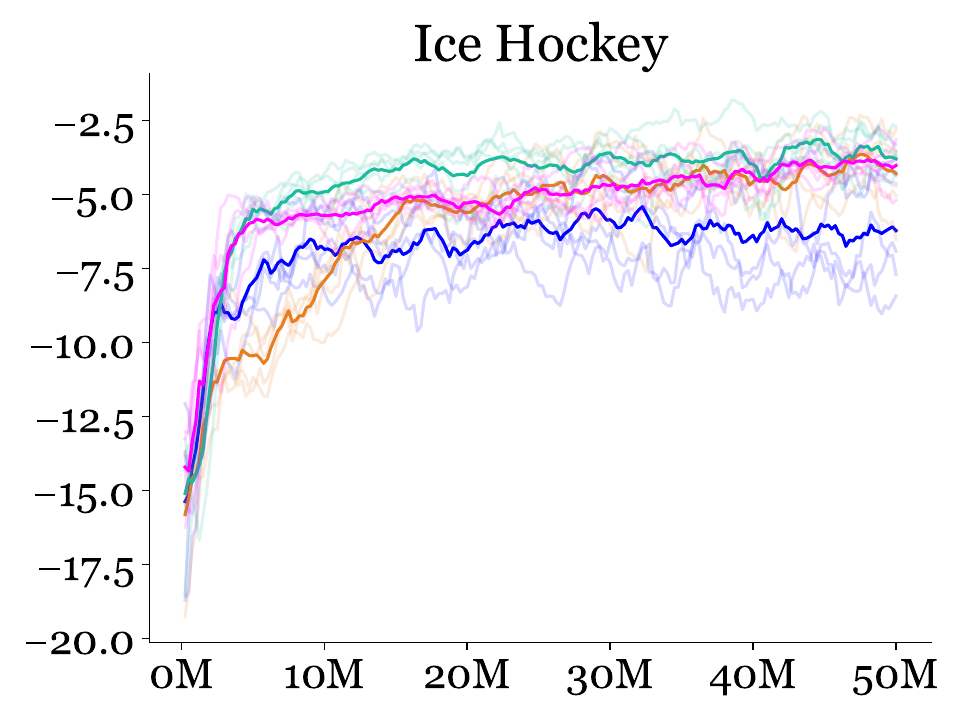} 
	\includegraphics[width=0.22\linewidth]{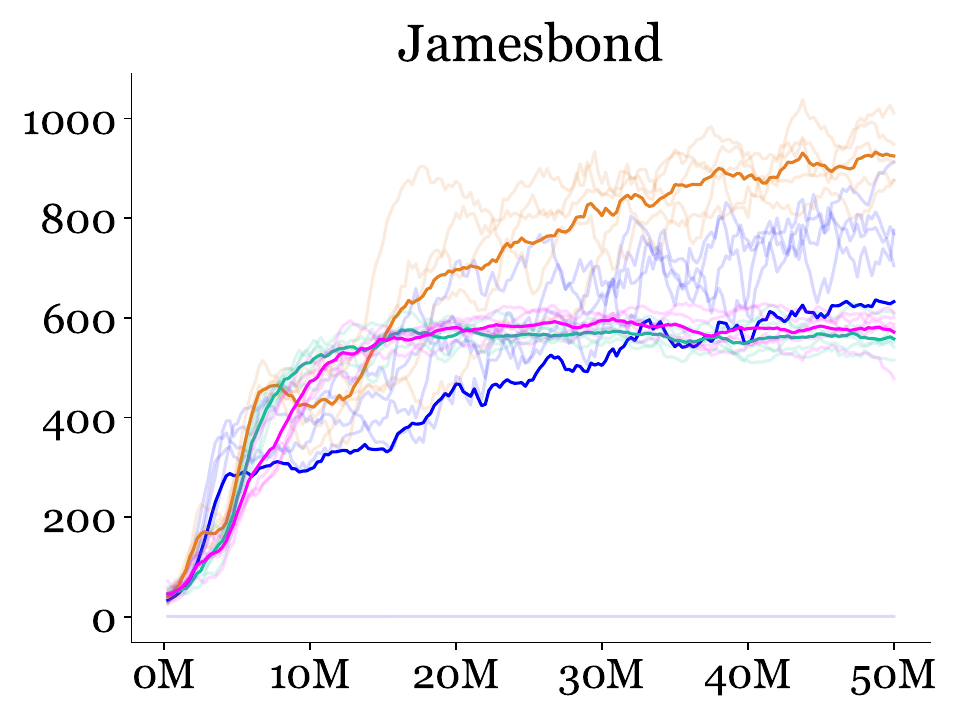} 
\end{figure}

\clearpage
\begin{table*}[!t] \centering
    \resizebox{0.85\columnwidth}{!}{
    \begin{tabular}{@{}l|c|c|c|c@{}}\toprule \textbf{Environment} & Double DQN & DDQL & DM-DDQL & DN-DDQL \\ 
 \midrule 
\textsc{Alien} & 2805.1& \textbf{4314.7} & 3483.1& 1811.0\\ 
\textsc{Amidar} & 508.9& \textbf{922.1} & 600.9& 366.6\\ 
\textsc{Assault} & 1888.3& 1870.4& 2290.9& \textbf{2462.5} \\ 
\textsc{Asterix} & 13371.6& 22896.3& 26851.5& \textbf{27554.4} \\ 
\textsc{Asteroids} & 1172.4& \textbf{1221.9} & 1176.6& 1071.7\\ 
\textsc{Atlantis} & 788923.3& 782803.3& 827005.0& \textbf{852533.3} \\ 
\textsc{BankHeist} & 1047.5& 964.3& 1148.2& \textbf{1242.7} \\ 
\textsc{BattleZone} & 21036.8& 28890.6& \textbf{30055.1} & 28458.9\\ 
\textsc{BeamRider} & 3988.6& 5110.9& \textbf{6209.2} & 6084.0\\ 
\textsc{Berzerk} & 579.2& \textbf{664.4} & 636.1& 555.5\\ 
\textsc{Bowling} & 1.1& \textbf{46.1} & 33.7& 31.1\\ 
\textsc{Boxing} & 93.6& 95.8& 97.2& \textbf{97.5} \\ 
\textsc{Breakout} & 163.4& 234.2& 292.8& \textbf{307.0} \\ 
\textsc{Centipede} & 4942.2& \textbf{4993.0} & 2623.3& 1875.2\\ 
\textsc{ChopperCommand} & 2204.7& \textbf{5620.7} & 4754.7& 4152.3\\ 
\textsc{CrazyClimber} & 110893.9& \textbf{122650.0} & 122173.1& 115661.2\\ 
\textsc{Defender} & 7330.3& 11725.4& 28707.9& \textbf{33471.0} \\ 
\textsc{DemonAttack} & 9060.2& 10496.3& 91866.8& \textbf{99210.5} \\ 
\textsc{DoubleDunk} & $-9.5$& \textbf{$-$5.6} & $-10.1$ & $-8.2$\\ 
\textsc{Enduro} & 1820.8& 2044.0& \textbf{2144.3} & 2133.7\\ 
\textsc{FishingDerby} & 26.9& 37.9& \textbf{40.3} & 39.7\\ 
\textsc{Freeway} & 17.9& \textbf{33.8} & 33.2& 32.1\\ 
\textsc{Frostbite} & \textbf{5889.5} & 5520.6& 4531.7& 3341.6\\ 
\textsc{Gopher} & 14134.3& 19253.2& 25712.6& \textbf{26542.9} \\ 
\textsc{Gravitar} & \textbf{1326.1} & 776.0& 801.5& 641.0\\ 
\textsc{Hero} & \textbf{27155.9} & 18281.5& 13710.1& 13595.8\\ 
\textsc{IceHockey} & $-6.1$& $-4.4$ & \textbf{$-$4.0} & \textbf{$-$4.0} \\ 
\textsc{Jamesbond} & 626.2& \textbf{934.6} & 555.7& 569.2\\ 
\textsc{Kangaroo} & 5040.0& 6560.8& \textbf{6722.1} & 5106.6\\ 
\textsc{Krull} & 8324.5& 8381.0& 8580.0& \textbf{9190.5} \\ 
\textsc{KungFuMaster} & 26818.1& 27777.2& \textbf{30671.6} & 25966.0\\ 
\textsc{MontezumaRevenge} & 0.0& 0.0& \textbf{53.3} & 0.0\\ 
\textsc{MsPacman} & 2728.3& 3834.2& \textbf{4304.7} & 4215.8\\ 
\textsc{NameThisGame} & 5810.1& 7618.7& 13759.8& \textbf{14972.4} \\ 
\textsc{Phoenix} & 6245.6& \textbf{14372.1} & 5120.8& 5084.8\\ 
\textsc{Pitfall} & $-37.8$ & $-40.1$ & $-4.8$ & \textbf{$-$1.7} \\ 
\textsc{Pong} & 10.8& \textbf{19.8} & 17.6 & 18.4\\ 
\textsc{PrivateEye} & $-137.1$& \textbf{100.4} & 81.6& 100.0\\ 
\textsc{Qbert} & 13104.1& \textbf{14269.0} & 12373.4& 10806.1\\ 
\textsc{Riverraid} & 13160.2& \textbf{14953.2} & 12309.0& 11859.0\\ 
\textsc{RoadRunner} & 52216.0& 56757.9& \textbf{57172.6} & 53529.3\\ 
\textsc{Robotank} & 61.5& 63.8& \textbf{65.2} & 61.0\\ 
\textsc{Seaquest} & 6414.8& 19617.2& \textbf{30658.2} & 9692.6\\ 
\textsc{Skiing} & $-19531.8$ & \textbf{$-$17661.6} & $-24623.8$ & $-29432.1$ \\ 
\textsc{Solaris} & 579.7& \textbf{805.7} & 222.0& 306.4\\ 
\textsc{SpaceInvaders} & 3544.4& 5609.2& \textbf{9559.0} & 7593.0\\ 
\textsc{StarGunner} & 53216.3& 59591.5& \textbf{76406.0} & 71549.3\\ 
\textsc{Surround} & $-5.3$ & 0.0 & \textbf{4.7} & $-3.6$\\ 
\textsc{Tennis} & 11.1& \textbf{22.6} & 13.3& $-3.5$\\ 
\textsc{TimePilot} & 9128.9& 10314.7& \textbf{12015.9} & 10932.3\\ 
\textsc{Tutankham} & \textbf{184.3} & 183.0& 159.8& 176.1\\ 
\textsc{UpNDown} & 11864.2& 15349.0& 19151.8& \textbf{23914.9} \\ 
\textsc{Venture} & 628.5& \textbf{1160.2} & 6.1& 0.0\\ 
\textsc{VideoPinball} & 222658.8& 360292.7& \textbf{364436.5} & 185664.2\\ 
\textsc{WizardOfWor} & 4204.4& 6142.2& \textbf{8681.3} & 6671.7\\ 
\textsc{YarsRevenge} & 63051.5& \textbf{69460.9} & 64182.2& 48159.0\\ 
\textsc{Zaxxon} & 7344.5& 11636.4& \textbf{11697.9} & 10771.2\\ 

    \bottomrule
    \end{tabular}
    }
    \captionsetup{width=0.85\linewidth}
    \caption{The mean evaluation score across the last three evaluations during training for different algorithms. The highest scores for an environment are bolded.}
    \label{table_results}
\end{table*}

\clearpage
\newpage
\section*{NeurIPS Paper Checklist}

\begin{enumerate}

\item {\bf Claims}
    \item[] Question: Do the main claims made in the abstract and introduction accurately reflect the paper's contributions and scope?
    \item[] Answer: \answerYes{}{} 
    \item[] Justification: The abstract and introduction claim that we adapt Double Q-learning to deep RL and that it outperforms Double DQN and reduces overestimation. Section 3 introduces this adaptation, DDQL, and Section 4 demonstrates DDQL's reduced overestimation and improved performance.

\item {\bf Limitations}
    \item[] Question: Does the paper discuss the limitations of the work performed by the authors?
    \item[] Answer: \answerYes{} 
    \item[] Justification: In the discussion and conclusion section, we discuss how the paper's scope is limited to discrete actions and that it does not incorporate many recent deep RL advances.

\item {\bf Theory assumptions and proofs}
    \item[] Question: For each theoretical result, does the paper provide the full set of assumptions and a complete (and correct) proof?
    \item[] Answer: \answerNA{} 
    \item[] Justification: The paper does not include theoretical results.

    \item {\bf Experimental result reproducibility}
    \item[] Question: Does the paper fully disclose all the information needed to reproduce the main experimental results of the paper to the extent that it affects the main claims and/or conclusions of the paper (regardless of whether the code and data are provided or not)?
    \item[] Answer: \answerYes{} 
    \item[] Justification: The appendix provides granular details for reproducibility purposes for readers as well as citations to work we have closely replicated. We additionally released our code, which runs on entirely open source software, ensuring reproducibility.

\item {\bf Open access to data and code}
    \item[] Question: Does the paper provide open access to the data and code, with sufficient instructions to faithfully reproduce the main experimental results, as described in supplemental material?
    \item[] Answer: \answerYes{} 
    \item[] Justification: We have provided a link to the code in the paper. We do not release the data.

\item {\bf Experimental setting/details}
    \item[] Question: Does the paper specify all the training and test details (e.g., data splits, hyperparameters, how they were chosen, type of optimizer) necessary to understand the results?
    \item[] Answer: \answerYes{} 
    \item[] Justification: Yes, our hyperparameters, optimizers, and games (our evaluation environments) are all clearly specified and we indicate where they come from.
    The appendices also provide detailed information on all hyperparameters, training details, and measurements.

\item {\bf Experiment statistical significance}
    \item[] Question: Does the paper report error bars suitably and correctly defined or other appropriate information about the statistical significance of the experiments?
    \item[] Answer: \answerYes{} 
    \item[] Justification: When appropriate (i.e., when we have sufficient runs), we use 95\% stratified bootstrap confidence intervals, following standard practice in the literature~\citep{statistical_precipice}. When we do not, we do not provide a statistically inaccurate summary or misuse error bars; rather we provide in the appendix the individual learning curves of each run.

\item {\bf Experiments compute resources}
    \item[] Question: For each experiment, does the paper provide sufficient information on the computer resources (type of compute workers, memory, time of execution) needed to reproduce the experiments?
    \item[] Answer: \answerYes{} 
    \item[] Justification: In a footnote in the experiments section we provide a rough calculation of the compute costs to run the experiments in the paper.
    
\item {\bf Code of ethics}
    \item[] Question: Does the research conducted in the paper conform, in every respect, with the NeurIPS Code of Ethics \url{https://neurips.cc/public/EthicsGuidelines}?
    \item[] Answer: \answerYes{} 
    \item[] Justification: We have reviewed the NeurIPS Code of Ethics and confirm that it conforms with the code. 
    Our research has no test subjects, human or environment data, or deployment into real world systems. Our work is algorithmic in nature and our data is entirely artificial, being game data.

\item {\bf Broader impacts}
    \item[] Question: Does the paper discuss both potential positive societal impacts and negative societal impacts of the work performed?
    \item[] Answer: \answerNA{} 
    \item[] Justification: This work examines fundamental deep RL algorithms on artificial environments (games) without direct societal impact. 
    While the broader field of deep RL may have societal impact, our paper does not create specific societal impacts that are not broadly applicable to the field.
    
\item {\bf Safeguards}
    \item[] Question: Does the paper describe safeguards that have been put in place for responsible release of data or models that have a high risk for misuse (e.g., pre-trained language models, image generators, or scraped datasets)?
    \item[] Answer: \answerNA{} 
    \item[] Justification: Our data is performance data on games and is not sensitive. Hence, safeguards are not needed.

\item {\bf Licenses for existing assets}
    \item[] Question: Are the creators or original owners of assets (e.g., code, data, models), used in the paper, properly credited and are the license and terms of use explicitly mentioned and properly respected?
    \item[] Answer: \answerYes{} 
    \item[] Justification: We use the PFRL library for our experiments. It is an open-source library with an associated paper, which we cite.

\item {\bf New assets}
    \item[] Question: Are new assets introduced in the paper well documented and is the documentation provided alongside the assets?
    \item[] Answer: \answerYes{} 
    \item[] Justification: We release our code, linked in the paper.

\item {\bf Crowdsourcing and research with human subjects}
    \item[] Question: For crowdsourcing experiments and research with human subjects, does the paper include the full text of instructions given to participants and screenshots, if applicable, as well as details about compensation (if any)? 
    \item[] Answer: \answerNA{} 
    \item[] Justification: This paper does not involve crowdsourcing nor research with human subjects.

\item {\bf Institutional review board (IRB) approvals or equivalent for research with human subjects}
    \item[] Question: Does the paper describe potential risks incurred by study participants, whether such risks were disclosed to the subjects, and whether Institutional Review Board (IRB) approvals (or an equivalent approval/review based on the requirements of your country or institution) were obtained?
    \item[] Answer: \answerNA{} 
    \item[] Justification: The paper does not have study participants nor crowdsourcing.

\item {\bf Declaration of LLM usage}
    \item[] Question: Does the paper describe the usage of LLMs if it is an important, original, or non-standard component of the core methods in this research? Note that if the LLM is used only for writing, editing, or formatting purposes and does \emph{not} impact the core methodology, scientific rigor, or originality of the research, declaration is not required.
    \item[] Answer: \answerNA{} 
    \item[] Justification: Our core method development in this research does not involve LLMs as any important, original, or non-standard components.

\end{enumerate}

\end{document}